\documentclass{article}

\usepackage{microtype}
\usepackage{graphicx}
\usepackage{subcaption}
\usepackage{booktabs} 
\usepackage{multirow}
\usepackage{hyperref}

\usepackage[preprint]{icml2026}

\usepackage{amsmath}
\usepackage{amssymb}
\usepackage{mathtools}
\usepackage{amsthm}
\usepackage{enumitem}

\usepackage[capitalize,noabbrev]{cleveref}

\usepackage{CJKutf8}

\theoremstyle{plain}

\theoremstyle{definition}

\theoremstyle{remark}

\usepackage[disable,textsize=tiny]{todonotes}

\icmltitlerunning{Spectral Surgery: Training-Free Refinement of LoRA via Gradient-Guided Singular Value Reweighting}

\begin{document}

\twocolumn[
  \icmltitle{Spectral Surgery: Training-Free Refinement of LoRA via \\  Gradient-Guided Singular Value Reweighting}
  



  \icmlsetsymbol{equal}{*}

\begin{icmlauthorlist}
  \icmlauthor{Zailong Tian}{smu}
  \icmlauthor{Yanzhe Chen}{affB}
  \icmlauthor{Zhuoheng Han}{affC}
  \icmlauthor{Lizi Liao}{smu}
\end{icmlauthorlist}

\icmlaffiliation{smu}{School of Computing and Information Systems, Singapore Management University}
\icmlaffiliation{affB}{School of Computing, National University of Singapore}
\icmlaffiliation{affC}{State Key Laboratory for Multimedia Information Processing, Peking University}
\icmlcorrespondingauthor{Lizi Liao}{lzliao@smu.edu.sg}

\icmlkeywords{Large Language Models, LoRA, Singular Value Decomposition, Parameter-Efficient Fine-Tuning}
  \vskip 0.3in
  ]



\printAffiliationsAndNotice{}  

\begin{abstract}
Low-Rank Adaptation (LoRA) improves downstream performance by restricting task updates to a low-rank parameter subspace, yet how this limited capacity is allocated within a trained adapter remains unclear. Through a geometric and empirical study across multiple tasks and backbones, we find that trained LoRA updates often exhibit an inefficient spectrum: task effects concentrate in a small subset of singular directions, while many remaining components are neutral or detrimental, motivating post-hoc refinement within the learned subspace. We propose \textbf{Spectral Surgery}, a simple training-free refinement that decomposes a LoRA update with SVD, estimates per-component sensitivity using gradients on a small calibration set, and reweights singular values under a magnitude constraint while keeping the learned directions fixed. Across Llama-3.1-8B and Qwen3-8B on four benchmarks, Spectral Surgery yields consistent gains (up to +4.4 points on CommonsenseQA and +2.4 pass@1 on HumanEval) by adjusting only $\approx 1{,}000$ scalar coefficients. These results demonstrate that SVD-structured, low-cost parameter editing can serve as a practical route to improving trained LoRA adapters in a purely post-hoc manner.
\end{abstract}


\section{Introduction}
Low-Rank Adaptation (LoRA) has become a standard for task-specific adaptation of large language models (LLMs) due to its strong empirical performance and favorable efficiency profile: instead of updating the full weight space, LoRA injects a low-rank update $\Delta W$ into selected linear layers while keeping the backbone frozen \citep{hu2022lora}. In practice, a LoRA adapter is often treated as a static endpoint of training: once optimization converges, the resulting low-rank matrix is deployed as-is and rarely revisited.

This “train-then-freeze” convention hides a natural question. From the task-vector view of adaptation, a trained adapter is a compact representation of a task-induced displacement in parameter space \citep{ilharco2023editing}. Yet, the convergence of stochastic gradient descent (SGD) does not guarantee that the limited representational budget of a rank-$r$ update is used efficiently: even within a fixed low-rank subspace, different allocations can encode dramatically different behaviors. This raises a basic efficiency gap that is orthogonal to choosing a larger rank or a better training recipe: \emph{given a converged LoRA adapter, is the capacity within its learned rank being allocated in the most useful way?}

Inspired by the finding of \citet{sharma2024the} that singular-structure manipulation can improve model performance, we adopt the singular value decomposition (SVD; \citealp{eckart1936approximation}) of LoRA updates as a lens for probing this internal allocation. Our empirical study reveals a striking dichotomy between these two ingredients. In residual-writing projections---notably the attention output projection and the MLP down projection---the learned singular subspaces often exhibit strong alignment across layers and even across module types, suggesting that optimization is comparatively reliable at discovering task-aligned directions (~Figure~\ref{fig:align_intra}). In contrast, the spectral allocation is often inefficient: substantial energy is assigned to neutral or harmful components that dilute the signal. Put differently, even when the adapter finds the right directions, it may assign the wrong spectral weights. This reframes a trained LoRA adapter not as a uniformly useful rank-$r$ object, but as a mixture in which only part of the low-rank capacity carries task-relevant signal.

This motivates a natural question: can we improve a \emph{trained} LoRA adapter \emph{without re-training}, by reallocating capacity \emph{within} its learned low-rank space? We answer this question with \textbf{Spectral Surgery}, a \emph{training-free} and \emph{post-hoc} refinement method that edits LoRA adapters after convergence. The core principle is simple: \emph{keep the subspace, fix the spectrum}. That is, we preserve the learned directions $U$ and $V$ to maintain the observed geometric alignment, and adjust only the spectrum $\Sigma$ to redistribute energy across components under conservative magnitude/energy constraints.

Spectral Surgery proceeds in three steps: (1) \textbf{Decompose}: compute the SVD of the trained update $\Delta W = U \Sigma V^\top$; (2) \textbf{Estimate}: using a small calibration set, compute lightweight gradient-based signals to estimate the sensitivity of each singular component (i.e., how changes along that component affect the calibration objective); and (3) \textbf{Reweight}: preserve $U$ and $V$ but reweight the singular values in $\Sigma$ under magnitude control, yielding an edited update that stays within the learned subspace while reallocating spectral energy. Importantly, this refinement requires no additional fine-tuning and modifies only $O(r)$ scalar coefficients per edited module (often $\approx 10^3$ scalars in total).

We evaluate Spectral Surgery on two 8B-class backbones (Llama-3.1-8B~\citep{Dubey2024TheL3} and Qwen3-8B~\citep{Yang2025Qwen3TR}) across four benchmarks spanning reasoning, code generation, instruction following, and commonsense question answering. Despite its simplicity, spectrum-only refinement yields clear but task- and model-dependent improvements, with gains as large as $\approx +4.2$--$+4.4$ points on CommonsenseQA~\citep{commonsenseqa} and up to $\approx +2.4$ points on HumanEval~\cite{chen2021evaluating} pass@1, while adjusting only $\approx 1{,}000$ scalar coefficients.
We further include a random reweighting baseline that randomly increases some singular values while decreasing others, entirely ignoring sensitivity. Notably, such random spectrum edits can occasionally surpass the unedited adapter, suggesting a form of spectral brittleness in standard LoRA solutions: the learned spectrum may contain overfit or noisy allocations that even unguided regularization can partially correct.

Existing work that touches LoRA's structure largely falls into two regimes.
Training-time interventions improve low-rank adaptation by modifying how the adapter is learned---e.g., altering optimization dynamics, reallocating rank/budget, or shaping initialization via decomposition---but necessarily require re-training and do not answer how to improve an already-converged $\Delta W$ \citep{hayou2024loraplus,zhang2023adalora,meng2024pissa,corda2025}.
In parallel, diagnostic analyses use spectral or geometric lenses to reveal qualitative structure and potential pathologies in low-rank updates, but typically stop at interpretation rather than providing an actionable correction mechanism \citep{shuttleworth2024lora,bidermanlora}.
Our work targets an underexplored practical middle ground: \emph{post-training refinement}---a lightweight procedure that treats a trained adapter as an editable object and improves it after convergence, under explicit structural constraints.

In summary, our contributions are:
\begin{enumerate}
\item \textbf{Perspective.} We uncover a consistent \emph{subspace--spectrum dichotomy} in trained LoRA updates: in residual-writing projections, the learned singular \emph{subspaces} are comparatively stable and task-aligned, while the learned \emph{spectrum} can be inefficient or even detrimental, emerging as a primary post-training bottleneck.
\item \textbf{Method.} We propose \textbf{Spectral Surgery}, a \emph{training-free} refinement framework that keeps the learned directions fixed and reallocates capacity \emph{within} the low-rank space by reweighting singular values using lightweight calibration signals (gradient-projection sensitivities) under conservative magnitude/energy control.
\item \textbf{Findings.} Across multiple backbones and benchmarks, we show that spectrum-only editing can yield clear task-dependent gains while modifying only $O(r)$ scalars per module, and we further diagnose \emph{spectral brittleness} of standard LoRA solutions via randomized spectrum reweighting controls that can occasionally outperform the unedited baseline.
\end{enumerate}

\section{Related Work}

\paragraph{Training-time PEFT and decomposition-based low-rank adapters.}
LoRA ~\citep{hu2022lora} adapts LLMs by injecting a low-rank update $\Delta W=BA$ while freezing the backbone.
A large line of work improves low-rank adaptation during training by modifying optimization dynamics or the parameterization of updates.
For example, LoRA+ ~\citep{hayou2024loraplus} revisits LoRA’s optimization and improves efficiency by addressing the scale/learning-rate imbalance between the two factors.
AdaLoRA ~\citep{zhang2023adalora} adaptively allocates parameter budget (effectively rank/degree-of-freedom) across modules based on importance, using an SVD-structured formulation.
PiSSA ~\citep{meng2024pissa} leverages the principal components of the pre-trained weights to initialize and update the “signal” subspace while freezing the residual parts, improving convergence and final quality.
Context-aware decomposition methods such as CorDA~\cite{yang2024corda} and its extensions (e.g., CorDA++~\citep{corda2025}) orient the decomposition using activation statistics from a few samples and then train selected components for adaptation/retention.
MAP ~\citep{map2025} takes a complementary view by rigorously decoupling update direction and magnitude via vector normalization and a small number of scaling coefficients.
In contrast to these training-time designs, we intentionally keep LoRA training fixed and study what can be improved after convergence by editing the realized update.

\paragraph{Post-hoc singular-value optimization and training-free interventions.}
Several works also exploit spectral structure beyond standard SGD training, but differ in objective and signal.
ESSA ~\citep{essa2025} performs black-box evolutionary search for alignment, and makes the search scalable by restricting optimization to singular values of LoRA-style adapters.
GRASP ~\citep{grasp2025} uses gradient-based attribution on a small calibration set to retain sensitivity-critical singular components for compression (e.g., replacing redundant layers with adaptive singular parameters).
Our setting is different: we target post-training refinement of a trained LoRA adapter for downstream capability, not reward-driven alignment or model compression.
Methodologically, we use a white-box gradient-projection sensitivity signal to reweight (not select/prune) the spectrum, while freezing singular vectors to preserve the learned subspace geometry.

\paragraph{Task vectors, spectral diagnostics, and turning analysis into intervention.}
Fine-tuning updates can be interpreted as task vectors whose composition enables task arithmetic; most such methods treat an update as an atomic object at model/layer granularity.
We instead perform intra-adapter editing by decomposing a single LoRA update and modulating its internal spectral components.
Meanwhile, recent analyses compare LoRA and full fine-tuning through spectral lenses and reveal qualitative structural differences (e.g., “intruder” directions) ~\citep{shuttleworth2024lora,bidermanlora}.
Our work operationalizes this perspective by introducing a concrete, lightweight post-hoc procedure—spectrum-only editing under a fixed subspace—together with random controls and failure-case analysis that clarify when spectral perturbations help or hurt.
\section{Methodology}
\begin{figure*}[t!]
    \centering
    \begin{subfigure}[b]{0.29\textwidth}
        \centering
        \includegraphics[width=\linewidth]{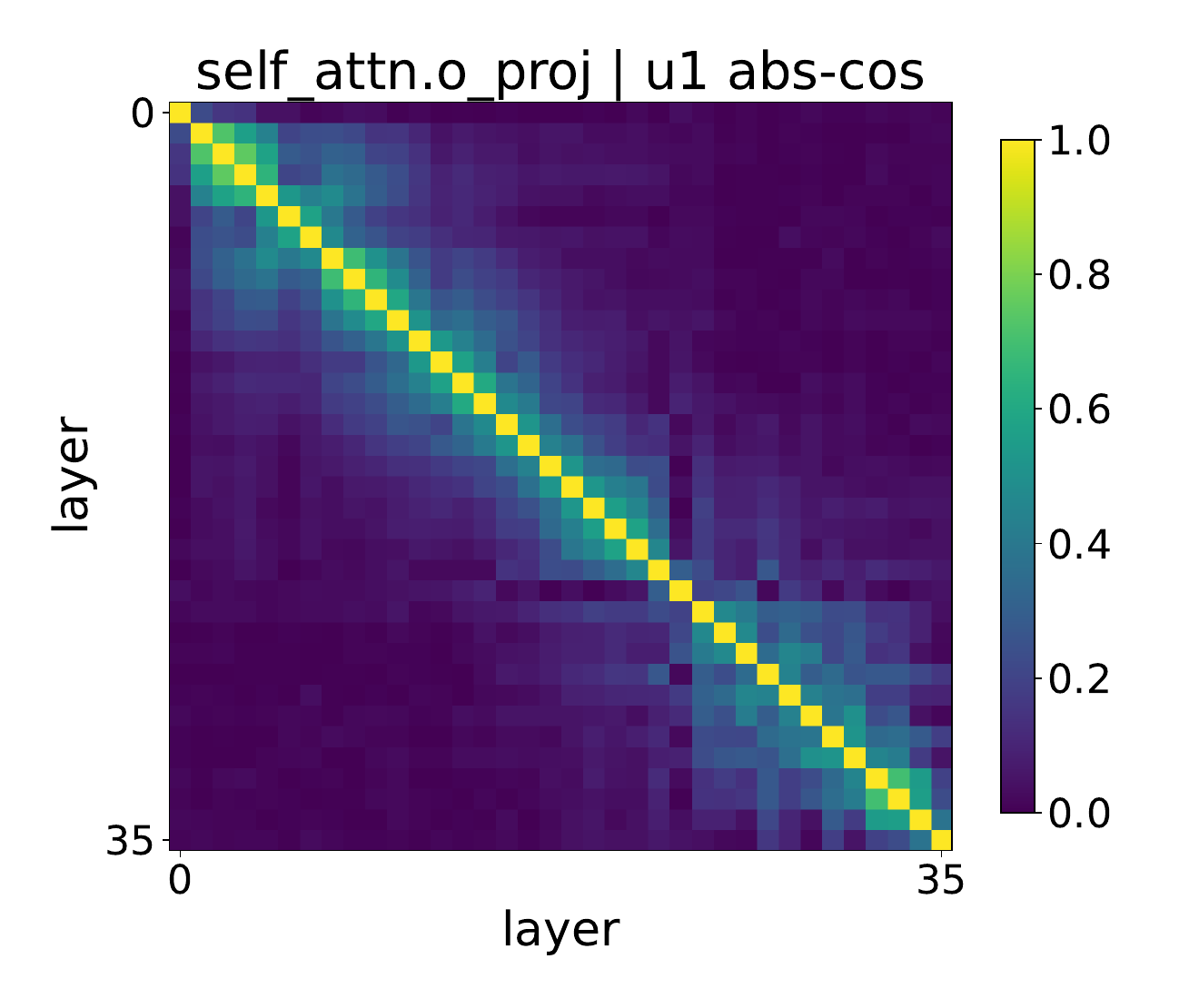}
        \caption{Principal direction similarity\linebreak ($|u_1^\top u_1|$).}
        \label{fig:align_u1}
    \end{subfigure}%
    \hspace{0.012\textwidth}%
    \begin{subfigure}[b]{0.29\textwidth}
        \centering
        \includegraphics[width=\linewidth]{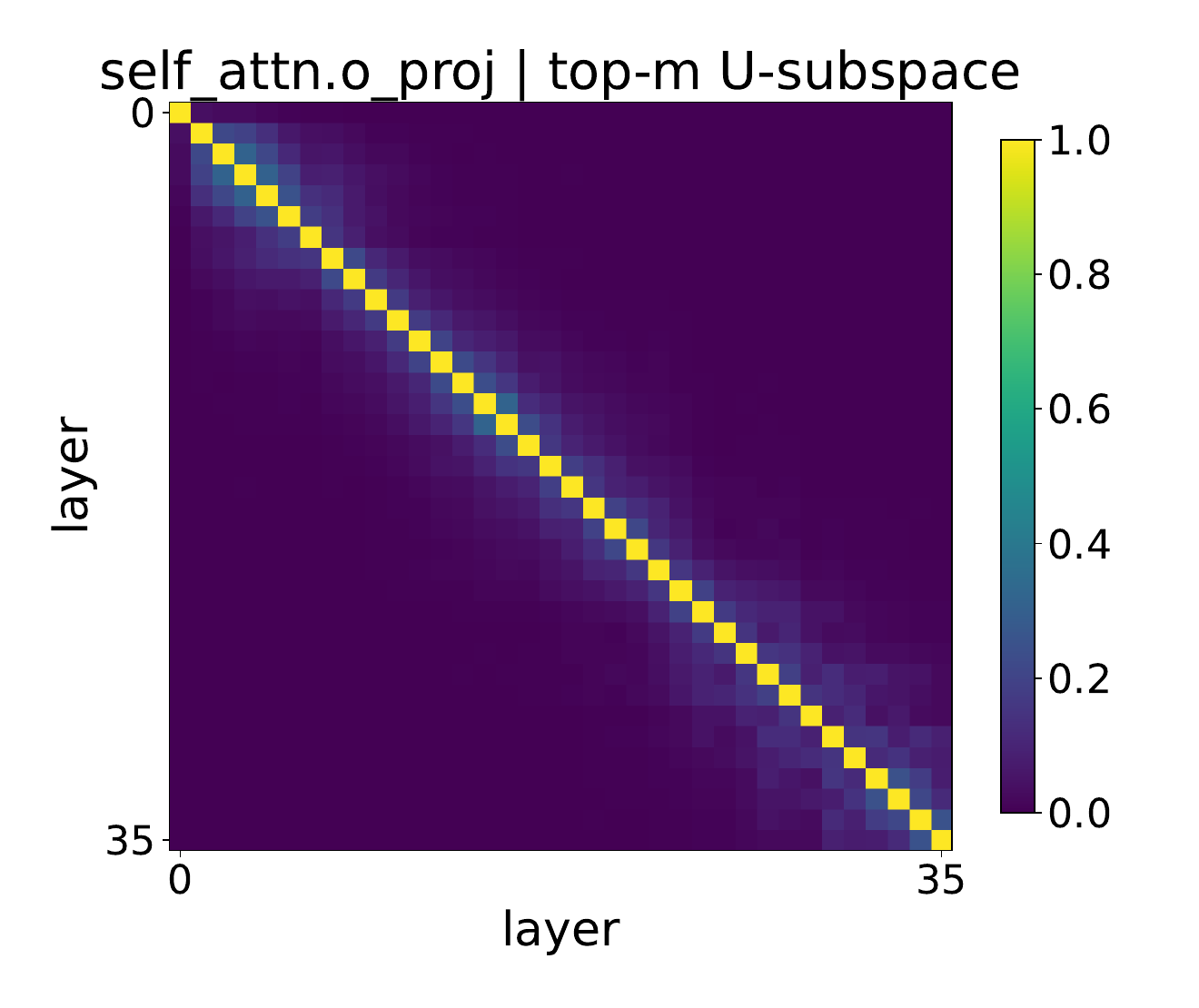}
        \caption{Top-$m$ output-subspace overlap ($\mathrm{Align}_U$, Eq.~\ref{eq:align_metric}).}
        \label{fig:align_subspace}
    \end{subfigure}%
    \hfill
\begin{subfigure}[b]{0.40\textwidth}
    \centering
    \raisebox{1.8ex}{\includegraphics[width=\linewidth]{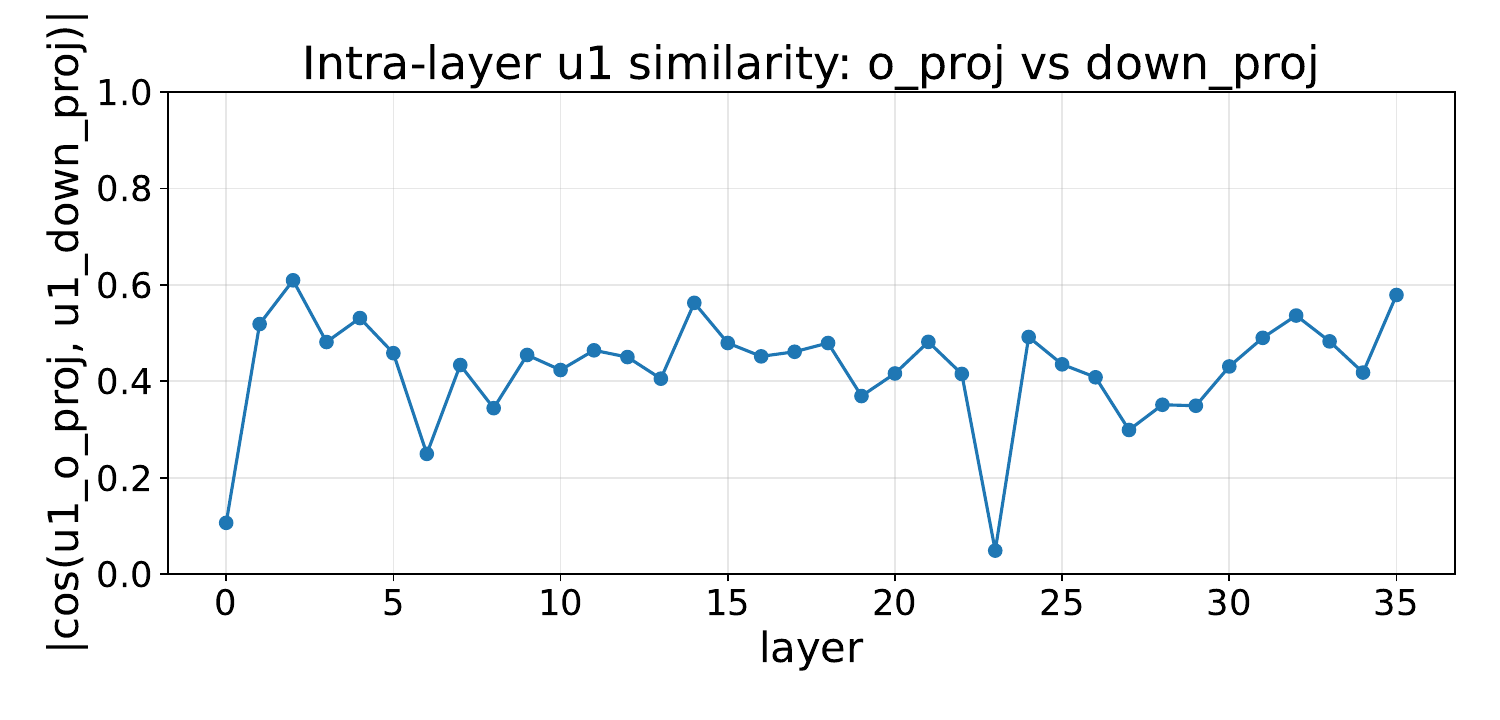}}
    \caption{Intra-layer synergy: \texttt{o\_proj} vs \texttt{down\_proj}.}
    \label{fig:align_intra}
\end{subfigure}

    \caption{\textbf{Geometric structure of LoRA updates in residual-writing modules.}
    We analyze LoRA updates of Qwen3-8B finetuned on the Alpaca dataset and visualize the alignment of the \texttt{o\_proj} module in the shared residual output space.
    \textbf{(a)} The leading output direction ($u_1$) exhibits consistently high similarity across layers.
    \textbf{(b)} The full top-4 output subspace $U^{(4)}$ (with rank $r=16$) is also highly stable across layers, indicating a shared update manifold in the residual stream.
    \textbf{(c)} Within each layer, \texttt{o\_proj} and \texttt{down\_proj} are strongly aligned relative to a random-subspace baseline ($m/d_{\text{model}}$).}
    \label{fig:alignment_analysis}
\end{figure*}

\subsection{Preliminaries: Spectral Decomposition of LoRA}
Consider a pre-trained linear layer with weight $W \in \mathbb{R}^{d_{\text{out}}\times d_{\text{in}}}$. 
LoRA parameterizes the weight update as a low-rank product $\Delta W = BA$, where $B \in \mathbb{R}^{d_{\text{out}}\times r}$ and $A \in \mathbb{R}^{r \times d_{\text{in}}}$ ($r \ll \min(d_{\text{out}}, d_{\text{in}})$).
We analyze the spectral structure of this update by computing the thin SVD of the product matrix:
\begin{equation}
    \Delta W = U \Sigma V^\top, \quad \Sigma = \mathrm{diag}(\sigma),
\end{equation}
where $U \in \mathbb{R}^{d_{\text{out}}\times r}$ and $V \in \mathbb{R}^{d_{\text{in}}\times r}$ have orthonormal columns, and $\sigma \in \mathbb{R}^r_{\ge 0}$ denotes the singular values.

\subsection{Geometric Motivation: Subspace Alignment in Residual Projections}
\label{sec:motivation}

We focus our spectral editing on the attention output projection (\texttt{o\_proj}) and the MLP down projection (\texttt{down\_proj}).
Unlike input-side projections (e.g., \texttt{q\_proj}, \texttt{up\_proj}), these modules write directly back into the residual stream $\mathbb{R}^{d_{\text{model}}}$.
This shared output space allows a clean geometric comparison of LoRA update directions across layers and module types.

\paragraph{Empirical Observation.}
For each LoRA module, we consider the low-rank update $\Delta W = BA$ and compute a thin SVD:
\begin{equation}
\Delta W = U \Sigma V^\top,
\end{equation}
where the columns of $U$ lie in the residual output space (for \texttt{o\_proj} and \texttt{down\_proj}, $U \in \mathbb{R}^{d_{\text{model}}\times r}$).
We use two complementary alignment measures across layers:
(i) the similarity of the leading output direction $u_1$ (Figure~\ref{fig:align_u1}),
and (ii) the overlap of the top-$m$ output subspaces (Figure~\ref{fig:align_subspace}).
Concretely, letting $U^{(m)}\in\mathbb{R}^{d_{\text{model}}\times m}$ denote the first $m$ columns of $U$, we define:
\begin{equation}
\label{eq:align_metric}
\mathrm{Align}_U(U_a, U_b)
=
\frac{1}{m}\left\lVert (U_a^{(m)})^\top U_b^{(m)} \right\rVert_F^2
\in [0,1],
\end{equation}
which equals the average $\cos^2$ of principal angles between the two $m$-dimensional subspaces.
For reference, the expected overlap between two random $m$-dimensional subspaces in $\mathbb{R}^{d_{\text{model}}}$ is approximately $m/d_{\text{model}}$
(e.g., $16/4096 \approx 0.004$ for Llama-3.1-8B as in ~\citep{Dubey2024TheL3}).

Our analysis reveals two geometric properties of residual-writing updates, which we observe to be a \textbf{general phenomenon} across different modules (visualized for \texttt{o\_proj} in Figure~\ref{fig:alignment_analysis}; see Appendix~\ref{app:heatmap_wall} for other modules):

\begin{itemize}
    \item \textbf{Layer invariance (inter-layer).}
    Across layers, both the leading direction ($u_1$) and the full top-$m$ output subspace ($U^{(m)}$) exhibit consistently high similarity.
    The strong off-diagonal structure in Figures~\ref{fig:align_u1}--\ref{fig:align_subspace} indicates that LoRA updates concentrate on an approximately layer-invariant manifold in the residual stream.

    \item \textbf{Module synergy (intra-layer).}
    Within the same layer, \texttt{o\_proj} and \texttt{down\_proj} share a strongly aligned output subspace (Figure~\ref{fig:align_intra}),
    substantially exceeding the random-subspace baseline ($m/d_{\text{model}} \approx 0.004$).
    This suggests that attention and FFN blocks coordinate updates to shared residual features.
\end{itemize}

\paragraph{Implication.}
These phenomena provide a geometric motivation for our method.
LoRA updates in residual-writing modules do not wander arbitrarily; instead, they target a shared and stable low-rank manifold in the residual stream.
Consequently, we opt to \emph{edit strictly the spectrum} ($\Sigma$) while preserving the empirically stable singular subspaces (in particular, the output scaffold $U$).
This allows us to modulate update intensity along valid residual directions without disrupting the geometric coherence established during training.

\subsection{Sensitivity Estimation via Gradient Projections}
To estimate the importance of each spectral component, we use a small calibration dataset $\mathcal{D}$ and loss function $\mathcal{L}$. 
Let $G = \frac{\partial \mathcal{L}}{\partial \Delta W}$ denote the gradient of the loss with respect to the accumulated update matrix.
The sensitivity of the $k$-th singular component is derived from the directional derivative along the unit matrix $u_k v_k^\top$:
\begin{equation}
    g_k = \langle G, u_k v_k^\top \rangle = u_k^\top G v_k.
\end{equation}
We aggregate the scalar sensitivity magnitude $s_k = |g_k|$ over calibration examples. 
Intuitively, a large $s_k$ indicates that perturbing $\sigma_k$ would strongly affect the task loss.

\subsection{Singular Value Reweighting (Spectral Editing)}
\label{sec:editing}

\begin{figure*}[ht]
    \centering
    \includegraphics[width=\linewidth]{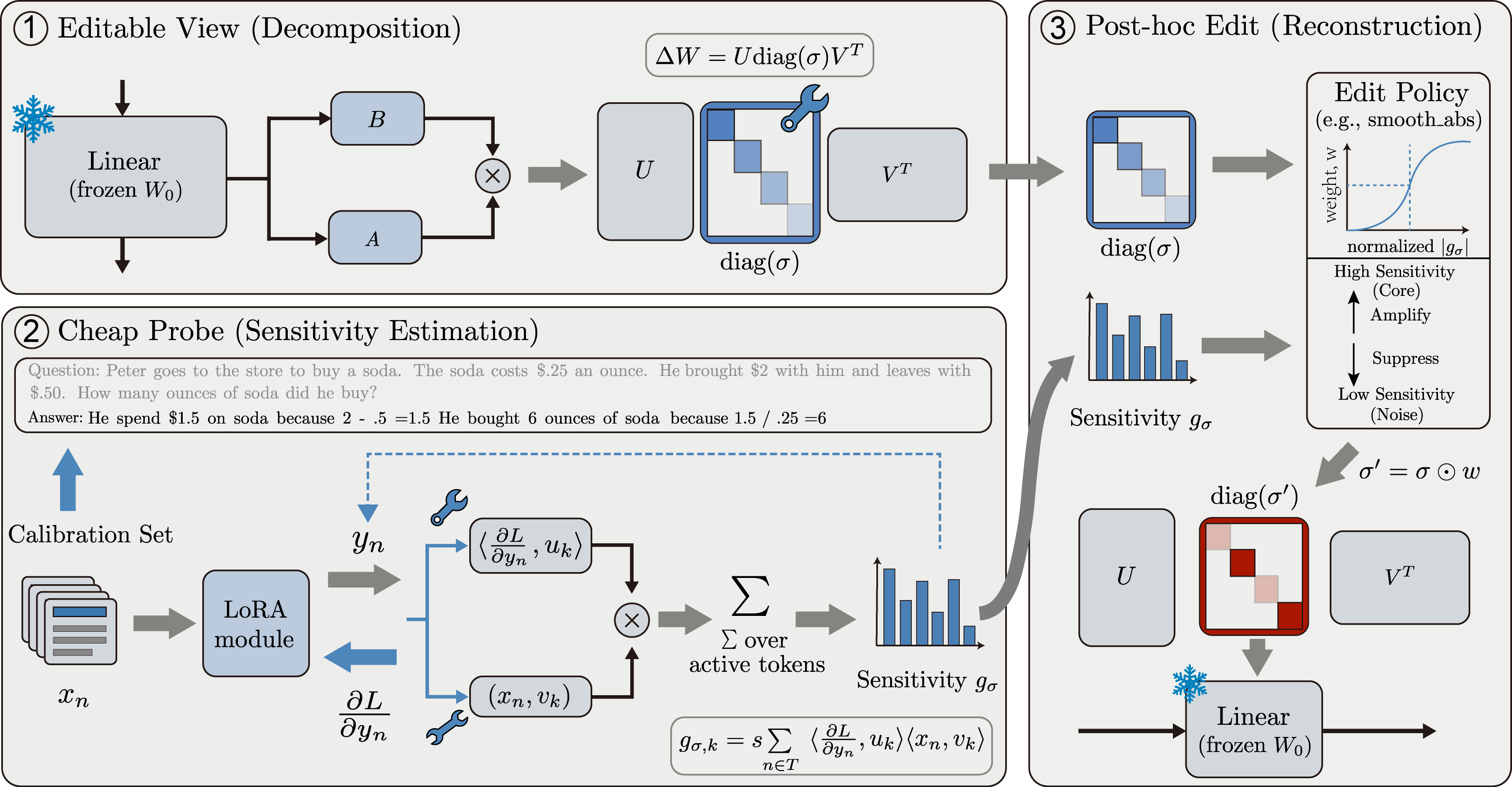}
    \caption{\textbf{Overview of Spectral Editing.}
    We decompose the LoRA update $\Delta W$ into singular components ($U, \Sigma, V^\top$).
    We estimate a sensitivity score $s_k$ for each singular component using gradient projections on a calibration set,
    and then reweight the singular values in $\Sigma$ (via hard selection or continuous reweighting) to amplify task-relevant directions while suppressing noise.
    This reconstructs an edited update $\Delta W'$ without altering the singular subspaces.
   }
    \label{fig:method_lora}
\end{figure*}

We edit only the spectrum of $\Delta W$ while keeping the singular subspaces fixed.
Let $\Delta W = U \Sigma V^\top$ with singular values $\{\sigma_k\}_{k=1}^r$.
We produce edited singular values via a per-component scaling,
\begin{equation}
\sigma'_k = \alpha_k \sigma_k.
\end{equation}
The scaling $\alpha_k$ is derived from a sensitivity profile $\{g_k\}_{k=1}^r$ computed on a small calibration set.
For magnitude-based strategies, we use $s_k = |g_k|$ and apply a simple within-module normalization
(default: mean-absolute normalization), yielding normalized magnitudes $\{x_k\}$.

\vspace{2pt}
\paragraph{1. Hard Selection (\texttt{abs\_select}).}
We rank $x_k$ within each module and form a core set and a noise set.
The implementation uses rounded counts with constraints:
$k_{\text{core}}=\min(r,\max(\lfloor rp\rceil, k_{\min}))$ and
$k_{\text{noise}}=\min(r-k_{\text{core}},\lfloor rq\rceil)$,
where $p=\texttt{core\_frac}$, $q=\texttt{noise\_frac}$, and $k_{\min}=\texttt{min\_core\_k}$.
We then apply a piecewise multiplicative gate:
top-$k_{\text{core}}$ indices receive $\gamma_{\text{amp}}$,
bottom-$k_{\text{noise}}$ indices receive $\gamma_{\text{sup}}$,
and all remaining indices receive $\gamma_{\text{mid}}$
(\texttt{amp\_factor}, \texttt{sup\_factor}, \texttt{mid\_factor}).

\vspace{2pt}
\paragraph{2. Continuous Reweighting (\texttt{smooth\_abs}).}
To avoid brittle hard thresholds, we use a smooth sigmoid gate on normalized magnitudes $x_k$:
\begin{equation}
\alpha_k
= \gamma_{\text{sup}}
+ (\gamma_{\text{amp}}-\gamma_{\text{sup}})\cdot
\operatorname{sigmoid}\!\left(\frac{x_k-\mu}{\tau}\right).
\end{equation}
We set the center to a quantile $\mu \leftarrow Q_c(x)$ with $c=\texttt{smooth\_center\_q}$ (default: median).
The temperature adapts to the spread of magnitudes using a quantile range:
$\tau = T\cdot\big(Q_{q_{\text{hi}}}(x)-Q_{q_{\text{lo}}}(x)\big)$,
where $T=\texttt{smooth\_temperature}$, $q_{\text{lo}}=\texttt{noise\_frac}$, and $q_{\text{hi}}=1-\texttt{core\_frac}$.
If $q_{\text{hi}}\le q_{\text{lo}}$, the implementation falls back to $(q_{\text{lo}},q_{\text{hi}})=(0.25,0.75)$.
If the magnitudes are nearly degenerate (very small range), we skip shaping and set $\alpha_k\equiv \gamma_{\text{mid}}$.
Optionally (\texttt{smooth\_align\_mid}), we shift $\mu$ so that the gate value at the center quantile equals a prescribed midpoint $\gamma_{\text{mid}}$. 
\vspace{2pt}

\paragraph{3. Random Control (\texttt{random\_index}).}
As a control, we keep $(k_{\text{core}},k_{\text{noise}})$ the same as \texttt{abs\_select}, but sample indices uniformly at random and apply the same three-level scaling ($\gamma_{\text{amp}}/\gamma_{\text{sup}}/\gamma_{\text{mid}}$). This matched-random baseline keeps the spectrum change magnitude fixed while removing sensitivity-based targeting, so improvements over it reflect the benefit of sensitivity-guided selection.

\vspace{2pt}
\paragraph{4. Signed Update (\texttt{grad\_direction}).}
Beyond magnitude-based gating, we support a signed update using normalized signed sensitivities $\tilde g_k$.
In the default asymmetric setting (\texttt{asymmetric\_update}), we treat positive and negative parts differently:
we form $g^+_k=\max(\tilde g_k,0)$ and $g^-_k=-\max(-\tilde g_k,0)$ (optionally applying a power transform to $g^+_k$),
and combine them as $g^{\text{eff}}_k=\eta_{\text{sup}} g^+_k + \eta_{\text{amp}} g^-_k$
(\texttt{eta\_suppress}, \texttt{eta\_enhance}).
We then apply a multiplicative update:
\begin{equation}
\sigma'_k = \sigma_k \exp\!\left(-g^{\text{eff}}_k\right).
\end{equation}
When \texttt{asymmetric\_update} is disabled, we revert to the standard multiplicative form with a single step size $\eta$:
$\sigma'_k = \sigma_k \exp(-\eta \tilde g_k)$.
\paragraph{Reconstruction and magnitude control.}
We reconstruct $\Delta W' = U \Sigma' V^\top$.
To prevent numerical issues, we clamp $\sigma'_k \ge \texttt{sigma\_clip\_min}$ (default: $0$).
Optionally, we preserve spectral energy by renormalizing $\sigma'$ to match $\sigma$ under $\ell_1$ (default) or no constraint).
Finally, $\Delta W'$ can be converted back to LoRA-compatible factors for standard inference pipelines.
\subsection{Compute and Editing Overhead}
\label{sec:overhead}

Spectral Surgery edits a trained LoRA update $\Delta W$ by computing a thin SVD
$\Delta W = U\,\mathrm{diag}(\sigma)\,V^\top$ and rescaling only the $r$ singular values $\sigma\in\mathbb{R}^r$ while keeping $(U,V)$ fixed.
Thus, the editable degrees of freedom are $r$ scalars per edited module.

If we edit the same module families $\mathcal{M}$ in every layer of an $L$-layer transformer, the total number of edited scalars is:
\begin{equation}
\#\text{edited scalars} = L\,|\mathcal{M}|\,r.
\end{equation}
In our default setting, $\mathcal{M}=\{\texttt{o\_proj},\texttt{down\_proj}\}$, giving $2Lr$ edited scalars (Table~\ref{tab:edit_overhead_scalars}).

\begin{table}[!htp]
\caption{\textbf{Editing overhead (editable scalars).} Default setting edits \texttt{o\_proj} and \texttt{down\_proj} in every layer, yielding $2Lr$ editable scalars.}
\centering
\small
\renewcommand{\arraystretch}{1.1}
\begin{tabular*}{\linewidth}{@{\extracolsep{\fill}} l c c c @{}}
\toprule
\textbf{Backbone} & \textbf{\#Layers $L$} & \textbf{Rank $r$} & \textbf{\#Edited Scalars} \\
\midrule
Llama-3.1-8B & 32 & 16 & 1024 \\
Qwen3-8B     & 36 & 16 & 1152 \\
\bottomrule
\end{tabular*}

\label{tab:edit_overhead_scalars}
\end{table}

\section{Experimental Setup}
\label{sec:exp_setup}

\subsection{Benchmarks and Tasks}
We evaluate spectral editing across four capabilities. For each capability, we train a LoRA adapter on a domain corpus and evaluate on a standard downstream benchmark.

\begin{itemize}[topsep=-1pt]
\setlength{\parskip}{1pt}
    \item \textbf{Mathematical reasoning:} train on \textsc{MetaMath~\citep{yu2023metamath}}, evaluate on \textsc{GSM8K~\citep{cobbe2021gsm8k}}. We report \textbf{exact-match accuracy} (answer-extraction based).
    \item \textbf{Code generation:} train on \textsc{Magicoder~\citep{wei2023magicoder}}, evaluate on \textsc{HumanEval}. We use \textbf{execution-based evaluation} and report \textbf{pass@1}.
    \item \textbf{Instruction following:} train on \textsc{Alpaca~\citep{alpaca}}, evaluate on \textsc{IFEval~\citep{ifeval}}. We report the benchmark's \textbf{strict prompt-level accuracy}, which requires satisfying all verifiable constraints in a prompt.
    \item \textbf{Commonsense reasoning:} both train and evaluate on \textsc{CommonsenseQA}, reporting \textbf{multiple-choice accuracy} as measure.
\end{itemize}

\subsection{Models and Training Settings}
We conduct experiments on two 8B-class decoder-only models: Llama-3.1-8B and Qwen3-8B.
All adapters are trained with a standardized recipe to ensure fair comparisons: AdamW optimizer, a fixed epoch budget ($E=3$), and task-specific maximum sequence lengths following common practice.
To control computational cost across tasks, we cap the training set size at 50k examples for large corpora (MetaMath and Magicoder).
Unless stated otherwise, adapters are trained on the standard set of Transformer projection modules, while editing is restricted to residual-writing projections (\texttt{o\_proj} and \texttt{down\_proj}), motivated by the geometric observations in Sec.~\ref{sec:motivation}.

\subsection{Sensitivity Estimation and Editing}
We estimate sensitivities on a small calibration set sampled from a fixed proxy calibration set. We build calibration batches via teacher forcing on prompt--answer concatenations: labels are masked as $-100$ on prompt tokens so the loss is computed only on the answer continuation (including an optional separator space and EOS).
Calibration examples are sampled by optional shuffling with a fixed seed and a contiguous range selection with a start offset for reproducibility.

Unless otherwise specified, we use $N_{\text{cal}}=128$ examples and aggregate sensitivities by mean absolute value as defined in Sec.~\ref{sec:editing}.
We evaluate four edit policies (\texttt{abs\_select}, \texttt{smooth\_abs}, \texttt{random\_index}, \texttt{grad\_direction}) defined in Sec.~\ref{sec:editing}.
By default, we preserve the $\ell_1$ mass of the singular values (nuclear-norm preservation) to prevent trivial gains from global rescaling.
All policy hyperparameters are fixed across tasks and reported in Appendix~\ref{app:implementation}.

\subsection{Evaluation Protocol}
We perform evaluations using the LM Evaluation Harness \citep{gao2023fewshot_eval_framework}.
We use deterministic decoding to reduce evaluation variance (temperature $T=0$) unless otherwise specified (Full harness configuration is in Appendix~\ref{sec:eval_setup}.).

\begin{itemize}[topsep=-1pt]
\setlength{\parskip}{1pt}
    \item \textbf{GSM8K:} 5-shot prompting with greedy decoding; we report exact-match accuracy.
    \item \textbf{HumanEval:} 0-shot prompting with execution-based scoring; we report \texttt{pass@1}. 
    \item \textbf{IFEval:} harness default prompting and strict scoring; we report strict prompt-level accuracy.
    \item \textbf{CSQA:} 0-shot likelihood-based answer selection; we report accuracy.
\end{itemize}

Full hyper-parameters, library versions, and command-level configurations are provided in Appendix~\ref{app:implementation}.

\section{Results and Analysis}
\label{sec:results}

\subsection{Experimental Goals}
Our experiments are designed to investigate four Research Questions (\textbf{RQ}) regarding the editability of LoRA adapters:
\begin{itemize}[topsep=-1pt]
\setlength{\parskip}{1pt}
    \item \textbf{(RQ1) Effectiveness:} Can we improve a trained LoRA adapter in a training-free manner?
    \item \textbf{(RQ2) Signal vs. Perturbation:} Are the gains attributable to our specific sensitivity signal, or can random spectral perturbations achieve similar results?
    \item \textbf{(RQ3) Stability:} How sensitive is the method to the calibration budget and energy constraints?
    \item \textbf{(RQ4) Locality:} Does restricting edits to specific module families matter?
\end{itemize}
Unless otherwise stated, we use \texttt{calib\_samples=128} and \texttt{preserve\_energy=L1} as the default setting based on our ablation studies (Table~\ref{tab:calib_sweep} and Table~\ref{tab:energy_ablation}).

\subsection{Main Results: Effectiveness and Signal Verification}

\begin{table*}[!htp]
\centering
\footnotesize
\setlength{\tabcolsep}{6.5pt}
\renewcommand{\arraystretch}{1.06}
\caption{\textbf{Main Results with Recommended Settings.} Editing policies on Llama-3.1-8B and Qwen3-8B under the default configuration (\texttt{calib=128}, \texttt{energy=L1}). \textbf{Baseline} is the unedited adapter.}
\label{tab:main_results}
\begin{tabular}{p{1.8cm} l c c c c c}
\toprule
\textbf{Model} & \textbf{Task} & \textbf{Baseline} & \textbf{Random\_index} & \textbf{Smooth\_abs} & \textbf{Abs\_select} & \textbf{Grad\_dir} \\
\midrule
\multicolumn{7}{l}{\textit{Llama-3.1-8B}} \\
& \quad GSM8K (acc)           & 0.657 & 0.661 & \textbf{0.668} & 0.659 & 0.614 \\
& \quad HumanEval (pass@1)    & 0.488 & 0.488 & \textbf{0.494} & 0.488 & 0.476 \\
& \quad IFEval (score)        & 0.305 & 0.306 & 0.312          & \textbf{0.321} & 0.223 \\
& \quad CSQA (acc)            & 0.740 & 0.743 & 0.735          & 0.736 & \textbf{0.784} \\
\midrule
\multicolumn{7}{l}{\textit{Qwen3-8B}} \\
& \quad GSM8K (acc)           & 0.815 & 0.810 & 0.802          & 0.805 & \textbf{0.816} \\
& \quad HumanEval (pass@1)    & 0.488 & 0.494 & 0.506          & 0.506 & \textbf{0.512} \\
& \quad IFEval (score)        & 0.590 & \textbf{0.597} & 0.552 & 0.535 & 0.173 \\
& \quad CSQA (acc)            & \textbf{0.855} & 0.852 & 0.851 & 0.852 & 0.853 \\
\bottomrule
\end{tabular}
\end{table*}

Table~\ref{tab:main_results} summarizes the overall picture across two model families and four representative tasks, directly addressing \textbf{(RQ1--RQ2)}.

\paragraph{Consistent gains in aligned settings (RQ1).}
Table~\ref{tab:main_results} shows that spectrum-only editing can improve a trained LoRA adapter without further training, supporting (RQ1).
Across eight (model, task) pairs, the best edited policy improves over the unedited baseline in seven cases, with the largest gain on Llama \textsc{CSQA} (+0.044 with \texttt{grad\_direction}).
Since the directions remain the same and we only reweigh their magnitudes, the results suggest that some useful components are under-emphasized after training and can be boosted post hoc. We also notice that when the calibration objective aligns with the downstream metric, spectral editing can improve an already trained adapter without further training. The clearest example in Table~\ref{tab:main_results} is CSQA on Llama-3.1-8B, where \texttt{grad\_direction} yields a $\sim$4.4\% absolute gain over the baseline (0.784 vs.\ 0.740). This suggests that gradient-guided reweighting can selectively amplify useful directions already present in the learned spectrum.


\begin{figure}[t]
    \centering
    \includegraphics[width=0.85\linewidth]{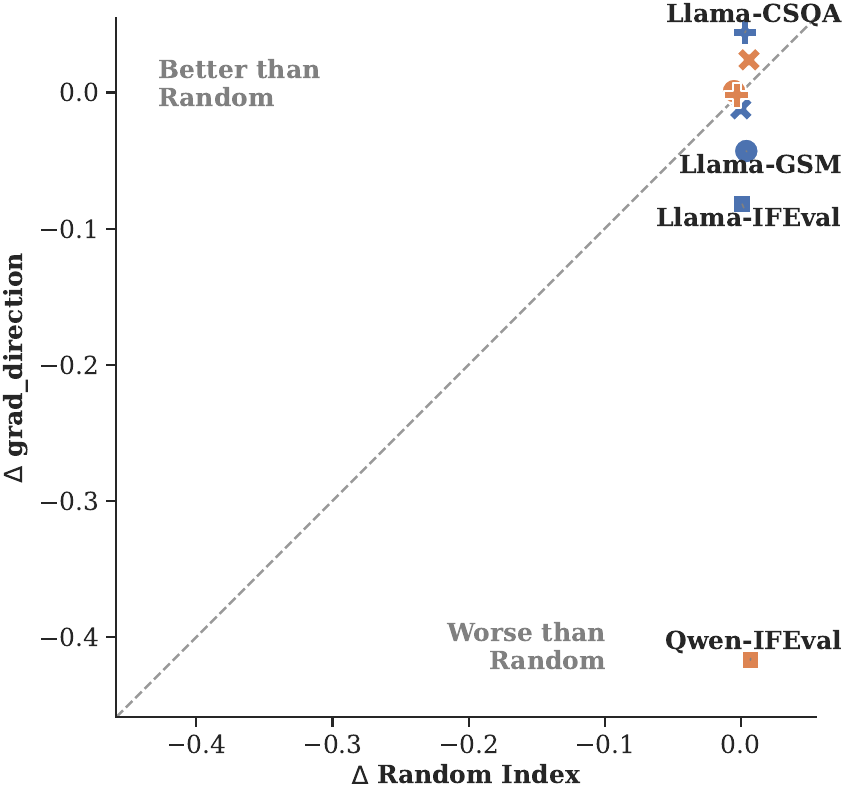}
    \caption{\textbf{Guided vs. random perturbations.} Each point is a (model, task) pair under the default setting. 
    The x-axis is the improvement of \texttt{random\_index} over the baseline, and the y-axis is the improvement of \texttt{grad\_direction}. 
    Points above the diagonal indicate genuine signal beyond random perturbation; the extreme failure on Qwen-\textsc{IFEval} illustrates the alignment tax of gradient-based editing.}
    \label{fig:guided_vs_random}
    \vspace{-2.0ex}
\end{figure}

\paragraph{Random controls separate signal from perturbation (RQ2).}
We treat \texttt{random\_index} as a required diagnostic baseline and summarize this comparison in Figure~\ref{fig:guided_vs_random}. Each point corresponds to a (model, task) pair under the default setting, with the x-axis showing the improvement of \texttt{random\_index} over the unedited adapter and the y-axis showing the improvement of \texttt{grad\_direction} (both measured as $\Delta$ over baseline). Points above the diagonal indicate that gradient-guided editing extracts a non-trivial sensitivity signal beyond generic perturbation, whereas points below the diagonal suggest that perturbation alone is more beneficial than the gradient-based ranking.

Figure~\ref{fig:guided_vs_random} shows that the benefit of gradient guidance is highly task-dependent. In an aligned setting such as Llama CSQA, \texttt{grad\_direction} lies above the diagonal and substantially outperforms \texttt{random\_index}, supporting the existence of a meaningful sensitivity signal beyond random spectral perturbation. In contrast, Qwen IFEval is a cautionary counterexample: \texttt{random\_index} is competitive and even best (0.597 vs.\ 0.590 baseline), while \texttt{grad\_direction} collapses far below the diagonal. This indicates that for strictly constrained instruction-following, a large fraction of the apparent improvement can come from structured regularization effects of perturbation rather than accurate importance ranking---and that gradient-guided amplification can be harmful when the calibration objective conflicts with formatting- and constraint-sensitive evaluation. Therefore, random controls are essential for distinguishing genuine signal extraction from perturbation-driven gains.

\subsection{The ``Alignment Tax'' of Gradient Editing}
\begin{figure}[!htp]
    \centering
    \includegraphics[width=0.86\linewidth]{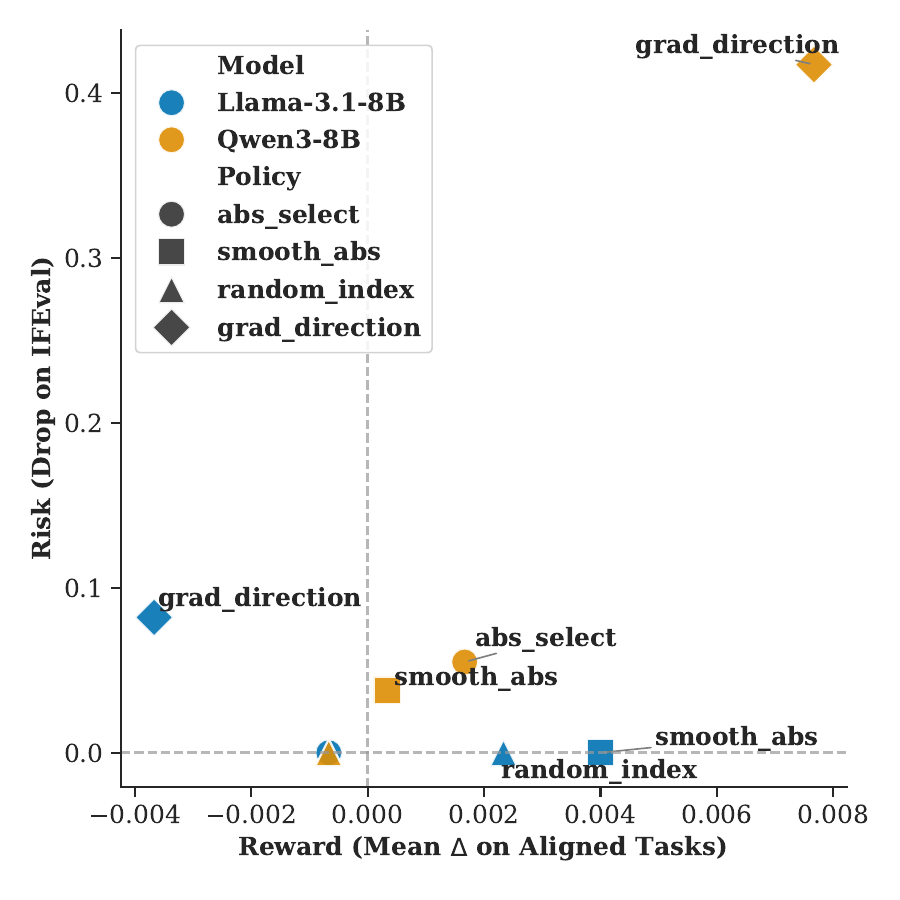}
    \caption{\textbf{Safety trade-off of editing policies under the default setting}. \textbf{Reward} is the mean improvement over aligned tasks (\textsc{GSM8K}, \textsc{HumanEval}, \textsc{CSQA}); 
    \textbf{Risk} is the performance drop on the constraint-sensitive benchmark \textsc{IFEval} (clipped at 0 if a policy does not decrease \textsc{IFEval}). 
    Gradient-based editing reaches the high-reward regime but can incur large risk, especially on \textsc{IFEval}.}
    \label{fig:risk_reward}
    \vspace{-2.0ex}
\end{figure}
Figure~\ref{fig:risk_reward} summarizes the safety--performance trade-off of spectral editing under the default setting by mapping each policy to its average reward on aligned tasks versus its risk on \textsc{IFEval}.
The plot highlights a clear alignment tax: policies that strongly exploit calibration gradients can yield high reward when objectives align, but they also expand the failure surface on strict instruction-following constraints.

\paragraph{Key observation.}
\texttt{grad\_direction} is the only policy that enters the high-reward regime, but it also incurs the largest risk.
On \textbf{Qwen3-8B}, \texttt{grad\_direction} achieves the highest average gain on aligned tasks yet suffers a catastrophic \textsc{IFEval} drop (risk $\approx$ 0.42), placing it in the extreme high-risk corner of Figure~\ref{fig:risk_reward}.
On \textbf{Llama-3.1-8B}, \texttt{grad\_direction} exhibits a different failure signature: despite a large improvement on CSQA, it becomes net-negative on the aligned-task average and still reduces \textsc{IFEval} (risk $\approx$ 0.08), indicating that gradient amplification can simultaneously hurt general aligned performance and constraint robustness.

\paragraph{Interpretation.}
This behavior is consistent with gradient signals prioritizing directions that reduce the calibration loss, which may trade off against strict formatting and instruction constraints.
In contrast, magnitude-based policies occupy the low-risk region in Figure~\ref{fig:risk_reward}: \texttt{smooth\_abs} provides modest positive reward with negligible-to-small \textsc{IFEval} degradation (risk $\le 0.04$ on Qwen and $\approx 0$ on Llama), making it a robust default.
\texttt{random\_index} stays essentially risk-free and can even improve \textsc{IFEval}, but its reward is small (and can be slightly negative), aligning with the view that some gains arise from perturbation-driven regularization rather than accurate importance ranking.

\subsection{Ablation Studies: Stability and Locality}

\paragraph{Energy constraints act as a stabilizer (RQ3).}
To isolate the role of energy preservation, we ablate \texttt{preserve\_energy} for \texttt{grad\_direction}. Table~\ref{tab:energy_ablation} shows that removing the constraint can further worsen the already unstable behavior on IFEval, supporting the view that $L_1$ energy preservation functions as a practical safety valve that limits over-amplification.

\begin{table}[h]
\centering
\small
\setlength{\tabcolsep}{3.4pt}
\renewcommand{\arraystretch}{1.05}
\caption{\textbf{Impact of Energy Preservation.} \texttt{grad\_direction} with and without energy constraints. The $L_1$ constraint mitigates extreme drift on strict instruction-following (IFEval), while leaving aligned reasoning gains largely intact.}
\label{tab:energy_ablation}
\begin{tabular}{@{}l l c c c@{}}
\toprule
& & \multicolumn{3}{c}{\textbf{\texttt{grad\_direction} Performance}} \\
\cmidrule(lr){3-5}
\textbf{Model} & \textbf{Task} & \textbf{Baseline} & \textbf{With L1 (Ours)} & \textbf{No Constraint} \\
\midrule
Llama-3.1 & CSQA   & 0.740 & 0.784 & \textbf{0.785} \\
Qwen3     & IFEval & \textbf{0.590} & 0.173 & \underline{0.017} \\
\bottomrule
\end{tabular}
\end{table}

\paragraph{Calibration efficiency and non-monotonicity (RQ3).}
We next sweep the calibration budget. Table~\ref{tab:calib_sweep} shows that increasing \texttt{calib\_samples} improves stability in the sense of reduced volatility, but it does not guarantee monotonic gains: for both a strongly aligned case (CSQA on Llama) and a misaligned constrained case (IFEval on Qwen), the best value can appear at smaller budgets. This supports using \texttt{calib=128} as a robust default that balances signal quality and compute.

\begin{table}[t]
\centering
\scriptsize
\setlength{\tabcolsep}{3.5pt}
\renewcommand{\arraystretch}{1.05}
\caption{\textbf{Effect of Calibration Size.} Performance consistency across varying \texttt{calib\_samples}. Larger budgets can stabilize outcomes but do not ensure monotonic improvements, motivating \texttt{calib=128} as a practical default.}
\label{tab:calib_sweep}
\resizebox{\columnwidth}{!}{%
\begin{tabular}{l c c c c c}
\toprule
\textbf{Task (Model)} & \textbf{Policy} & \textbf{32} & \textbf{64} & \textbf{128} & \textbf{256} \\
\midrule
\multirow{2}{*}{\shortstack[l]{CSQA\\(Llama)}} & \texttt{grad\_dir} & 0.785 & 0.782 & 0.784 & 0.783 \\
& \texttt{random} & 0.742 & 0.740 & 0.743 & 0.740 \\
\midrule
\multirow{2}{*}{\shortstack[l]{IFEval\\(Qwen)}} & \texttt{grad\_dir} & 0.179 & 0.181 & 0.173 & 0.169 \\
& \texttt{random} & 0.603 & 0.598 & 0.597 & 0.604 \\
\bottomrule
\end{tabular}%
}
\end{table}

\paragraph{Module sensitivity and locality (RQ4).}
Finally, we test whether restricting the edit locality matters. Table~\ref{tab:module_ablation} compares (i) editing attention input projections (Q/K/V), (ii) editing all modules, and (iii) restricting to residual-writing modules (our default). Editing all modules can unlock higher peaks (e.g., CSQA on Llama), but it also expands the misalignment surface and can severely harm constrained metrics (e.g., IFEval on Qwen). In contrast, residual-writing restriction typically offers a better safety--performance trade-off, improving or preserving reasoning-oriented metrics while limiting extreme failures.

\begin{table}[t]
\centering
\scriptsize
\setlength{\tabcolsep}{2.8pt}
\renewcommand{\arraystretch}{1.05}
\caption{\textbf{Module Selection Ablation.} \texttt{grad\_direction} under different edit localities (\texttt{calib=128}, \texttt{energy=L1}). ``Residual (Ours)'' denotes restricting edits to residual-writing projections (e.g., \texttt{o\_proj} and \texttt{down\_proj} in our main setup). We add ``MLP Internal'' (\texttt{up}, \texttt{gate}) for comparison. Editing MLP internals yields high peaks on code tasks but degrades arithmetic/instruction following.}
\label{tab:module_ablation}
\resizebox{\columnwidth}{!}{%
\begin{tabular}{l l c c c c}
\toprule
\textbf{Model} & \textbf{Task} & \textbf{Baseline} & \textbf{Attn Inputs} & \textbf{All Modules} & \textbf{\shortstack[c]{Residual\\(Ours)}} \\
\midrule
\multirow{2}{*}{Llama} & CSQA & 0.740 & 0.744 & \textbf{0.790} & 0.784 \\
& GSM8K & \textbf{0.657} & 0.637 & 0.538 & 0.614 \\
\midrule
\multirow{2}{*}{Qwen} & HumanEval & 0.488 & 0.512 & 0.524 & 0.512 \\
& IFEval & \textbf{0.590} & 0.499 & 0.071 & 0.173 \\
\bottomrule
\end{tabular}%
}
\end{table}

\section{Conclusion}

We reveal a subspace–spectrum dichotomy in trained LoRA: residual-writing modules often learn stable directions, while the spectrum can be inefficient or harmful. We propose Spectral Surgery, a training-free post-hoc method that preserves the learned subspace and reweights only singular values using lightweight calibration sensitivities under conservative constraints. Across two 8B backbones and four benchmarks, it yields task-dependent gains with only (O(r)) scalar edits, and random reweighting exposes spectral brittleness. Future work will improve objective-aligned sensitivity (especially for code) and extend spectrum-only refinement to decoding, safety, and multi-task settings.

\section{Impact Statement}

This paper presents work whose goal is to advance the field of Machine Learning, specifically in improving the parameter efficiency and interpretability of Large Language Models. Potential societal consequences include reducing the computational energy consumption required for model tuning (Green AI) and enhancing the transparency of model adaptation mechanisms. We do not foresee any specific negative ethical impacts that must be highlighted here.





\bibliography{example_paper}
\bibliographystyle{icml2026}

\newpage
\appendix
\onecolumn

\section{Implementation Details}\label{app:implementation}In this section, we provide the exact specifications for the LoRA fine-tuning process to ensure reproducibility. We focus on a standardized training recipe across tasks, modifying only task-specific hyperparameters (e.g., learning rate and batch size) where necessary.\subsection{LoRA Architecture and Training Infrastructure}We fine-tune LoRA adapters on a fixed set of Transformer projection modules: \texttt{{q, k, v, o, gate, up, down}\_proj}. The base model weights are kept frozen. Unless otherwise stated, we use the following LoRA configuration:\begin{itemize}\item \textbf{Rank ($r$):} 16\item \textbf{Alpha ($\alpha$):} 32\item \textbf{Dropout:} 0.05\item \textbf{Bias:} None\end{itemize}Training is performed using 8-way distributed data parallelism (DDP). We do not employ quantization during the training phase.\subsection{Dataset Processing and Formatting}We utilize the Hugging Face \texttt{datasets} library. To ensure consistent evaluation, we limit large training corpora (MetaMath, Magicoder) to 50k examples using a deterministic shuffle-and-select strategy (fixed seed).\paragraph{Formatting Logic.}\begin{itemize}\item \textbf{Math (MetaMathQA):} Constructed as strictly supervised pairs. We filter out malformed examples where query or response fields are missing.\item \textbf{Code (Magicoder):} We preserve the dataset's native conversation style by concatenating: $\texttt{instruction} + \texttt{"\textbackslash n"} + \texttt{response}$.\item \textbf{Instruction Following (Alpaca):} We employ a robust formatter to handle optional input fields. If an explicit \texttt{text} field is absent, we format the entry as:\begin{quote}\texttt{Instruction + "\textbackslash n\textbackslash nInput:\textbackslash n" + Input} (if Input exists)\end{quote}Empty instruction or output fields result in the example being dropped.\end{itemize}

\subsection{Training Hyperparameters}We utilize the AdamW optimizer with $\beta_1=0.9, \beta_2=0.95$, and no weight decay. All runs use a cosine learning rate schedule with a 10\% warmup period ($ratio=0.1$) and a minimum LR ratio of 0.01.\paragraph{Batch Size Configuration.}To maintain consistent training dynamics across different hardware setups, we define a target \textbf{Global Batch Size (GBS)}. The gradient accumulation steps ($N_{\text{accum}}$) are dynamically calculated based on the number of GPUs ($N_{\text{gpu}}=8$) and the per-device micro-batch size ($B_{\text{micro}}$):\begin{equation}N_{\text{accum}} = \left\lceil \frac{\text{GBS}}{N_{\text{gpu}} \times B_{\text{micro}}} \right\rceil.\end{equation}Table~\ref{tab:training_hyperparams} details the specific hyperparameters for each task family.
\begin{table}[ht]
\centering
\small
\caption{\textbf{Hyperparameter configurations.} "Global BS" denotes the effective batch size after gradient accumulation. We use a fixed epoch budget of 3 for all tasks.}
\renewcommand{\arraystretch}{1.2}
\begin{tabular}{l|ccc|ccc}
\toprule
\textbf{Task Family} & \textbf{Max Len} & \textbf{Epochs} & \textbf{Global BS} & \textbf{Peak LR} & \textbf{Min LR} & \textbf{Grad Clip} \\
\midrule
\textbf{Math} (MetaMath) & 1024 & 3 & 768 & $1{\times}10^{-4}$ & $1{\times}10^{-6}$ & 1.0 \\
\textbf{Commonsense} (CSQA) & 2048 & 3 & 256 & $2{\times}10^{-4}$ & $2{\times}10^{-6}$ & 1.0 \\
\textbf{Instruction} (Alpaca) & 2048 & 3 & 256 & $2{\times}10^{-4}$ & $2{\times}10^{-6}$ & 1.0 \\

\textbf{Code} (Magicoder) & 4096 & 3 & 192 & $2{\times}10^{-4}$ & $2{\times}10^{-6}$ & 1.0 \\
\bottomrule
\end{tabular}
\label{tab:training_hyperparams}
\end{table}
\subsection{Evaluation Specifics}
\label{sec:eval_setup}
While Section~\ref{sec:exp_setup} outlines the primary metrics, we provide additional configuration details here for precise reproduction:

\begin{itemize}
    \item \textbf{Mathematical Reasoning:} We employ a 5-shot prompt strategy with greedy decoding (\texttt{temperature=0}, \texttt{top\_p=1}). To ensure efficient processing, GPU memory utilization is set to 0.95.

    \item \textbf{Code Evaluation:} For HumanEval~\citep{chen2021evaluating}, we utilize a zero-shot setting to assess raw code generation capabilities. We enforce a sandboxed execution environment with \texttt{confirm\_unsafe\_code} enabled and strict timeouts to handle potentially unsafe outputs or infinite loops. GPU memory utilization is limited to 0.90.

    \item \textbf{Instruction Following (Alpaca):} We evaluate without few-shot examples, allowing for a maximum generation length of 2048 tokens. Decoding follows a greedy strategy (\texttt{temperature=0}, \texttt{top\_p=1}) with GPU memory utilization at 0.95.

    \item \textbf{CommonsenseQA:} The model is prompted with the question and options (A--E) in a zero-shot configuration. We evaluate the likelihood of the single-token option labels, with GPU memory utilization set to 0.85.
\end{itemize}

\section{Complete Subspace-Alignment Heatmaps for All Target Modules}
\subsection{Full-Module Alignment Heatmap Wall}
\label{app:heatmap_wall}

We provide a complete heatmap wall covering all seven LoRA target modules
(\texttt{q/k/v/o/gate/up/down\_proj}) for two base models and four task families.
For each (model, task, module), we visualize two alignment diagnostics:
(i) principal-direction similarity ($|u_1^\top u_1|$) and
(ii) top-$m$ output-subspace overlap ($\mathrm{Align}_U$, Eq.~\ref{eq:align_metric}).
All heatmaps are computed from trained rank-$r{=}16$ adapters with $m{=}4$ (Top-4 subspace).

\clearpage
\clearpage
\begin{figure*}[p]
\centering
\small
\setlength{\tabcolsep}{2pt}
\renewcommand{\arraystretch}{1.0}
\begin{tabular}{lcccc}
\toprule
\textbf{Module} &
\textbf{Math} & \textbf{Code} & \textbf{Instruction} & \textbf{Commonsense} \\
\midrule
\texttt{q\_proj} &
\includegraphics[width=0.195\textwidth]{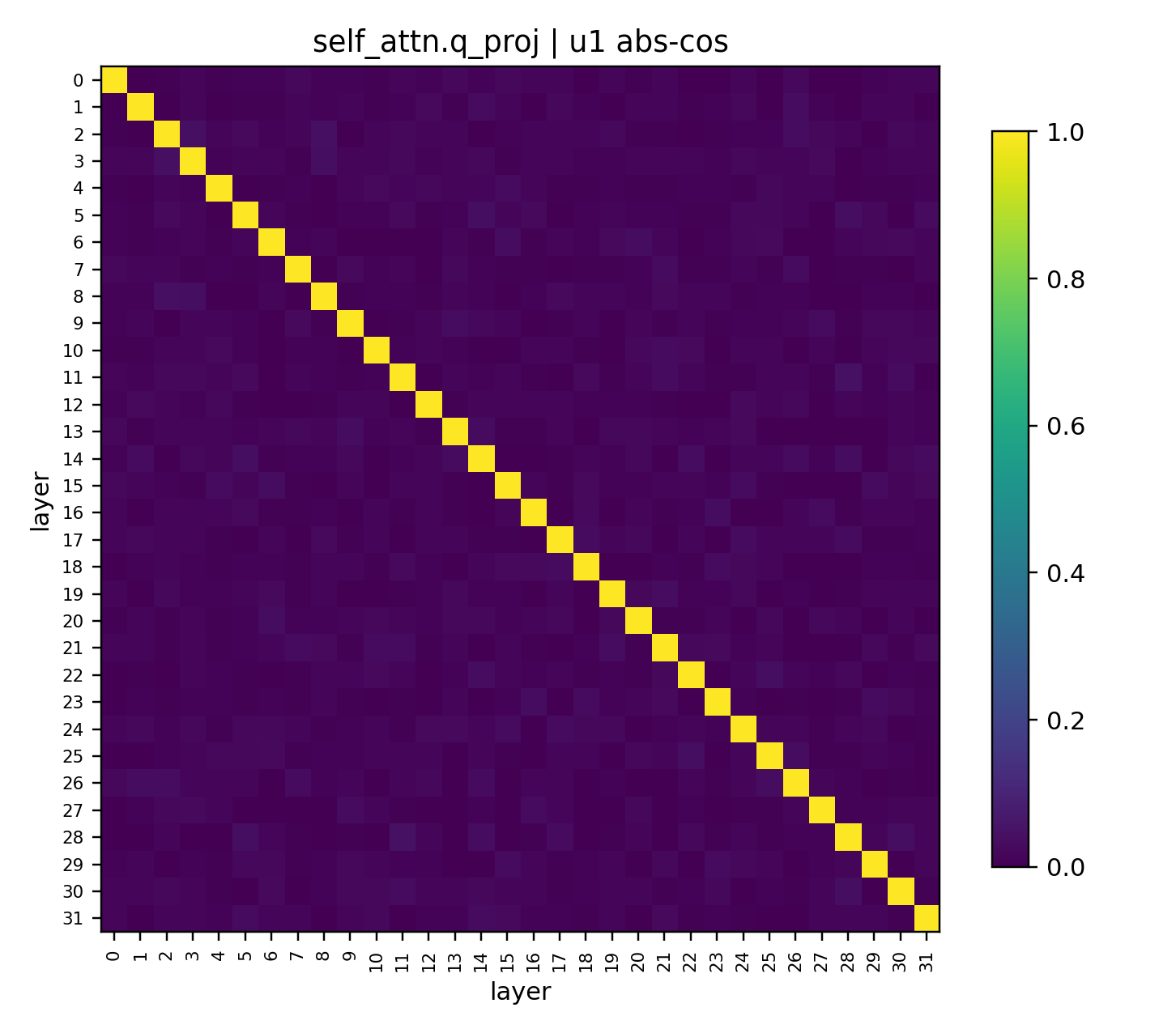} &
\includegraphics[width=0.195\textwidth]{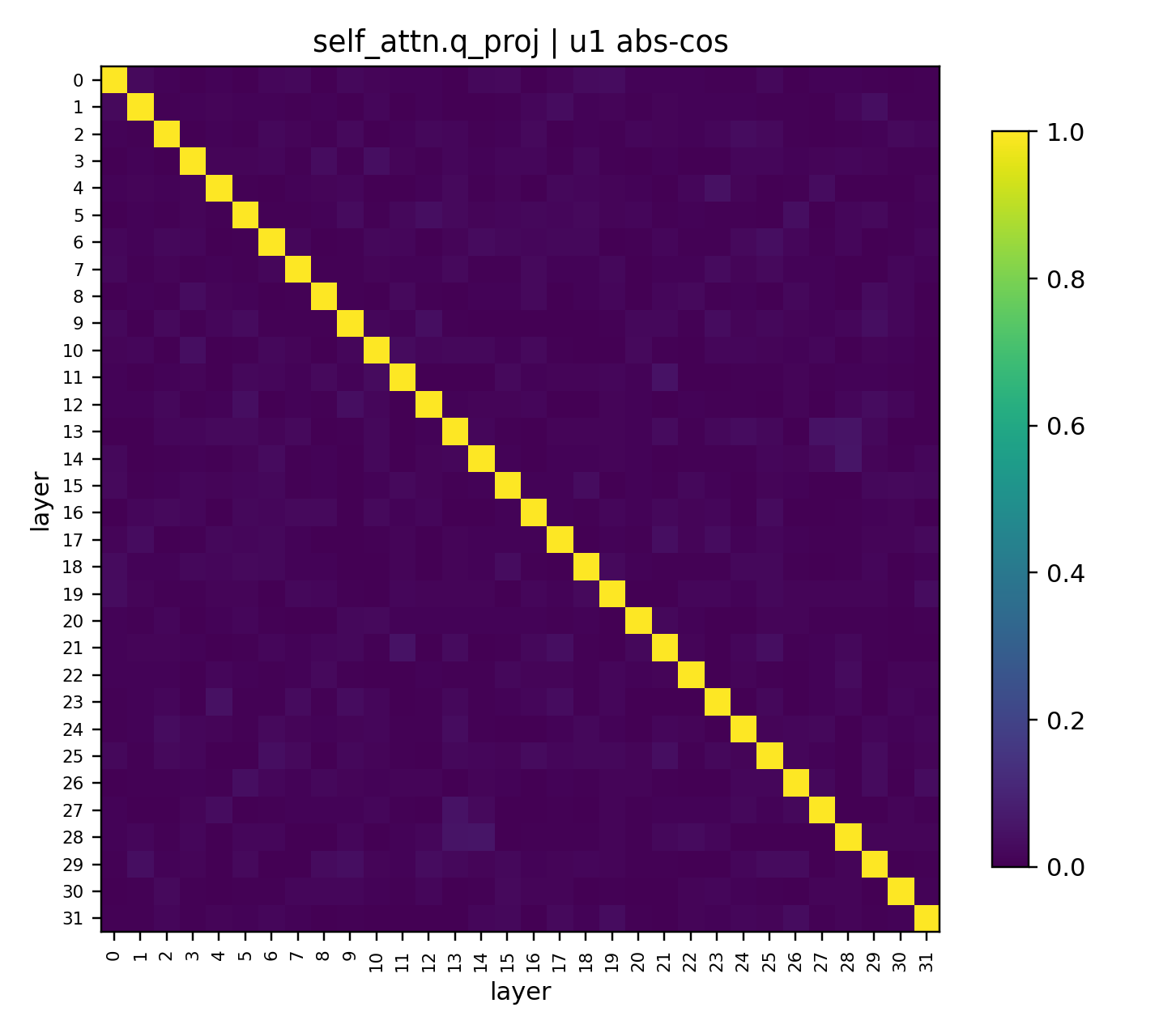} &
\includegraphics[width=0.195\textwidth]{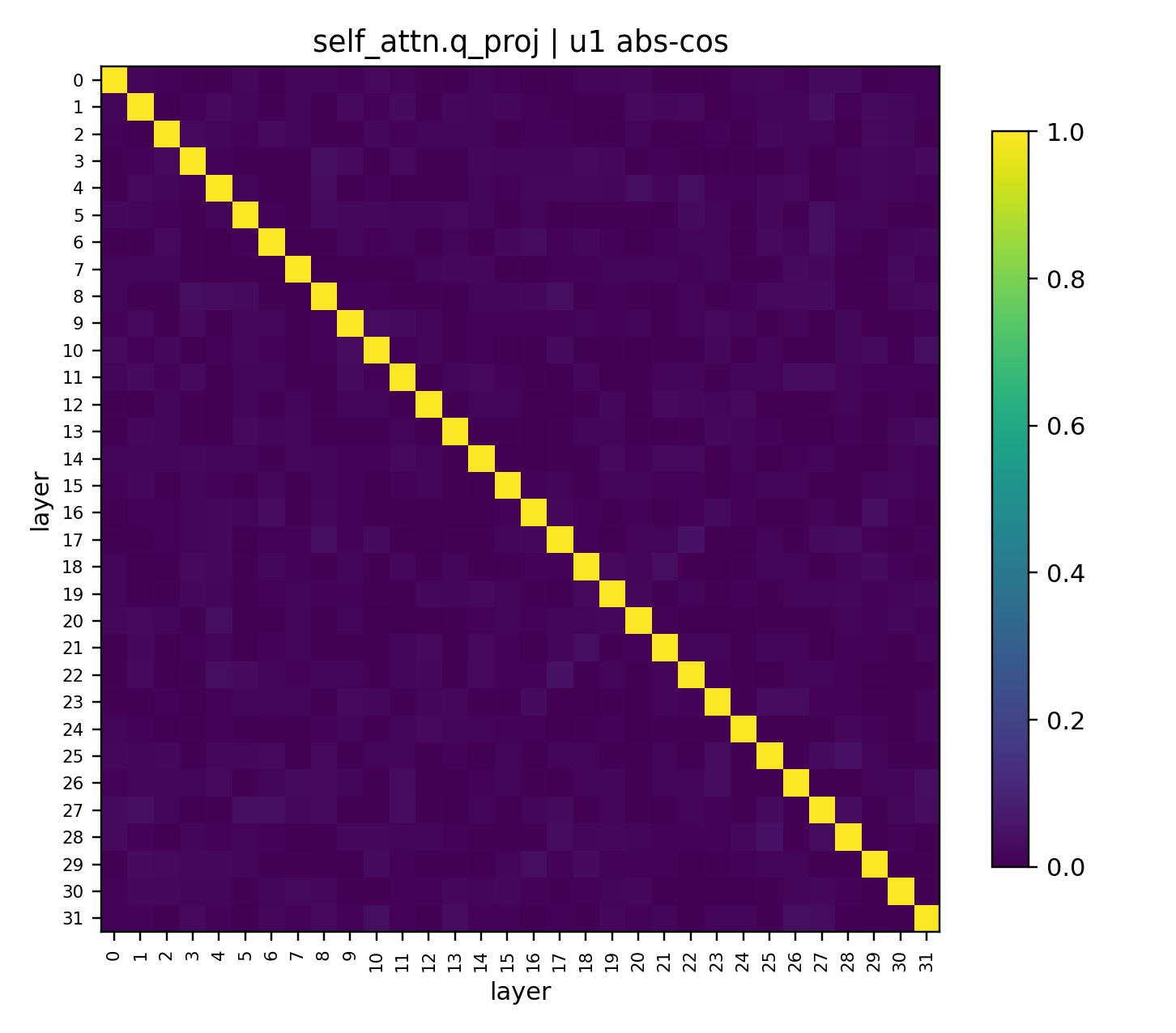} &
\includegraphics[width=0.195\textwidth]{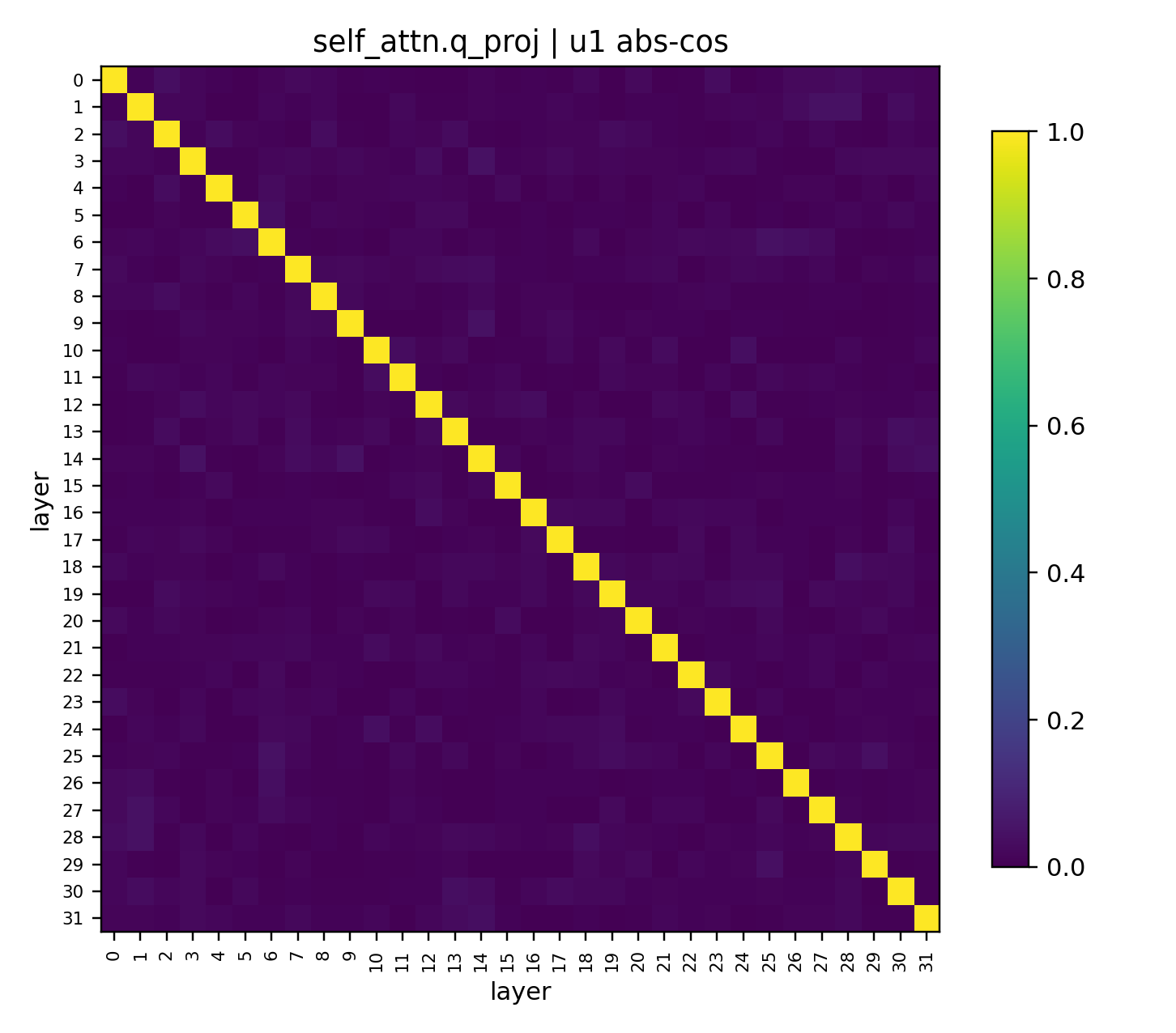} \\
\texttt{k\_proj} &
\includegraphics[width=0.195\textwidth]{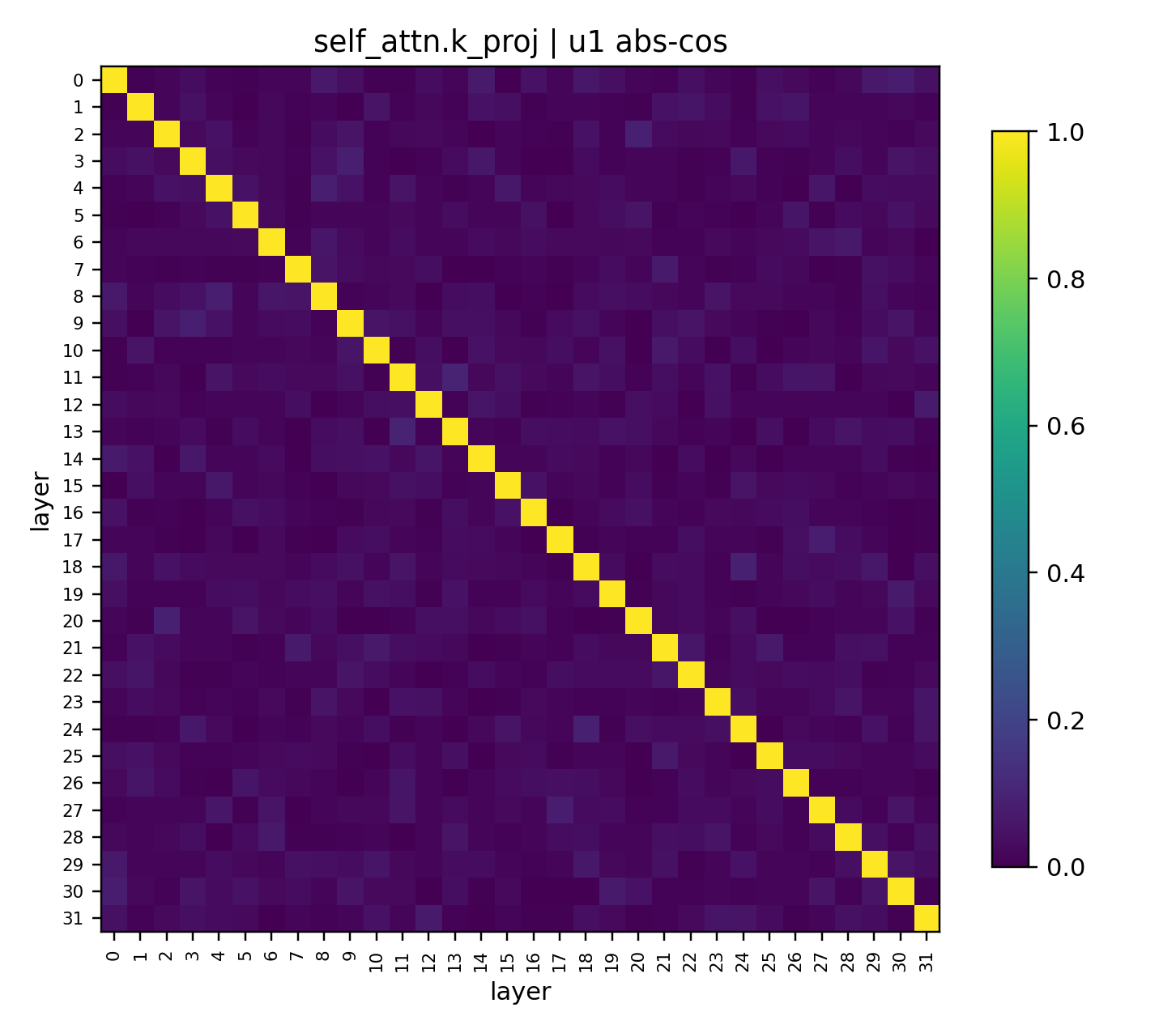} &
\includegraphics[width=0.195\textwidth]{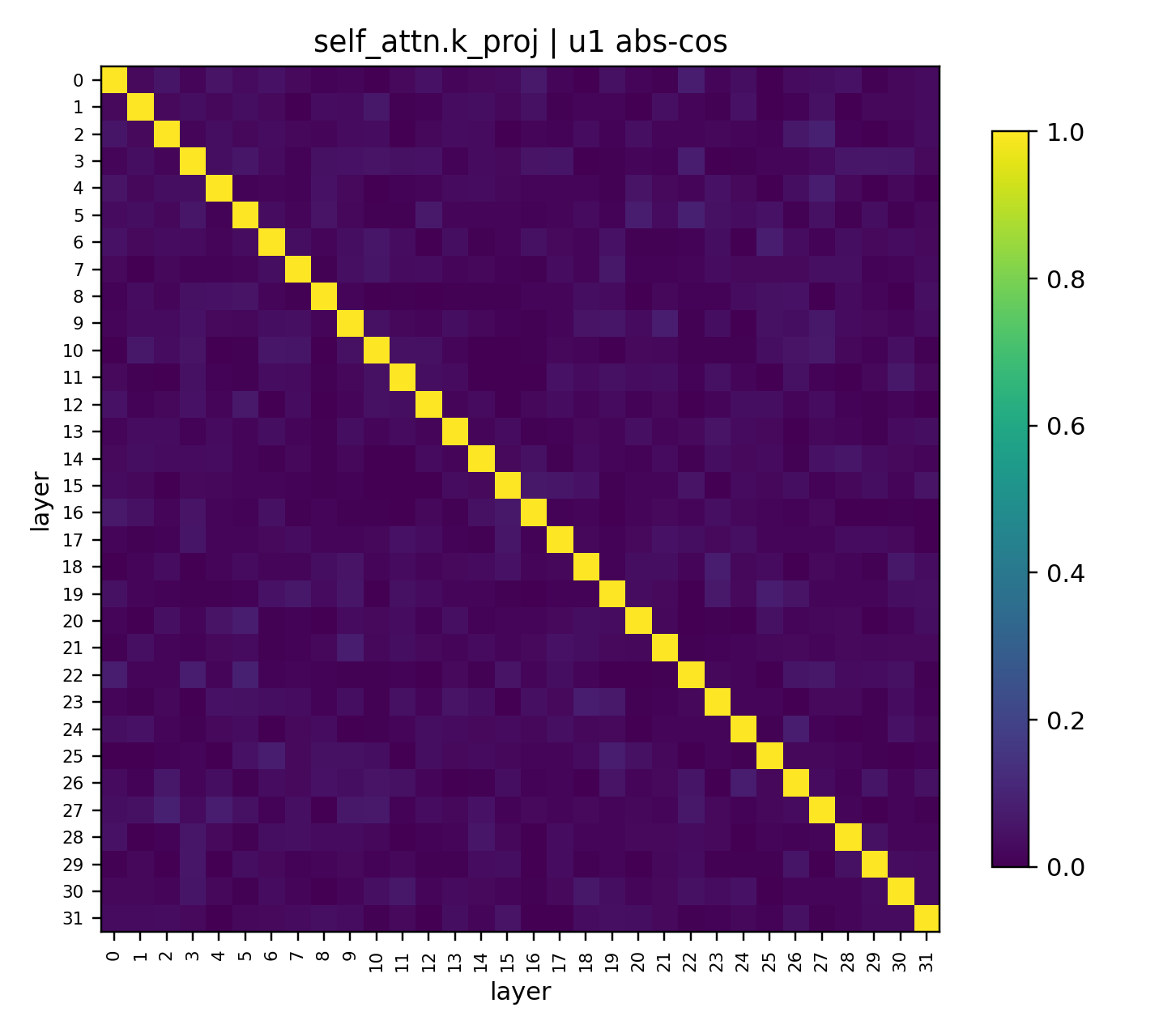} &
\includegraphics[width=0.195\textwidth]{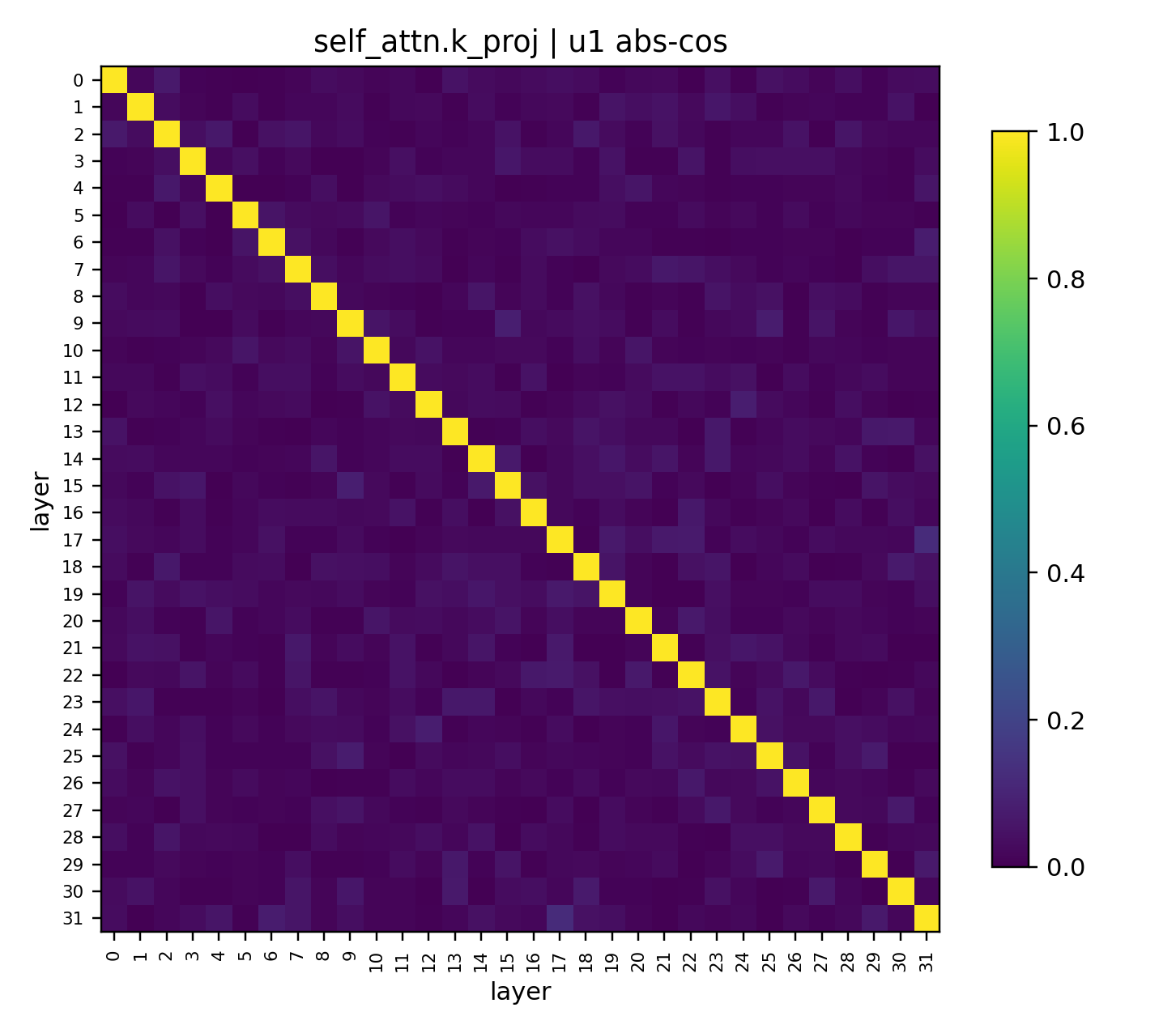} &
\includegraphics[width=0.195\textwidth]{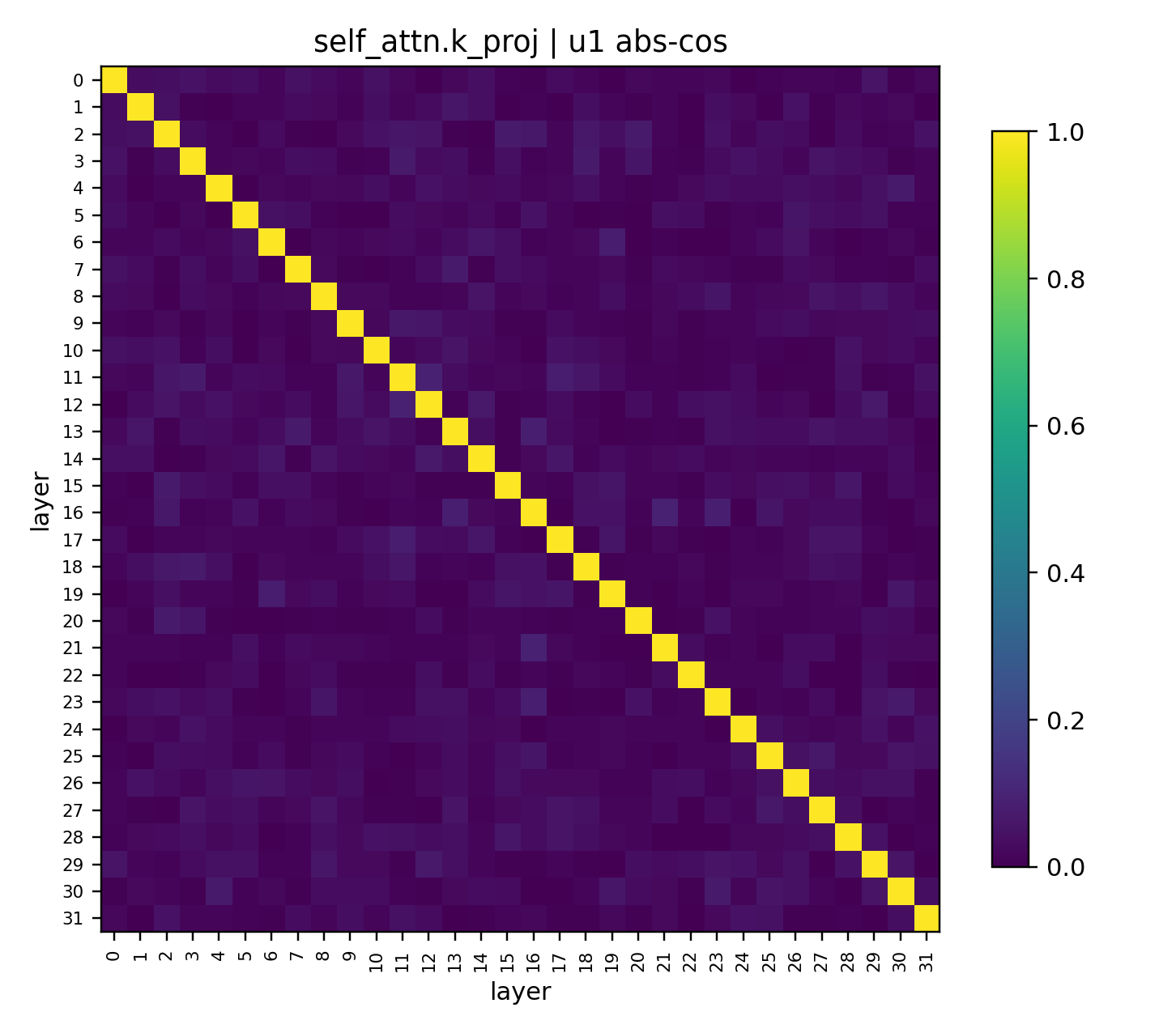} \\
\texttt{v\_proj} &
\includegraphics[width=0.195\textwidth]{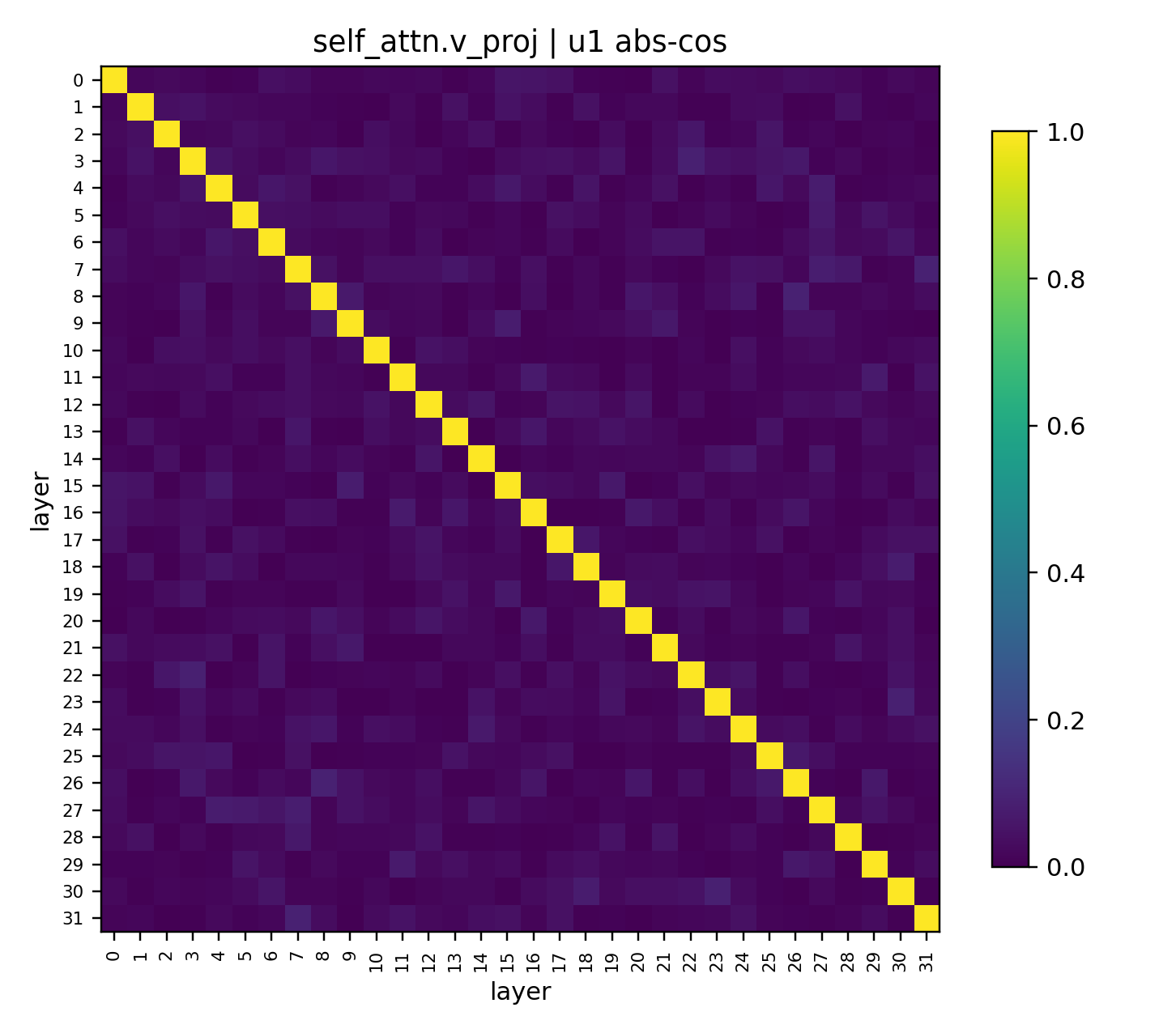} &
\includegraphics[width=0.195\textwidth]{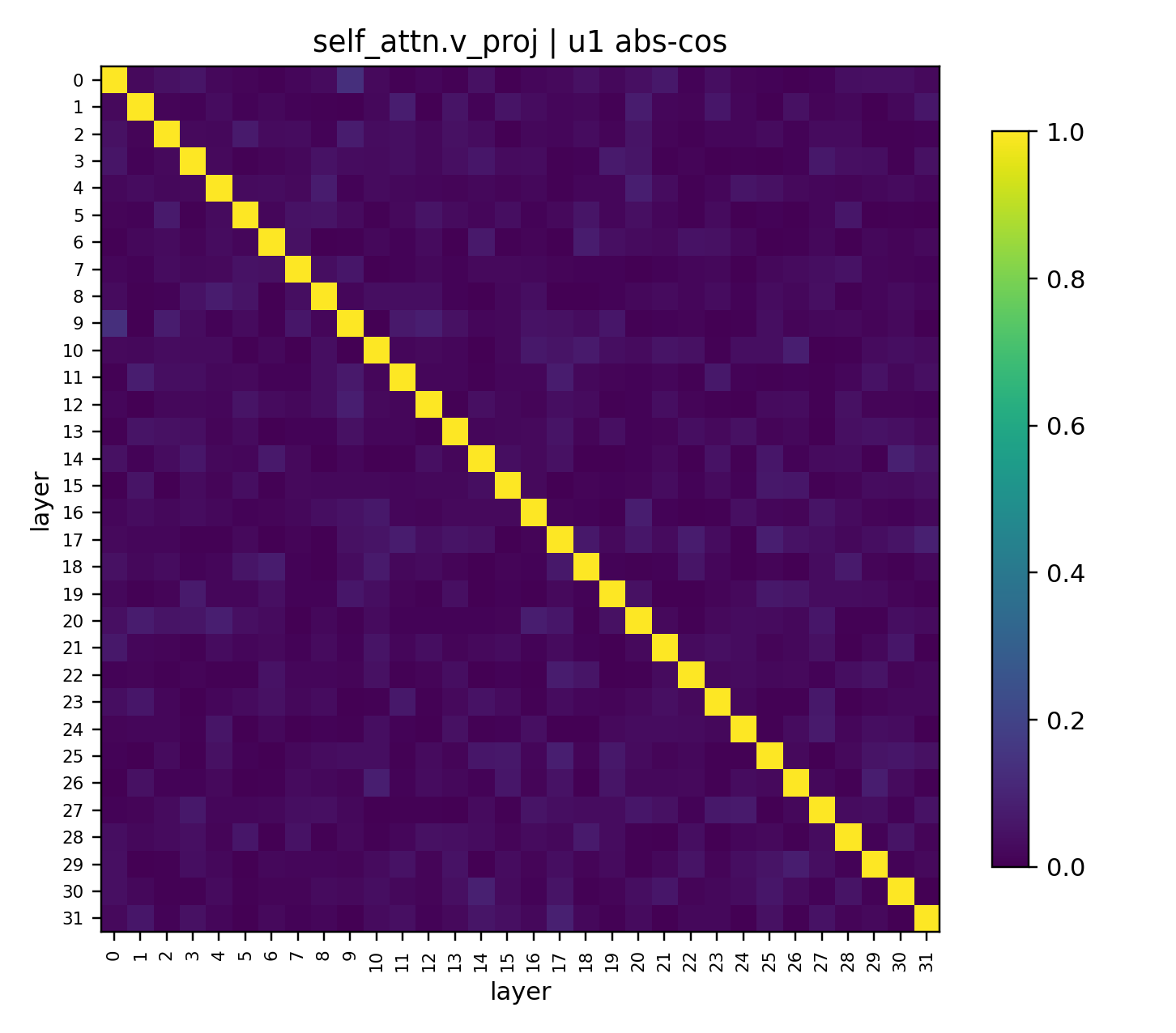} &
\includegraphics[width=0.195\textwidth]{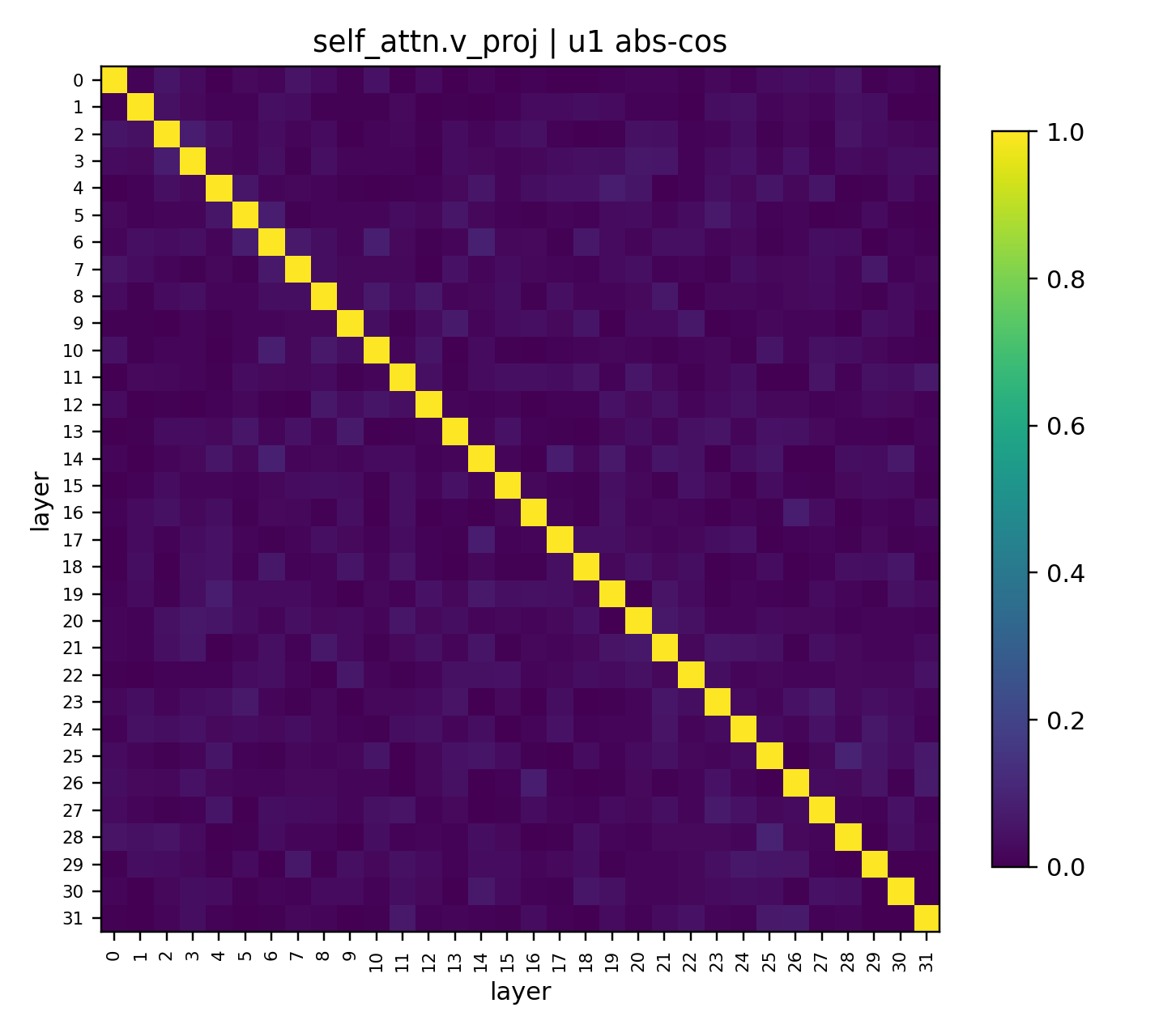} &
\includegraphics[width=0.195\textwidth]{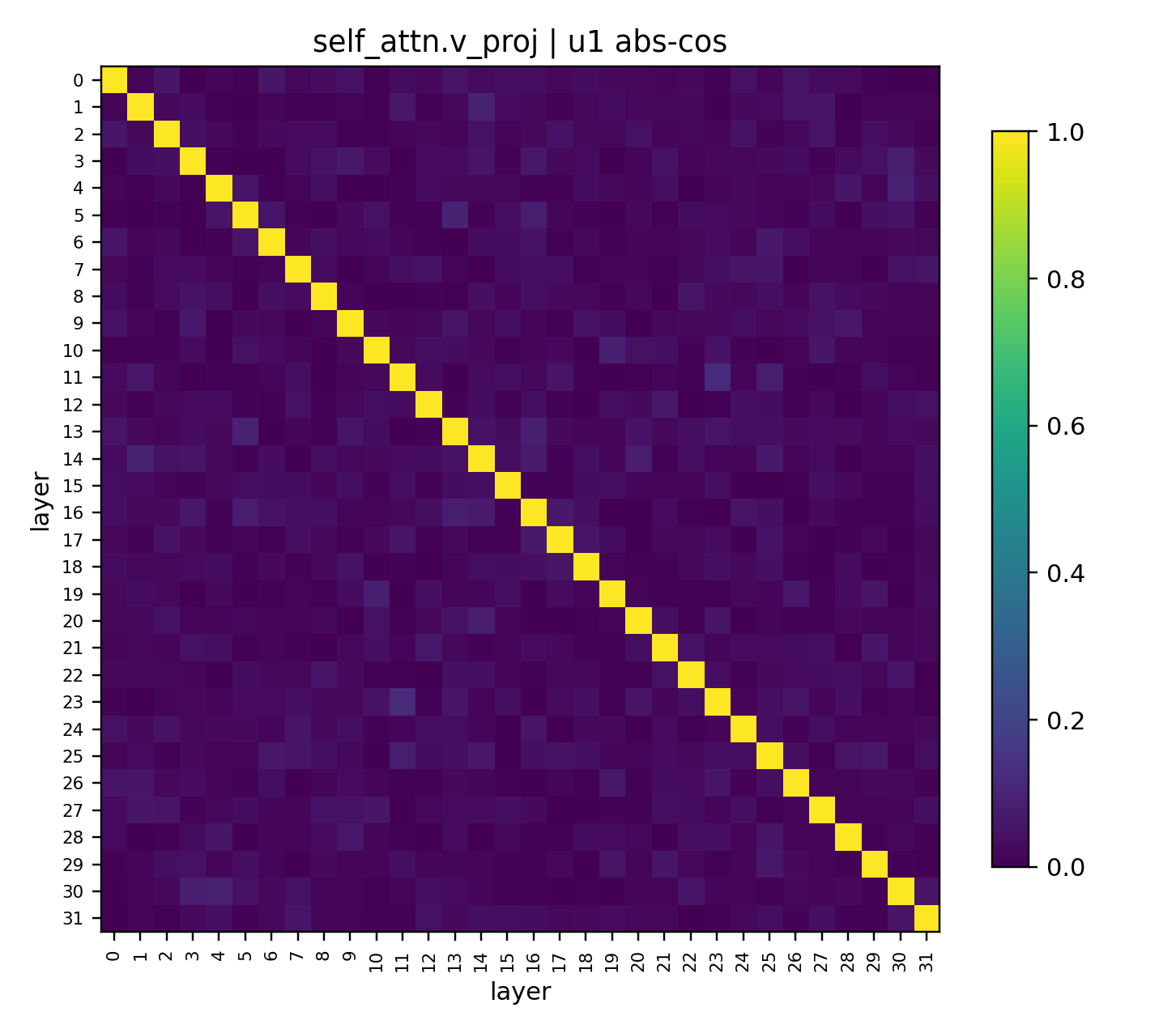} \\
\texttt{o\_proj} &
\includegraphics[width=0.195\textwidth]{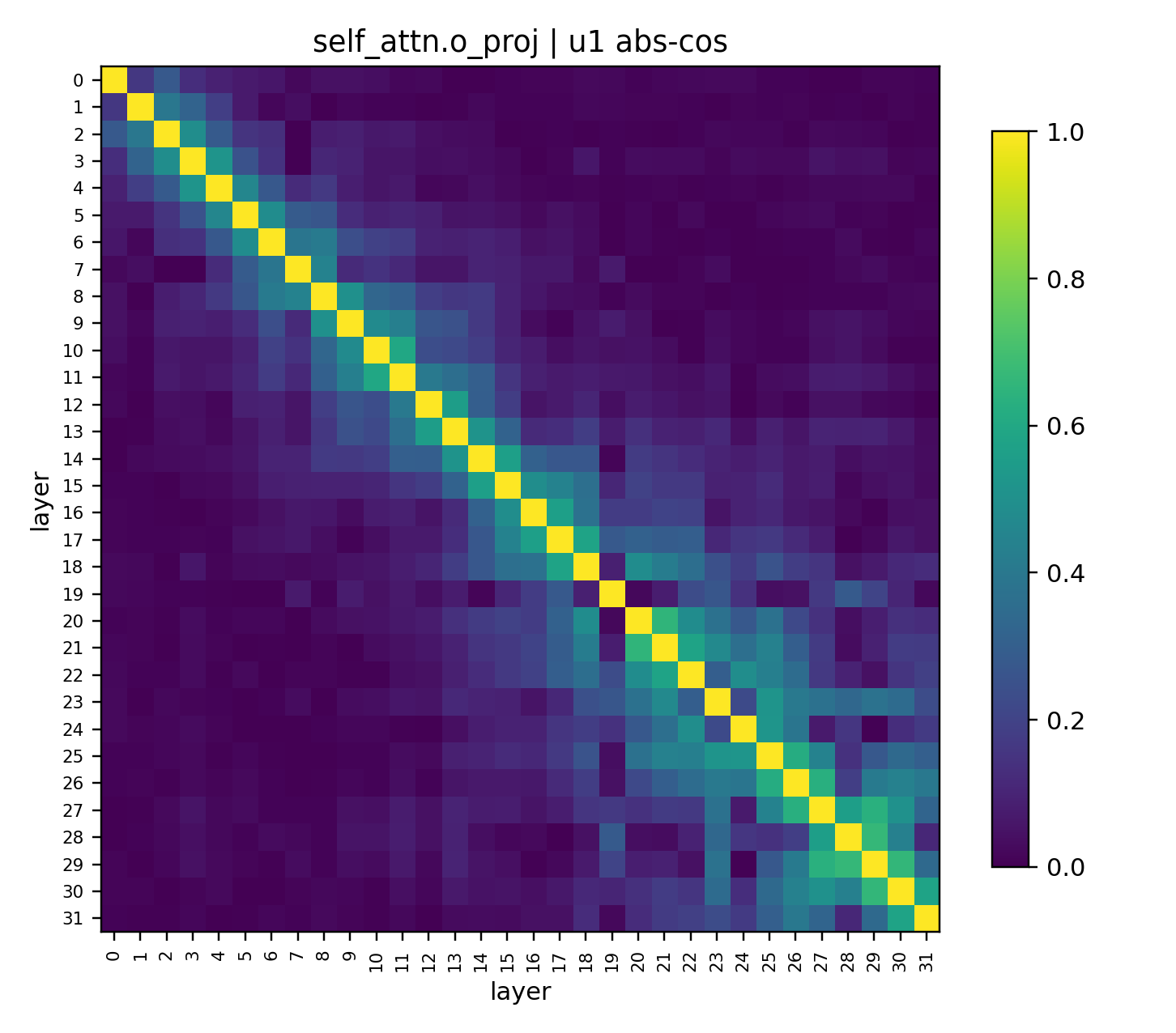} &
\includegraphics[width=0.195\textwidth]{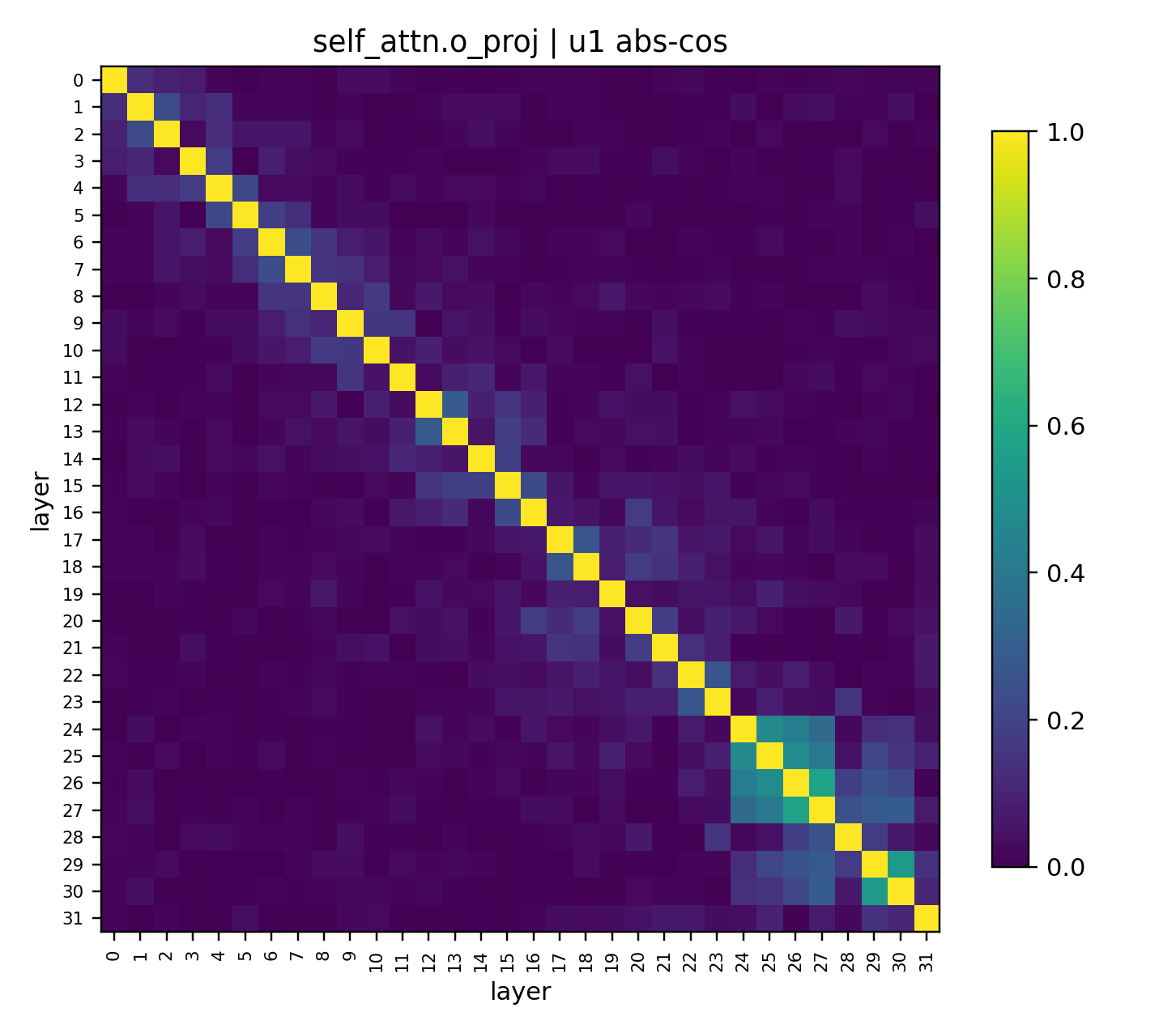} &
\includegraphics[width=0.195\textwidth]{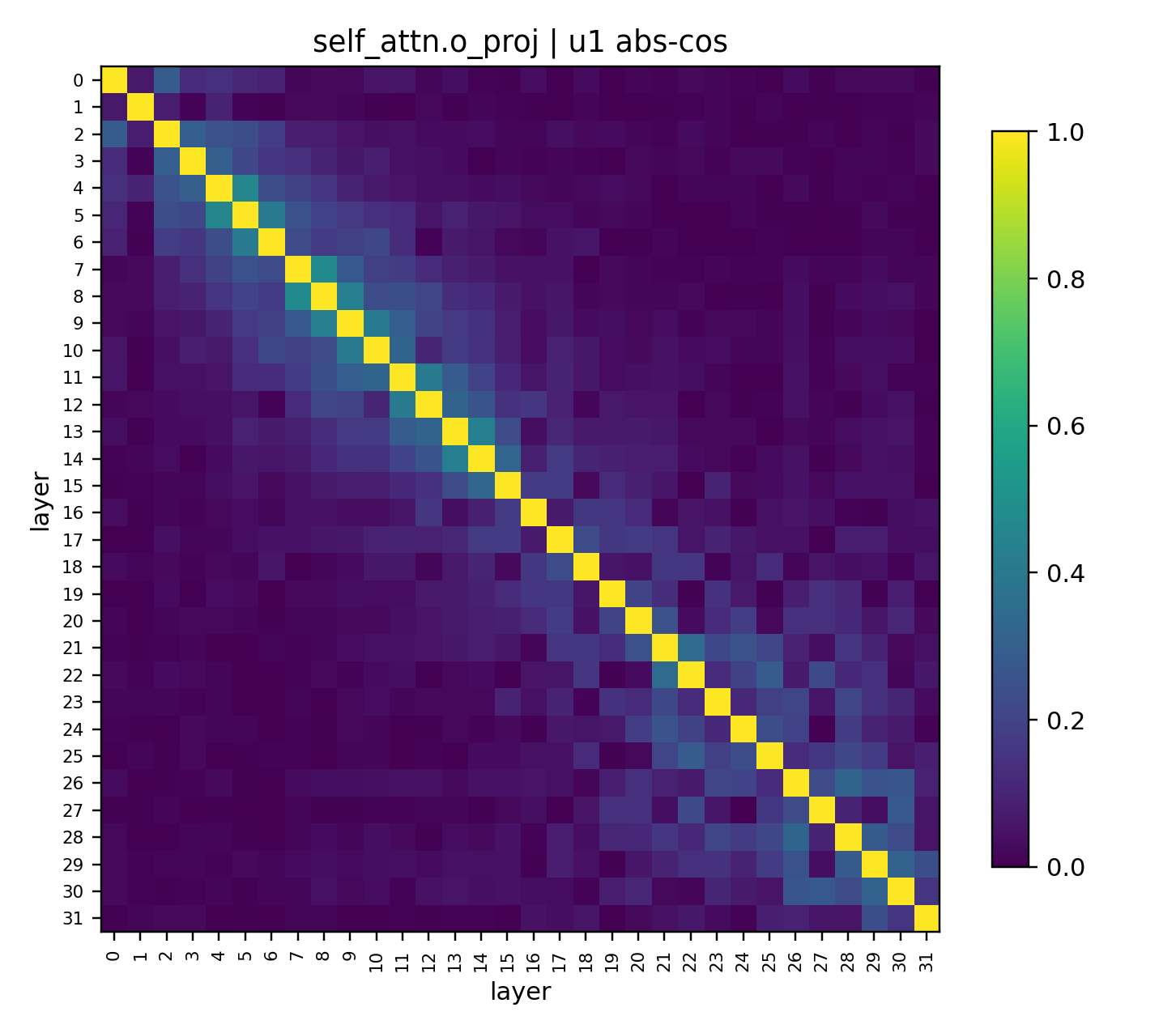} &
\includegraphics[width=0.195\textwidth]{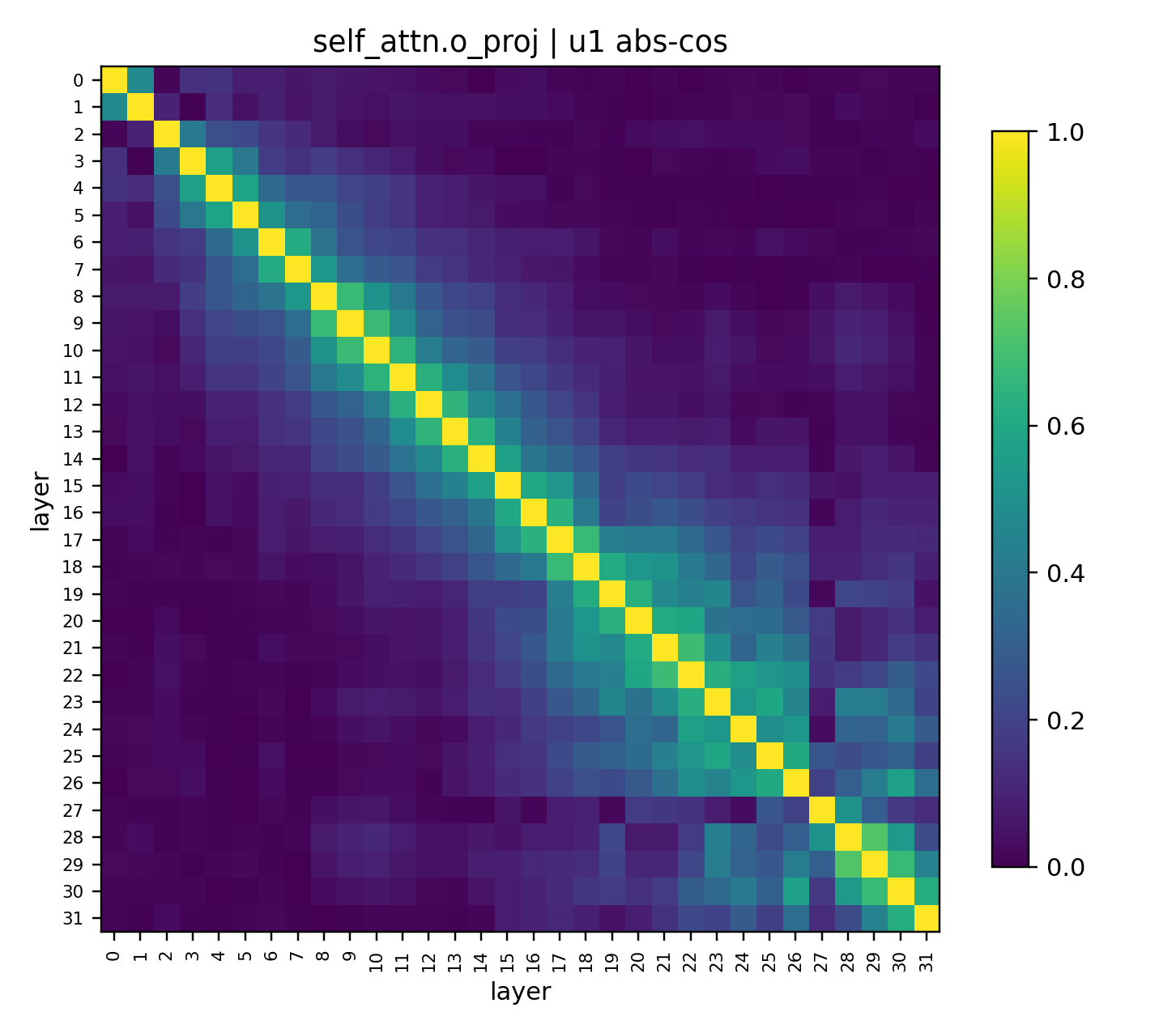} \\
\texttt{gate\_proj} &
\includegraphics[width=0.195\textwidth]{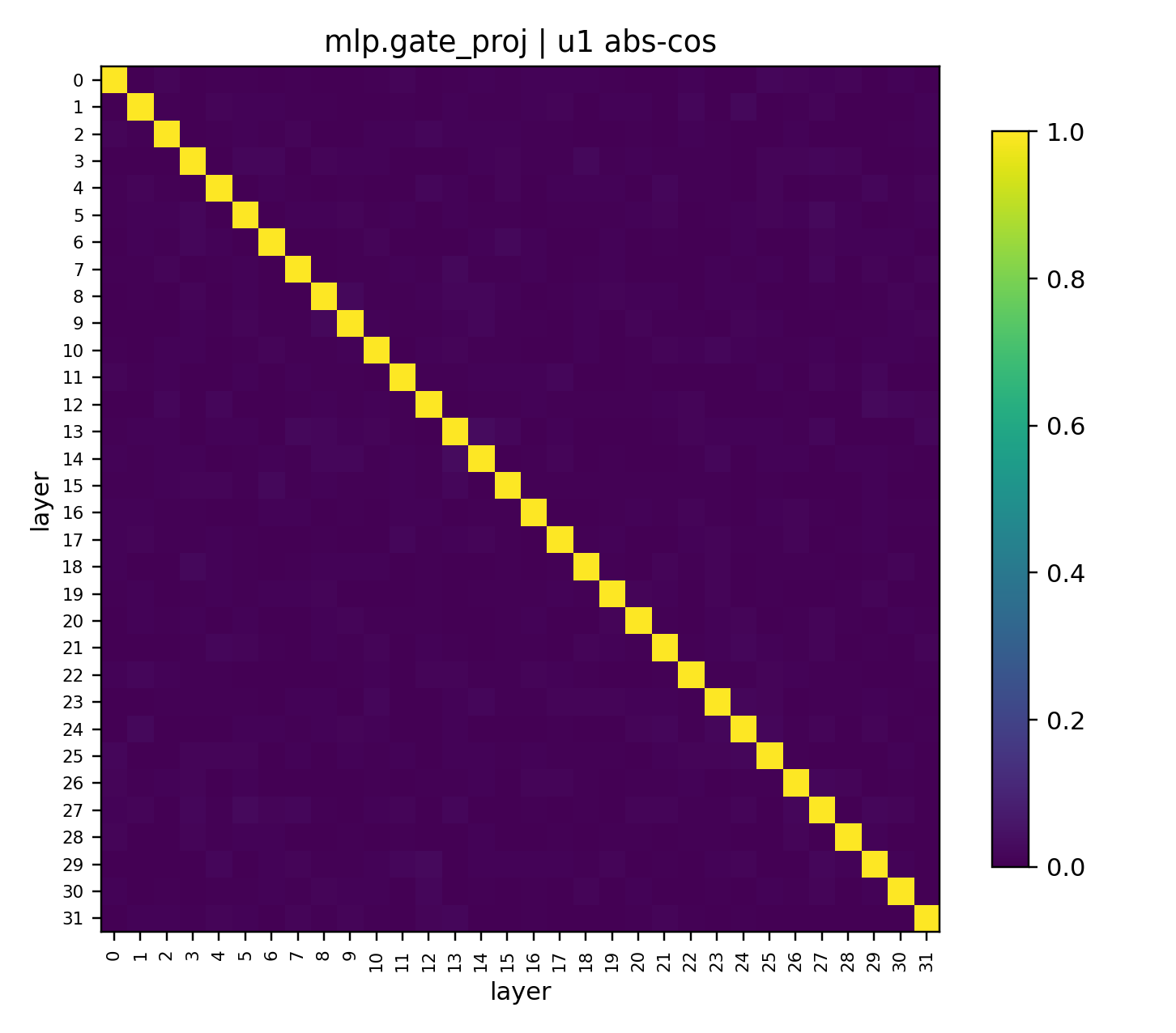} &
\includegraphics[width=0.195\textwidth]{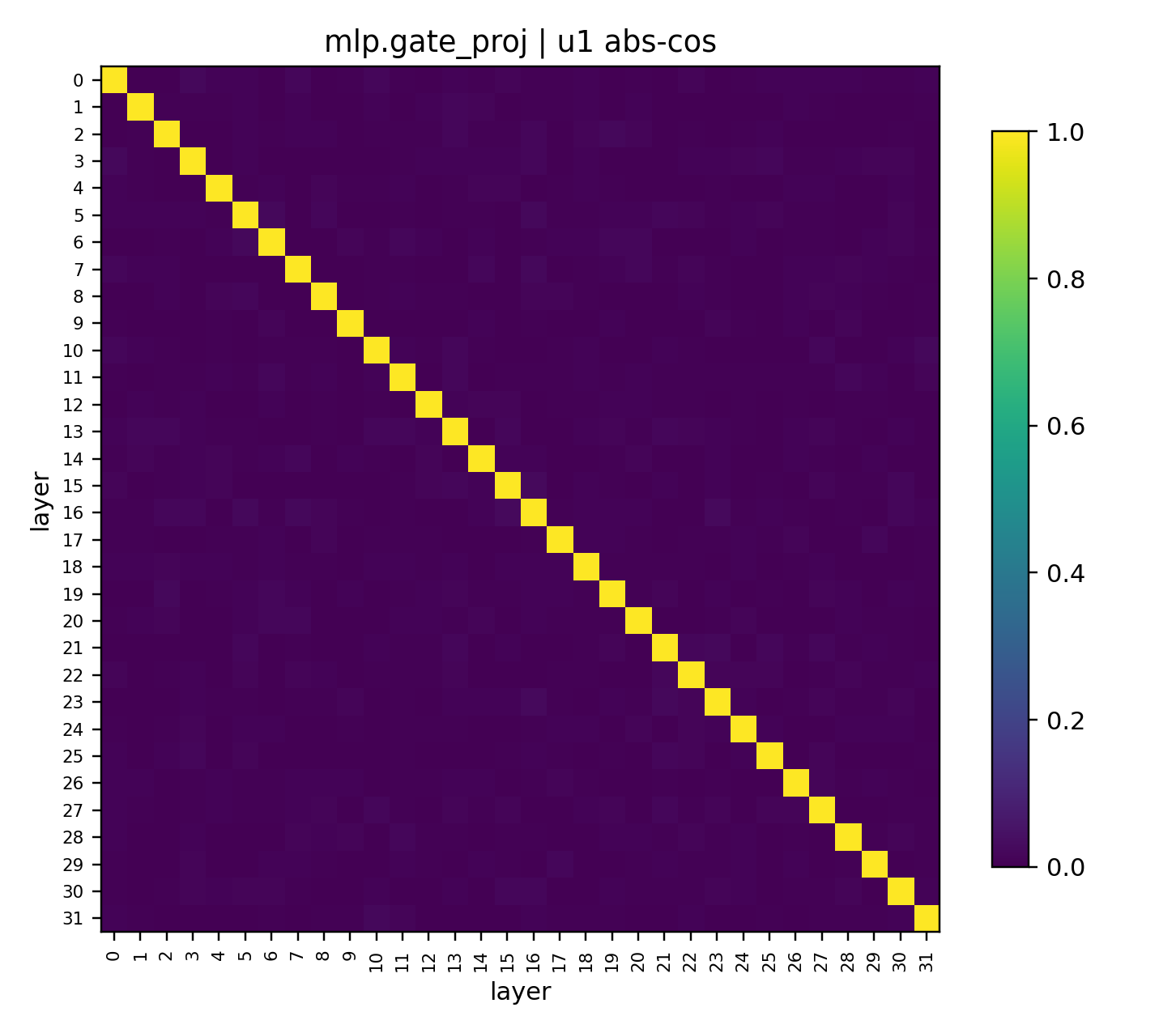} &
\includegraphics[width=0.195\textwidth]{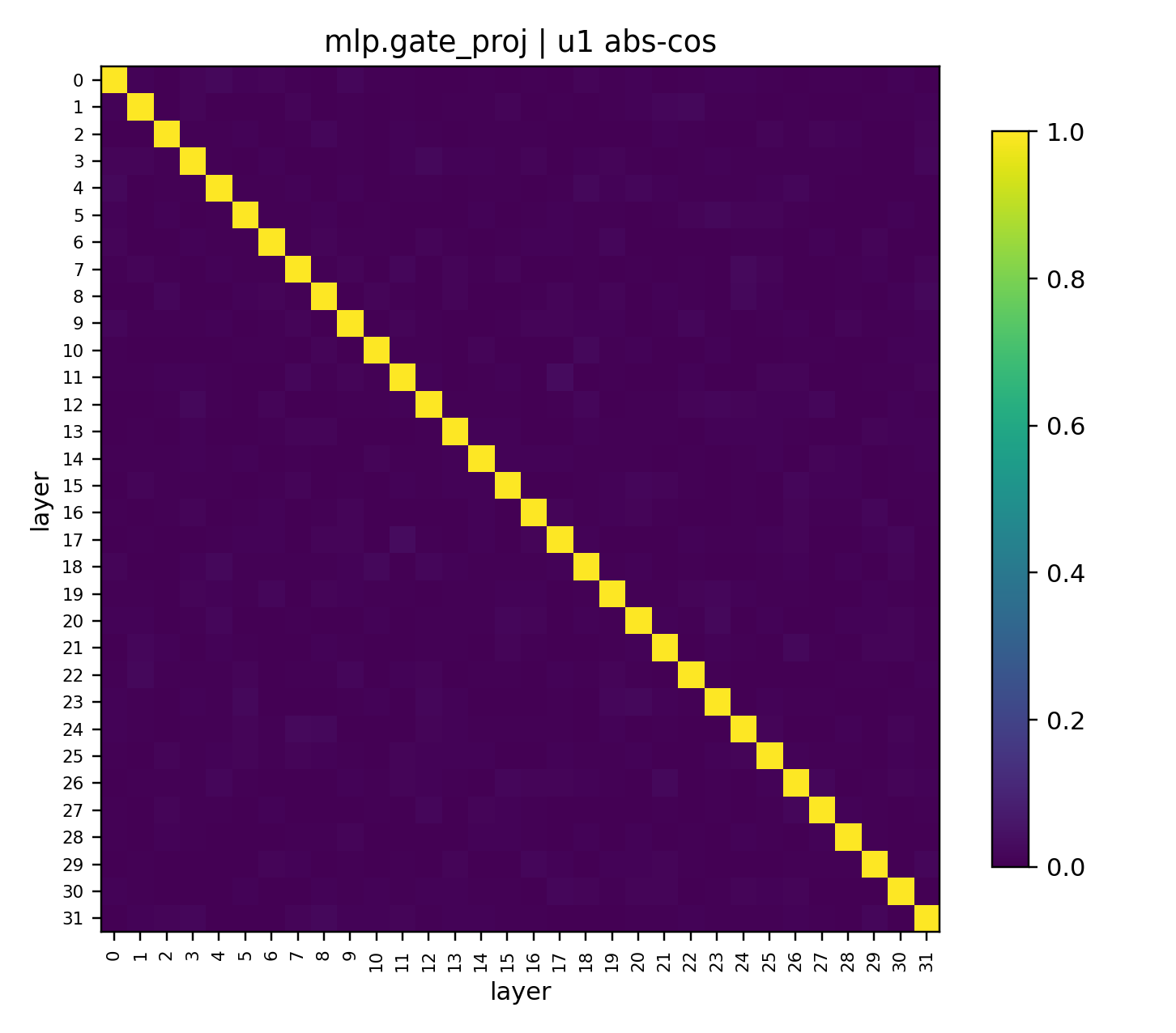} &
\includegraphics[width=0.195\textwidth]{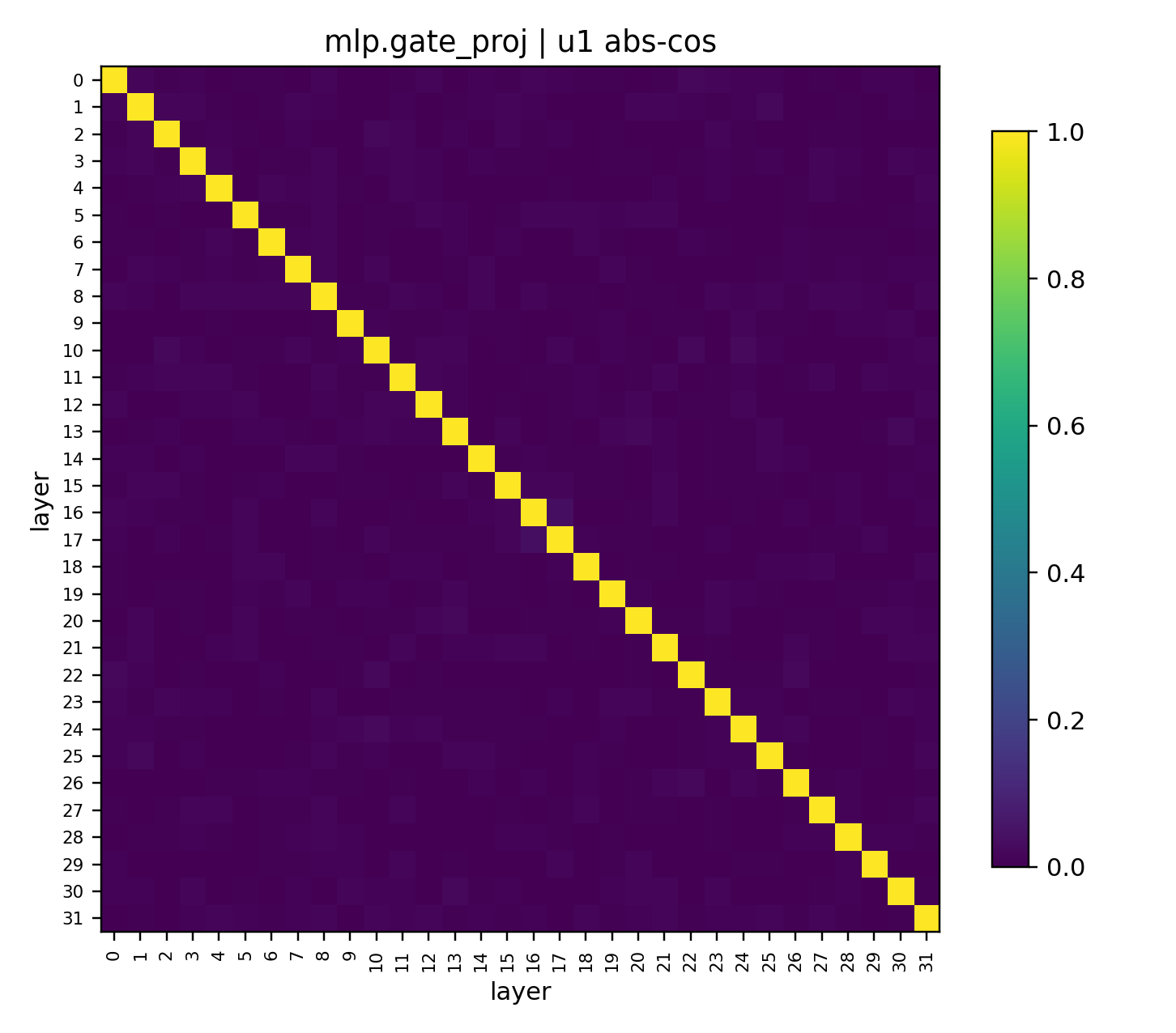} \\
\texttt{up\_proj} &
\includegraphics[width=0.195\textwidth]{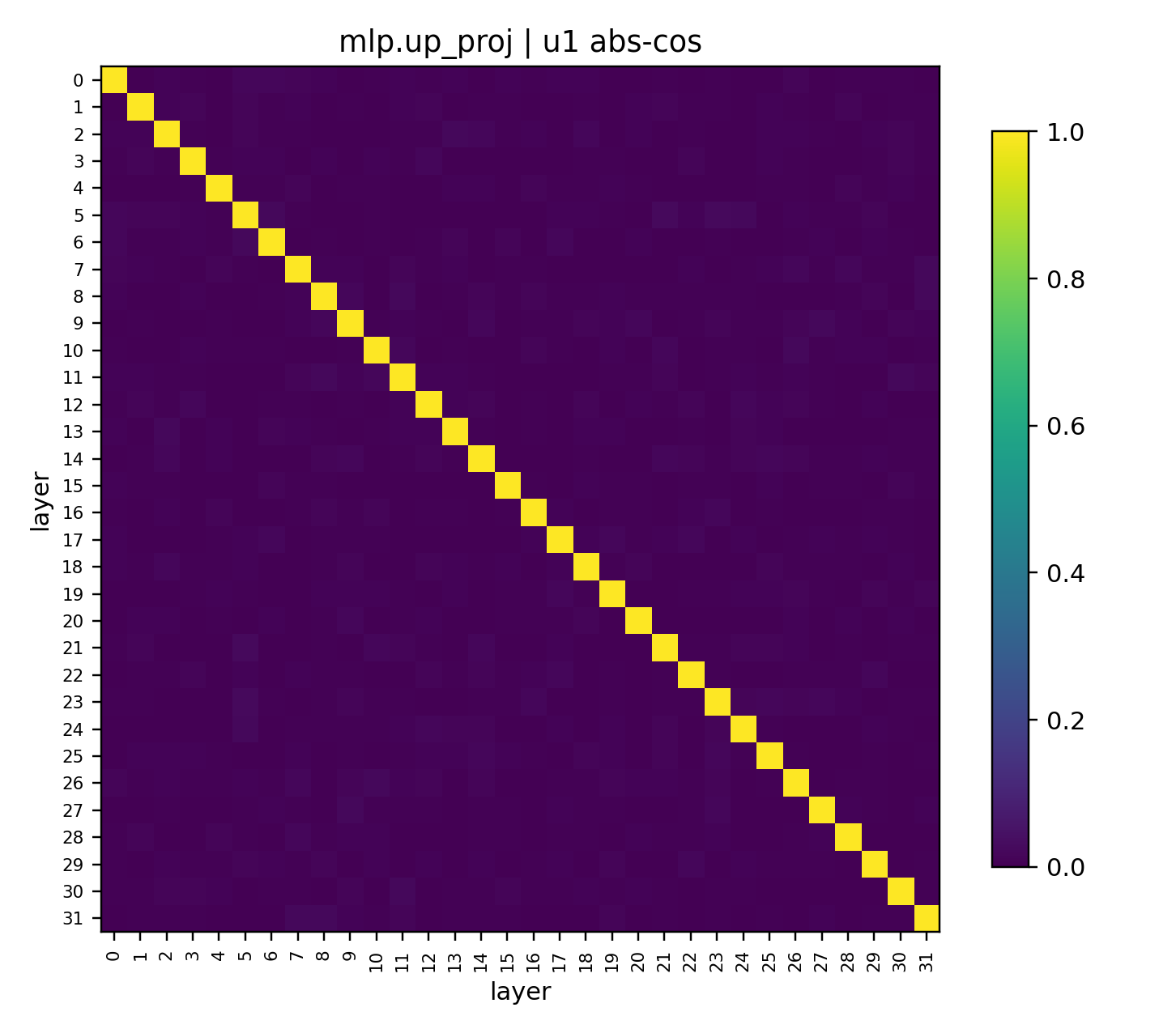} &
\includegraphics[width=0.195\textwidth]{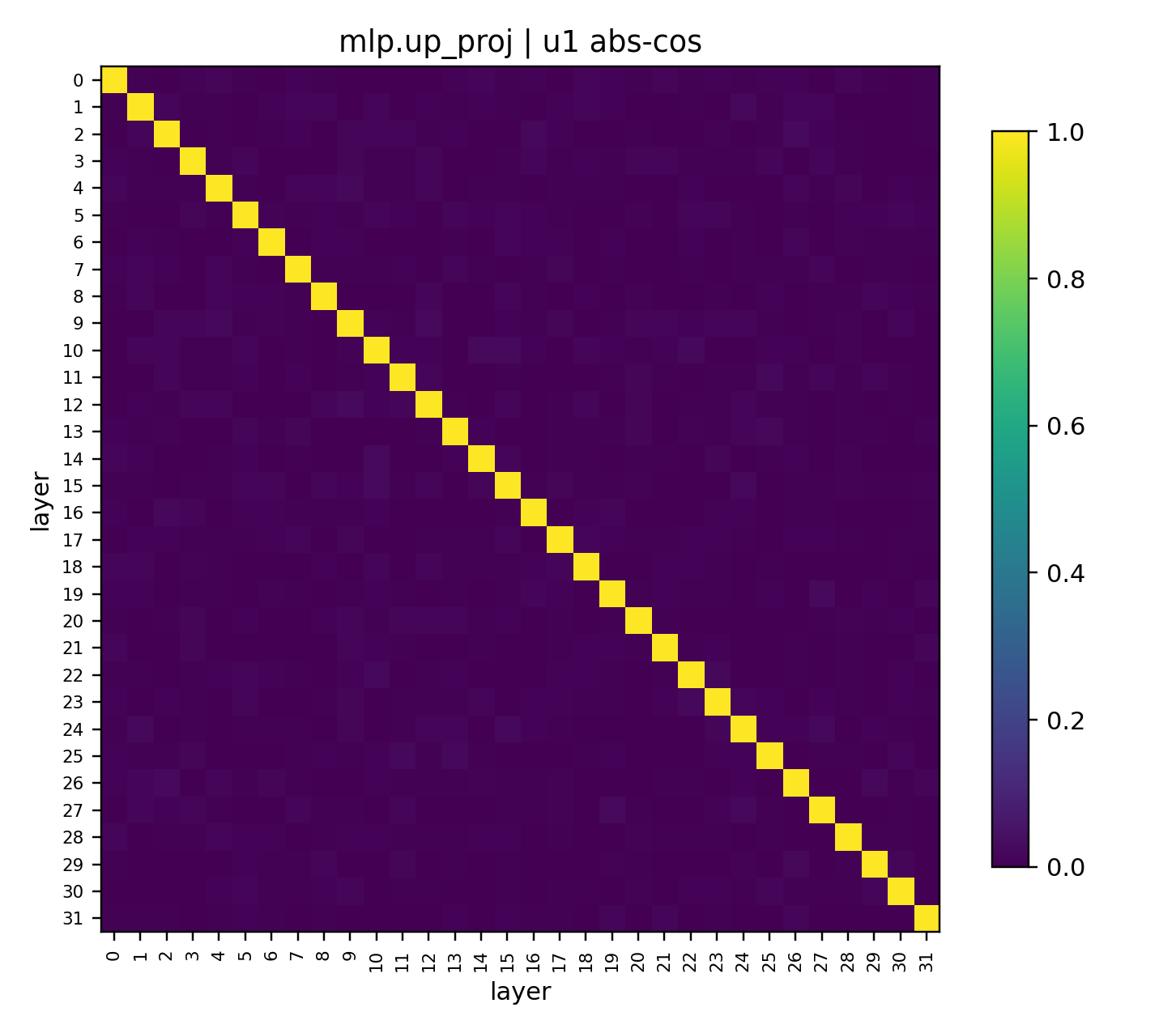} &
\includegraphics[width=0.195\textwidth]{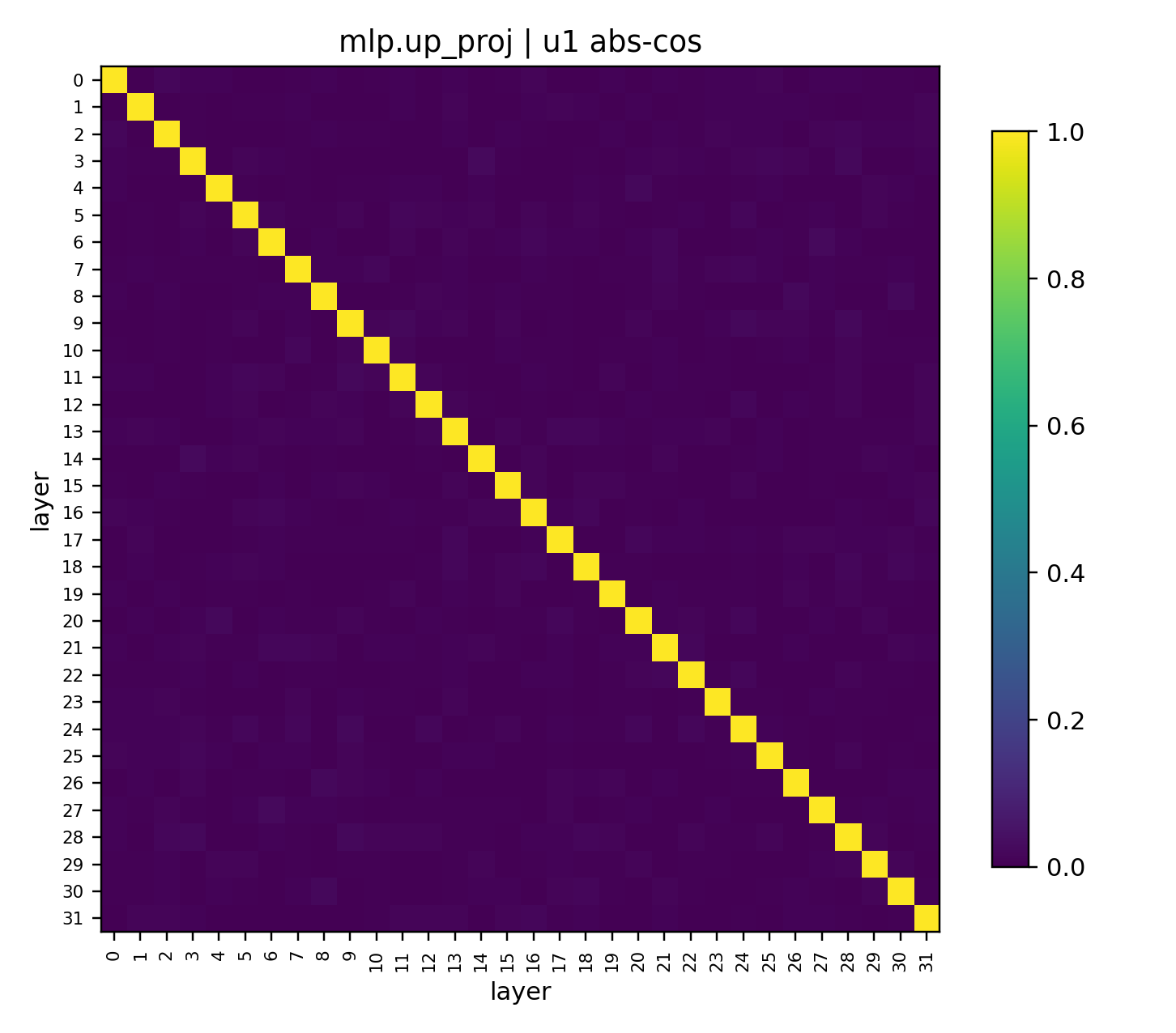} &
\includegraphics[width=0.195\textwidth]{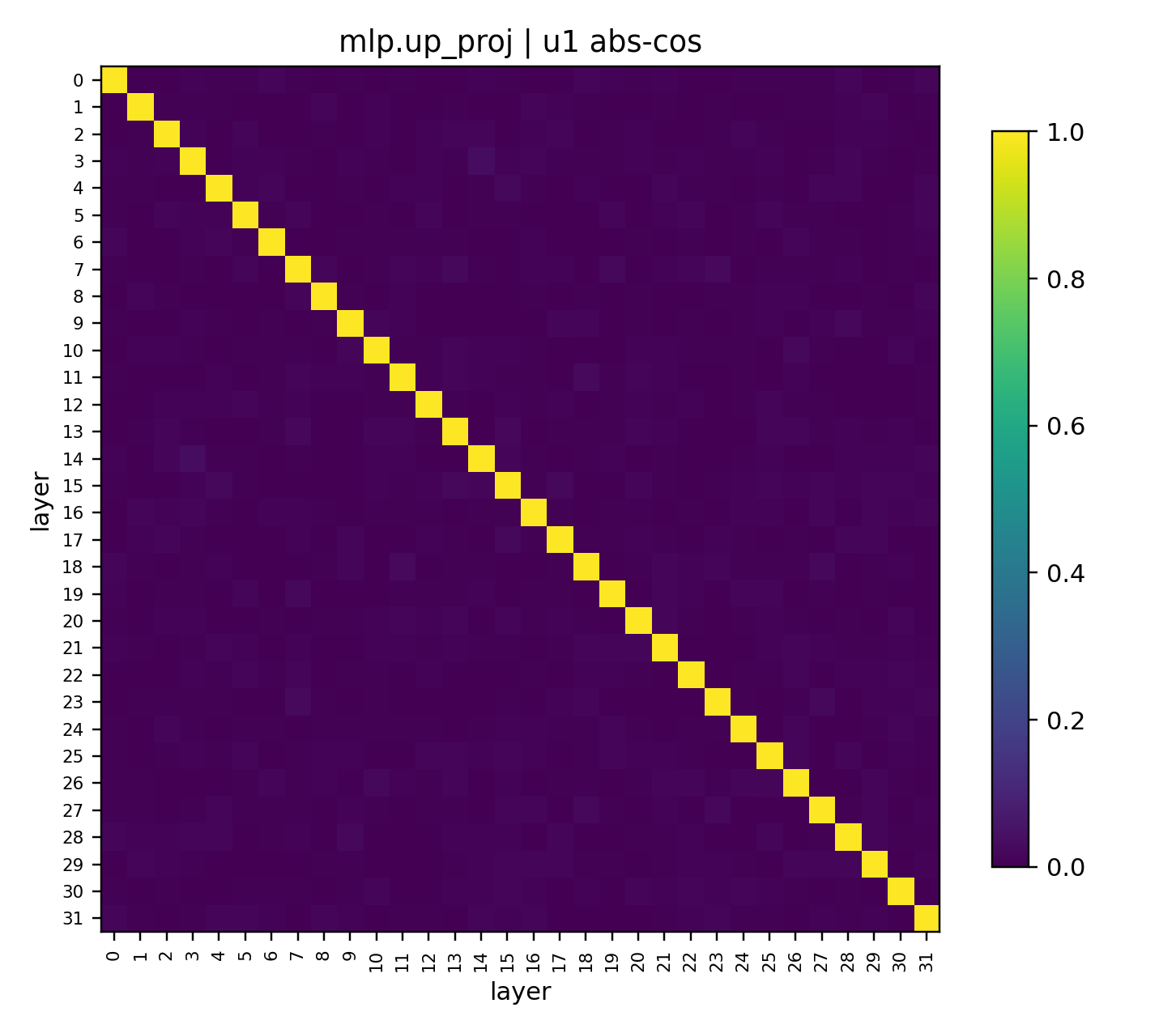} \\
\texttt{down\_proj} &
\includegraphics[width=0.195\textwidth]{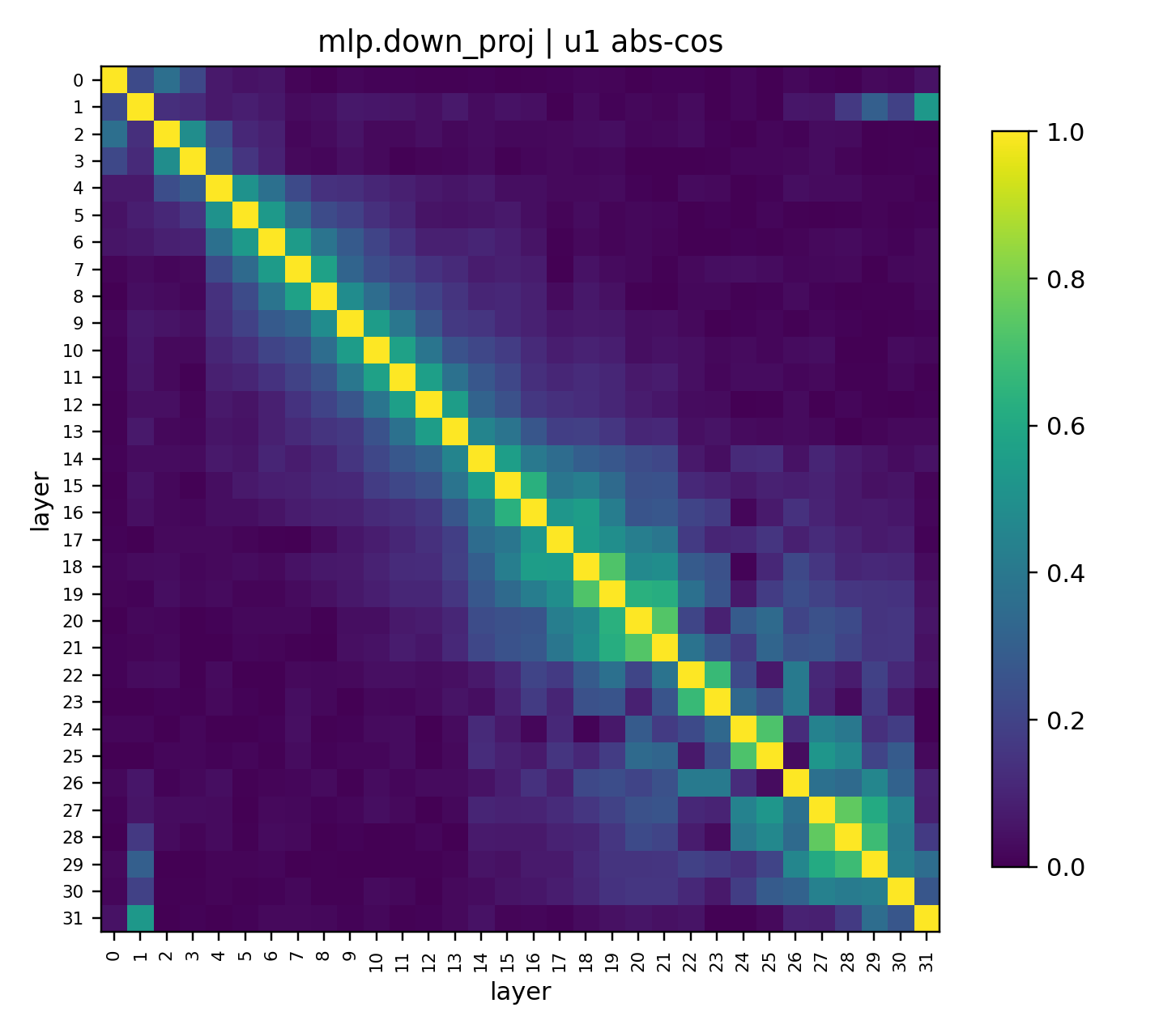} &
\includegraphics[width=0.195\textwidth]{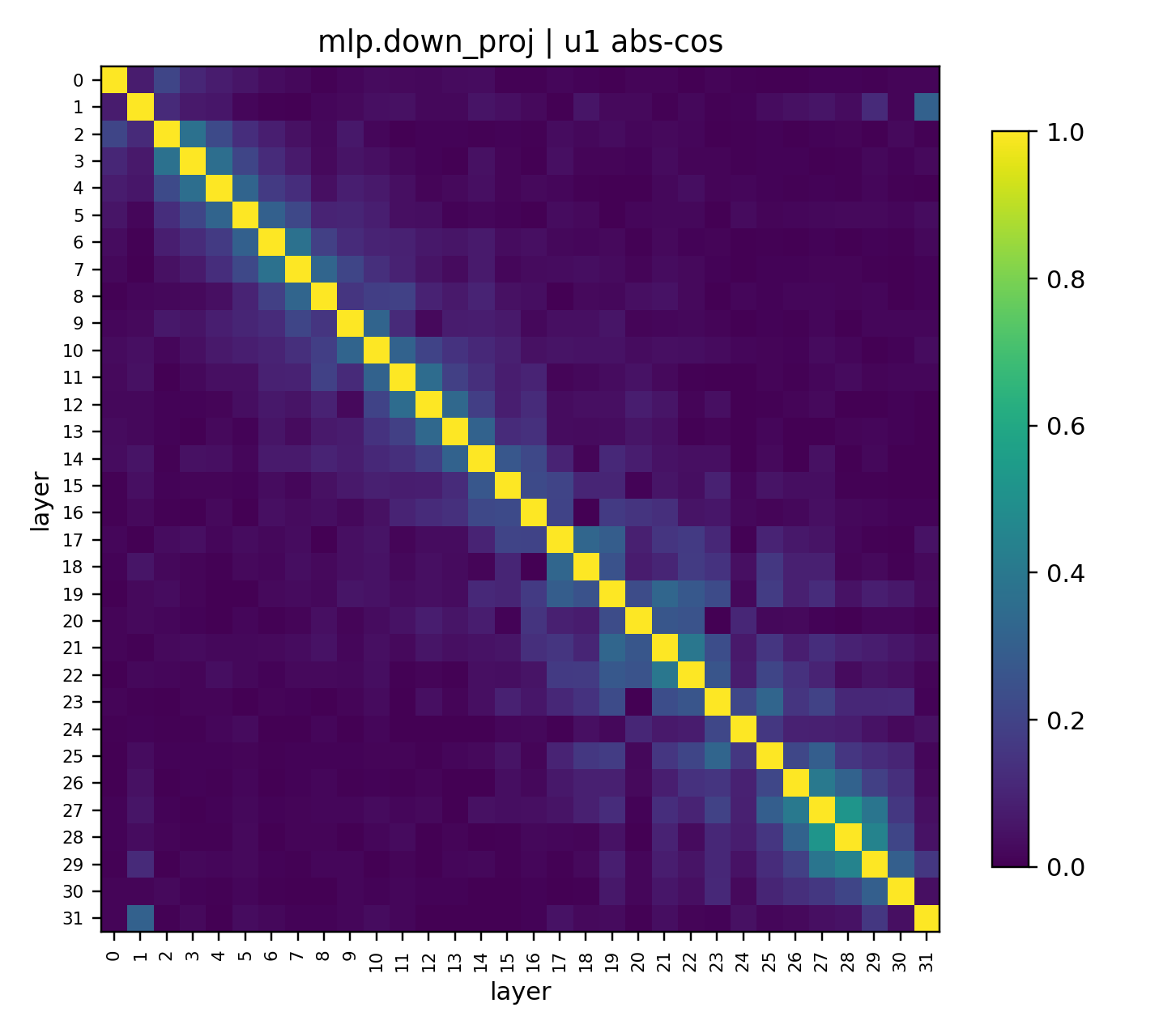} &
\includegraphics[width=0.195\textwidth]{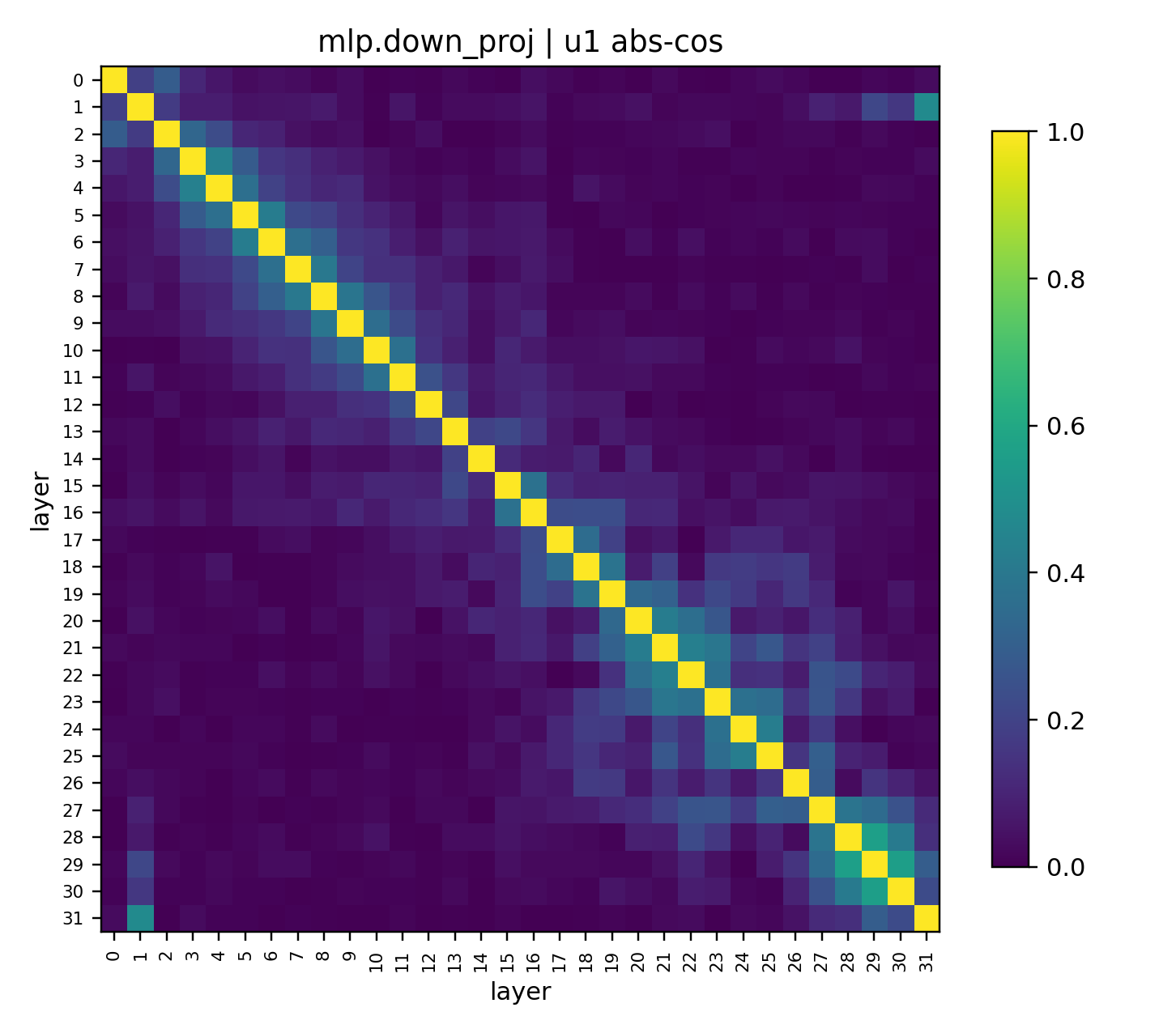} &
\includegraphics[width=0.195\textwidth]{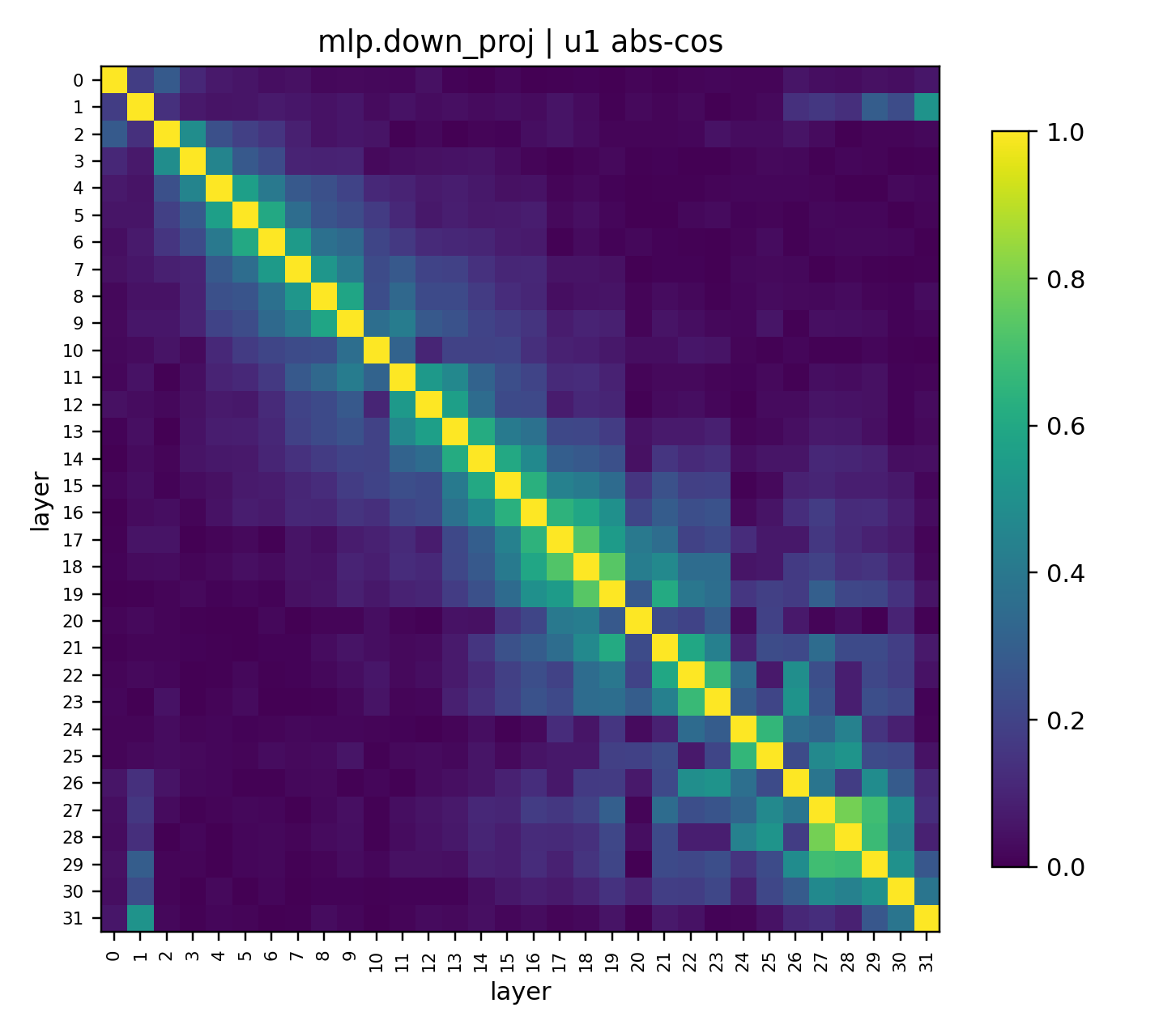} \\
\bottomrule
\end{tabular}
\caption{\textbf{Llama-3.1-8B: Principal-direction similarity heatmap wall ($|u_1^\top u_1|$).}
Each cell shows the inter-layer similarity heatmap for a specific (task, module).}
\label{fig:heatwall_llama_u1}
\end{figure*}

\clearpage
\begin{figure*}[p]
\centering
\small
\setlength{\tabcolsep}{2pt}
\renewcommand{\arraystretch}{1.0}
\begin{tabular}{lcccc}
\toprule
\textbf{Module} &
\textbf{Math} & \textbf{Code} & \textbf{Instruction} & \textbf{Commonsense} \\
\midrule
\texttt{q\_proj} &
\includegraphics[width=0.195\textwidth]{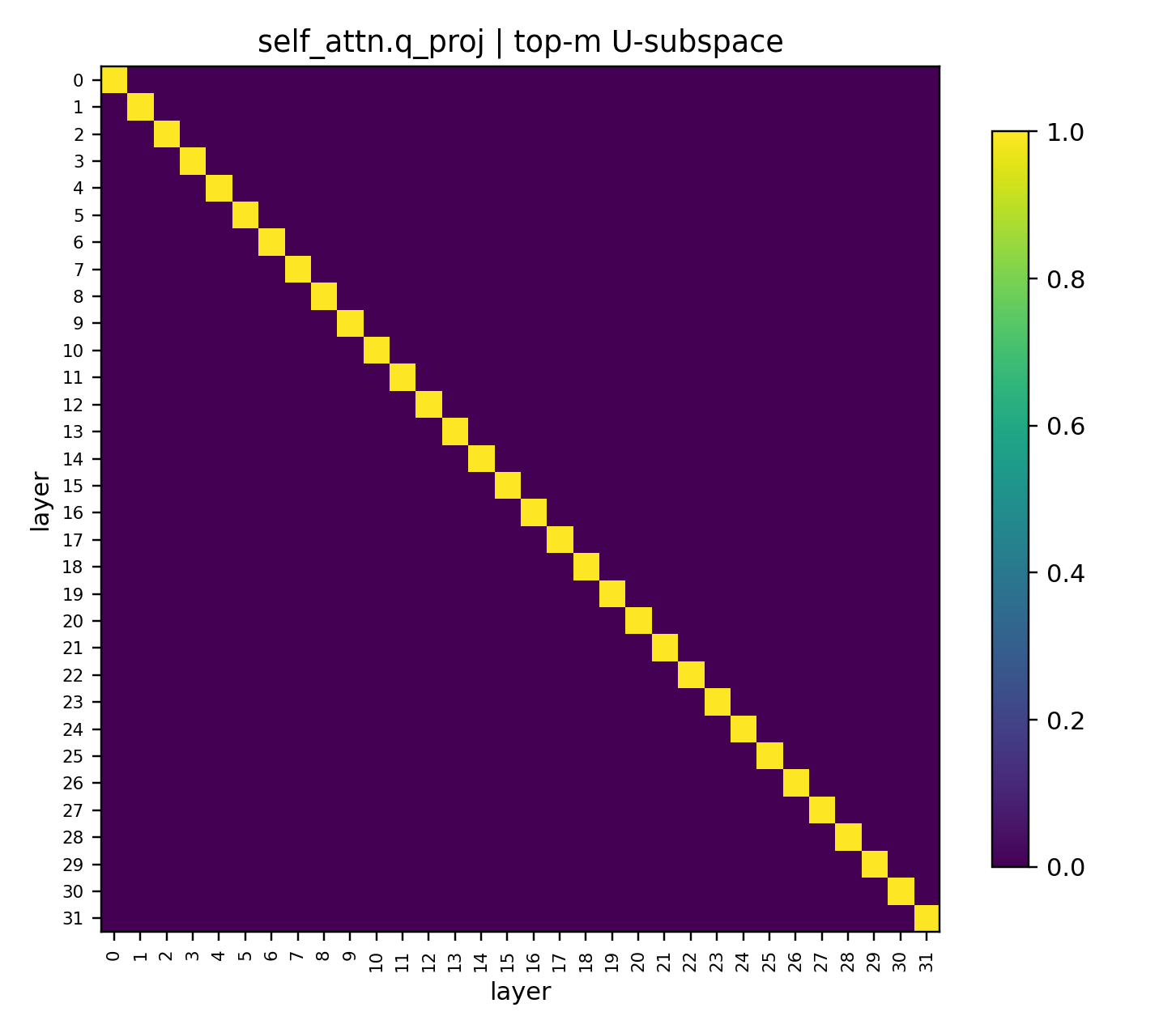} &
\includegraphics[width=0.195\textwidth]{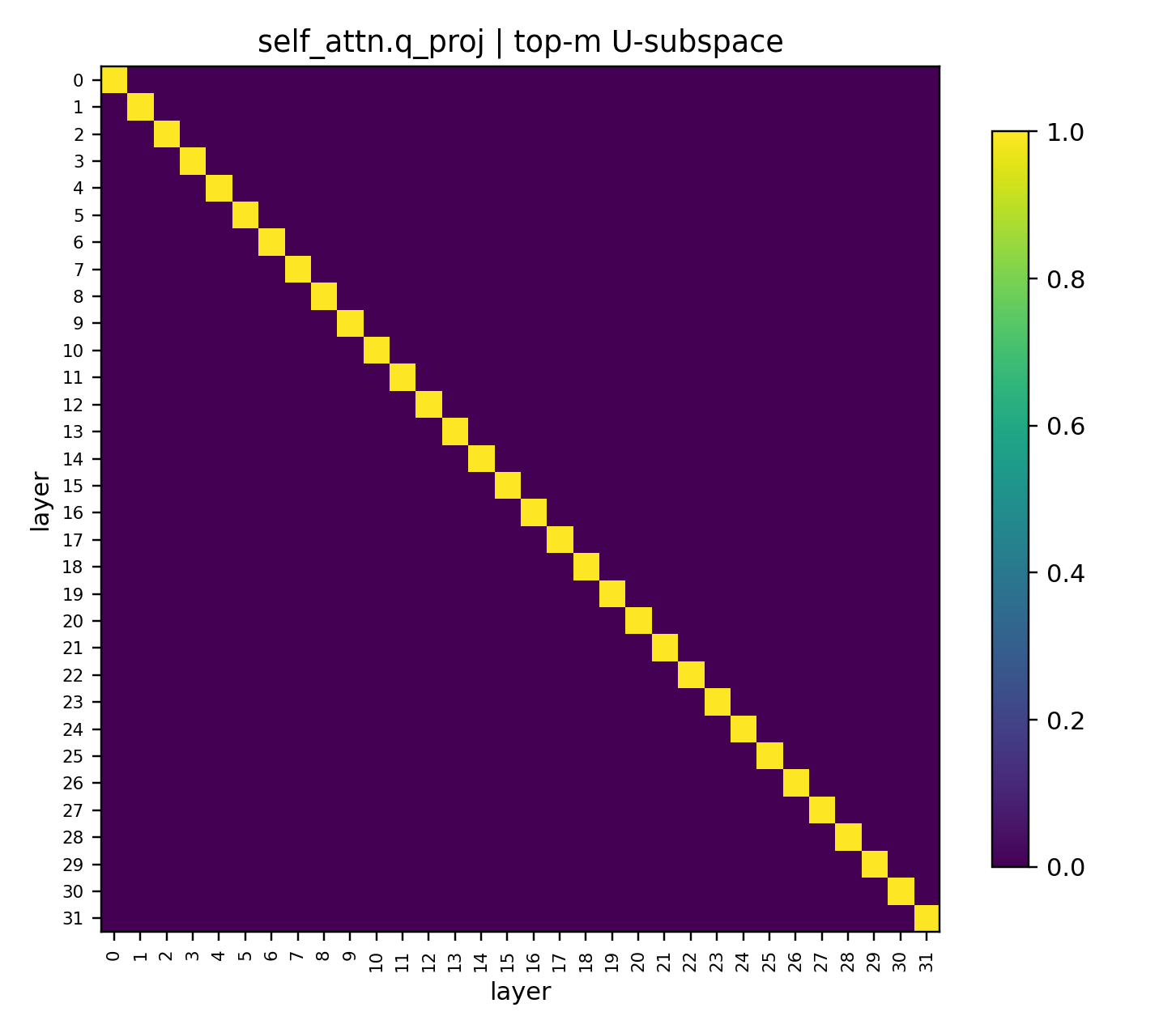} &
\includegraphics[width=0.195\textwidth]{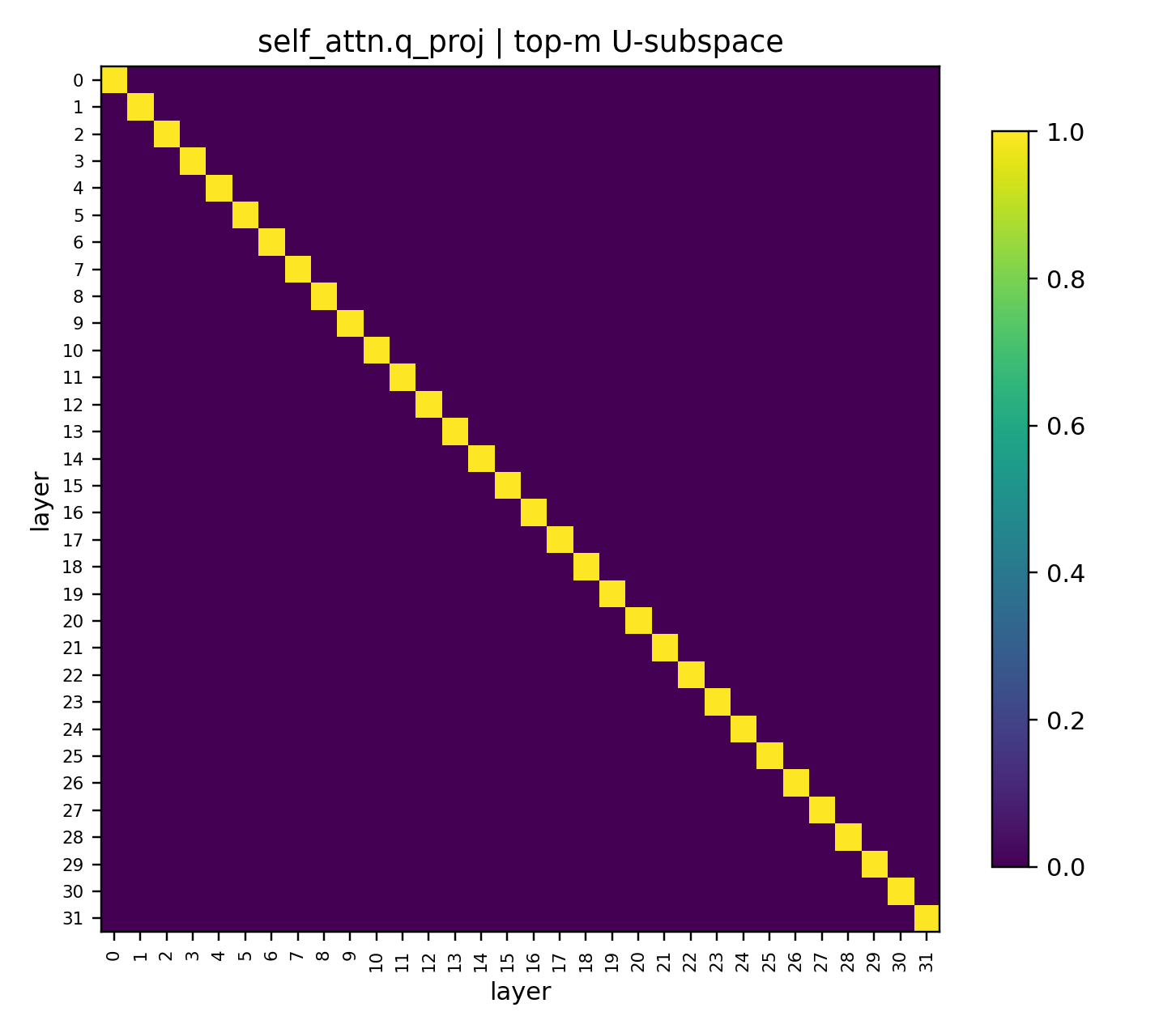} &
\includegraphics[width=0.195\textwidth]{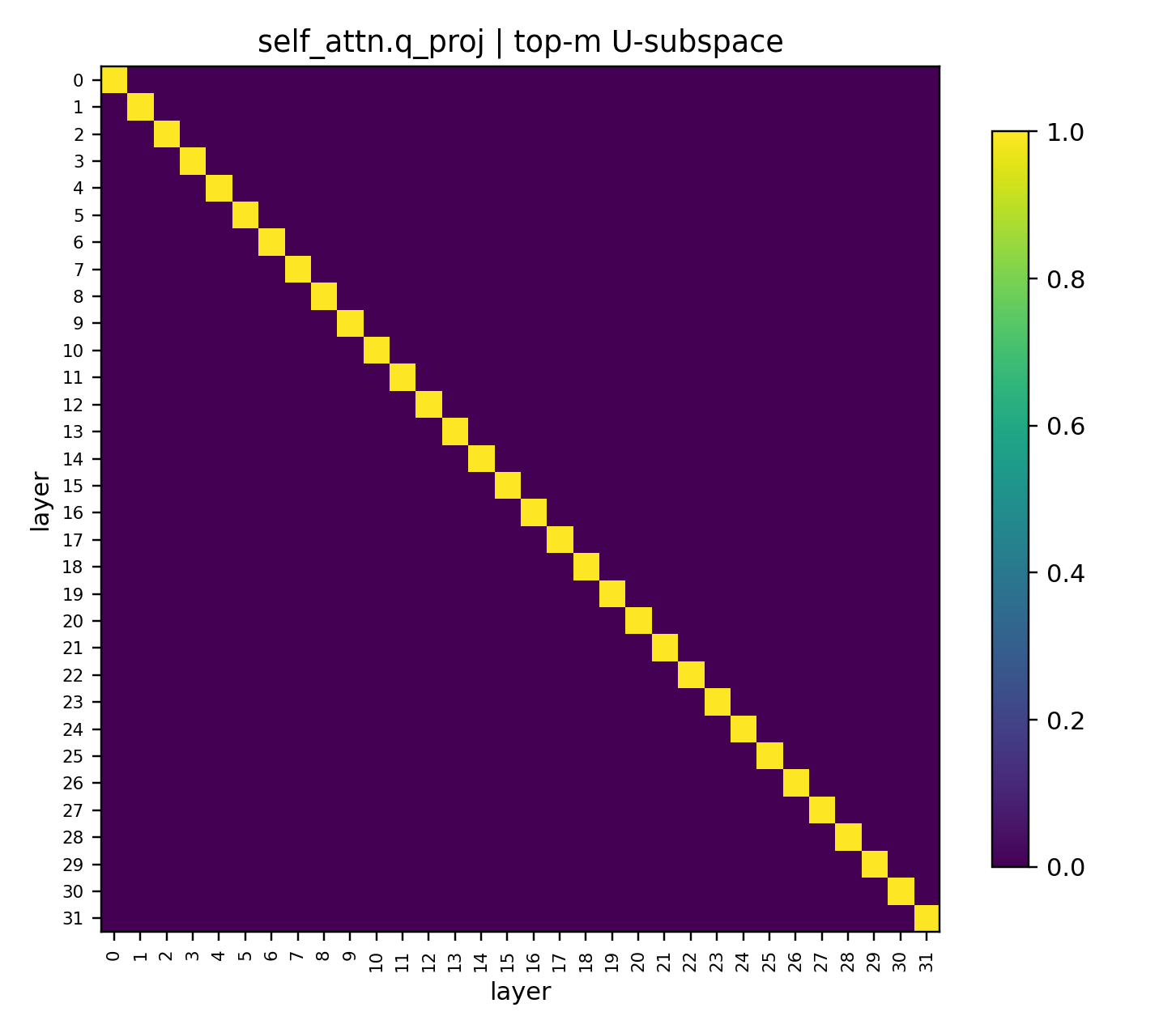} \\
\texttt{k\_proj} &
\includegraphics[width=0.195\textwidth]{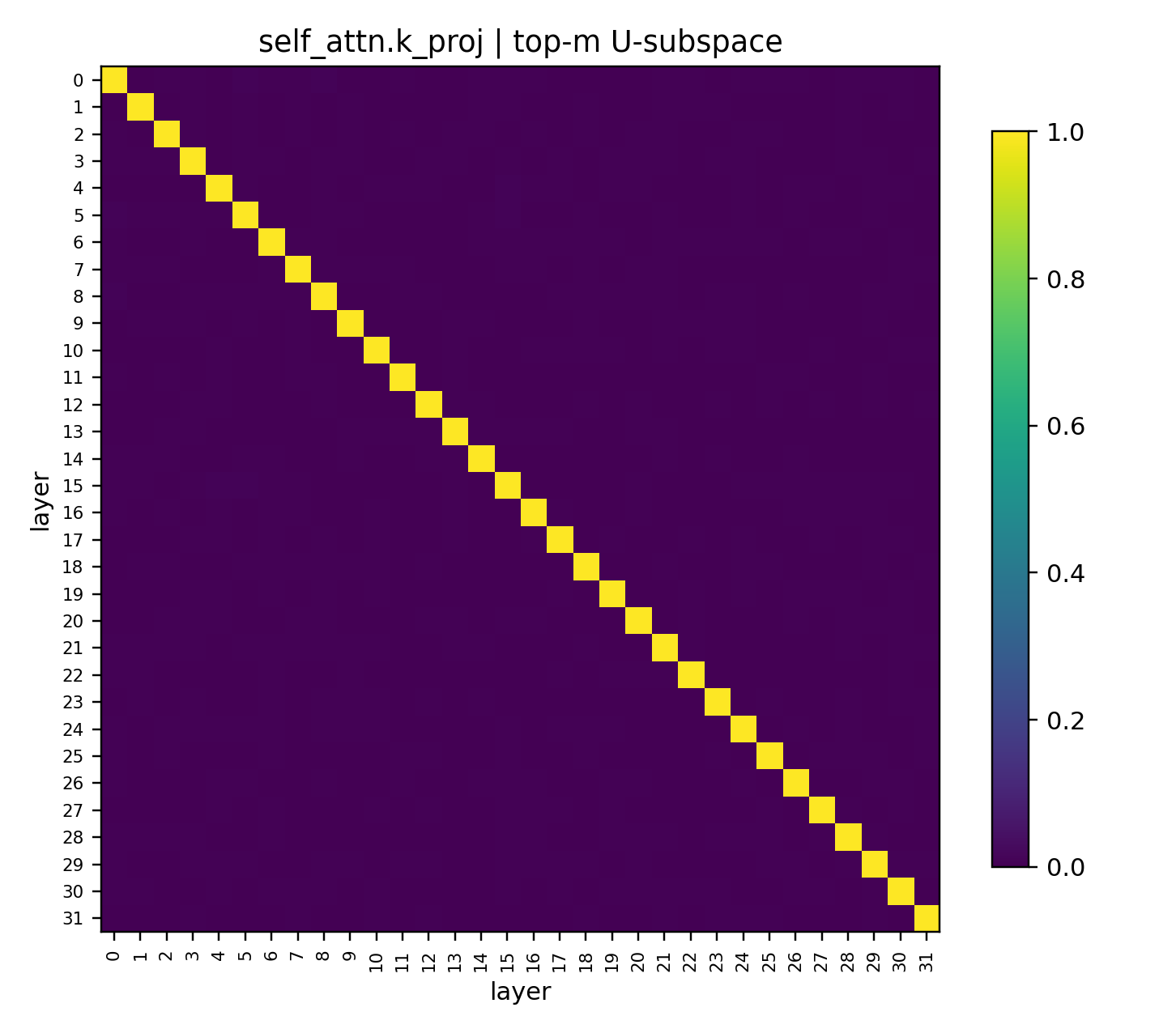} &
\includegraphics[width=0.195\textwidth]{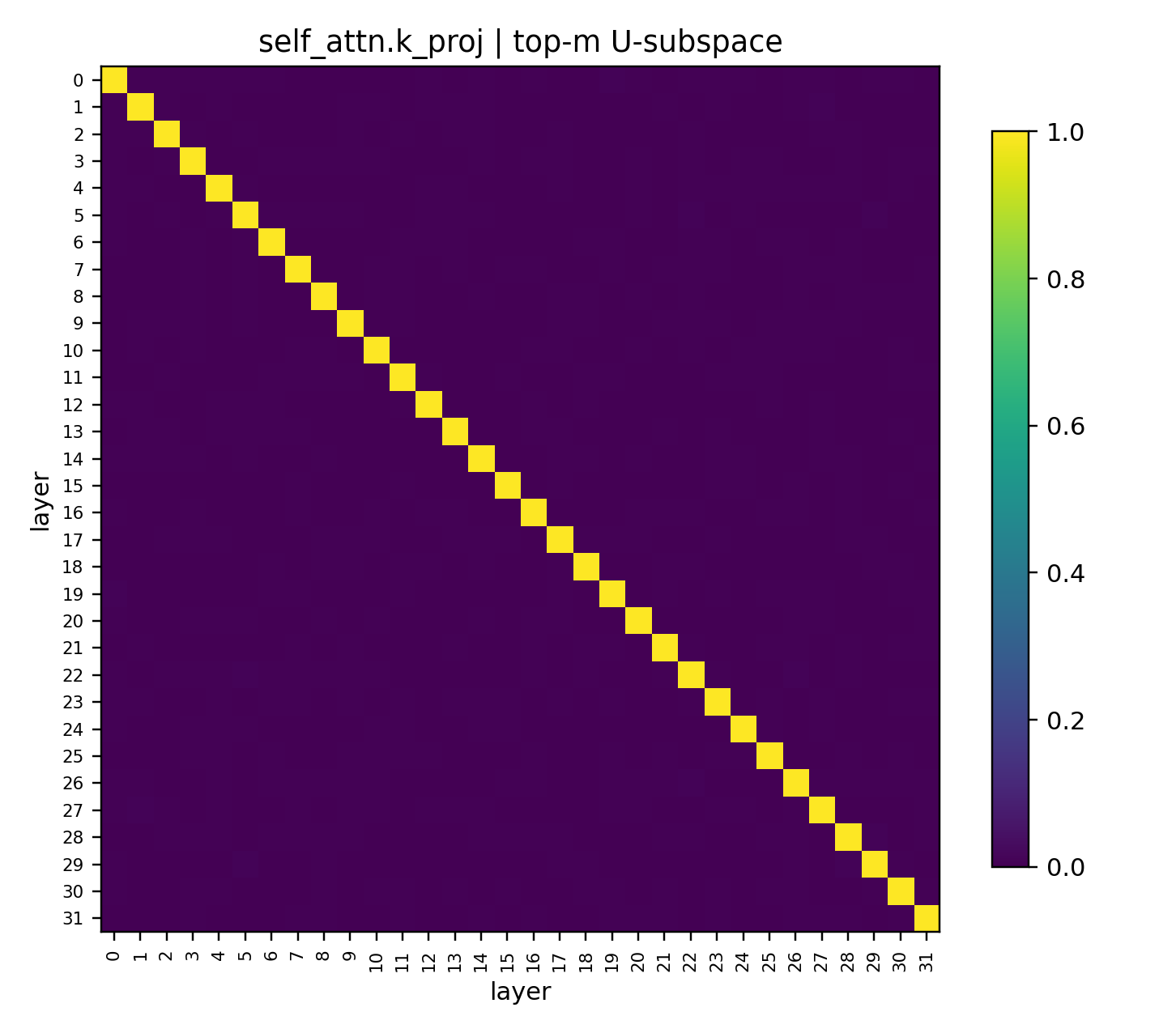} &
\includegraphics[width=0.195\textwidth]{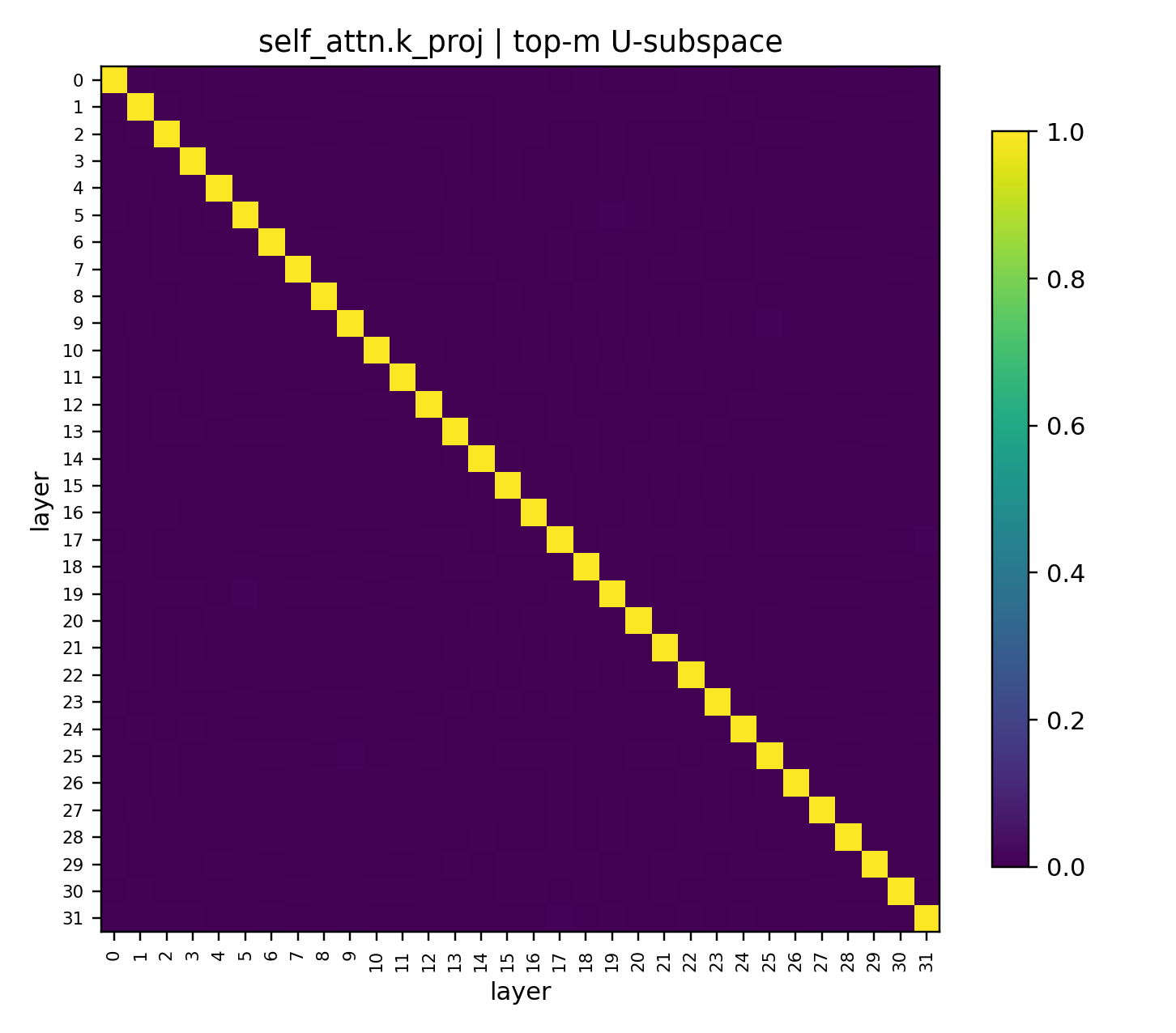} &
\includegraphics[width=0.195\textwidth]{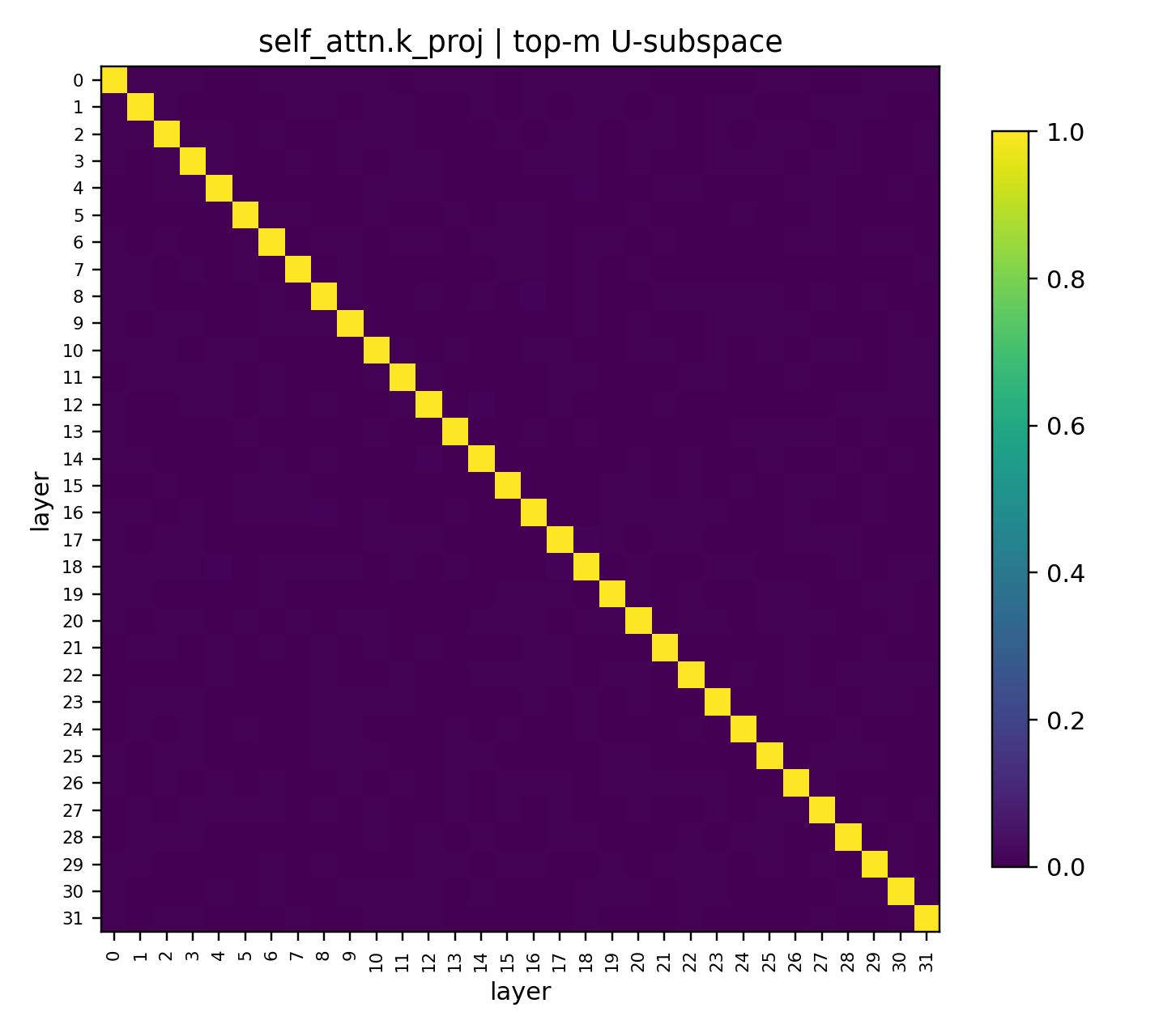} \\
\texttt{v\_proj} &
\includegraphics[width=0.195\textwidth]{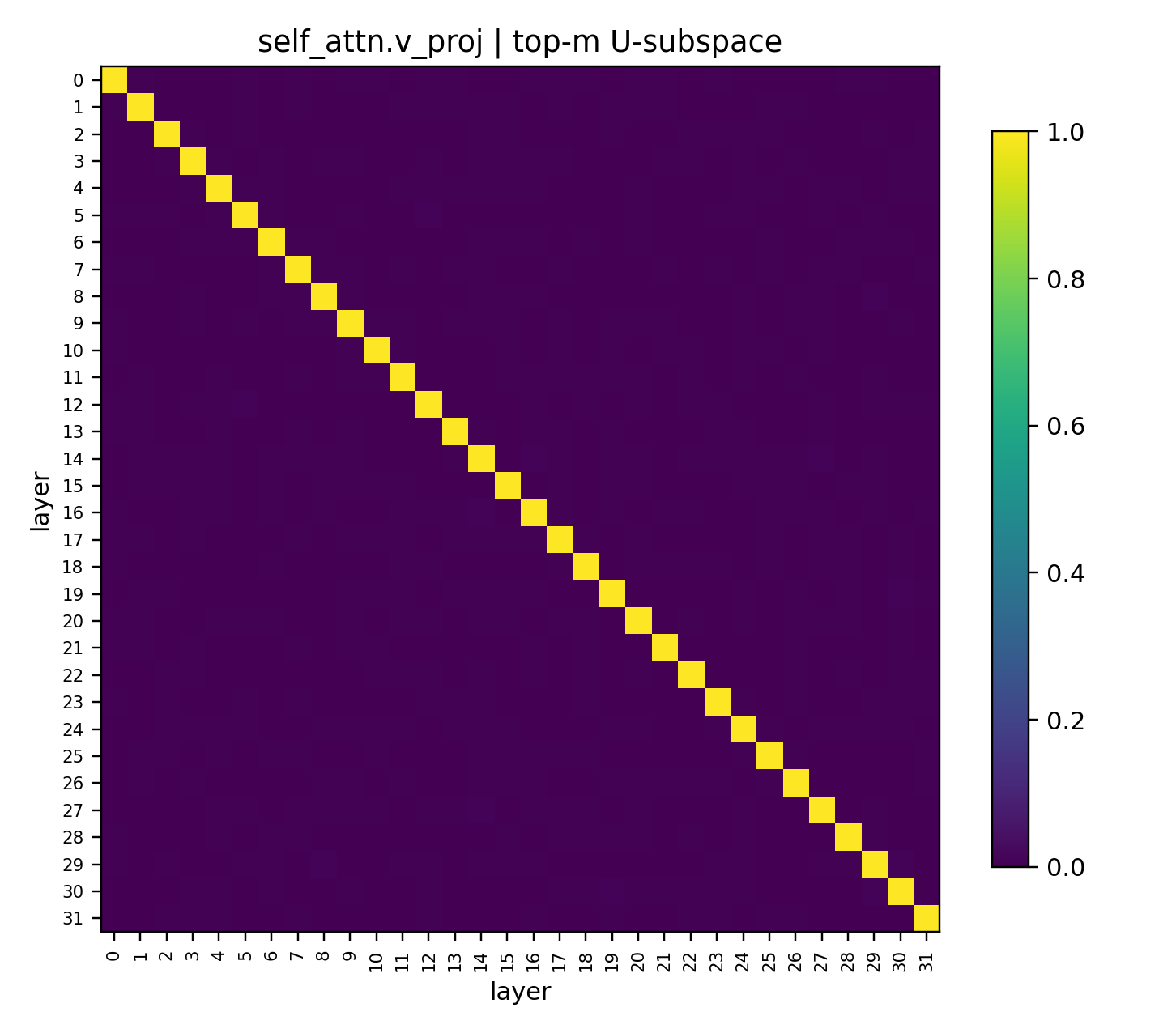} &
\includegraphics[width=0.195\textwidth]{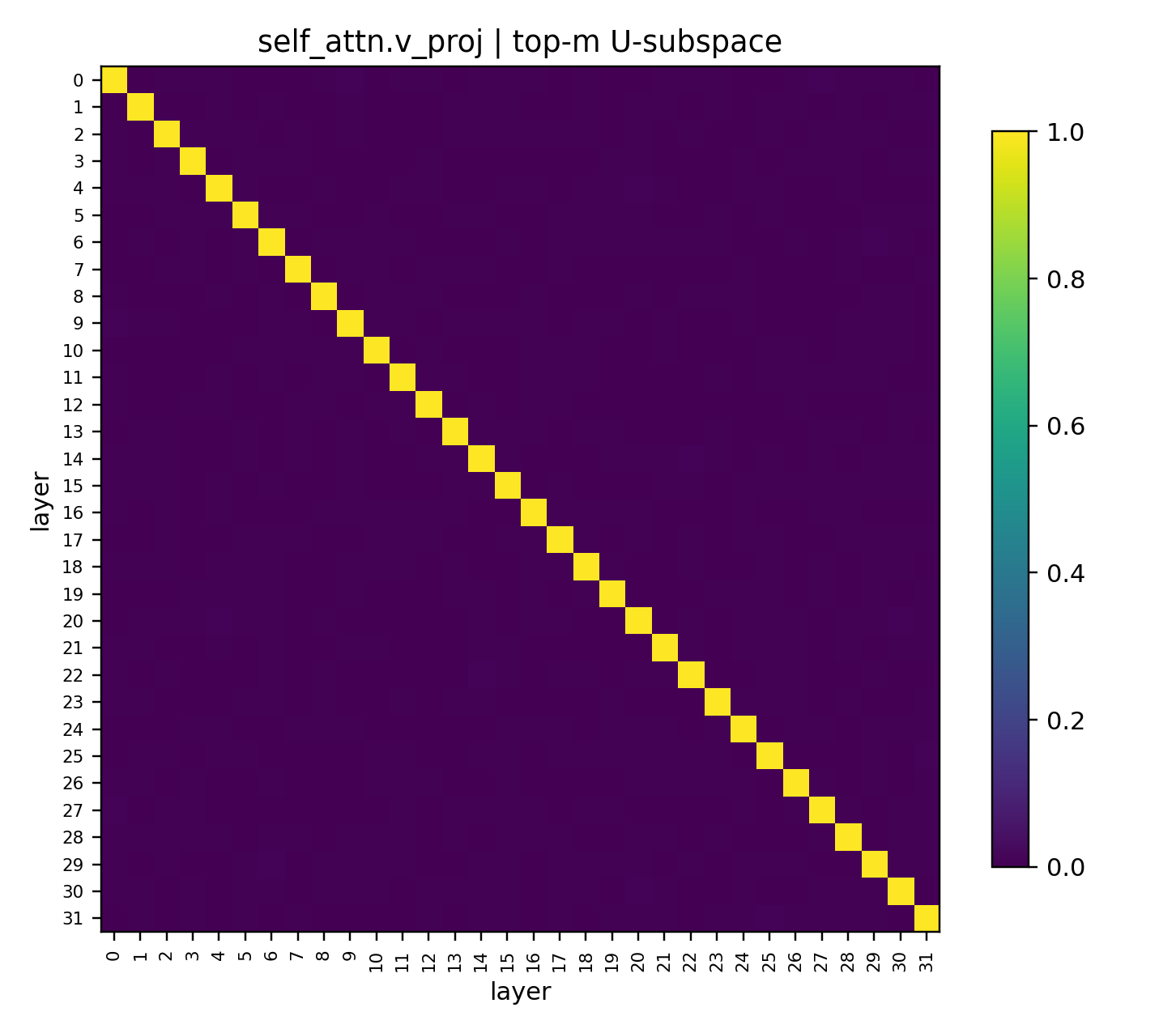} &
\includegraphics[width=0.195\textwidth]{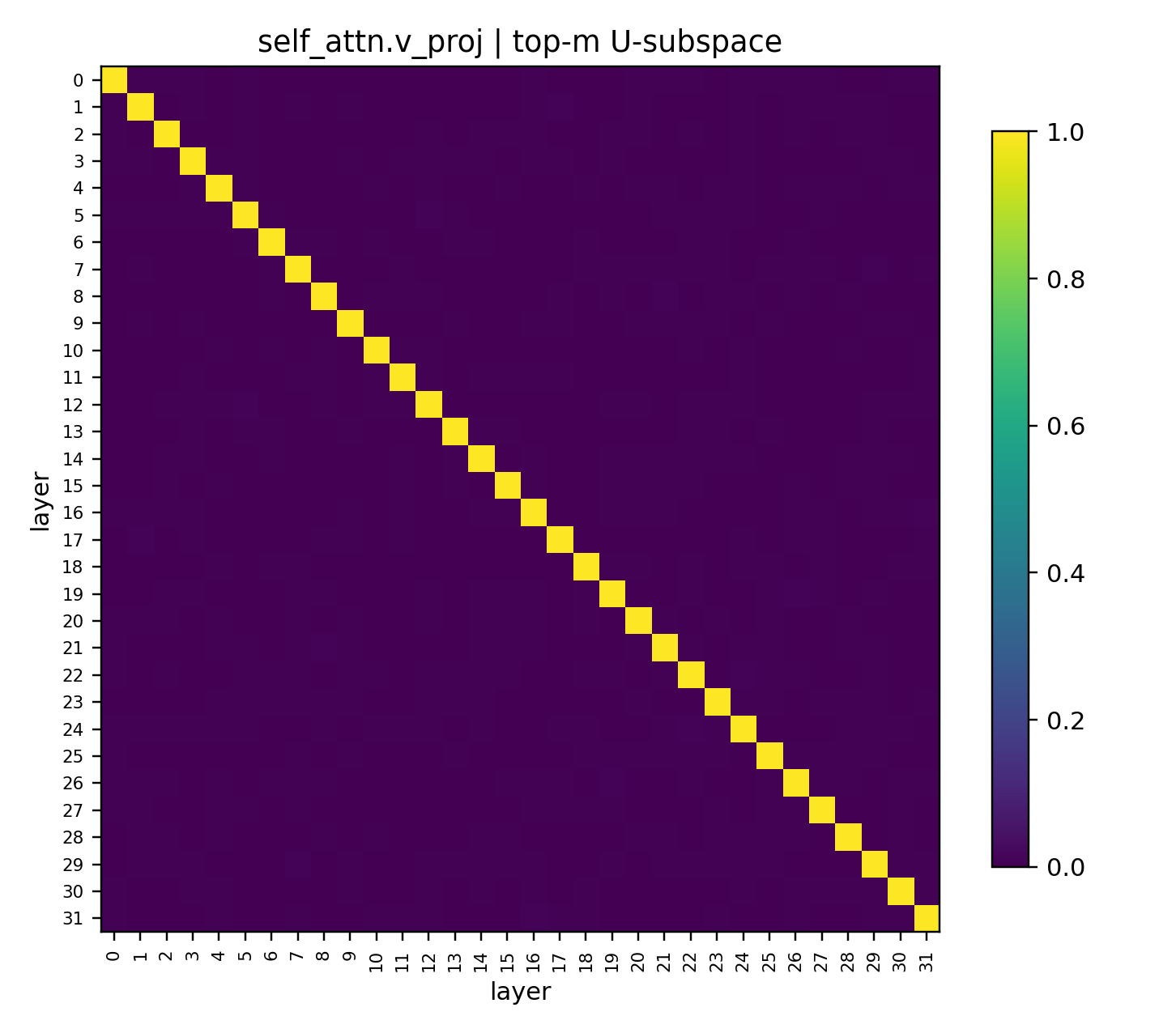} &
\includegraphics[width=0.195\textwidth]{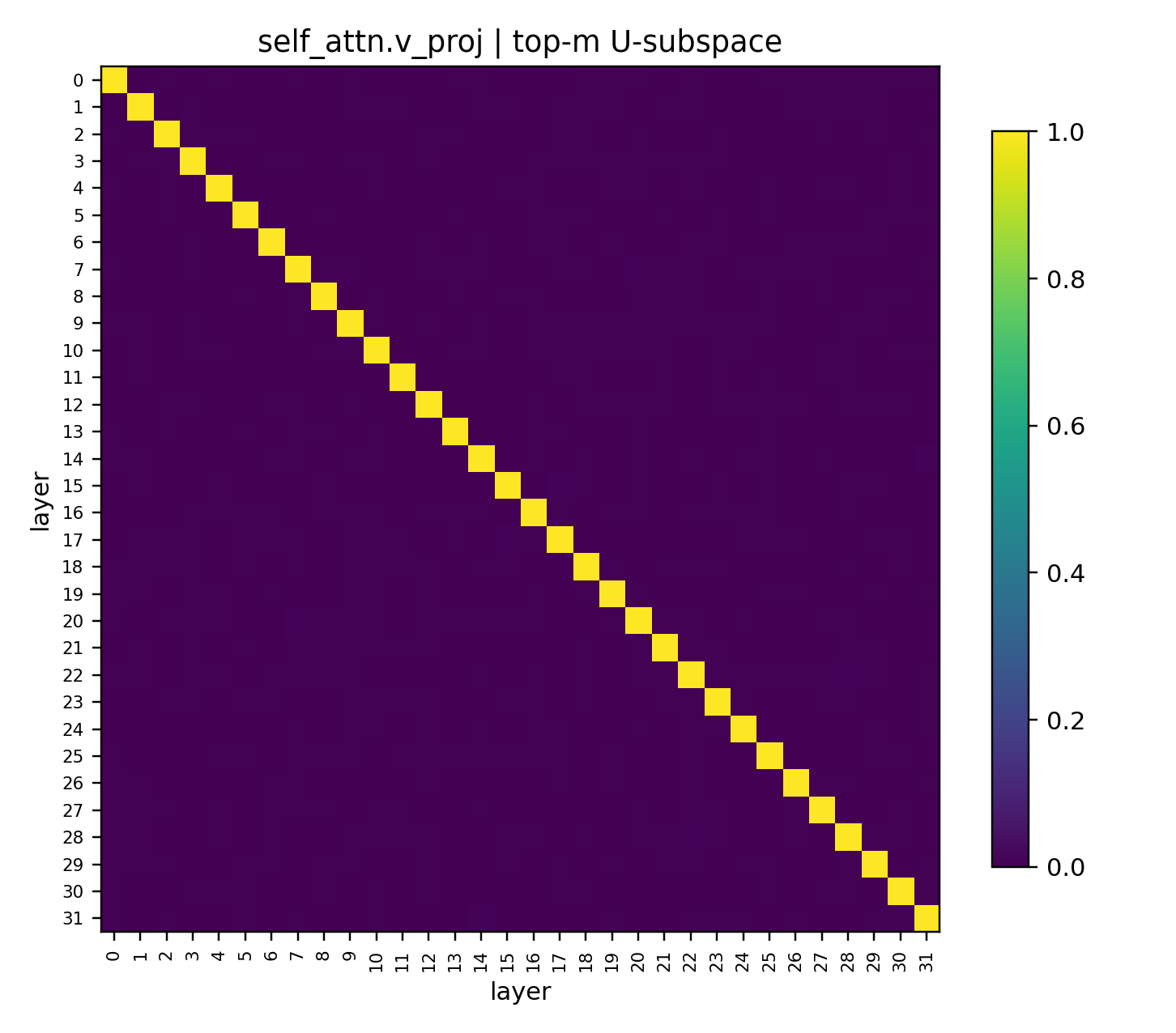} \\
\texttt{o\_proj} &
\includegraphics[width=0.195\textwidth]{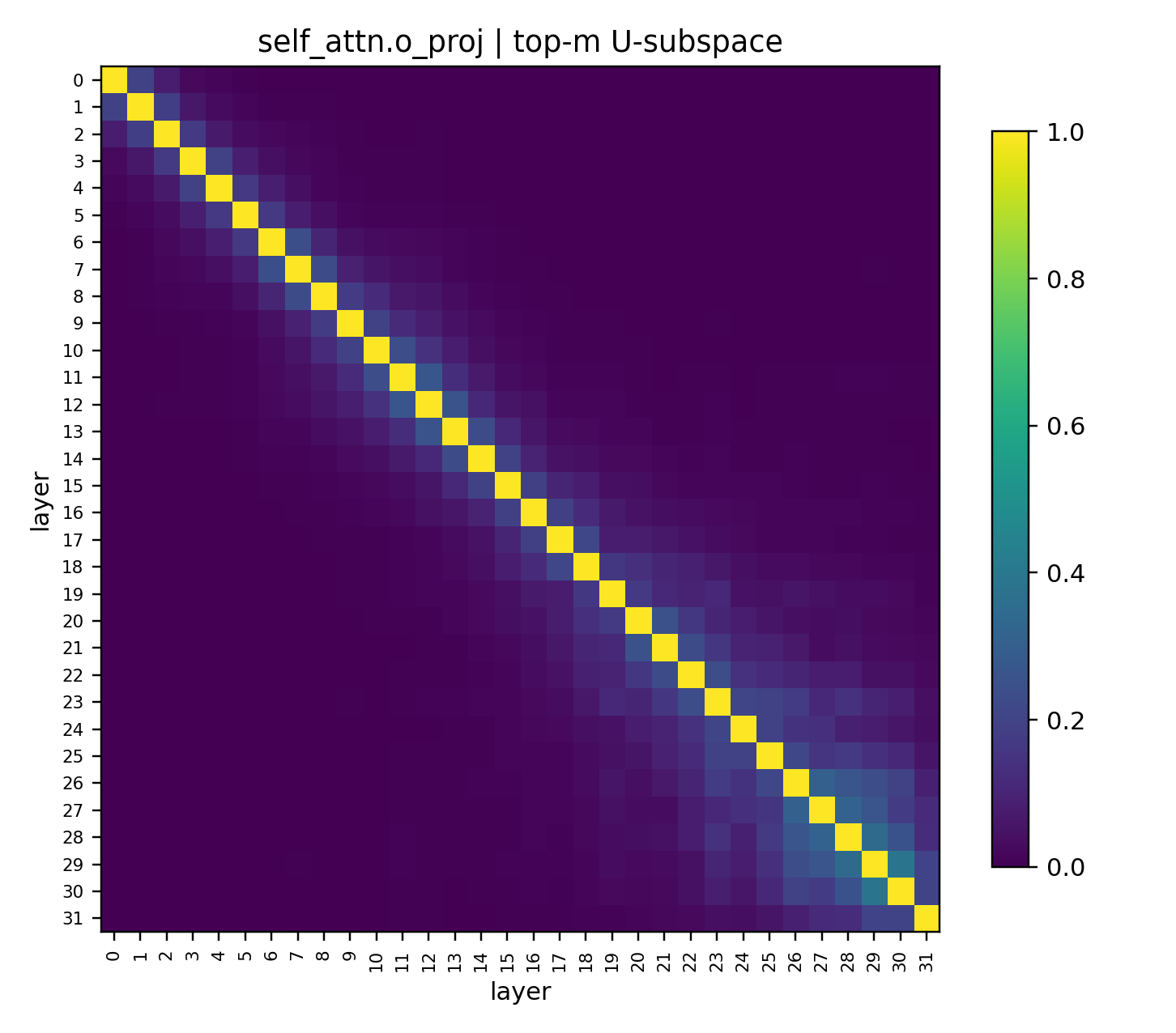} &
\includegraphics[width=0.195\textwidth]{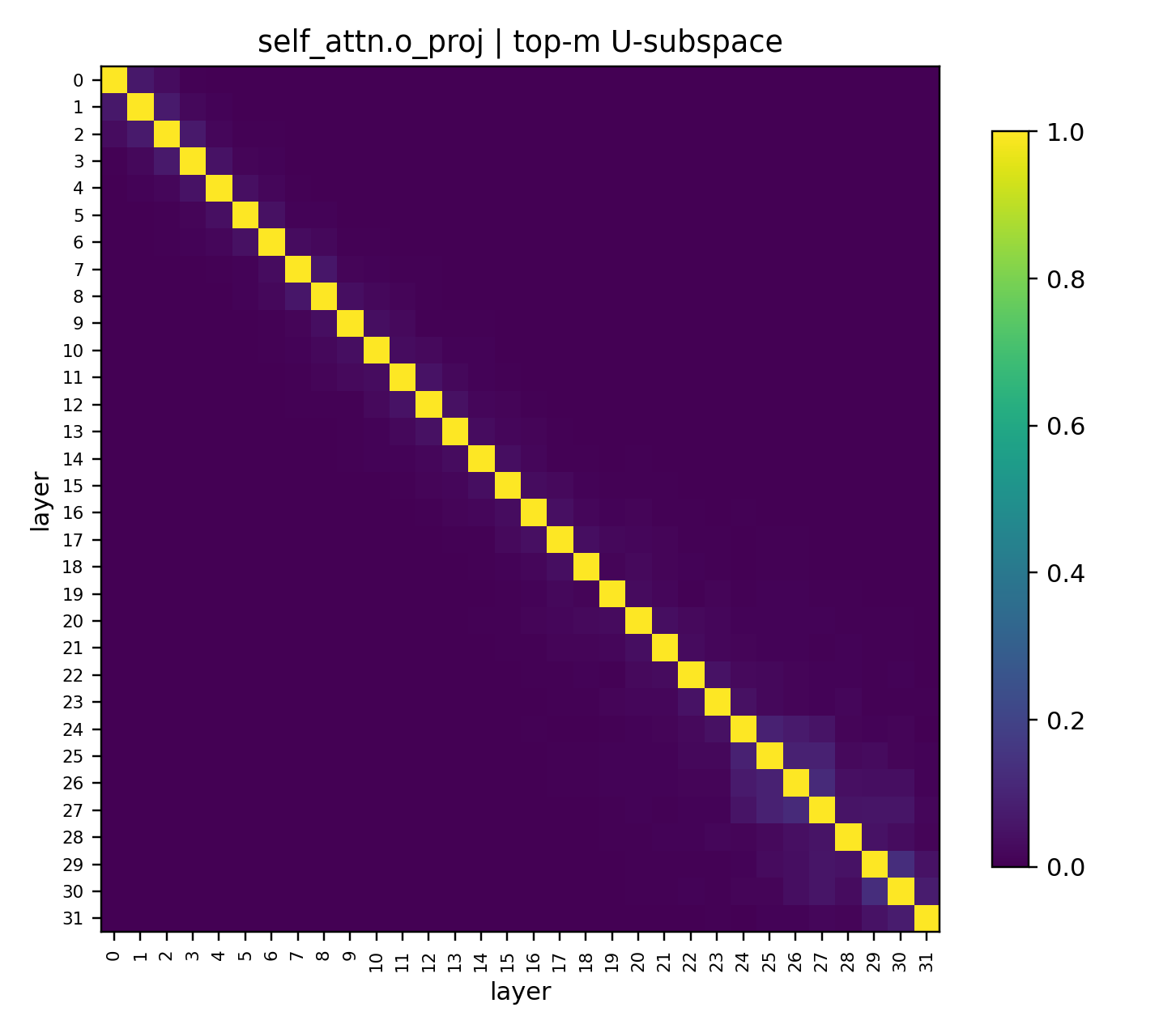} &
\includegraphics[width=0.195\textwidth]{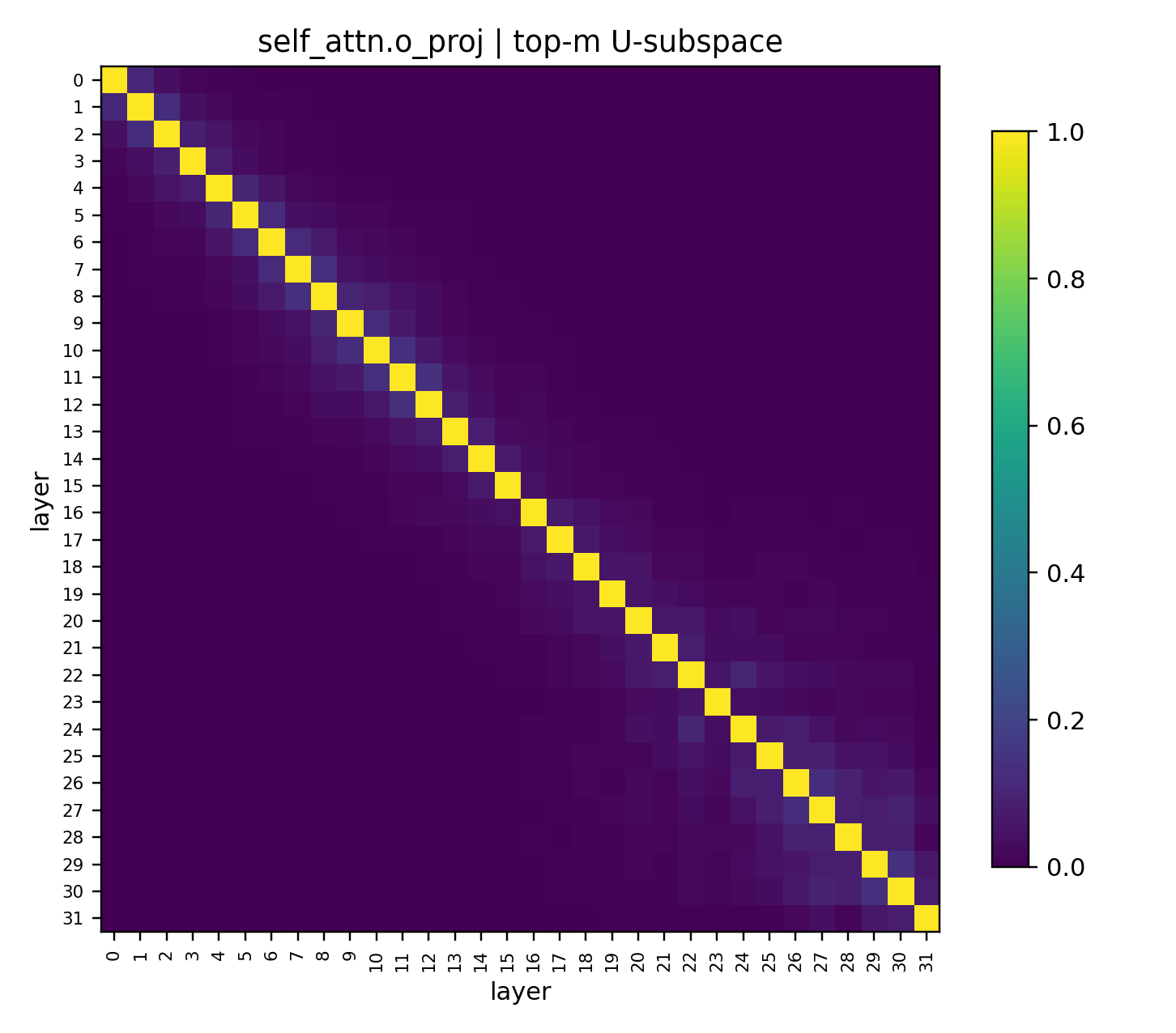} &
\includegraphics[width=0.195\textwidth]{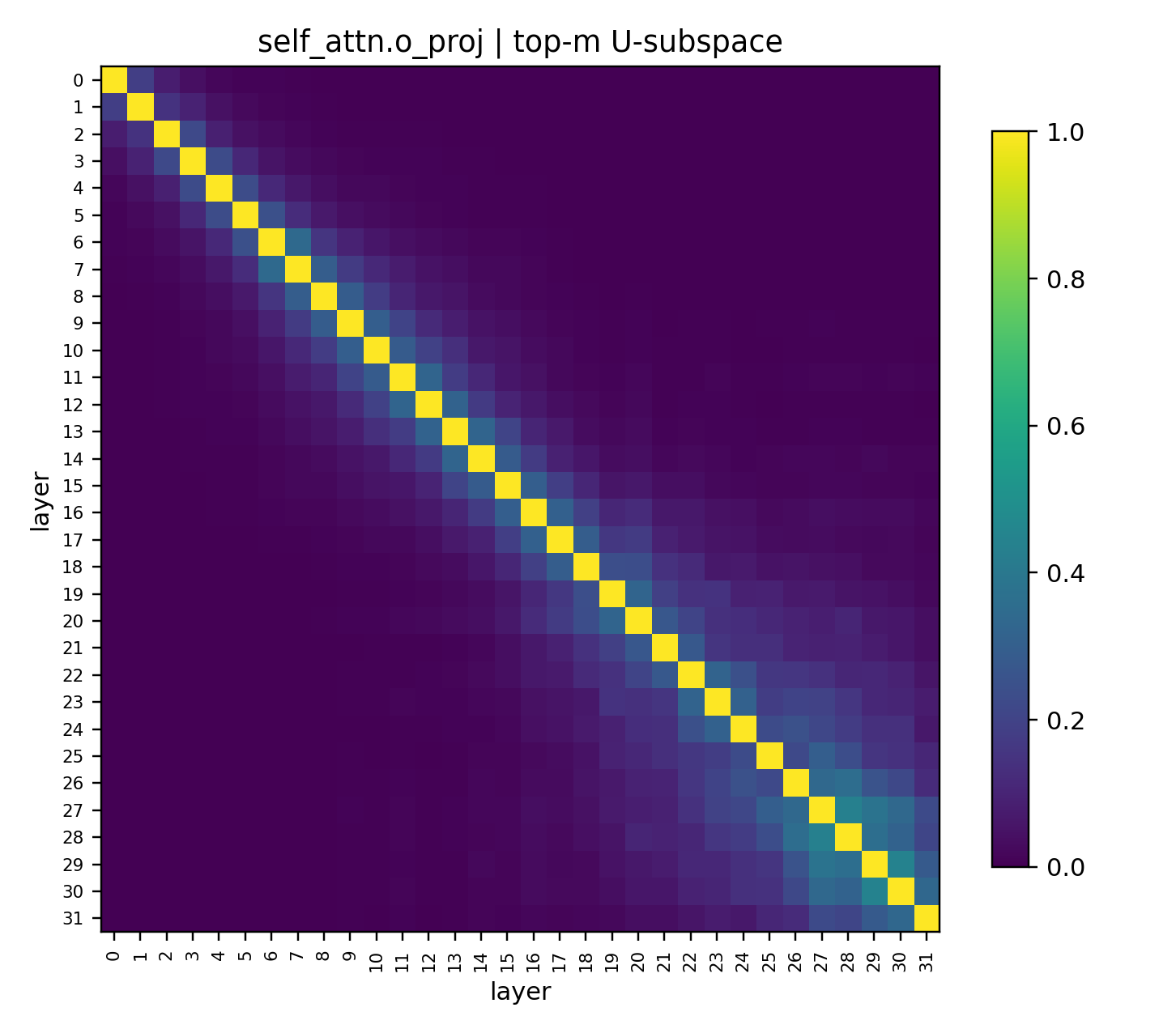} \\
\texttt{gate\_proj} &
\includegraphics[width=0.195\textwidth]{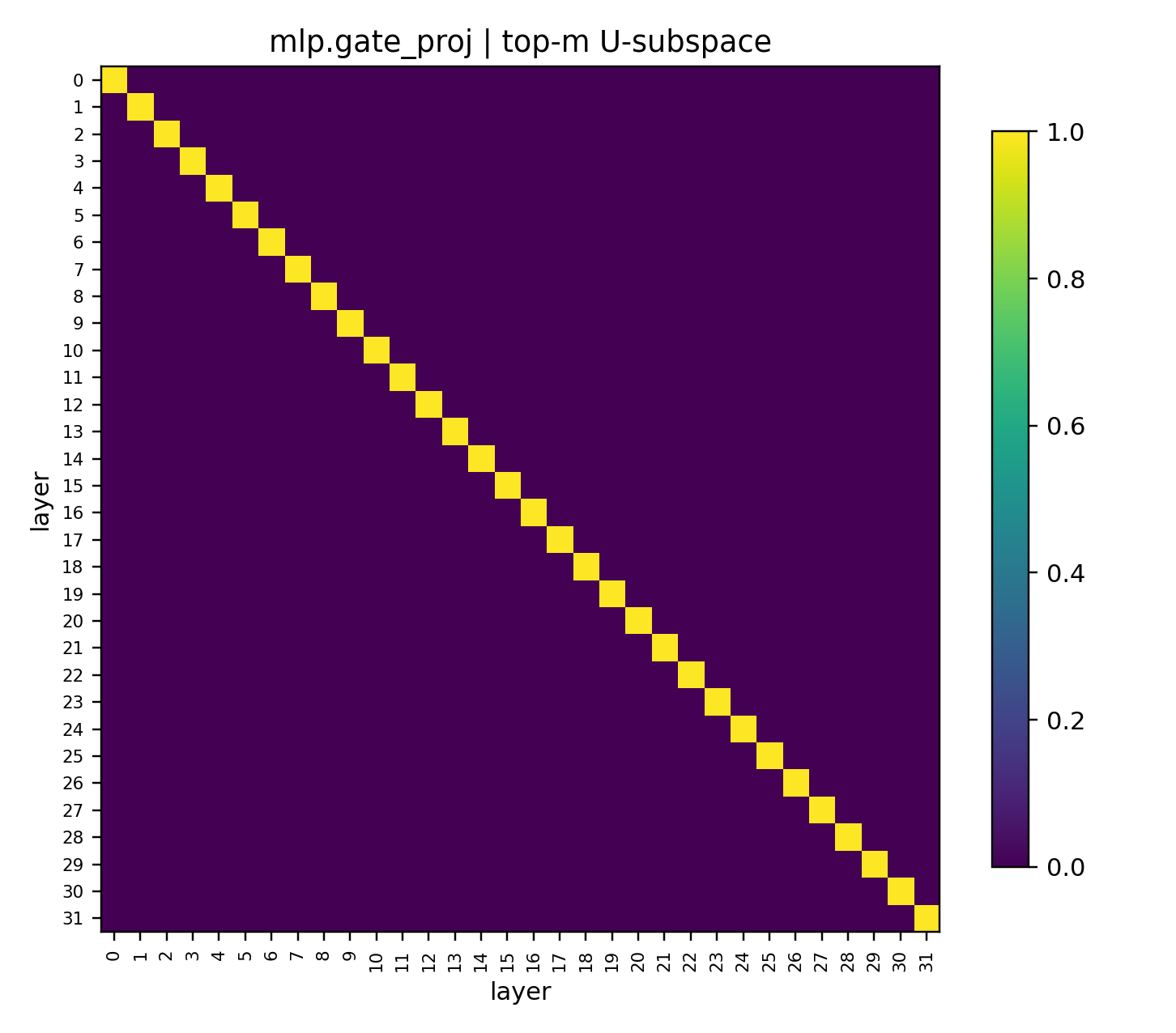} &
\includegraphics[width=0.195\textwidth]{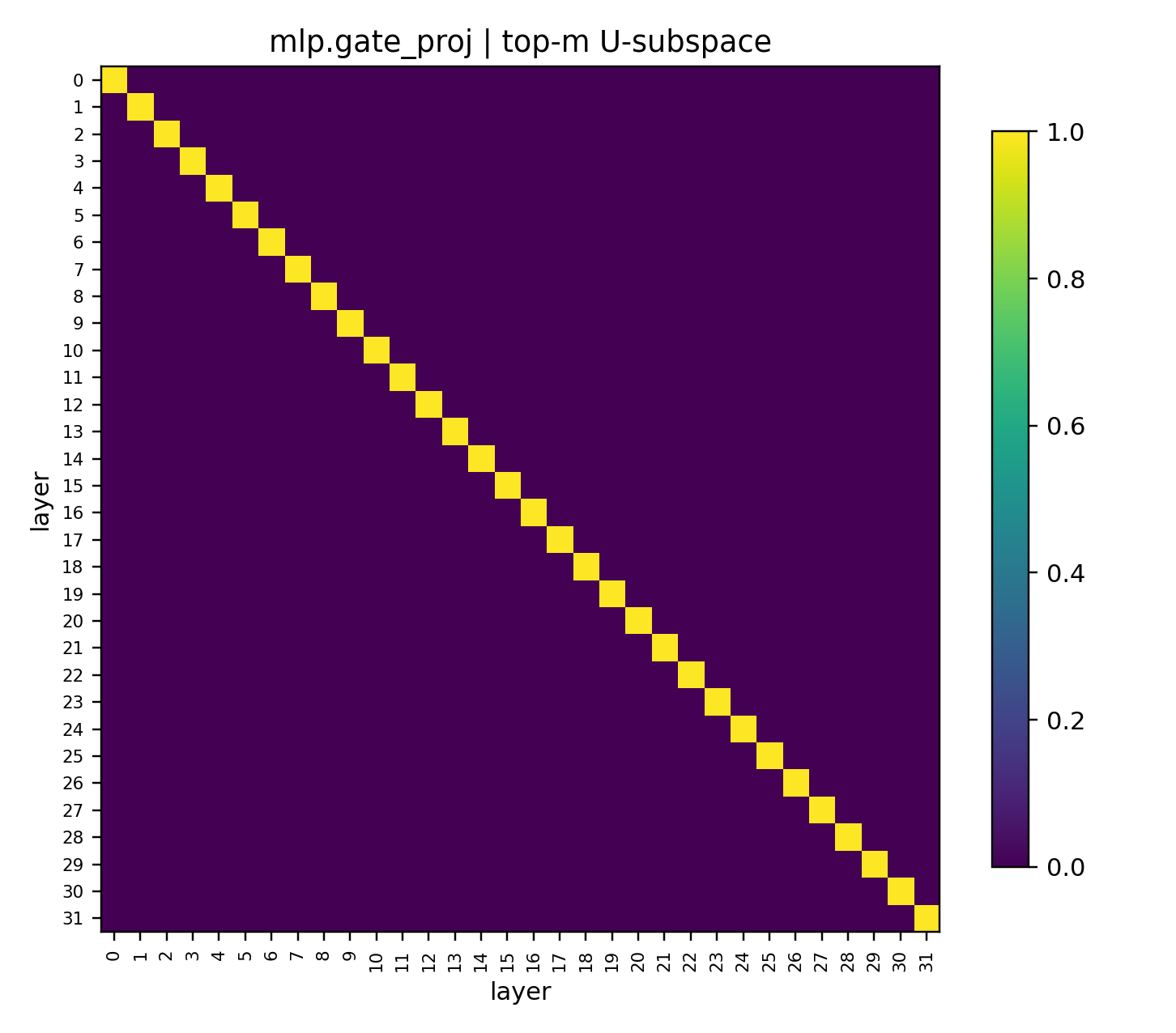} &
\includegraphics[width=0.195\textwidth]{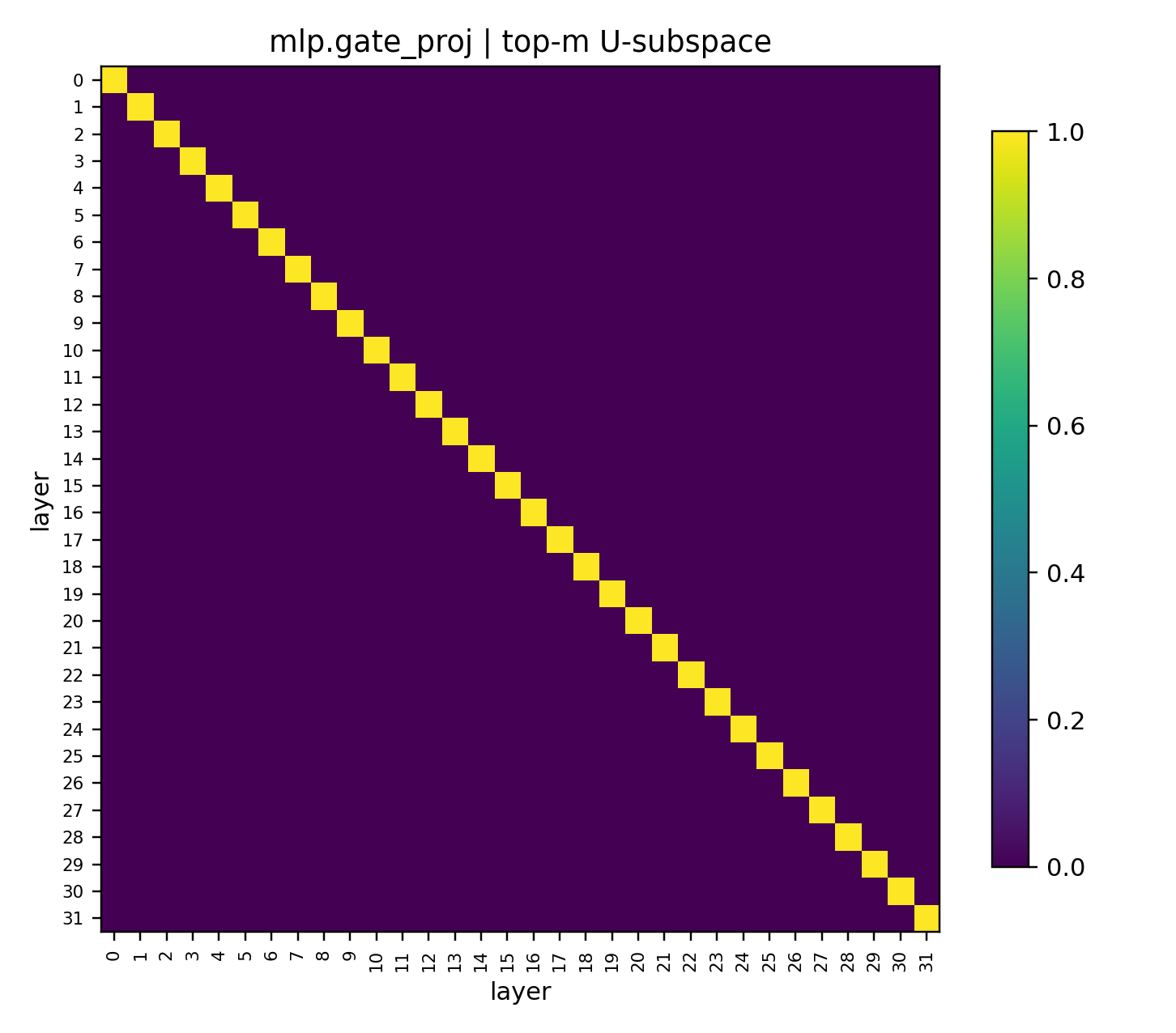} &
\includegraphics[width=0.195\textwidth]{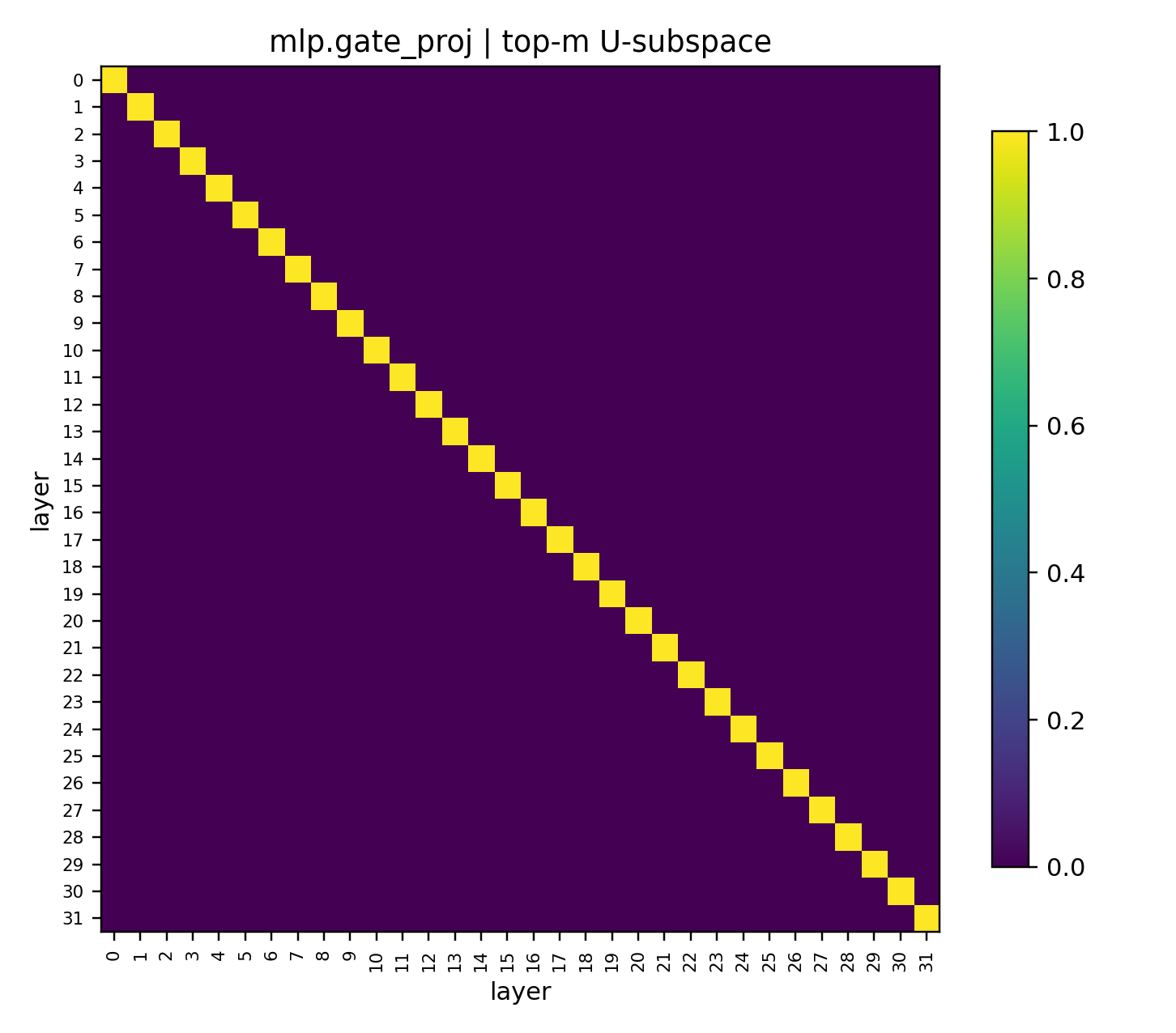} \\
\texttt{up\_proj} &
\includegraphics[width=0.195\textwidth]{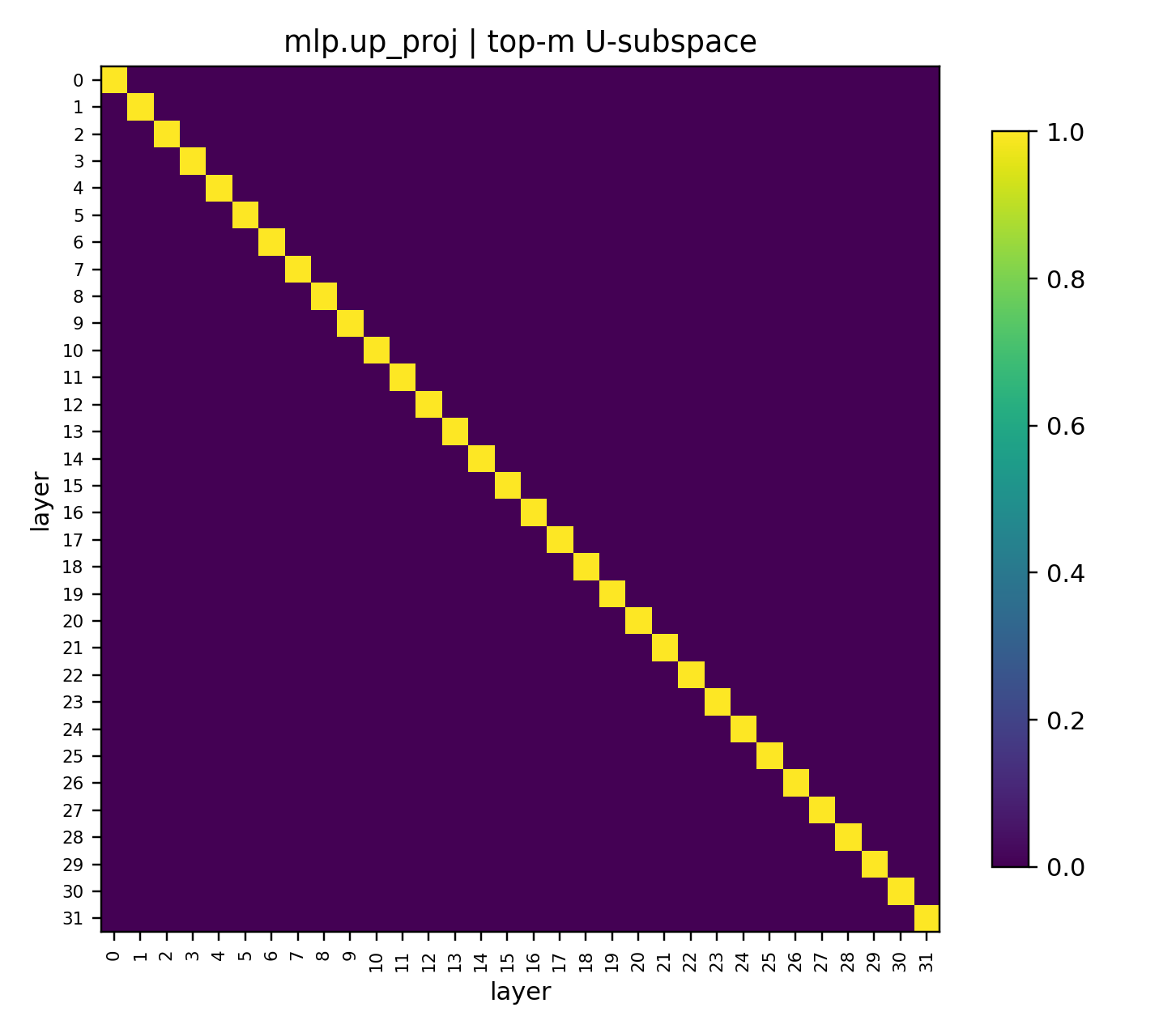} &
\includegraphics[width=0.195\textwidth]{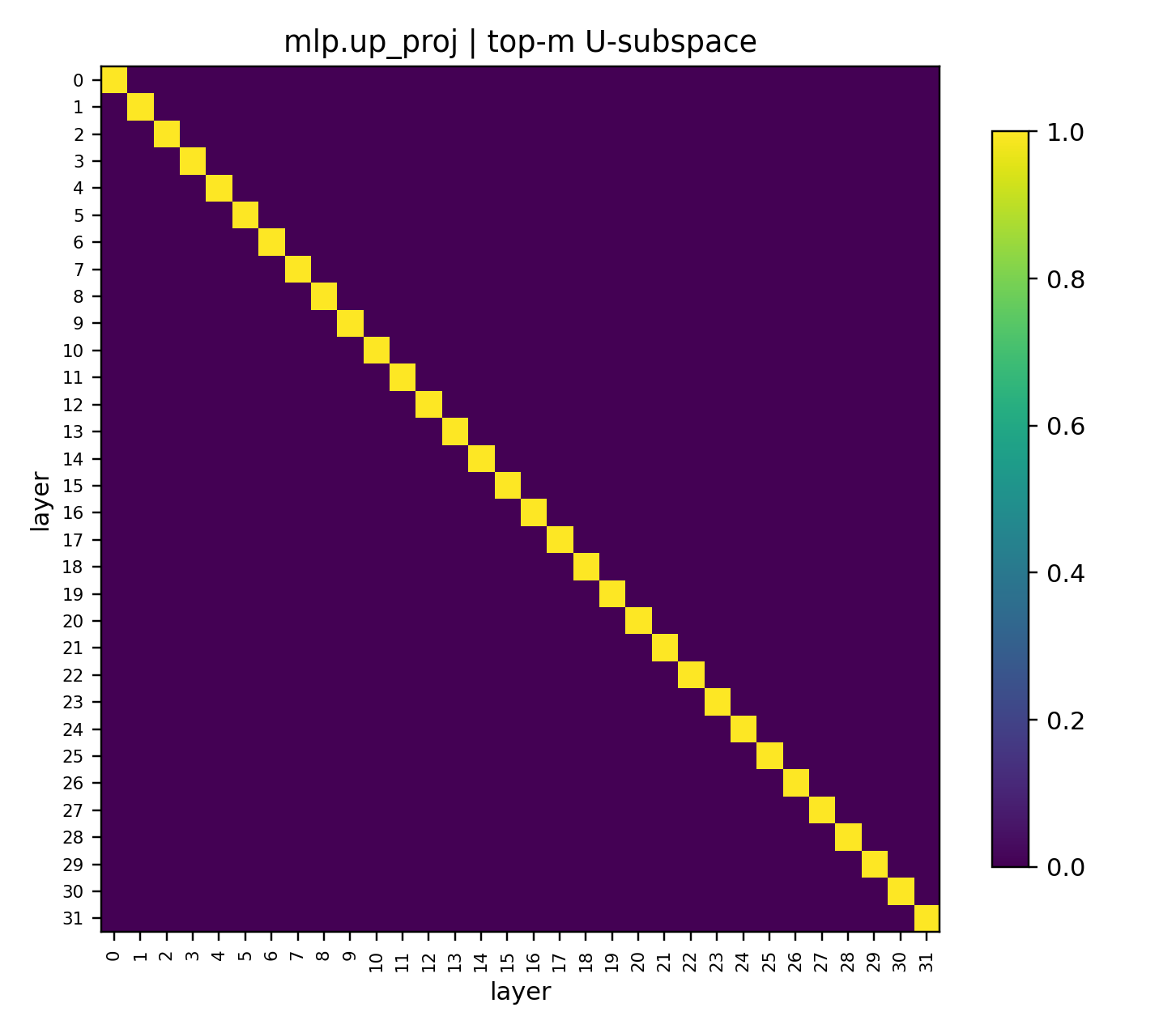} &
\includegraphics[width=0.195\textwidth]{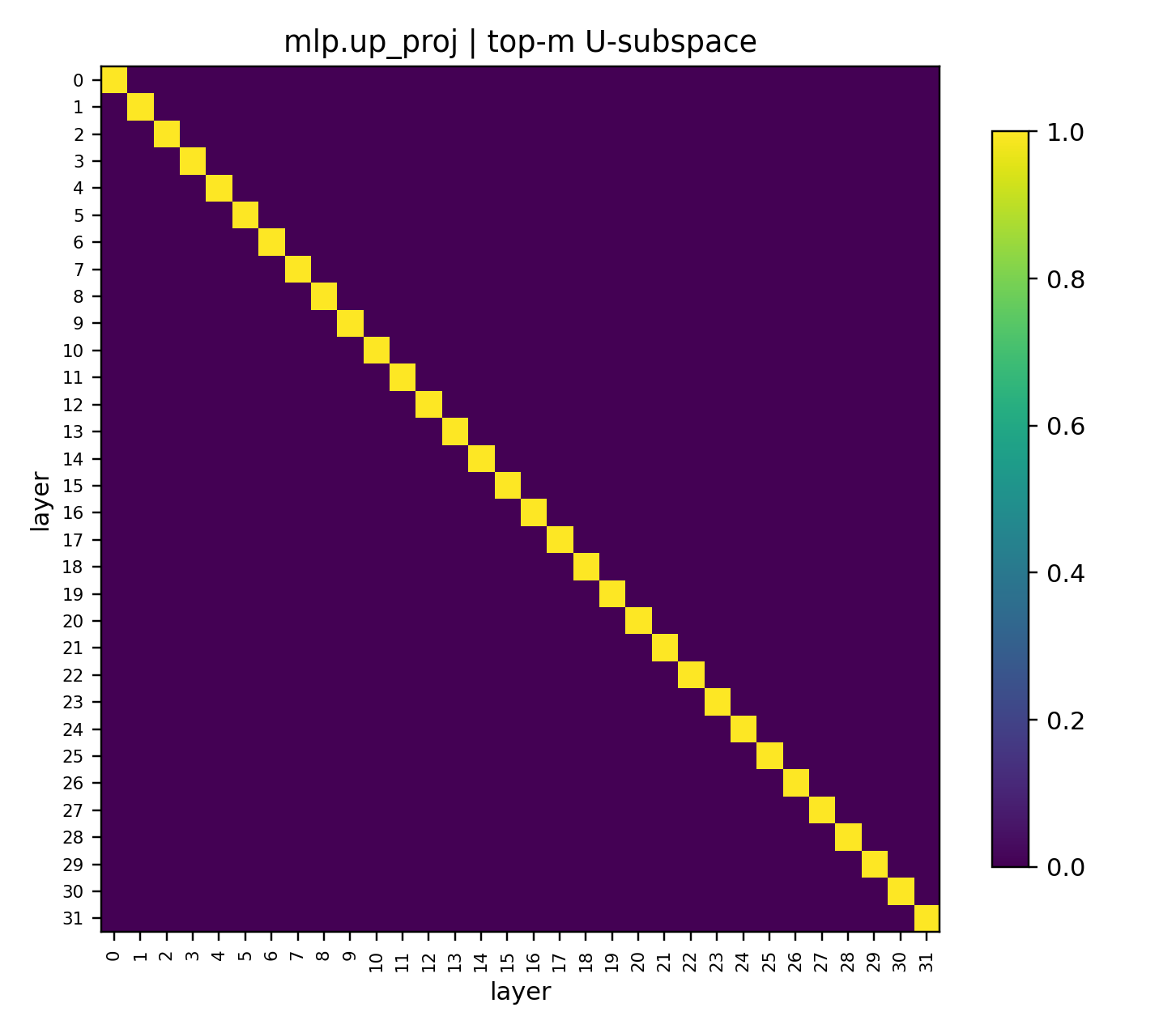} &
\includegraphics[width=0.195\textwidth]{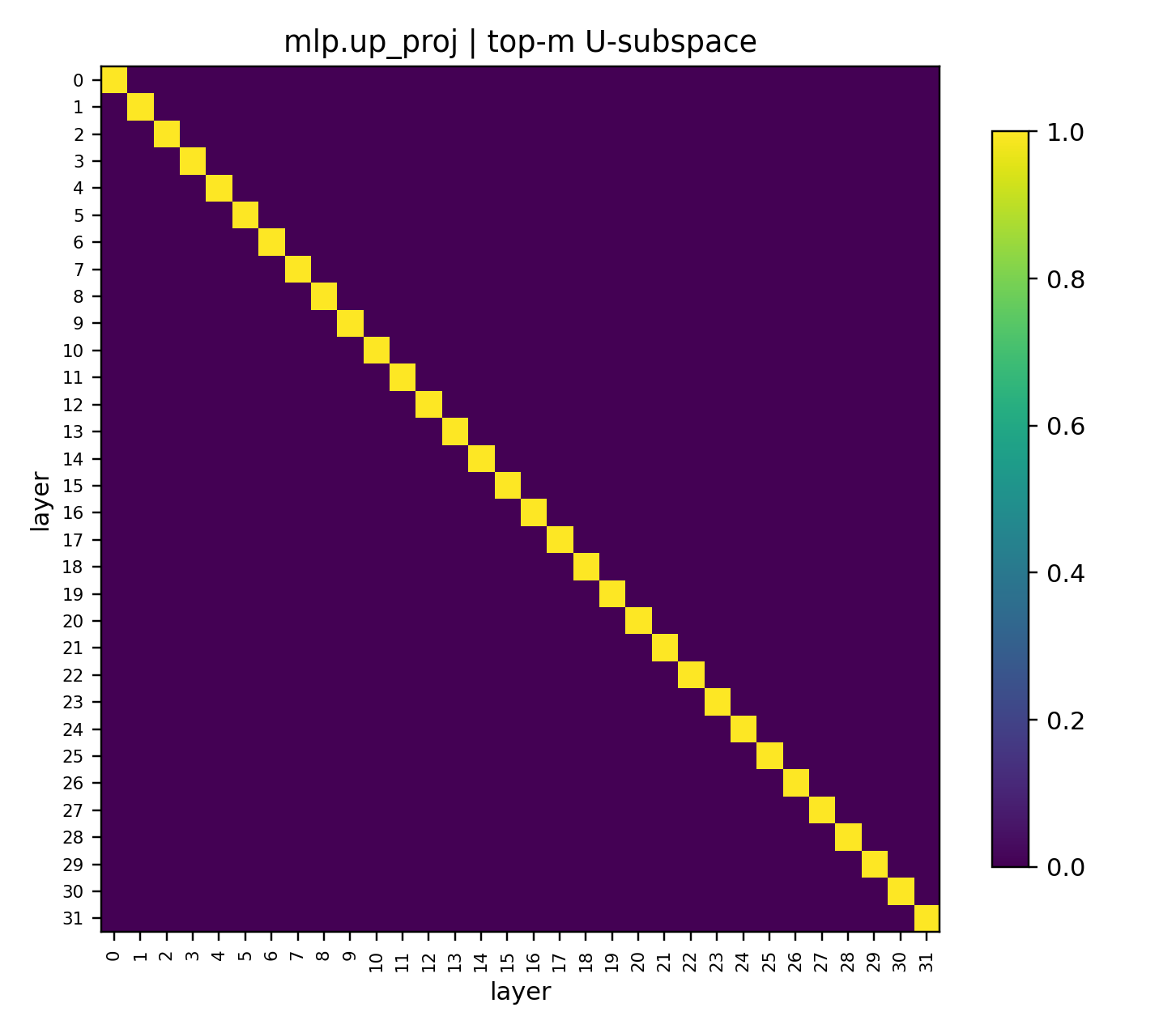} \\
\texttt{down\_proj} &
\includegraphics[width=0.195\textwidth]{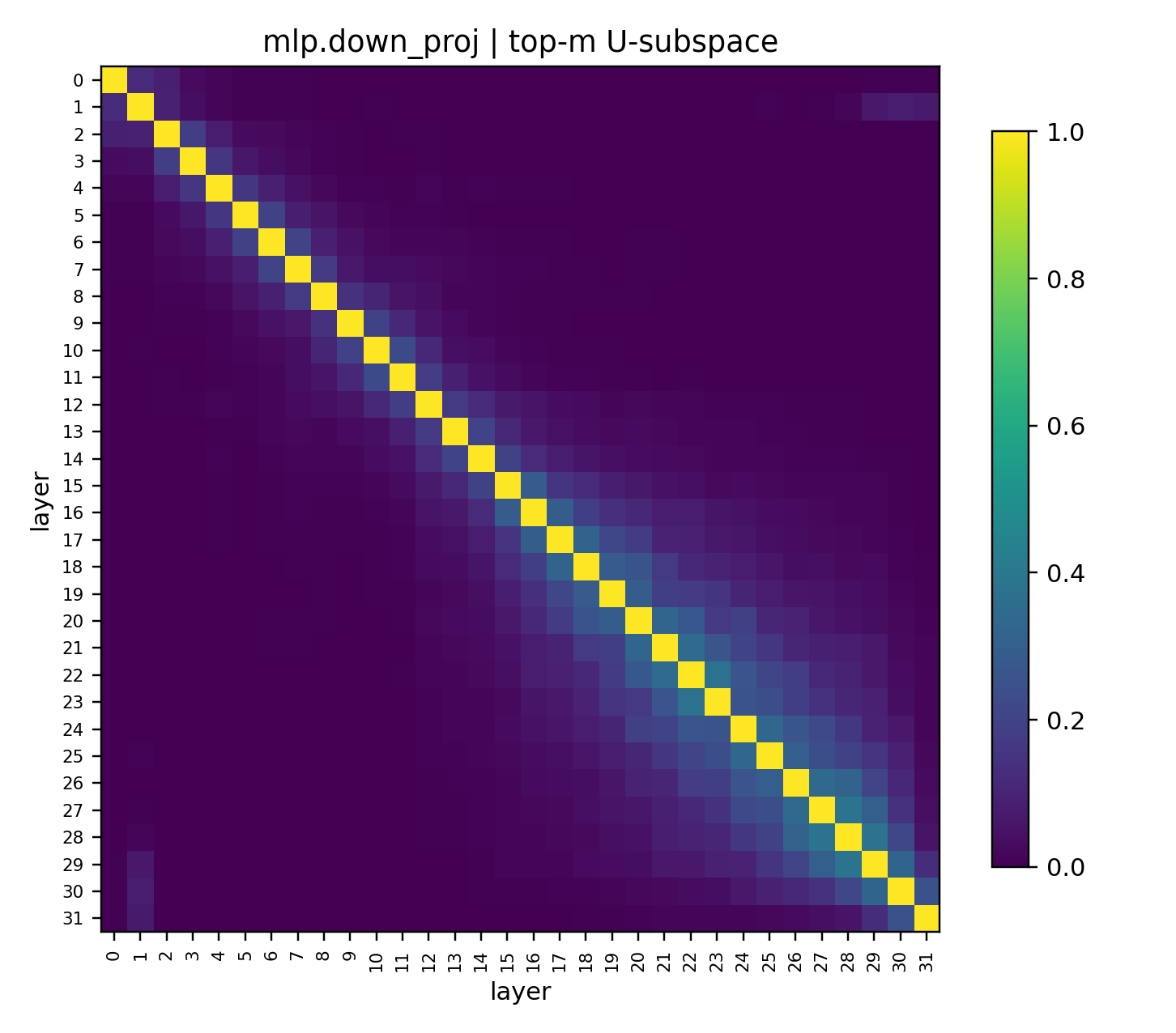} &
\includegraphics[width=0.195\textwidth]{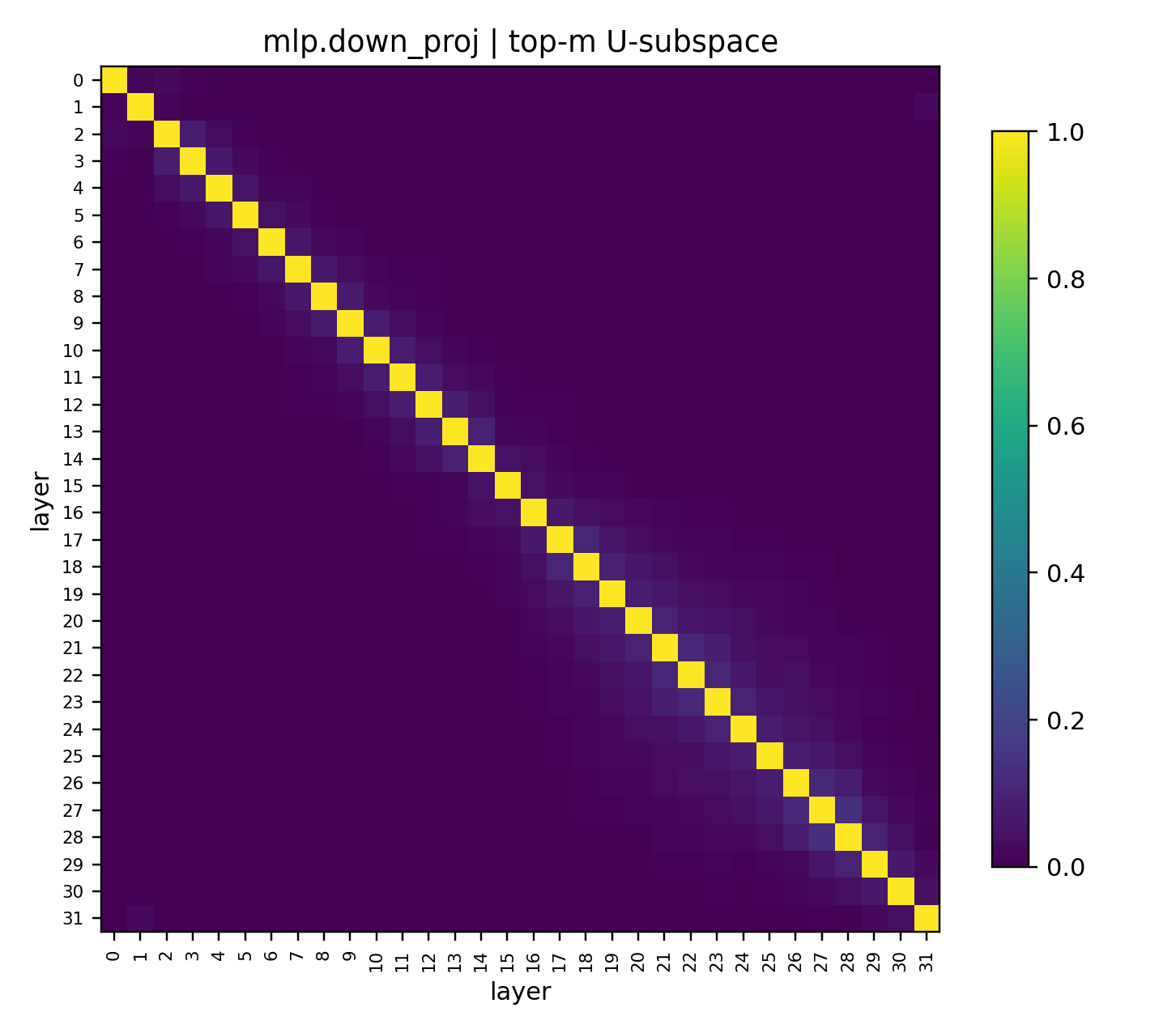} &
\includegraphics[width=0.195\textwidth]{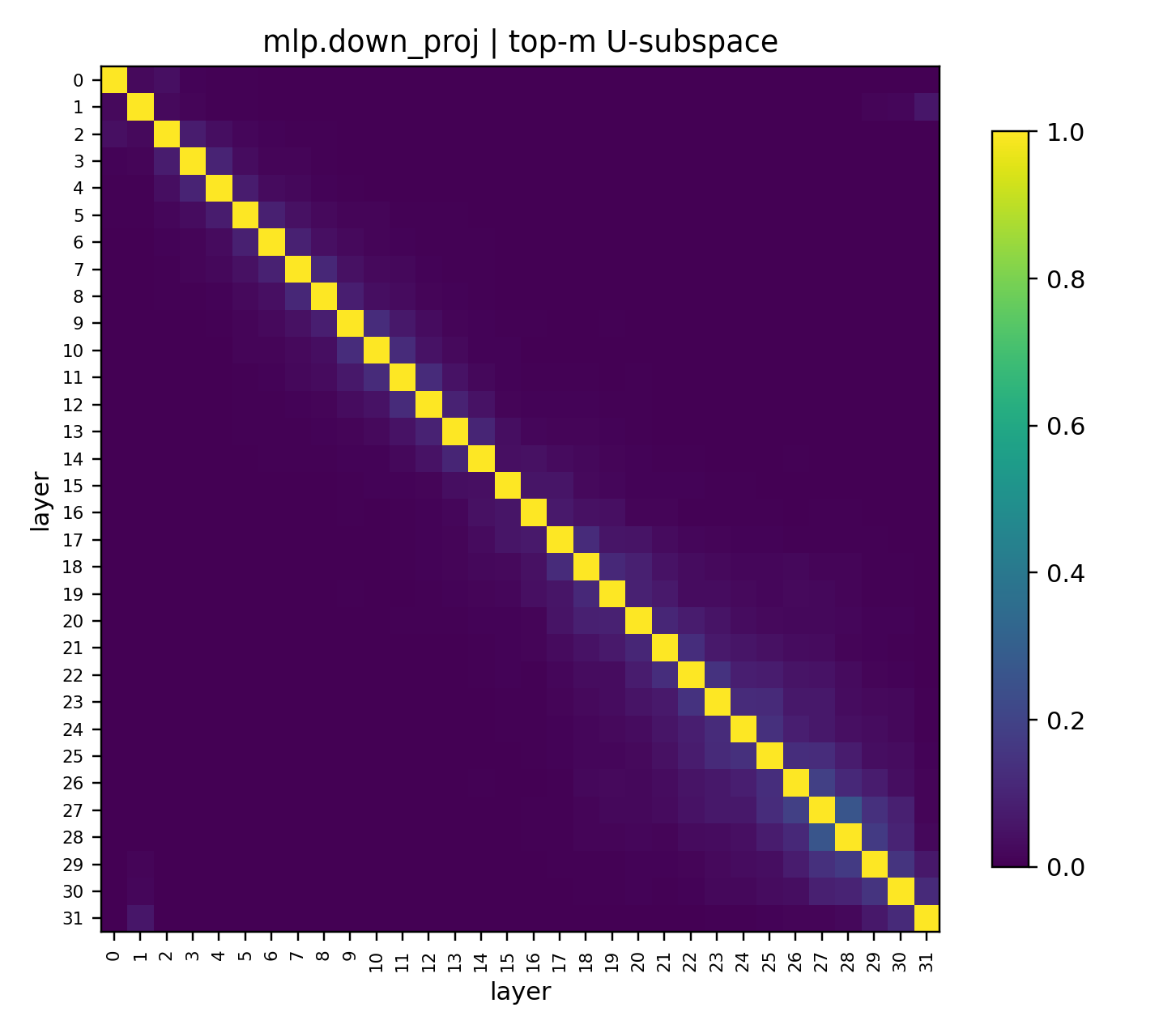} &
\includegraphics[width=0.195\textwidth]{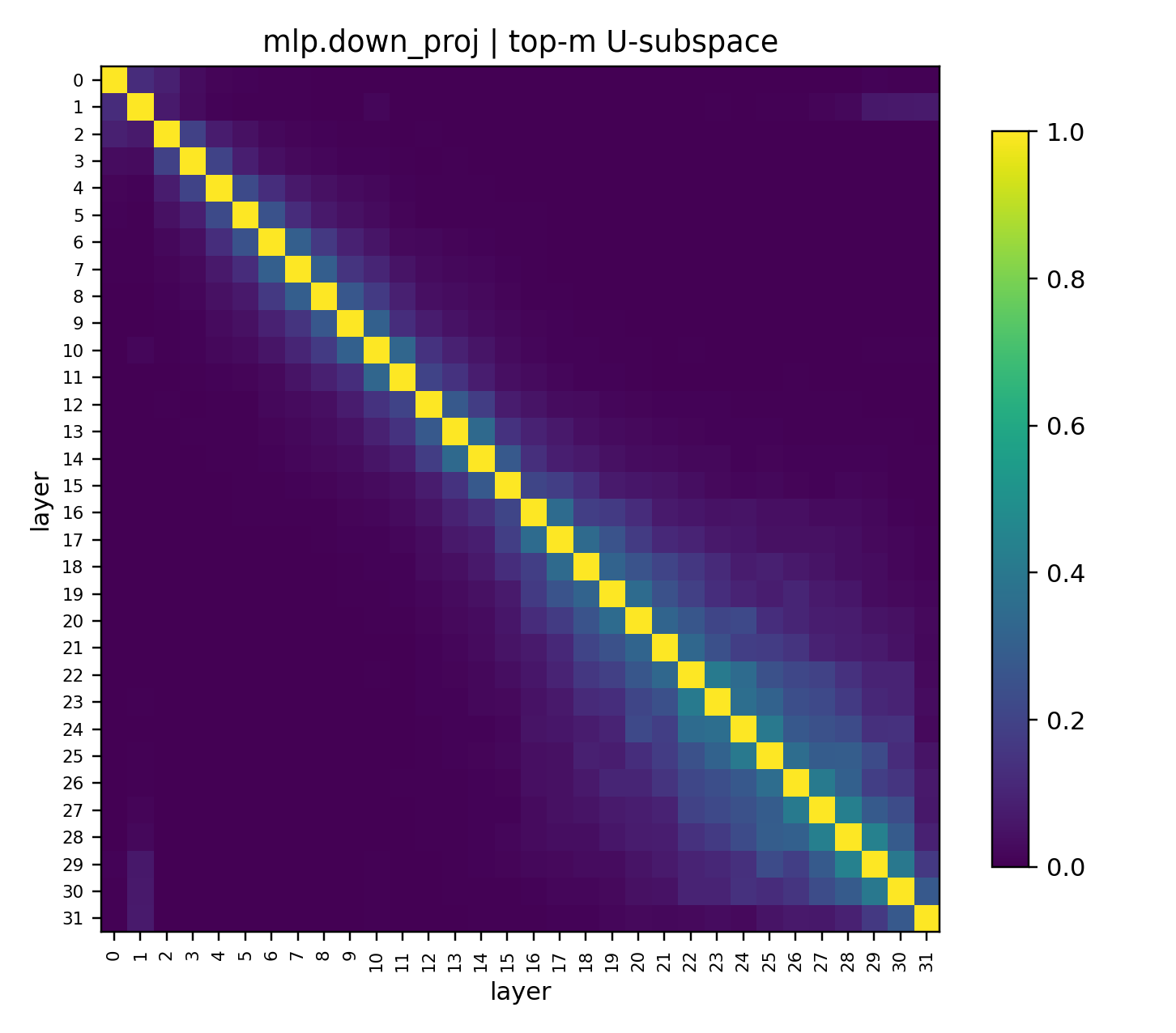} \\
\bottomrule
\end{tabular}
\caption{\textbf{Llama-3.1-8B: Top-$m$ output-subspace overlap heatmap wall ($\mathrm{Align}_U$, $m{=}4$).}
Each cell shows the inter-layer subspace-overlap heatmap for a specific (task, module).}
\label{fig:heatwall_llama_subU}
\end{figure*}

\clearpage
\begin{figure*}[p]
\centering
\small
\setlength{\tabcolsep}{2pt}
\renewcommand{\arraystretch}{1.0}
\begin{tabular}{lcccc}
\toprule
\textbf{Module} &
\textbf{Math} & \textbf{Code} & \textbf{Instruction} & \textbf{Commonsense} \\
\midrule
\texttt{q\_proj} &
\includegraphics[width=0.195\textwidth]{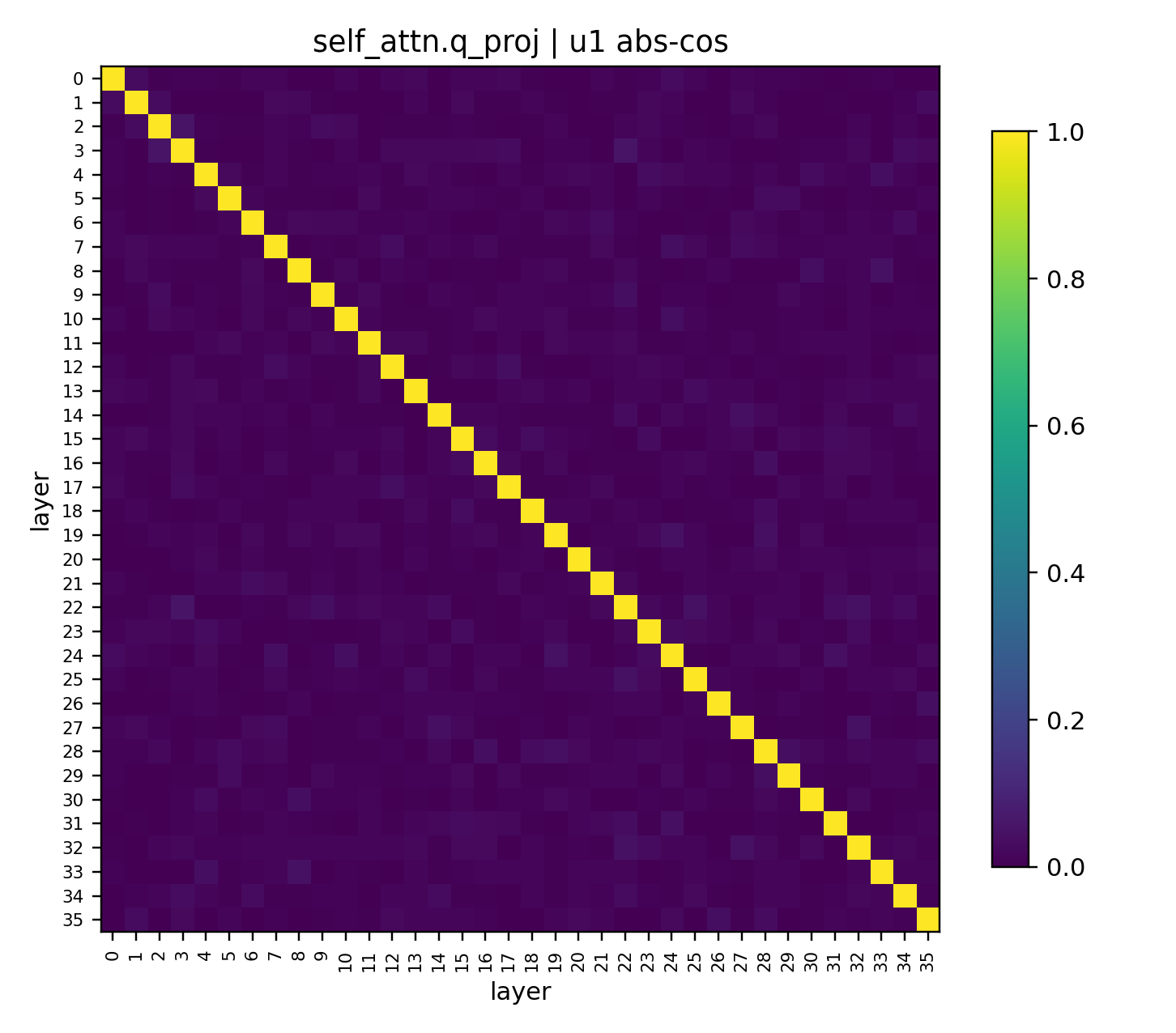} &
\includegraphics[width=0.195\textwidth]{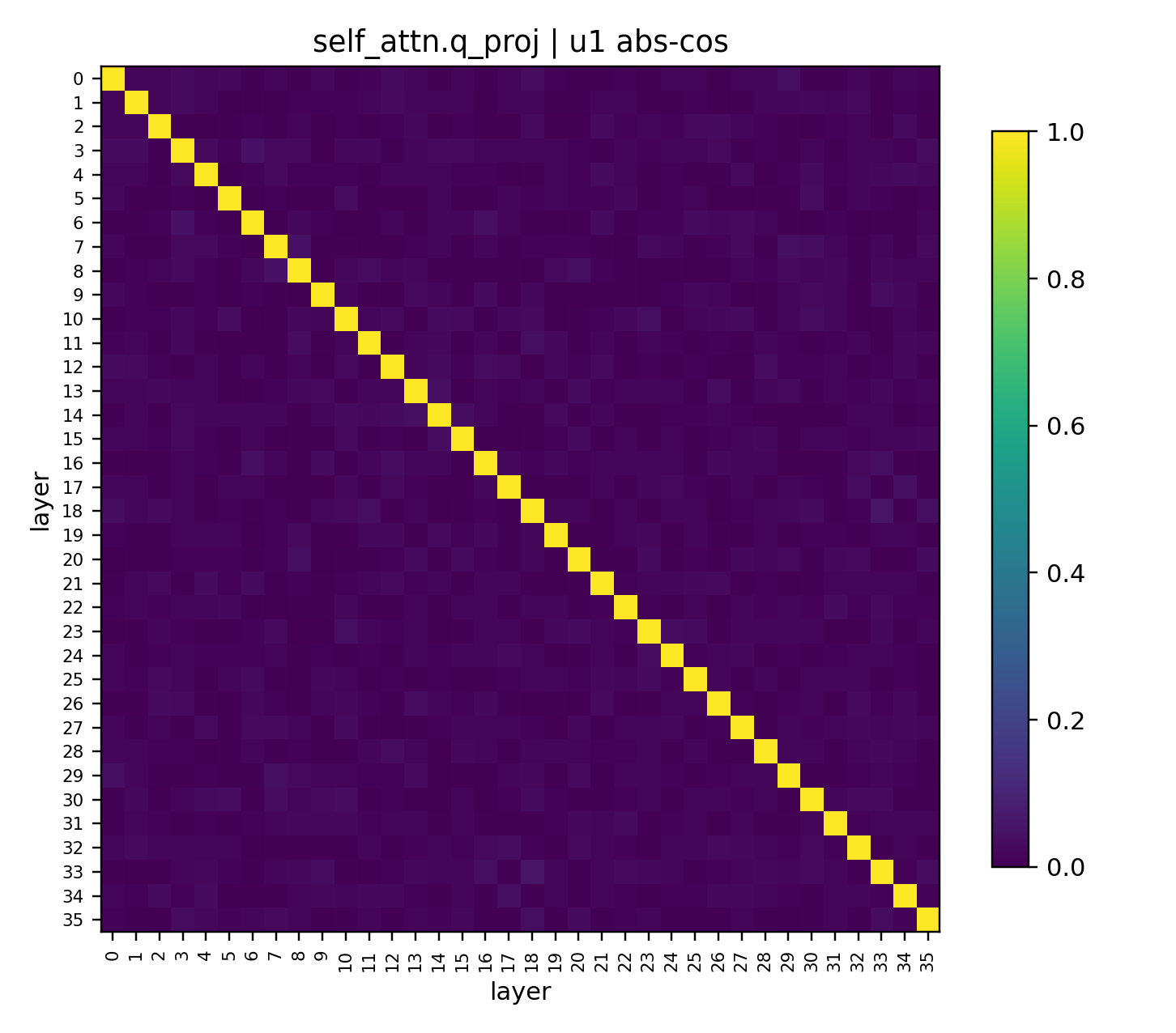} &
\includegraphics[width=0.195\textwidth]{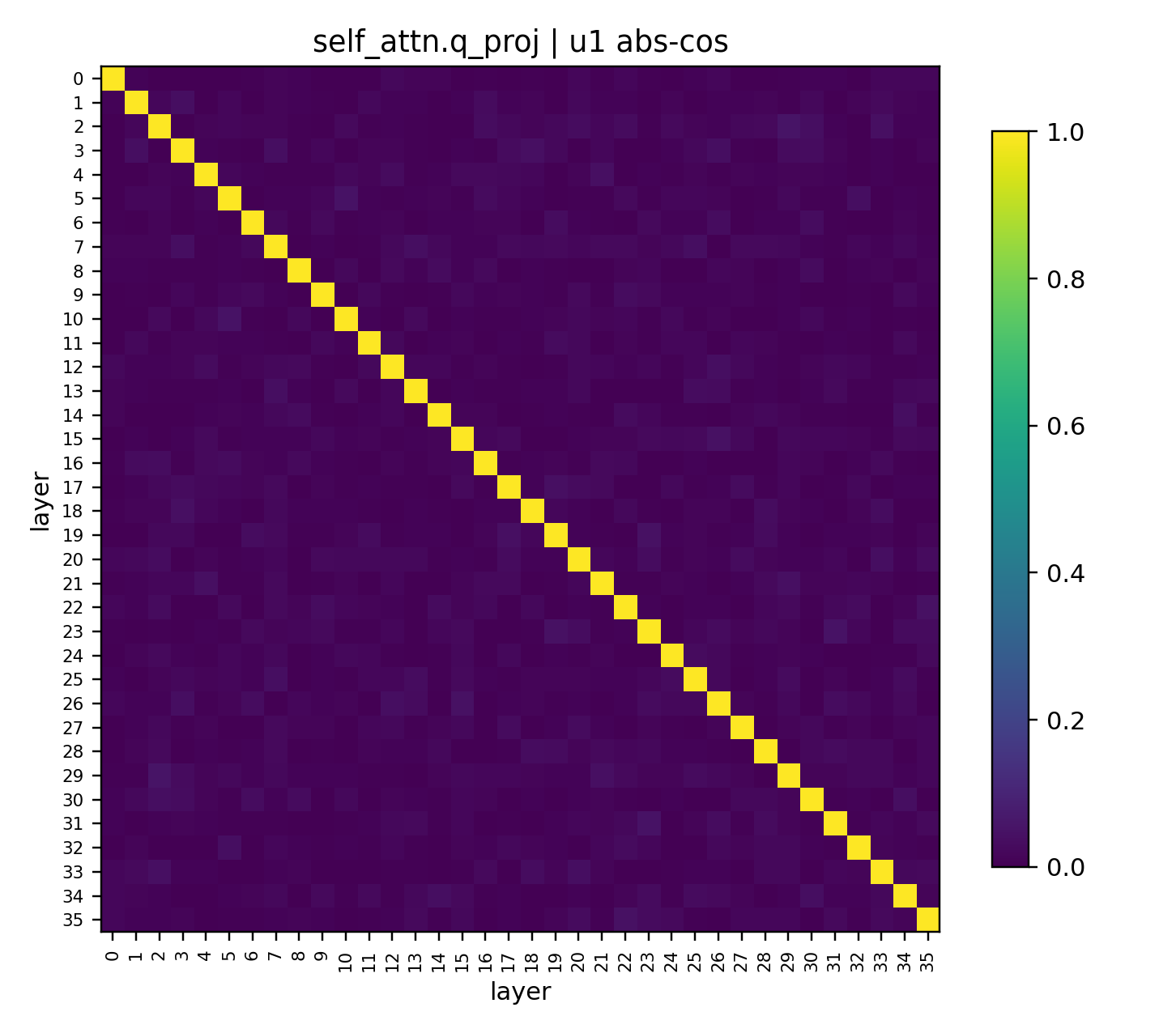} &
\includegraphics[width=0.195\textwidth]{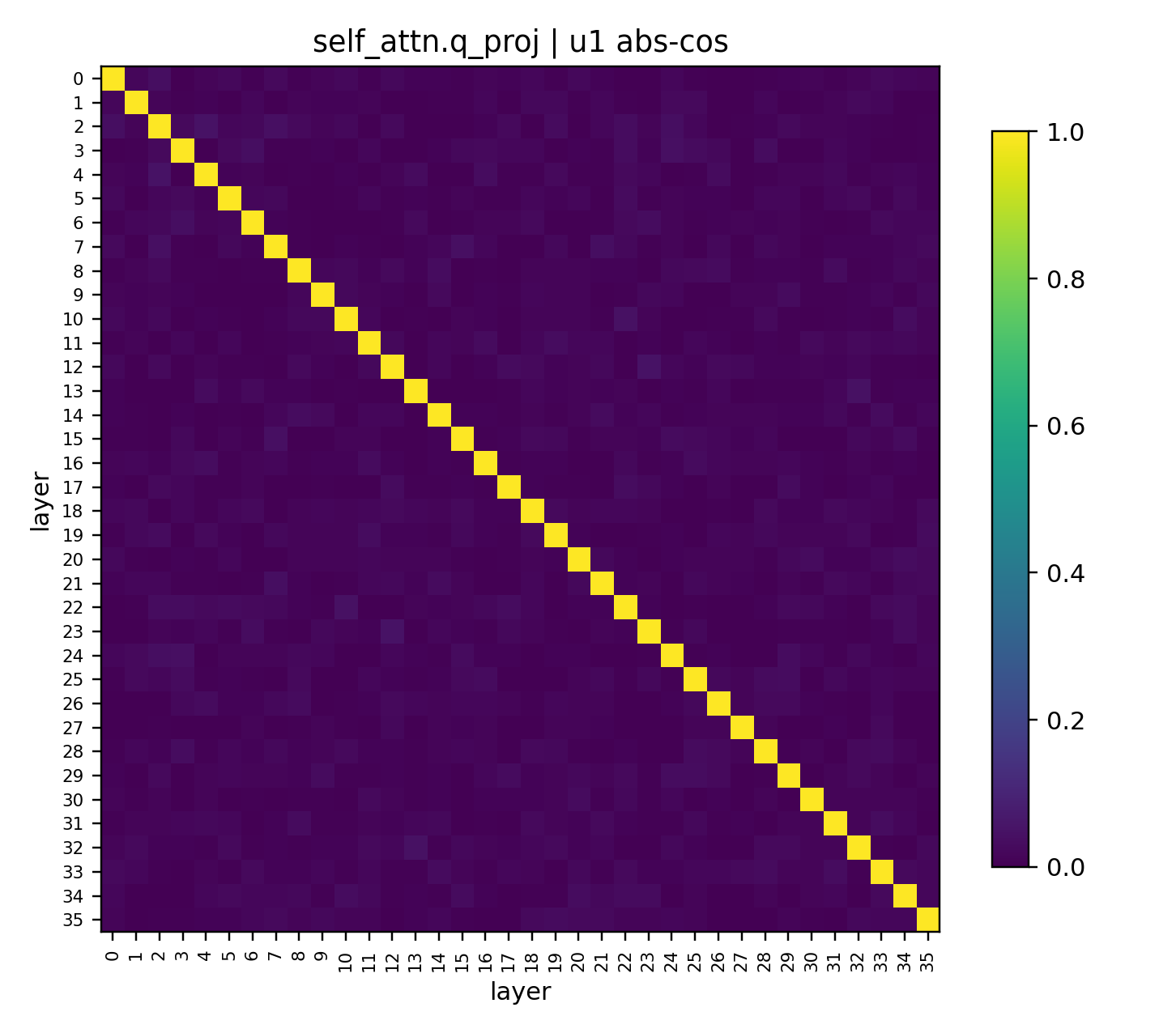} \\
\texttt{k\_proj} &
\includegraphics[width=0.195\textwidth]{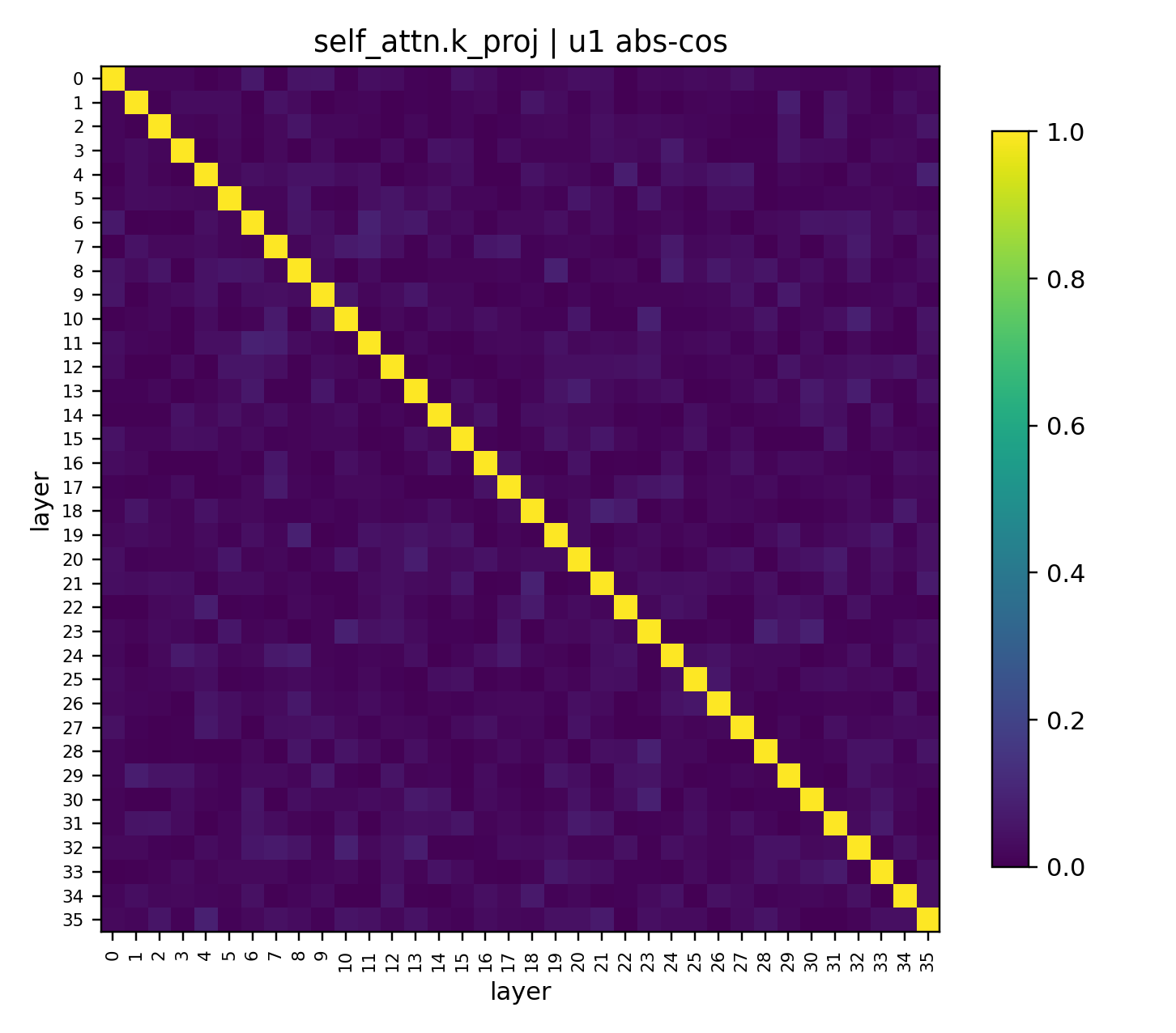} &
\includegraphics[width=0.195\textwidth]{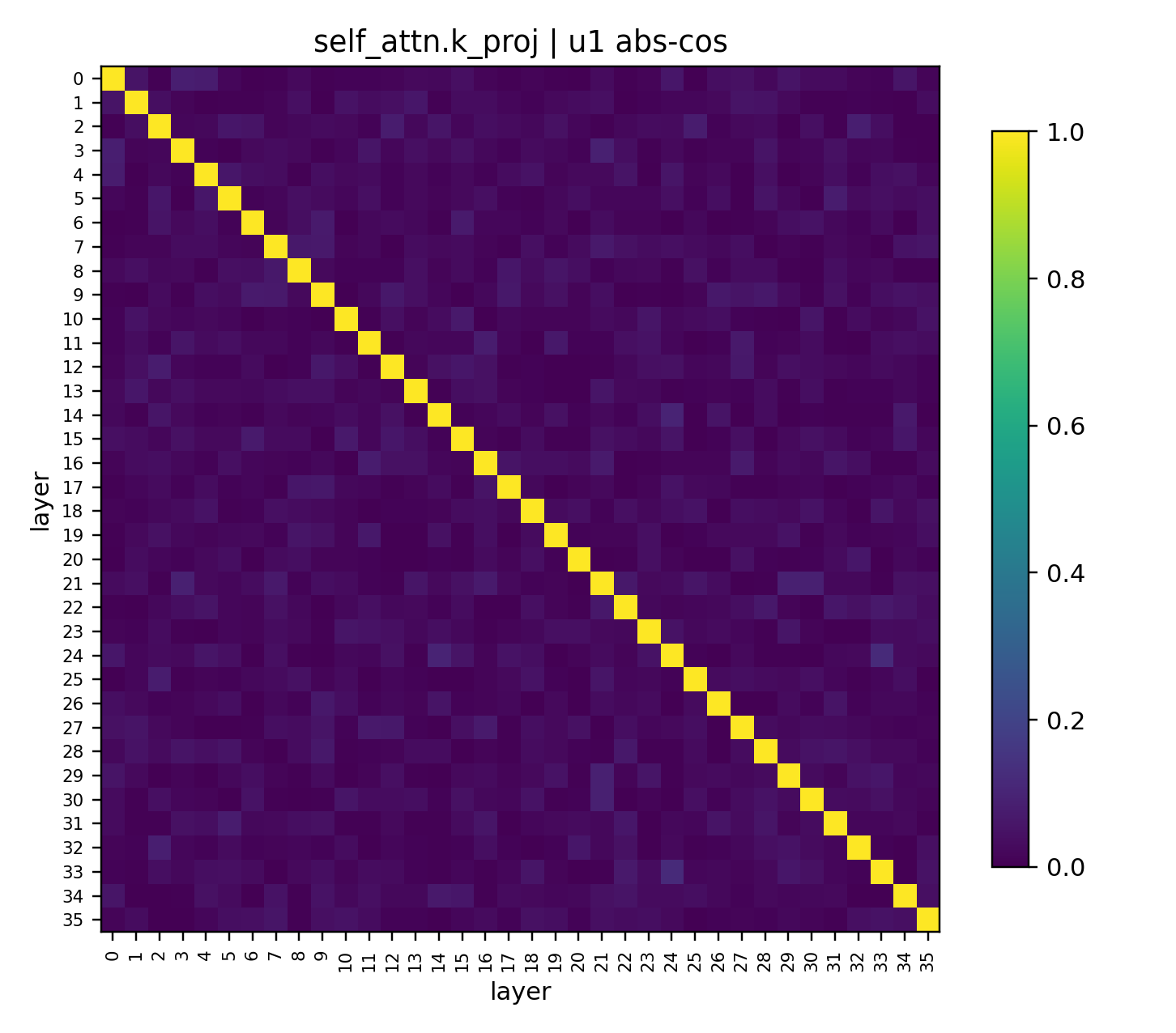} &
\includegraphics[width=0.195\textwidth]{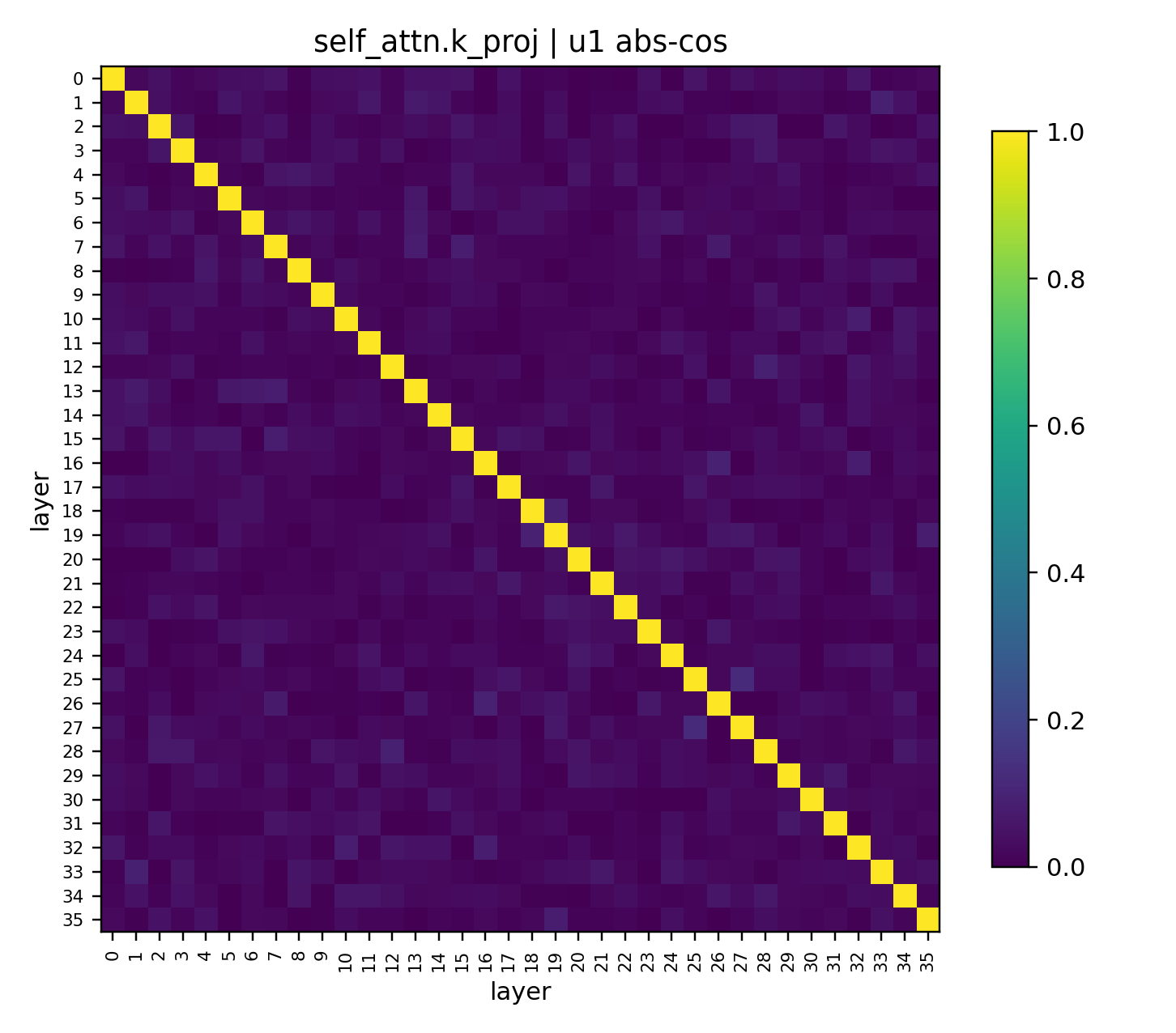} &
\includegraphics[width=0.195\textwidth]{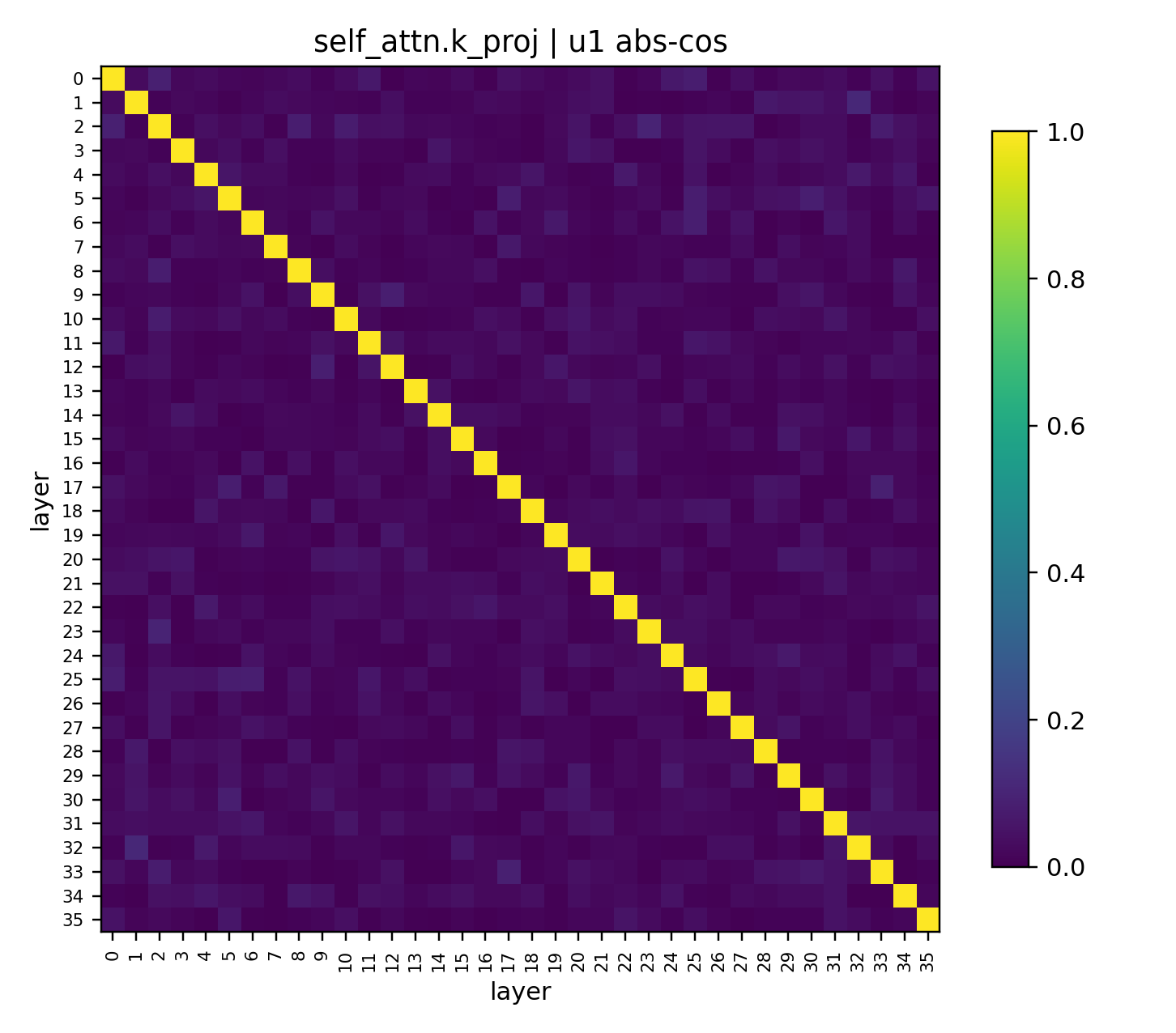} \\
\texttt{v\_proj} &
\includegraphics[width=0.195\textwidth]{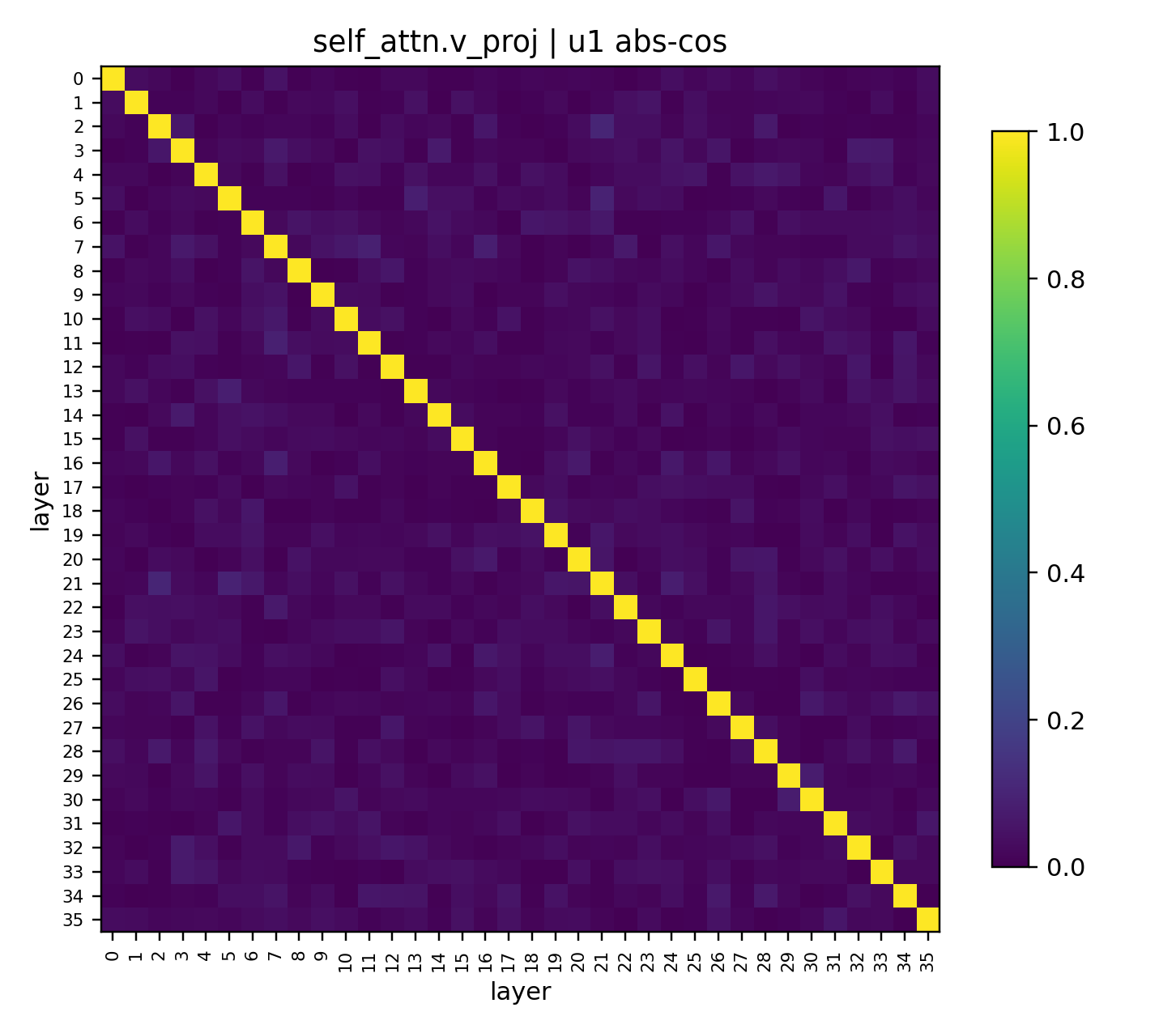} &
\includegraphics[width=0.195\textwidth]{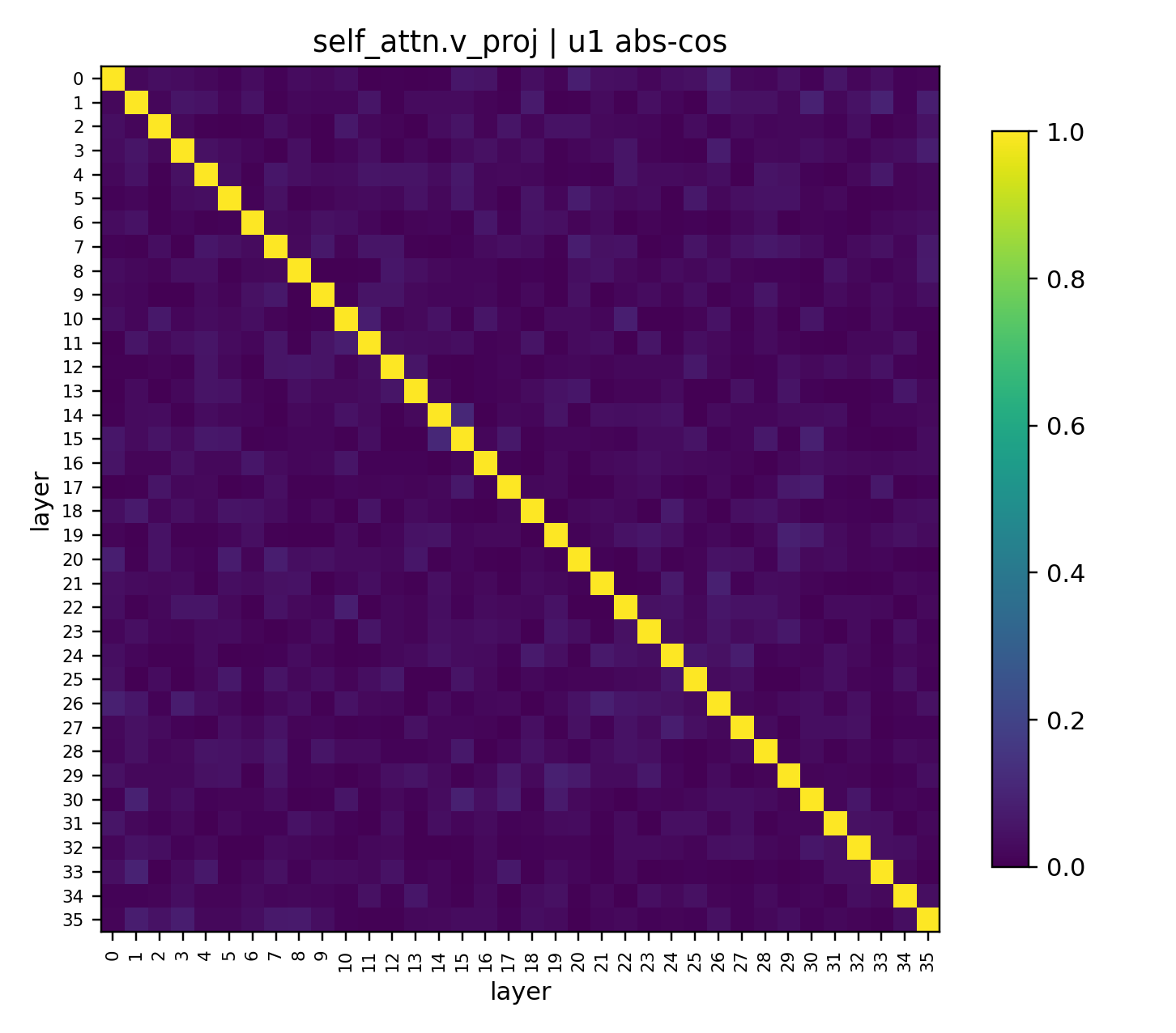} &
\includegraphics[width=0.195\textwidth]{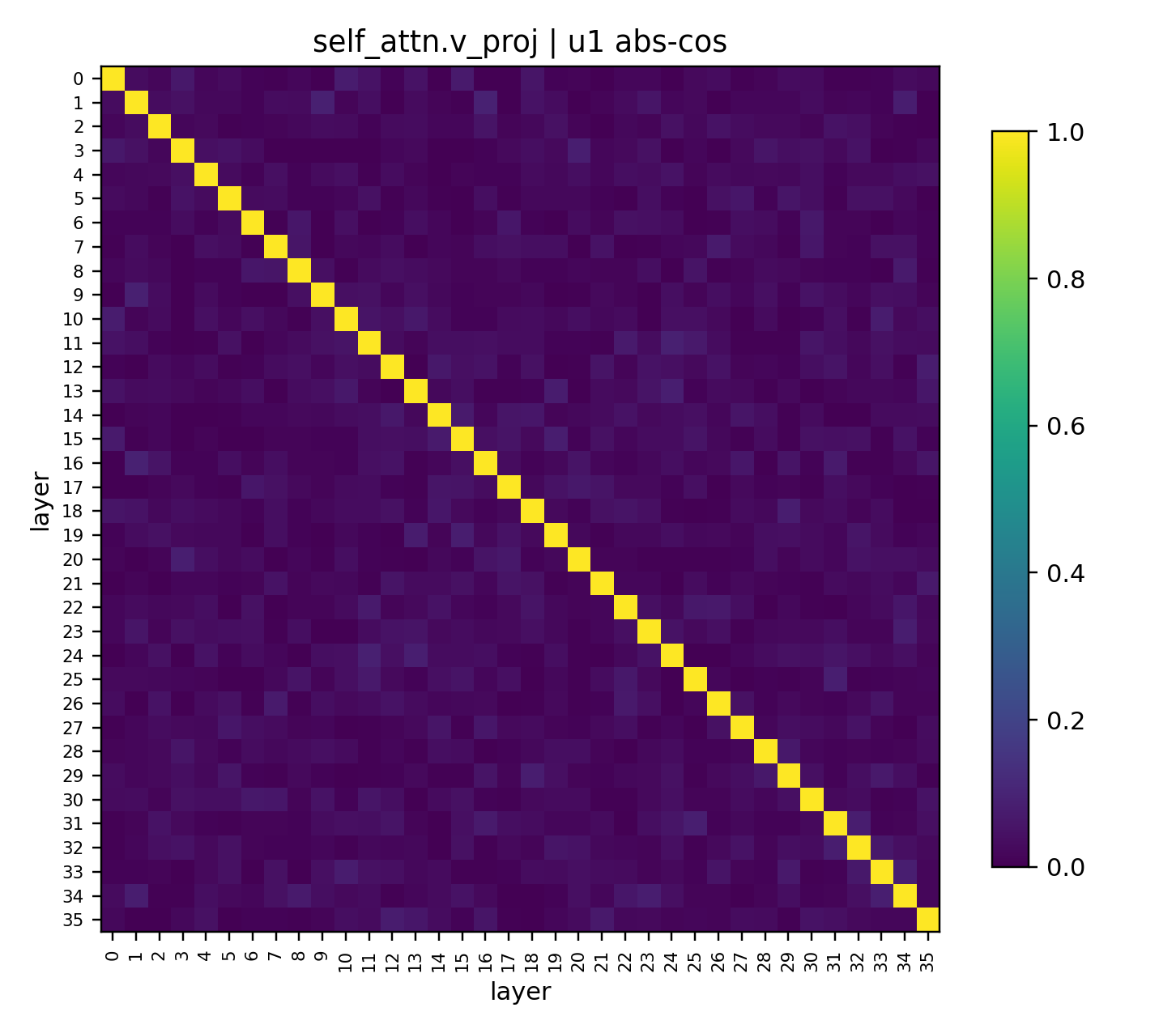} &
\includegraphics[width=0.195\textwidth]{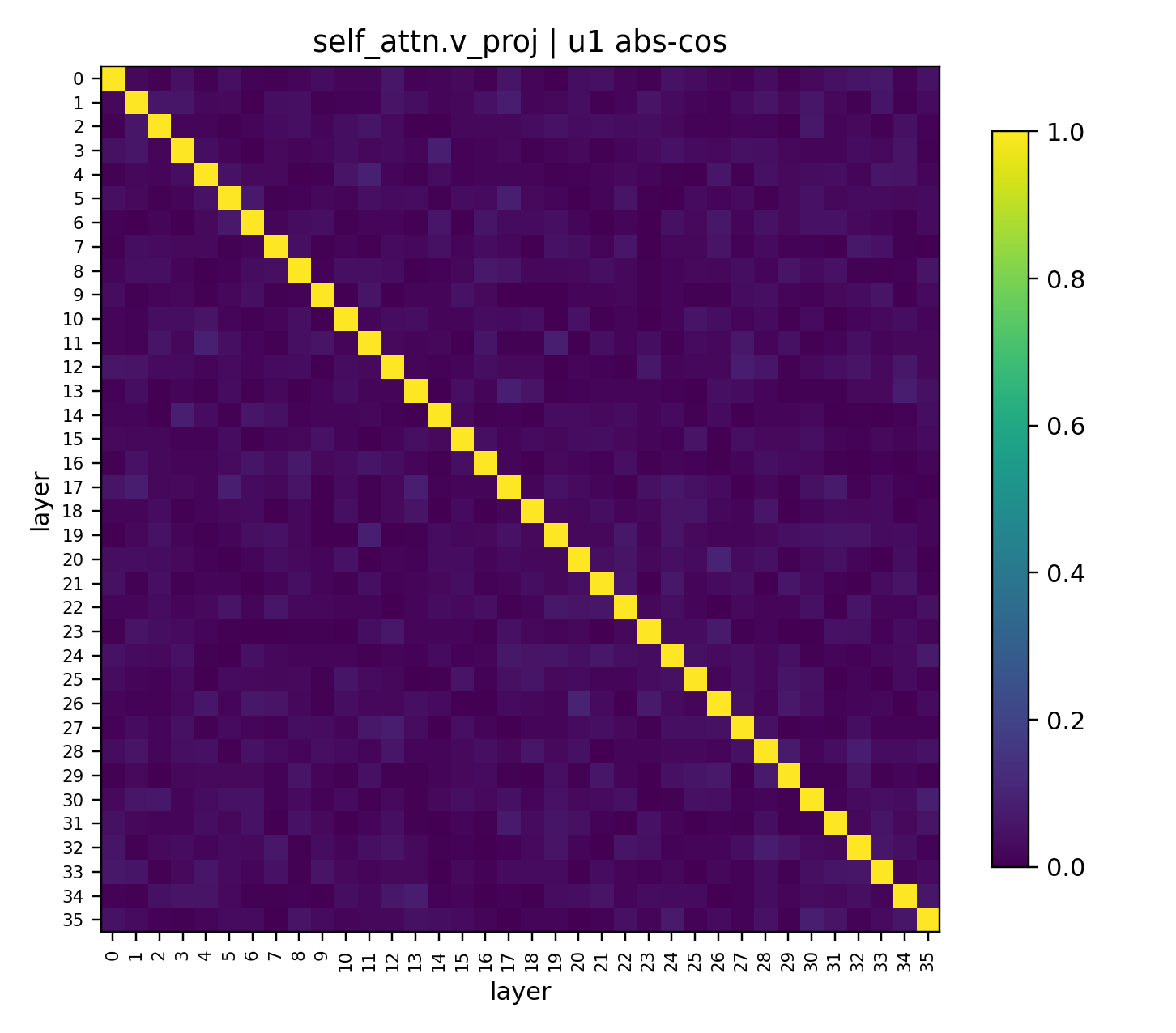} \\
\texttt{o\_proj} &
\includegraphics[width=0.195\textwidth]{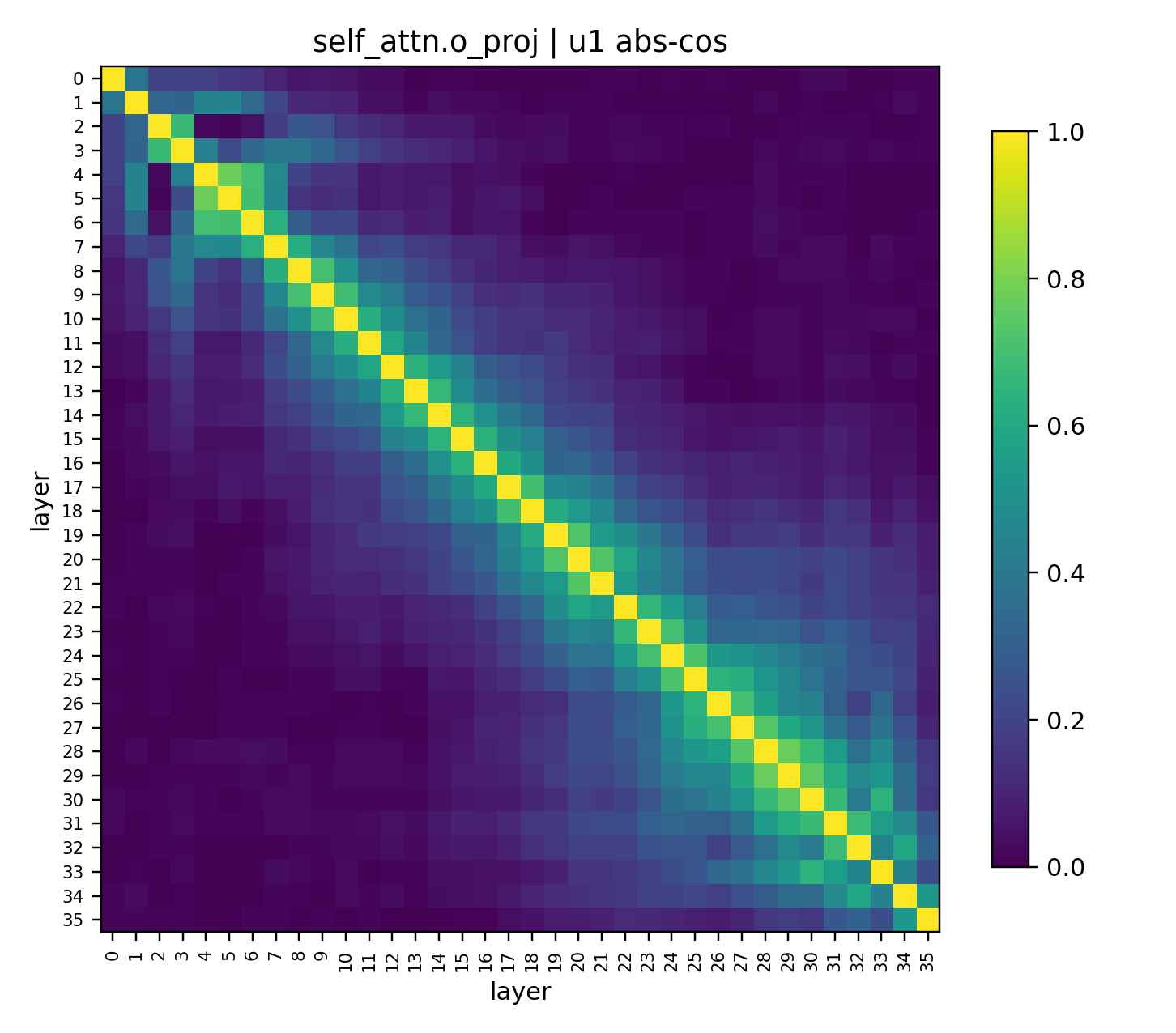} &
\includegraphics[width=0.195\textwidth]{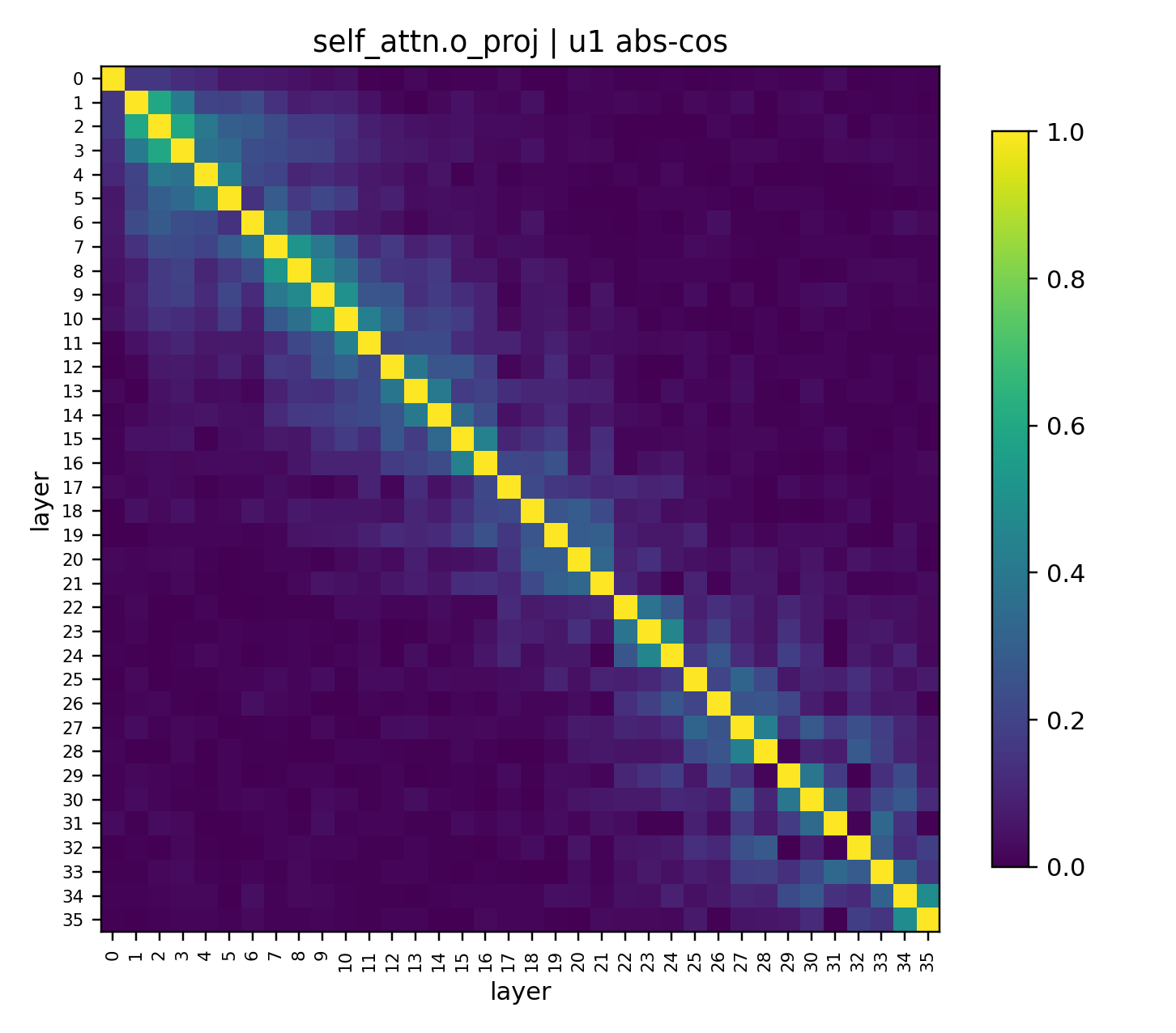} &
\includegraphics[width=0.195\textwidth]{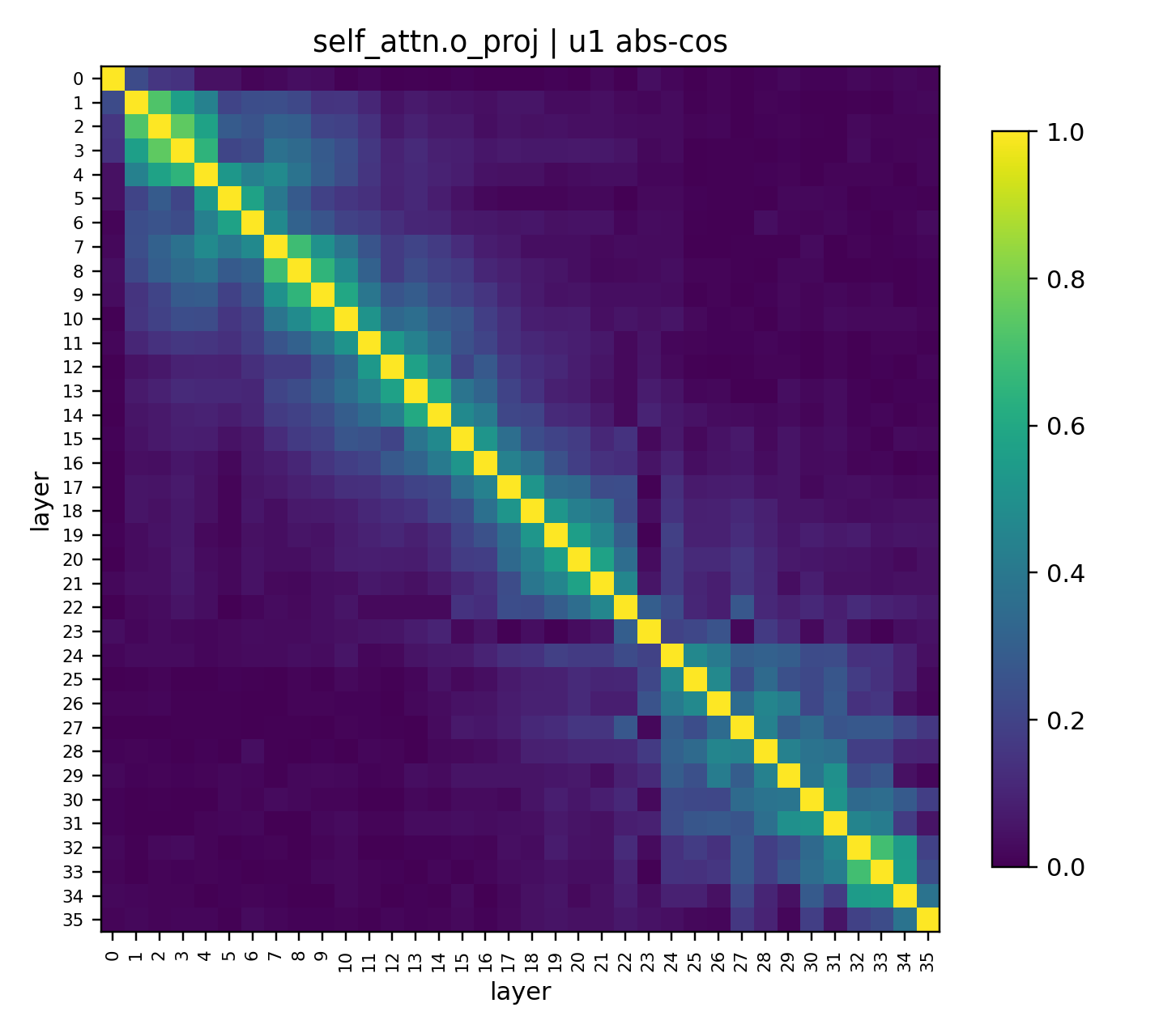} &
\includegraphics[width=0.195\textwidth]{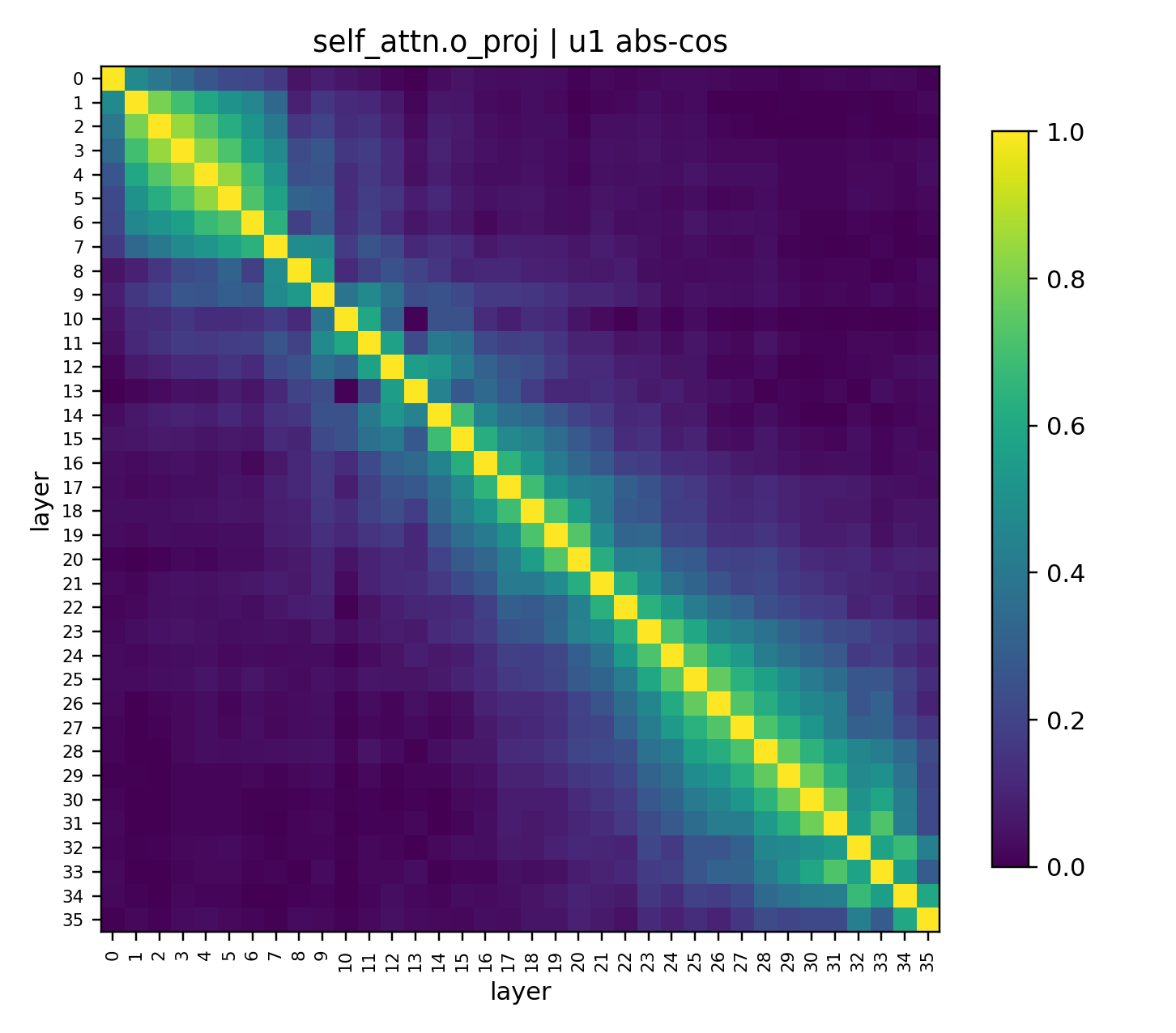} \\
\texttt{gate\_proj} &
\includegraphics[width=0.195\textwidth]{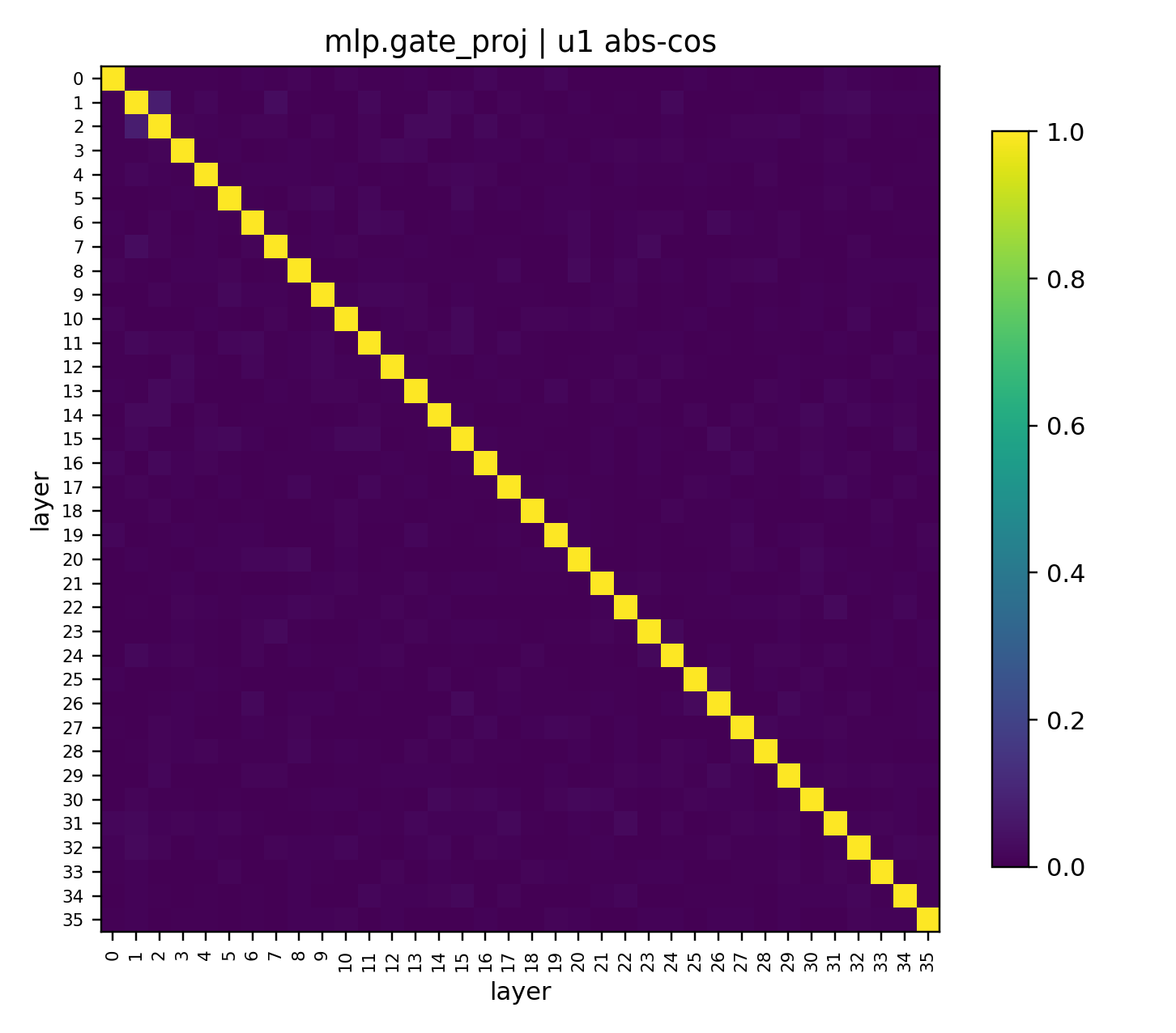} &
\includegraphics[width=0.195\textwidth]{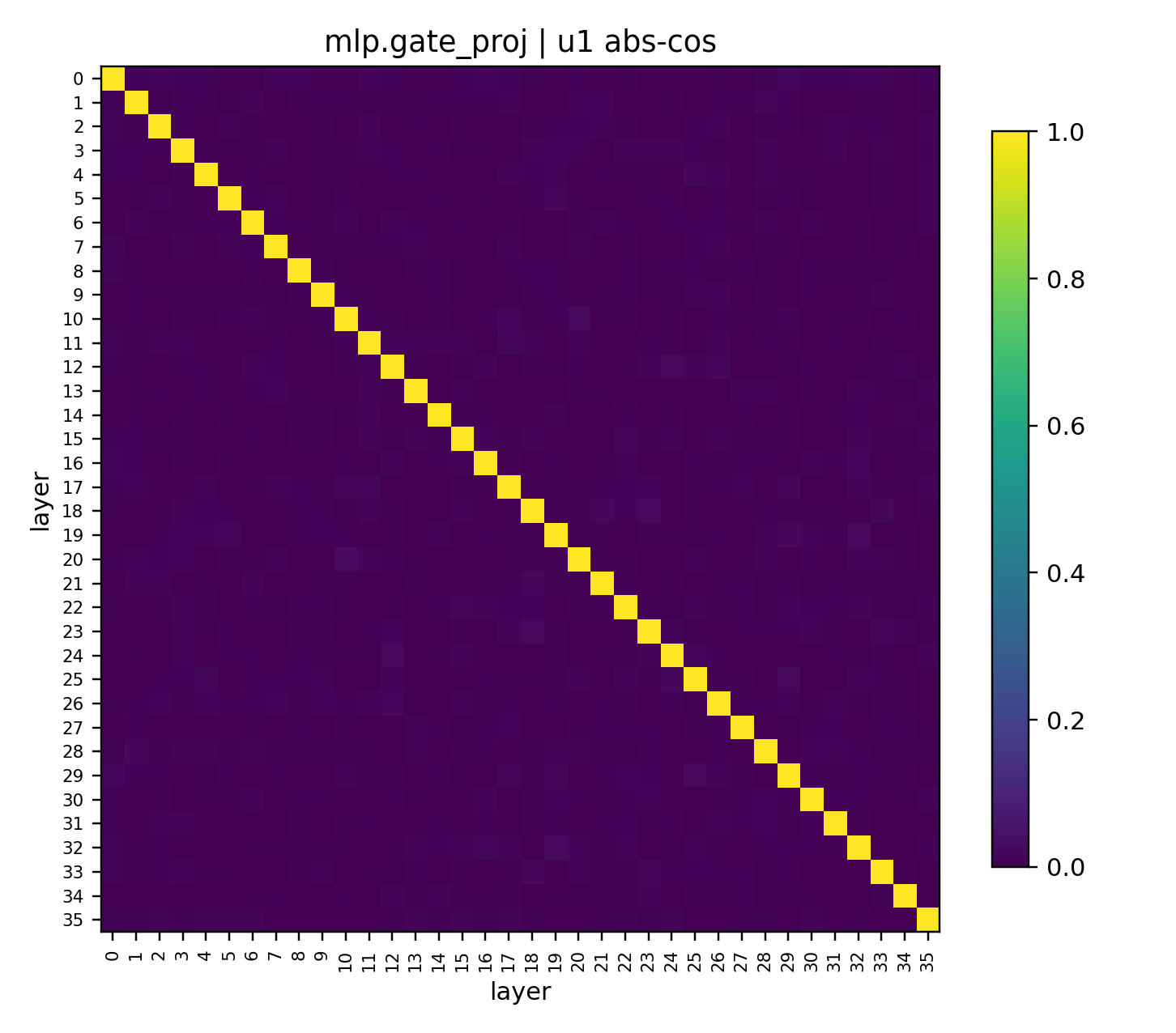} &
\includegraphics[width=0.195\textwidth]{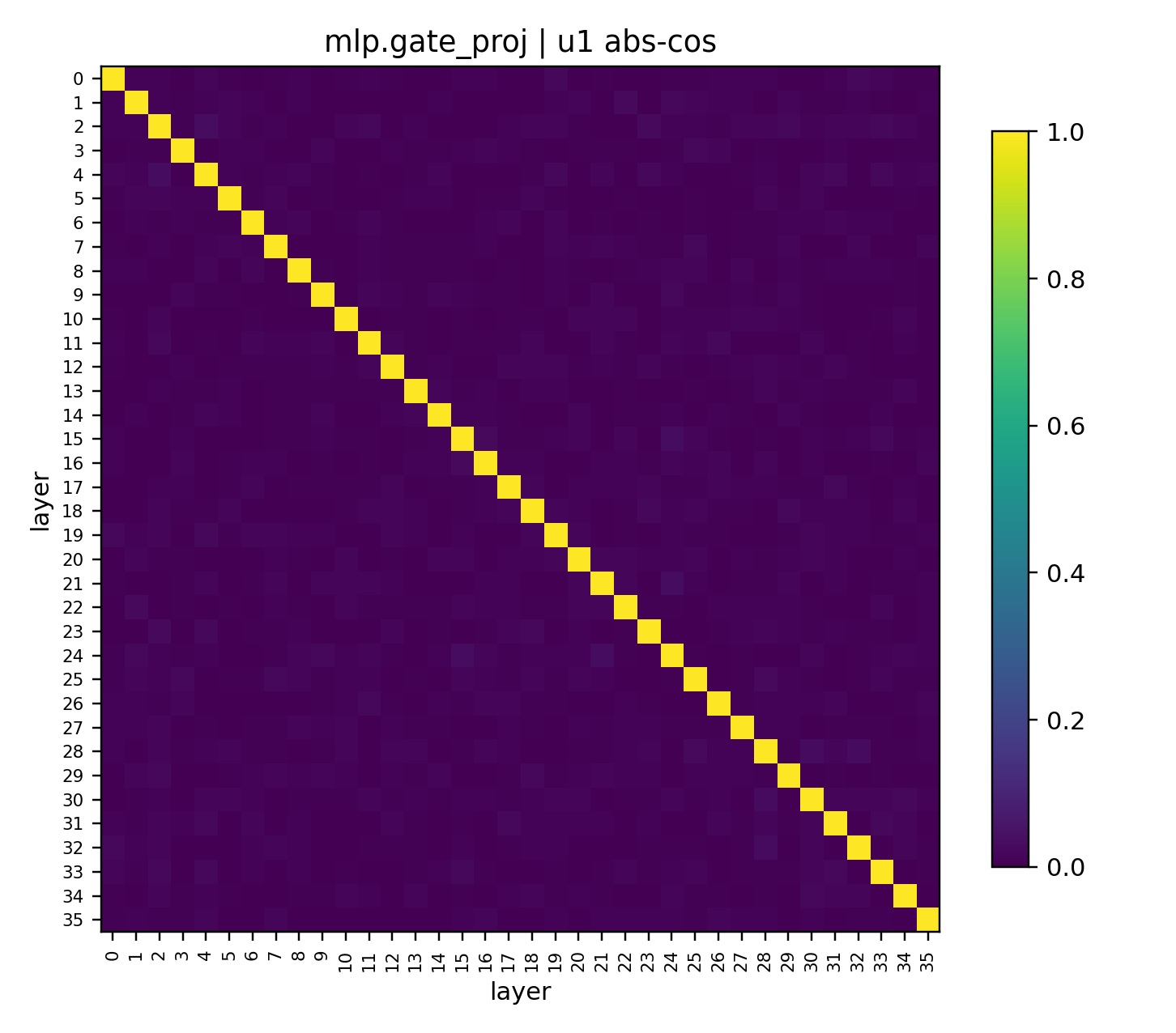} &
\includegraphics[width=0.195\textwidth]{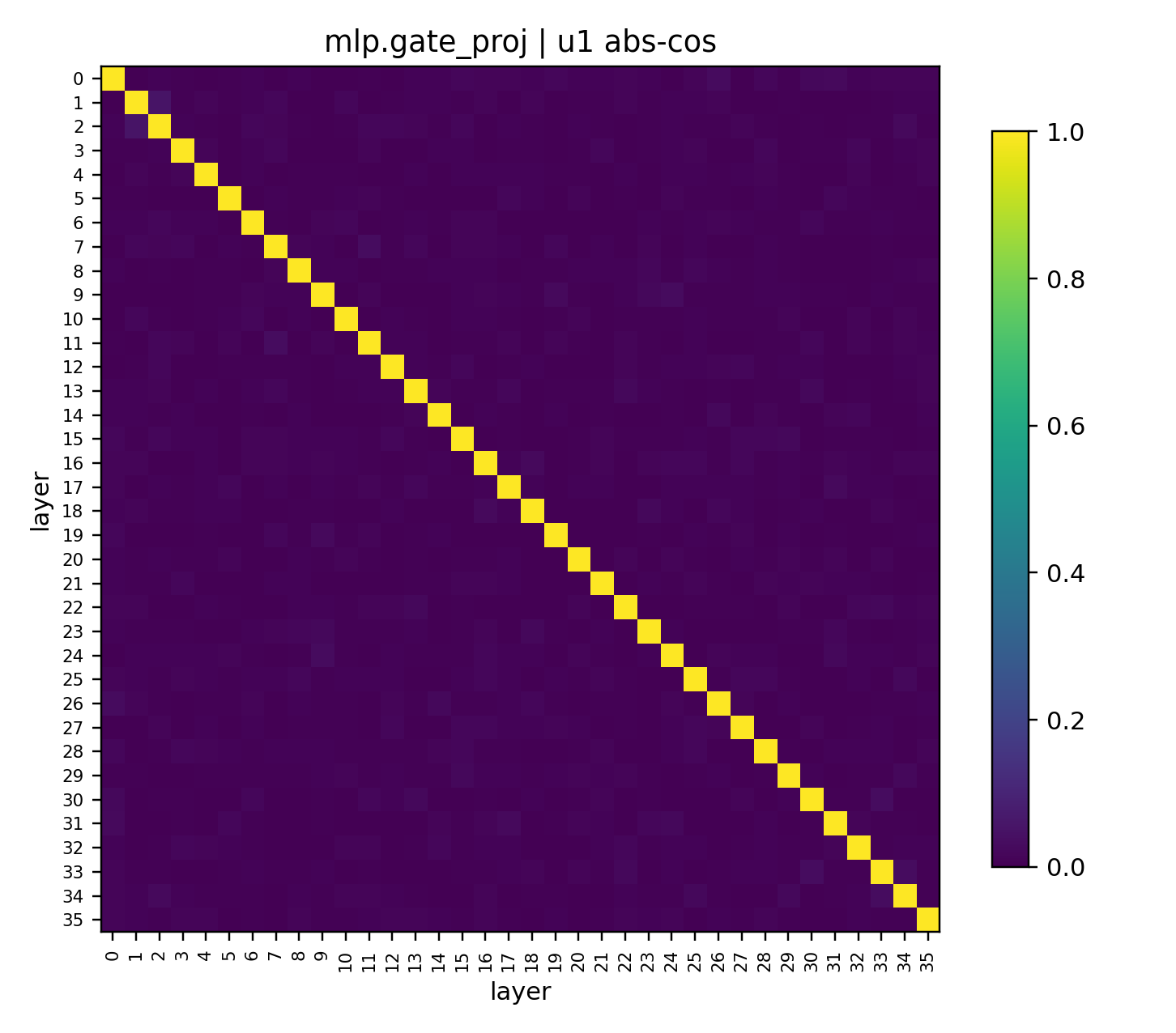} \\
\texttt{up\_proj} &
\includegraphics[width=0.195\textwidth]{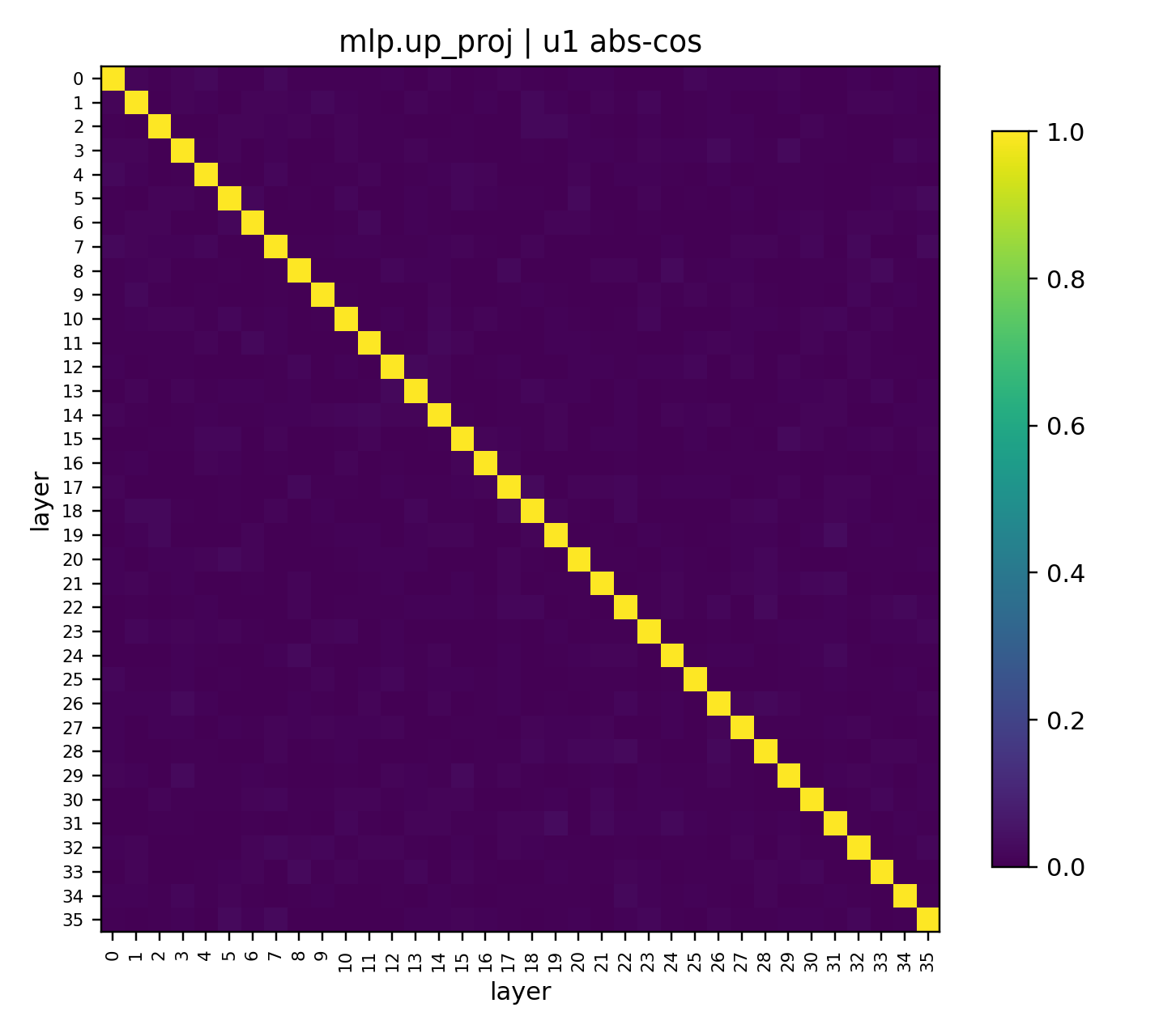} &
\includegraphics[width=0.195\textwidth]{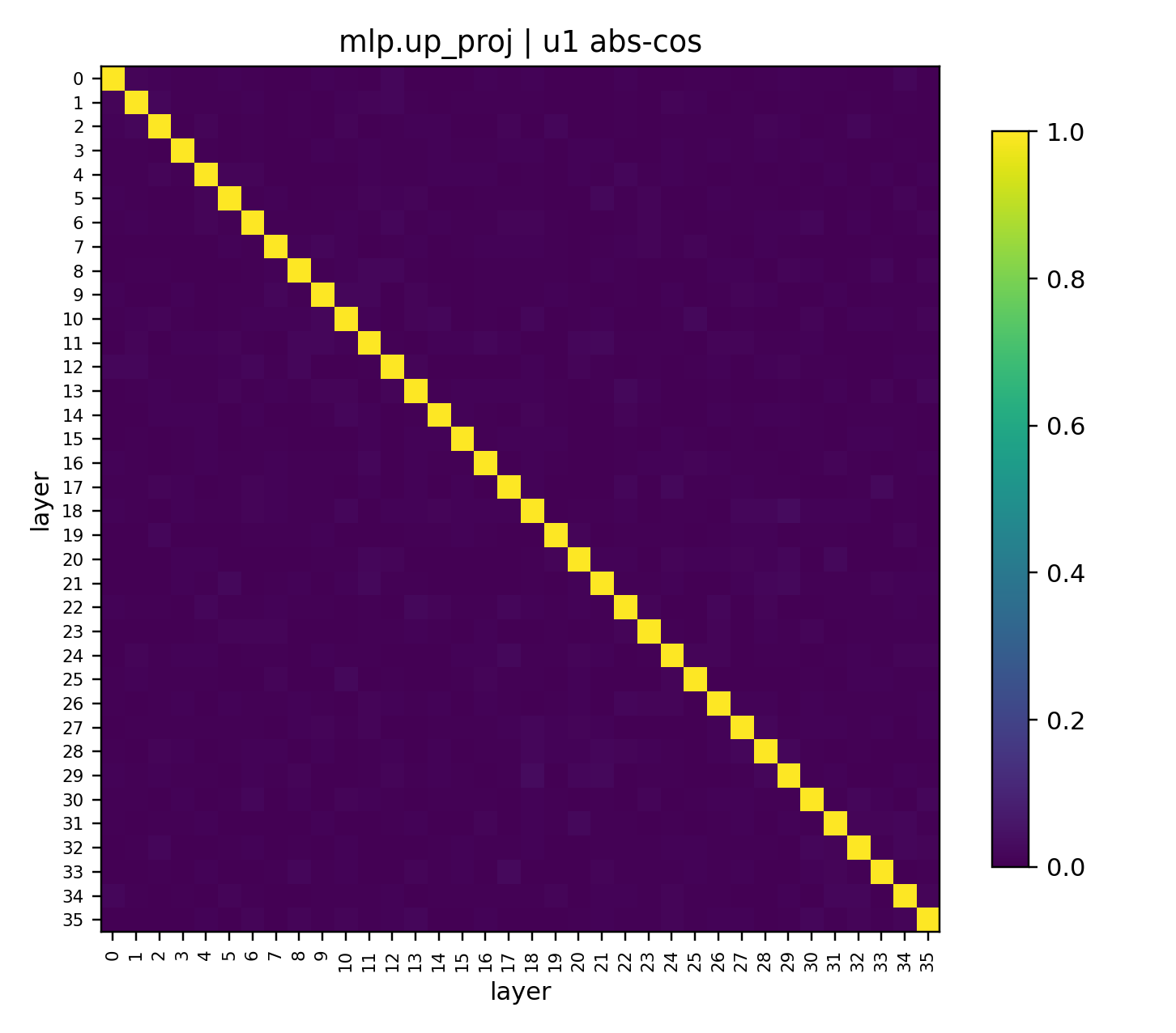} &
\includegraphics[width=0.195\textwidth]{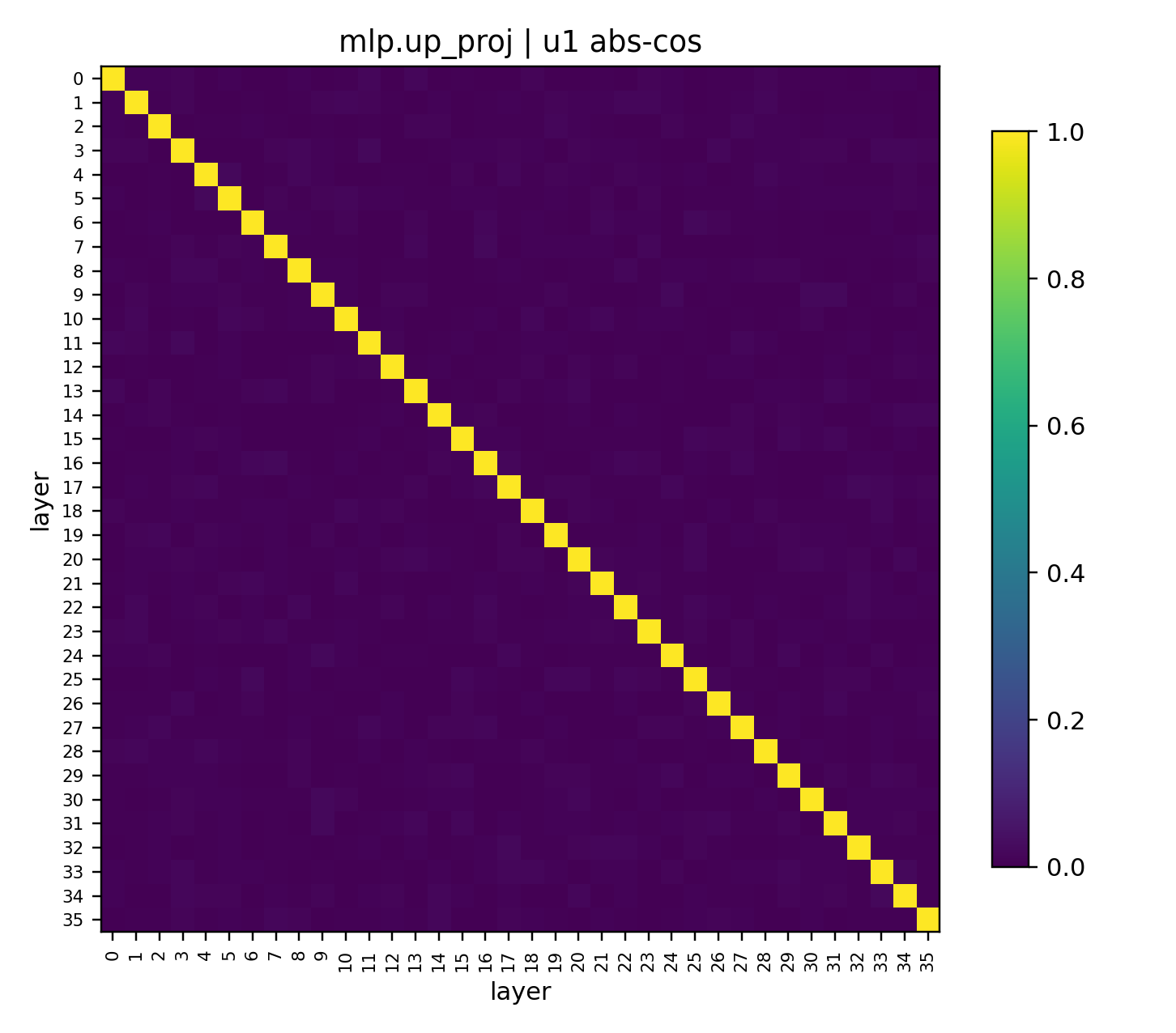} &
\includegraphics[width=0.195\textwidth]{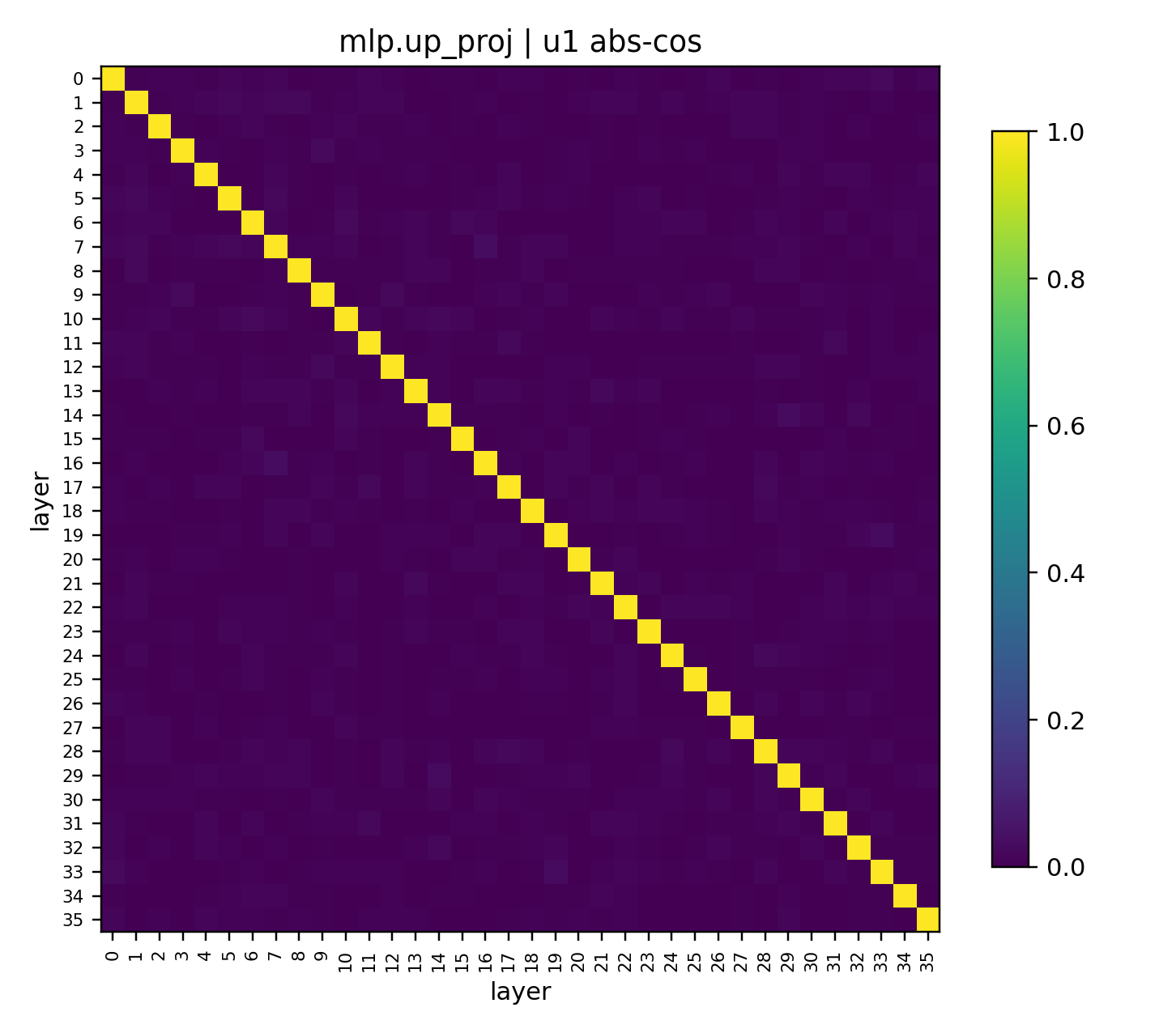} \\
\texttt{down\_proj} &
\includegraphics[width=0.195\textwidth]{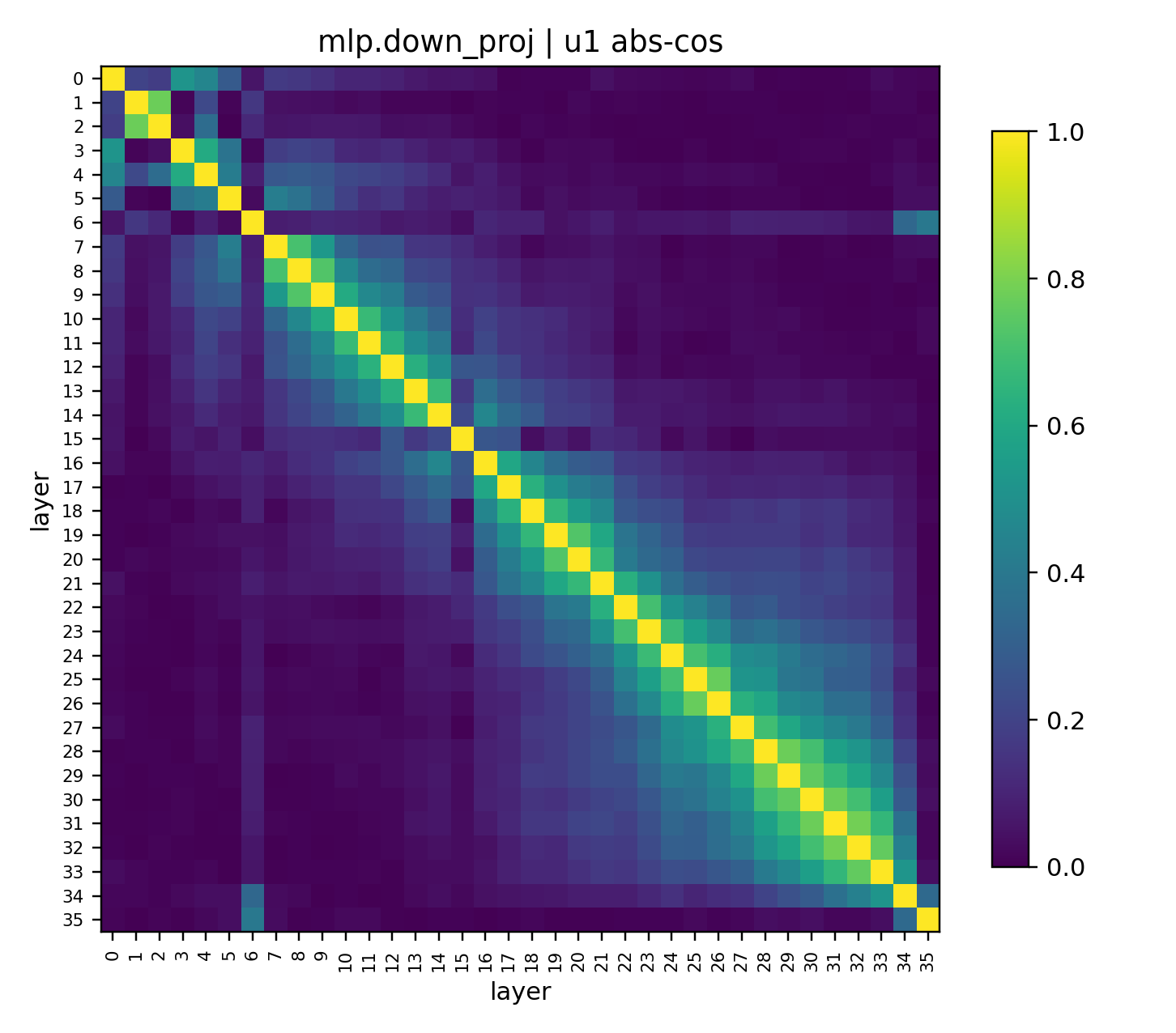} &
\includegraphics[width=0.195\textwidth]{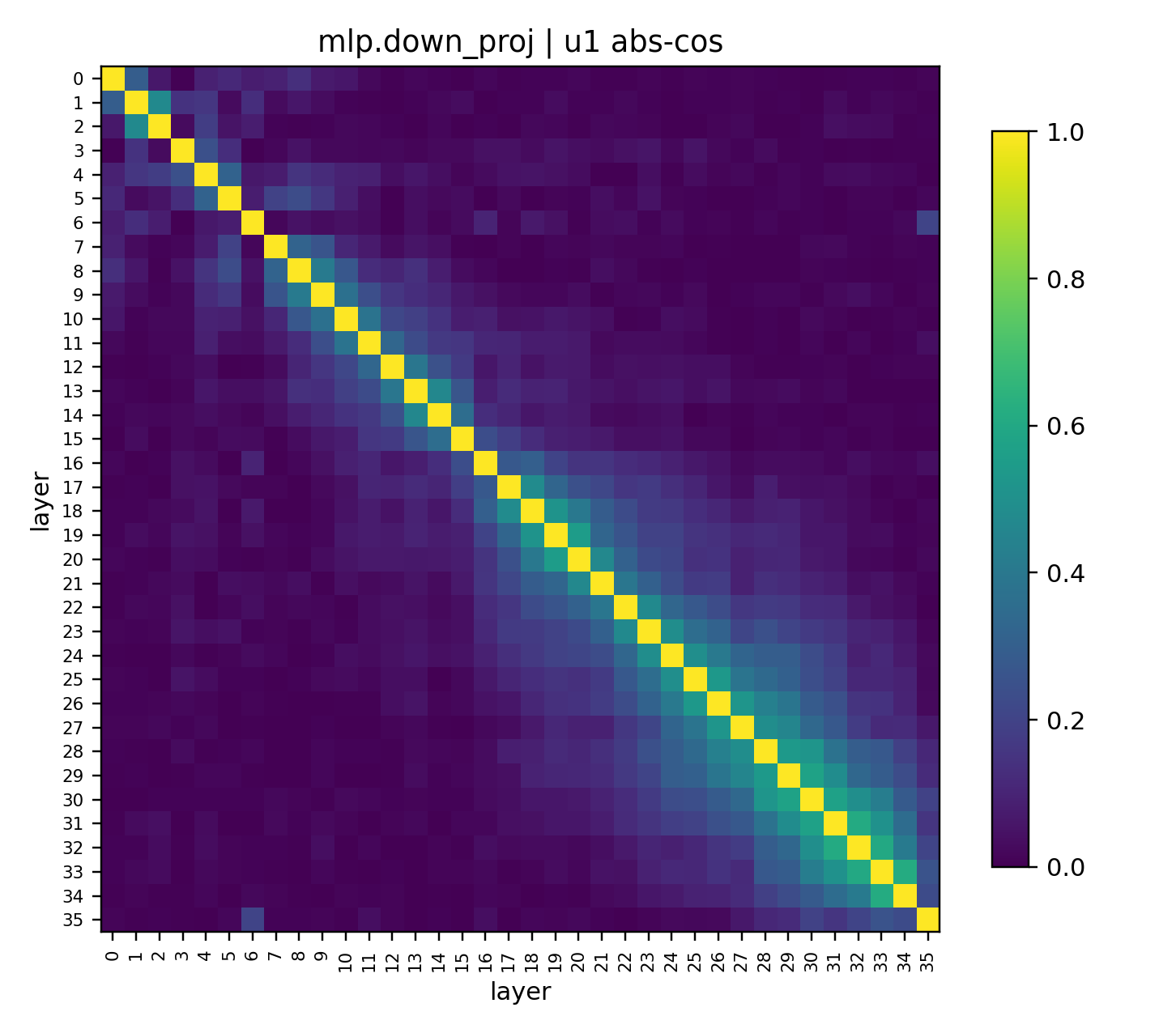} &
\includegraphics[width=0.195\textwidth]{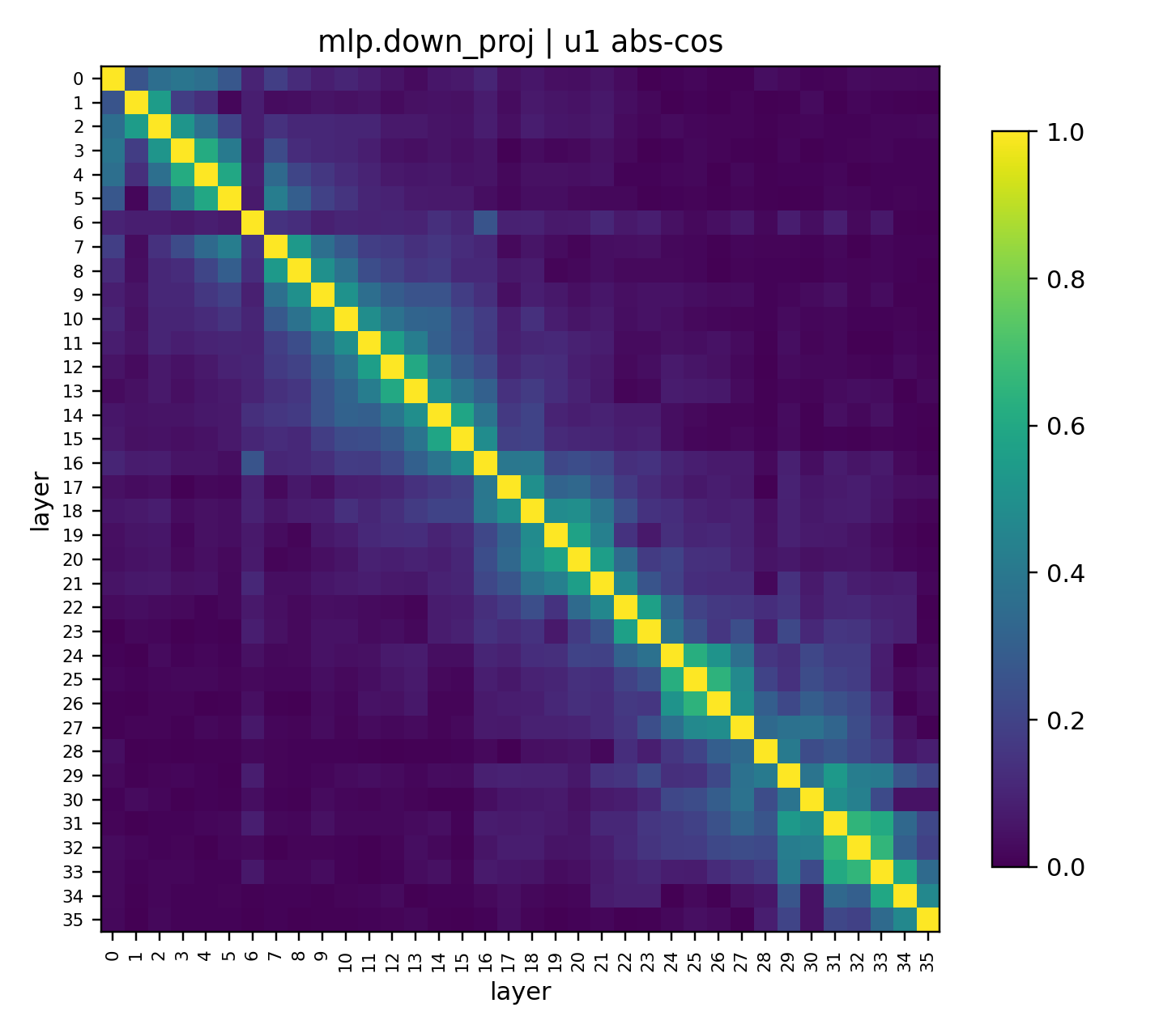} &
\includegraphics[width=0.195\textwidth]{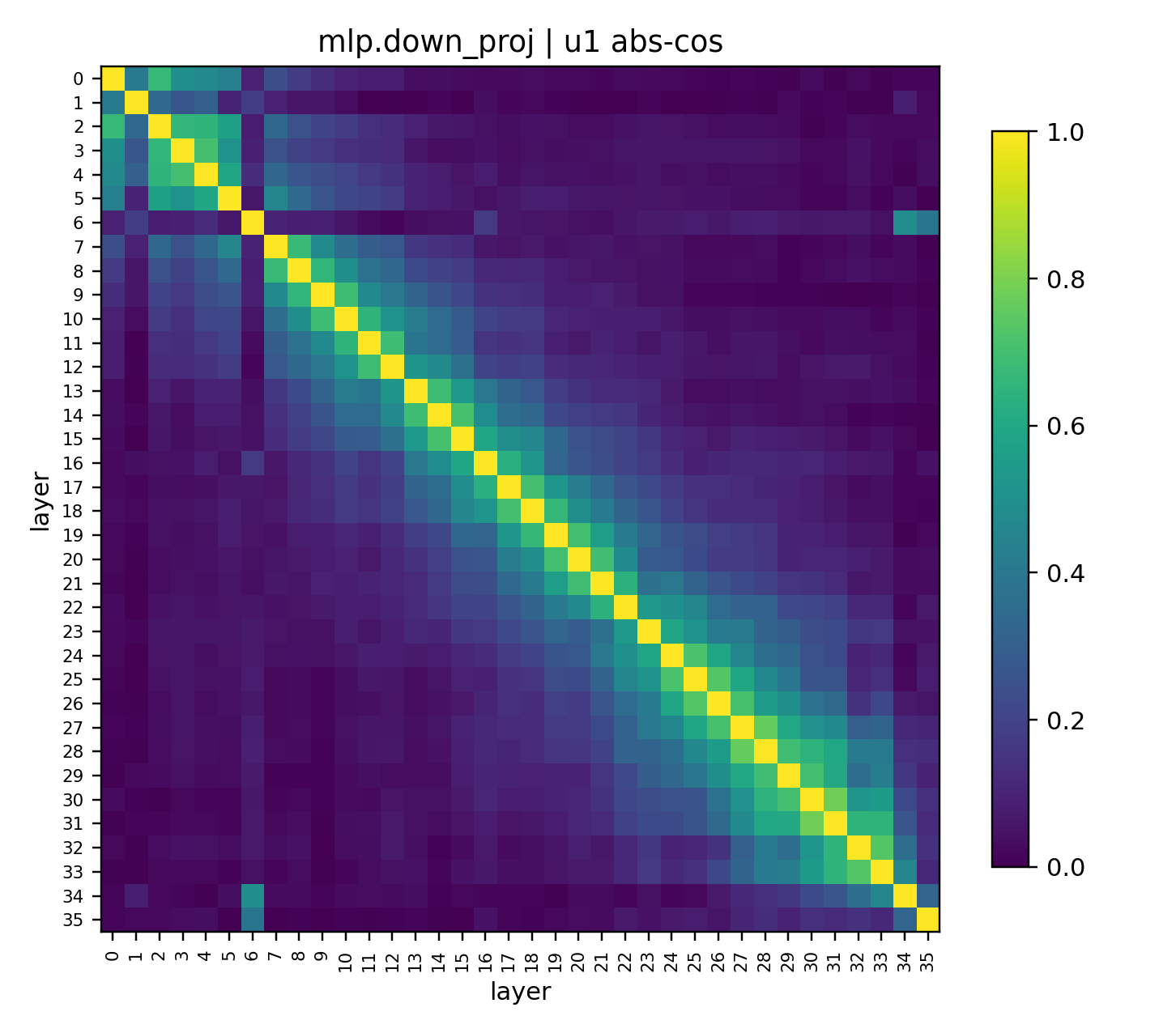} \\
\bottomrule
\end{tabular}
\caption{\textbf{Qwen3-8B: Principal-direction similarity heatmap wall ($|u_1^\top u_1|$).}
Each cell shows the inter-layer similarity heatmap for a specific (task, module).}
\label{fig:heatwall_qwen_u1}
\end{figure*}

\clearpage
\begin{figure*}[p]
\centering
\small
\setlength{\tabcolsep}{2pt}
\renewcommand{\arraystretch}{1.0}
\begin{tabular}{lcccc}
\toprule
\textbf{Module} &
\textbf{Math} & \textbf{Code} & \textbf{Instruction} & \textbf{Commonsense} \\
\midrule
\texttt{q\_proj} &
\includegraphics[width=0.195\textwidth]{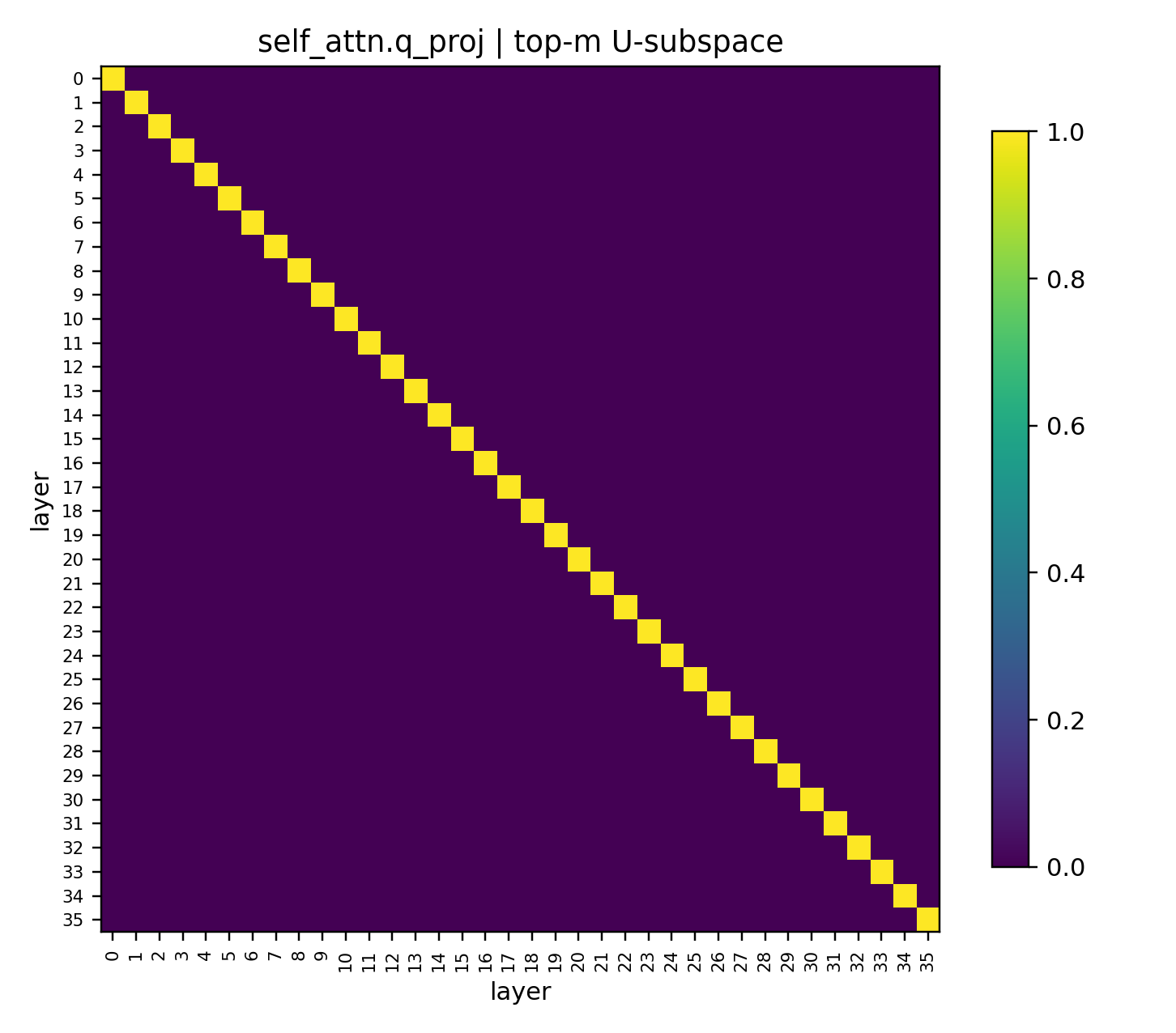} &
\includegraphics[width=0.195\textwidth]{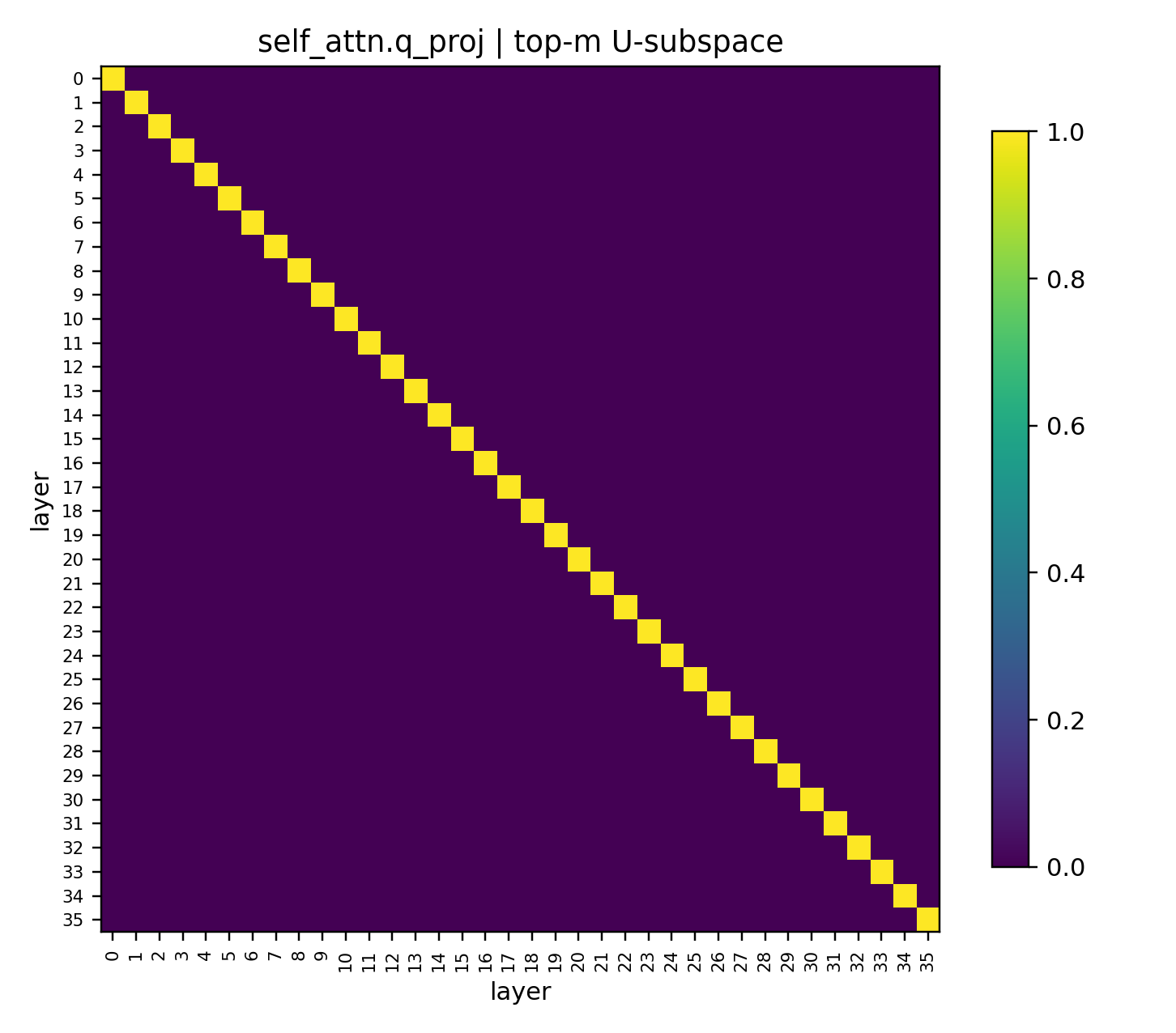} &
\includegraphics[width=0.195\textwidth]{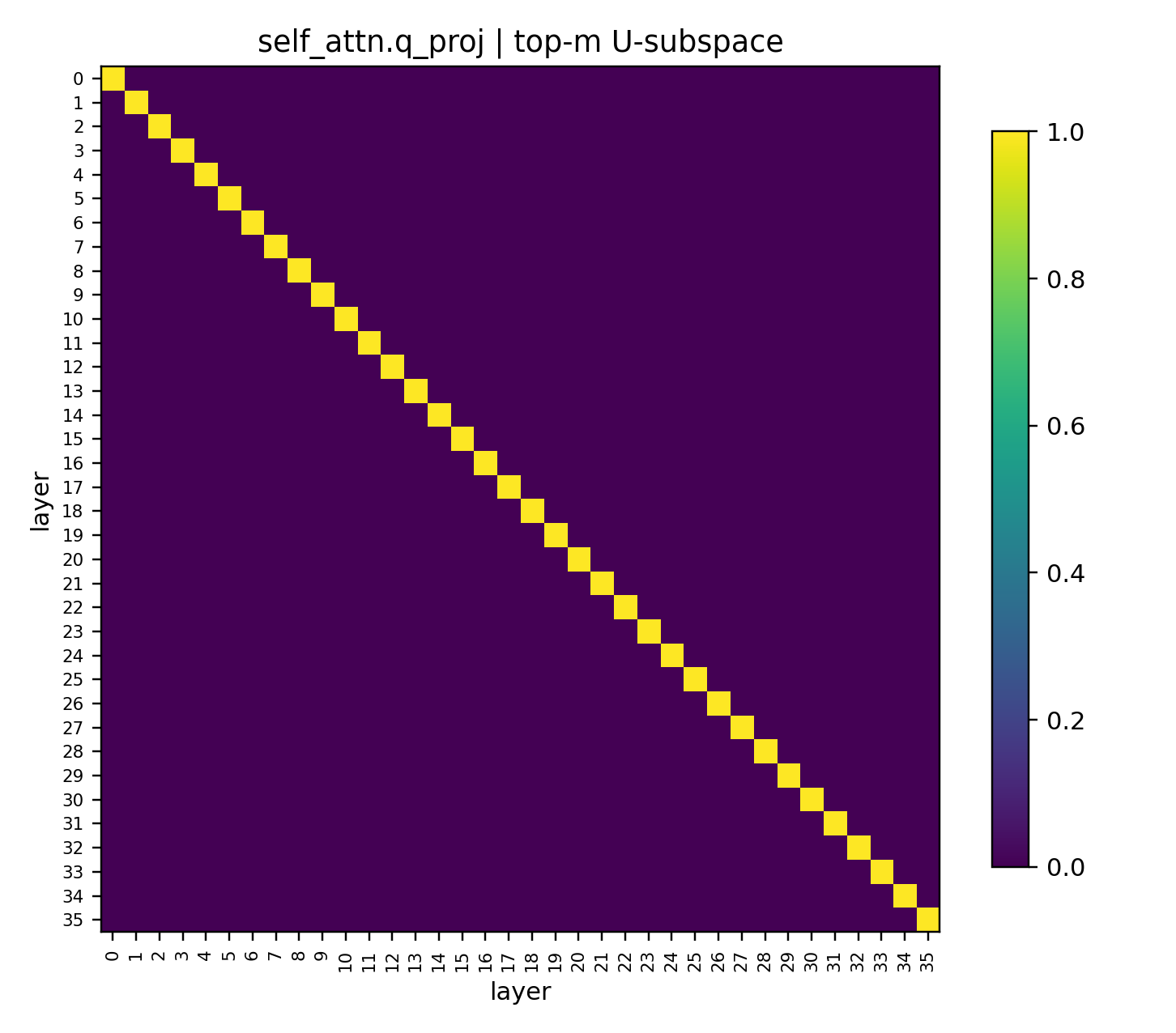} &
\includegraphics[width=0.195\textwidth]{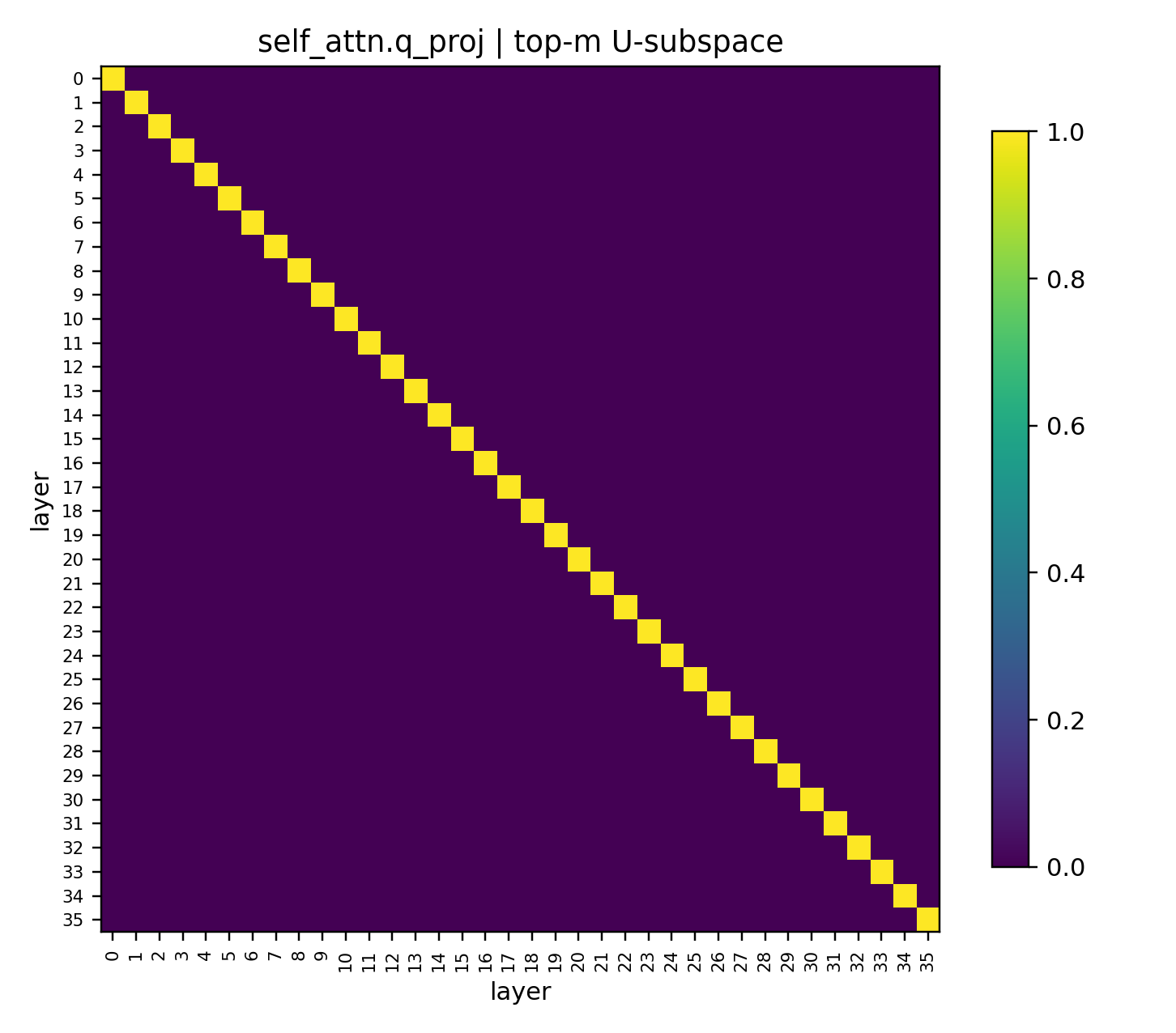} \\
\texttt{k\_proj} &
\includegraphics[width=0.195\textwidth]{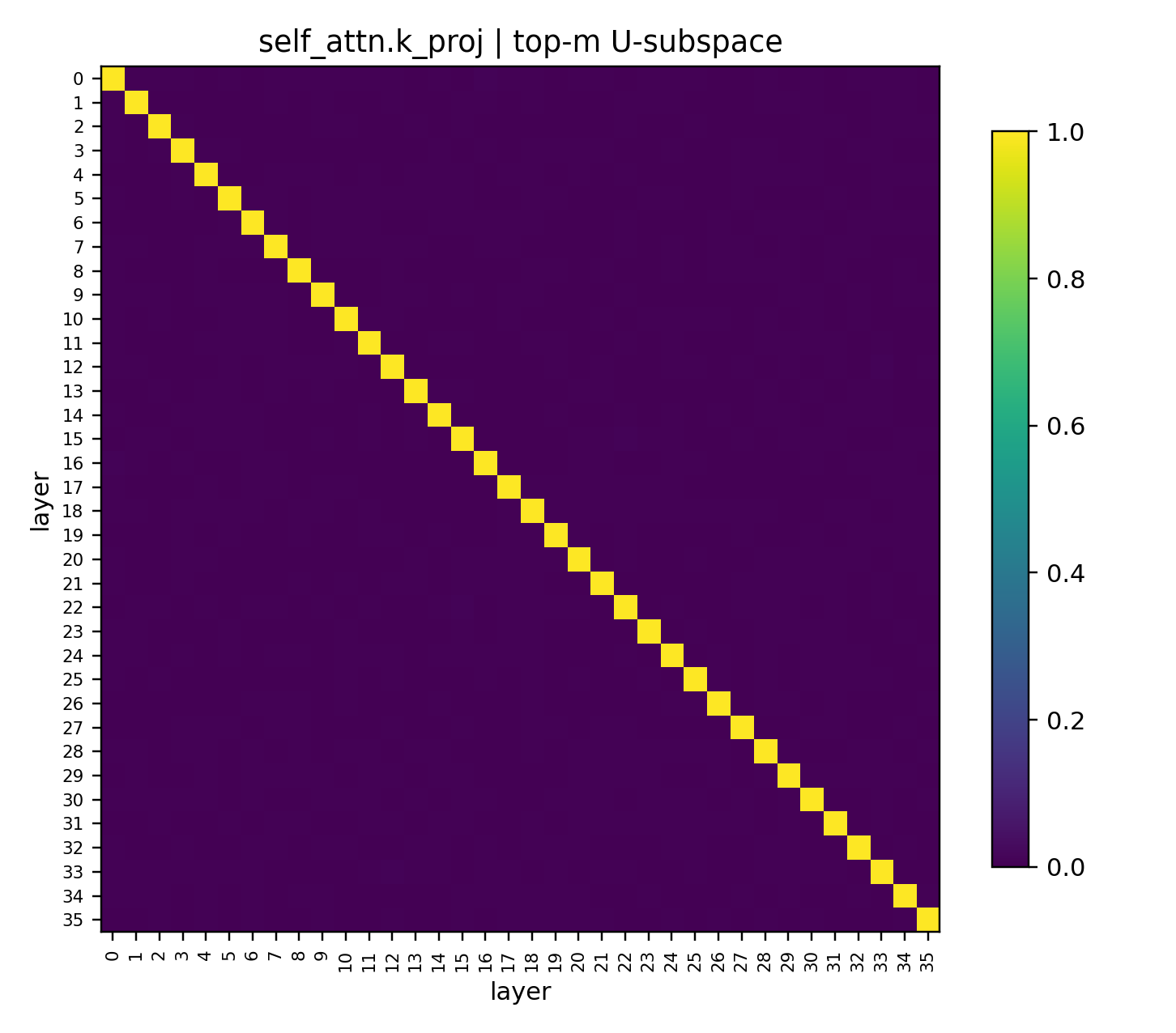} &
\includegraphics[width=0.195\textwidth]{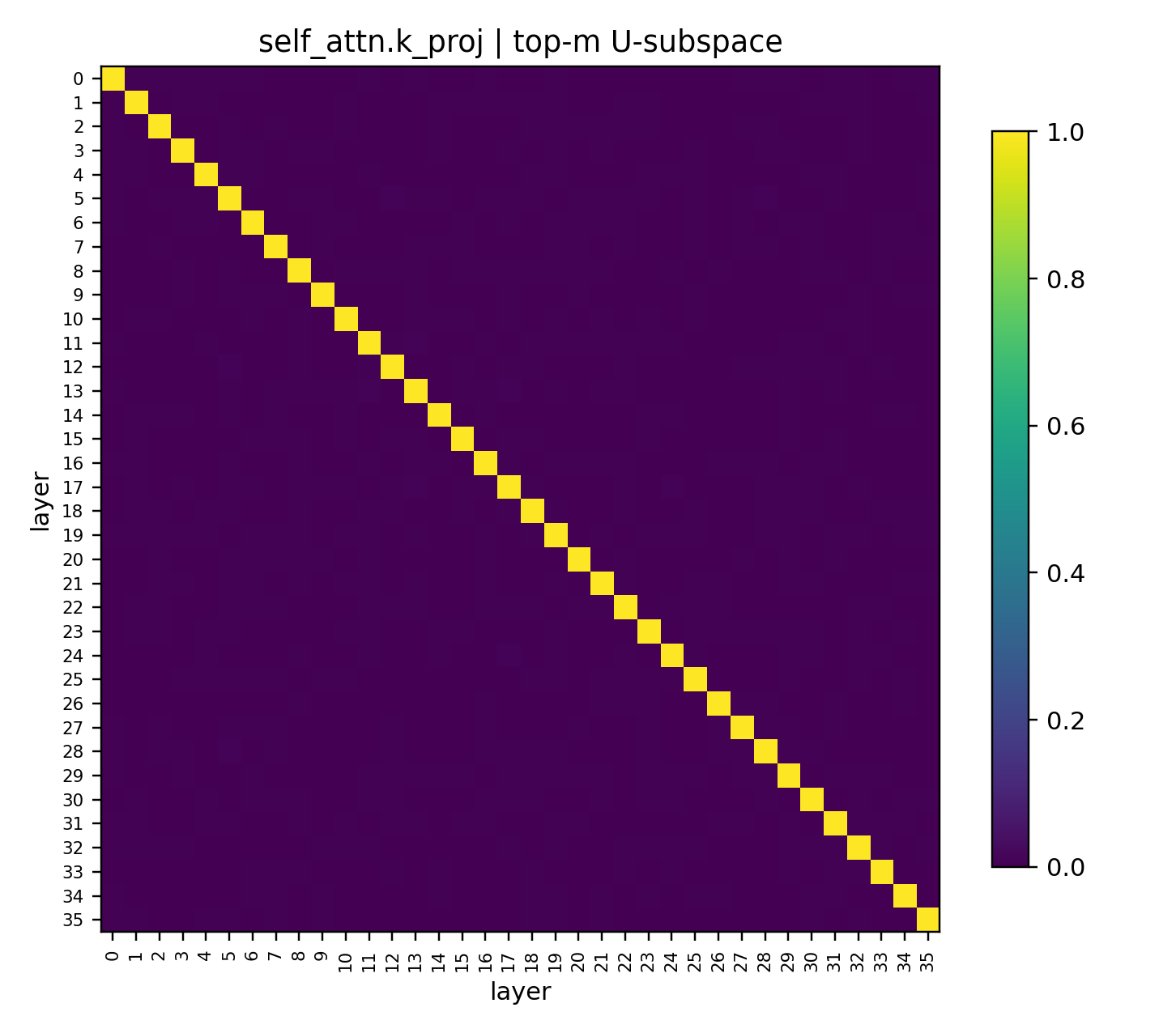} &
\includegraphics[width=0.195\textwidth]{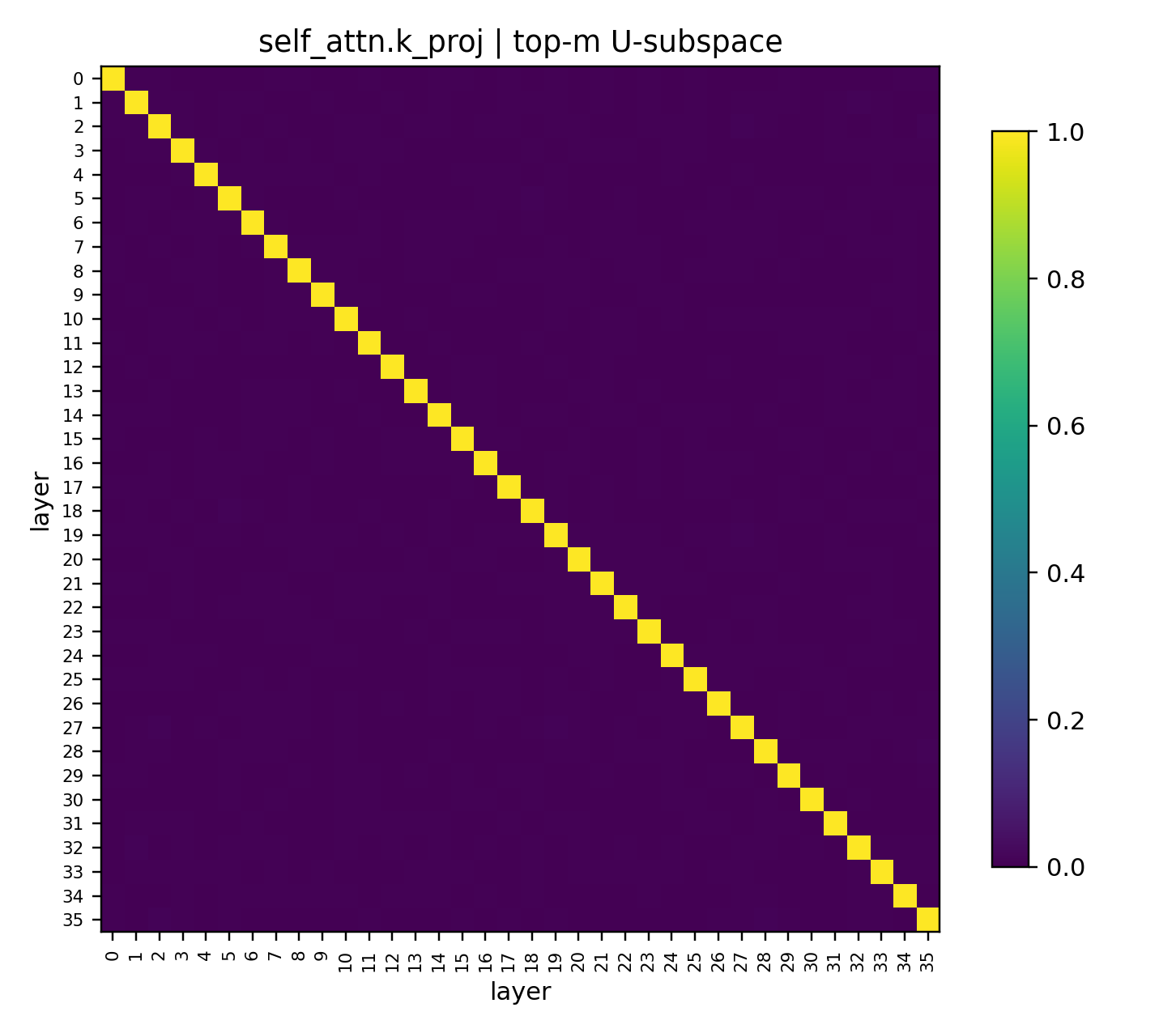} &
\includegraphics[width=0.195\textwidth]{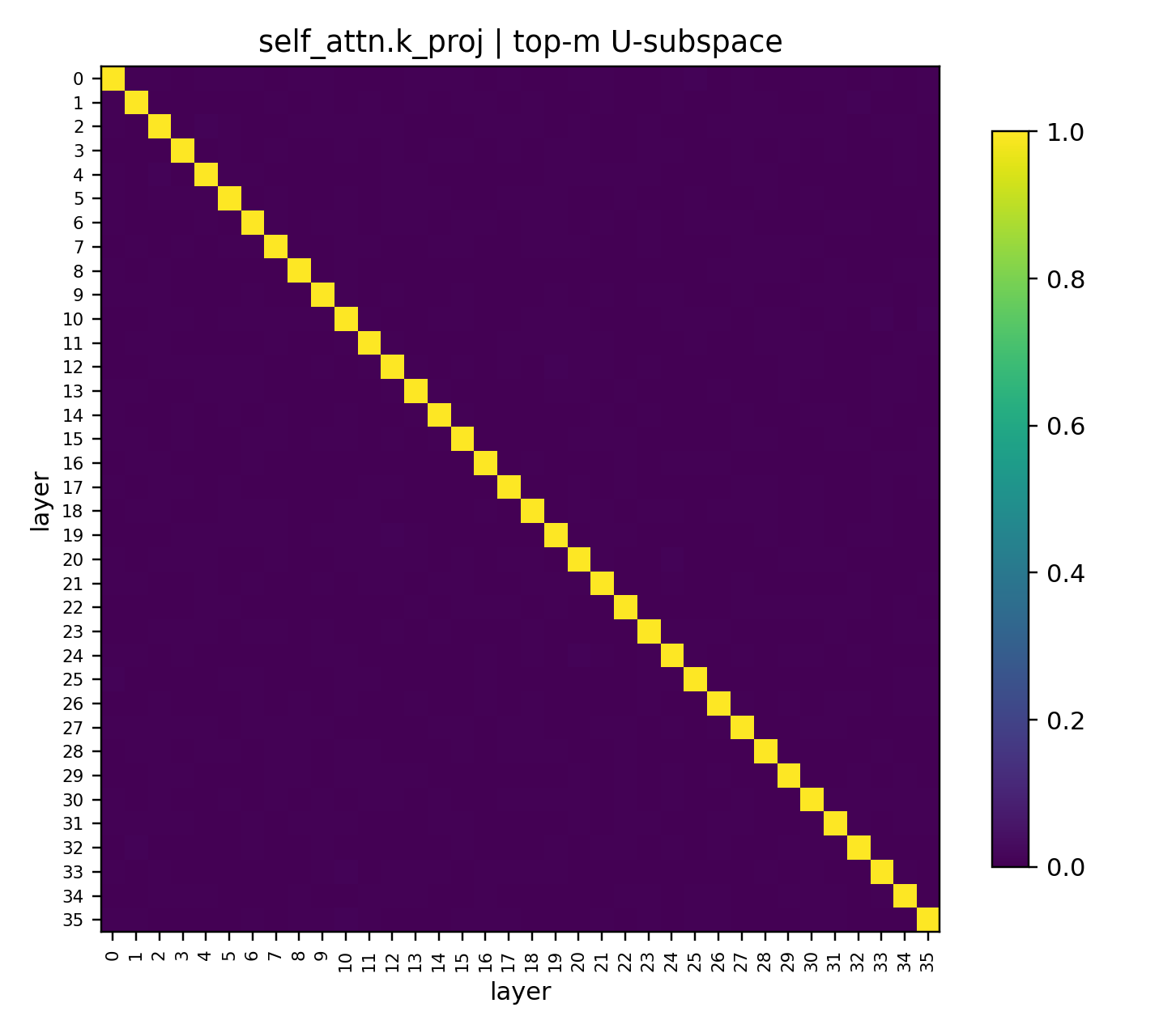} \\
\texttt{v\_proj} &
\includegraphics[width=0.195\textwidth]{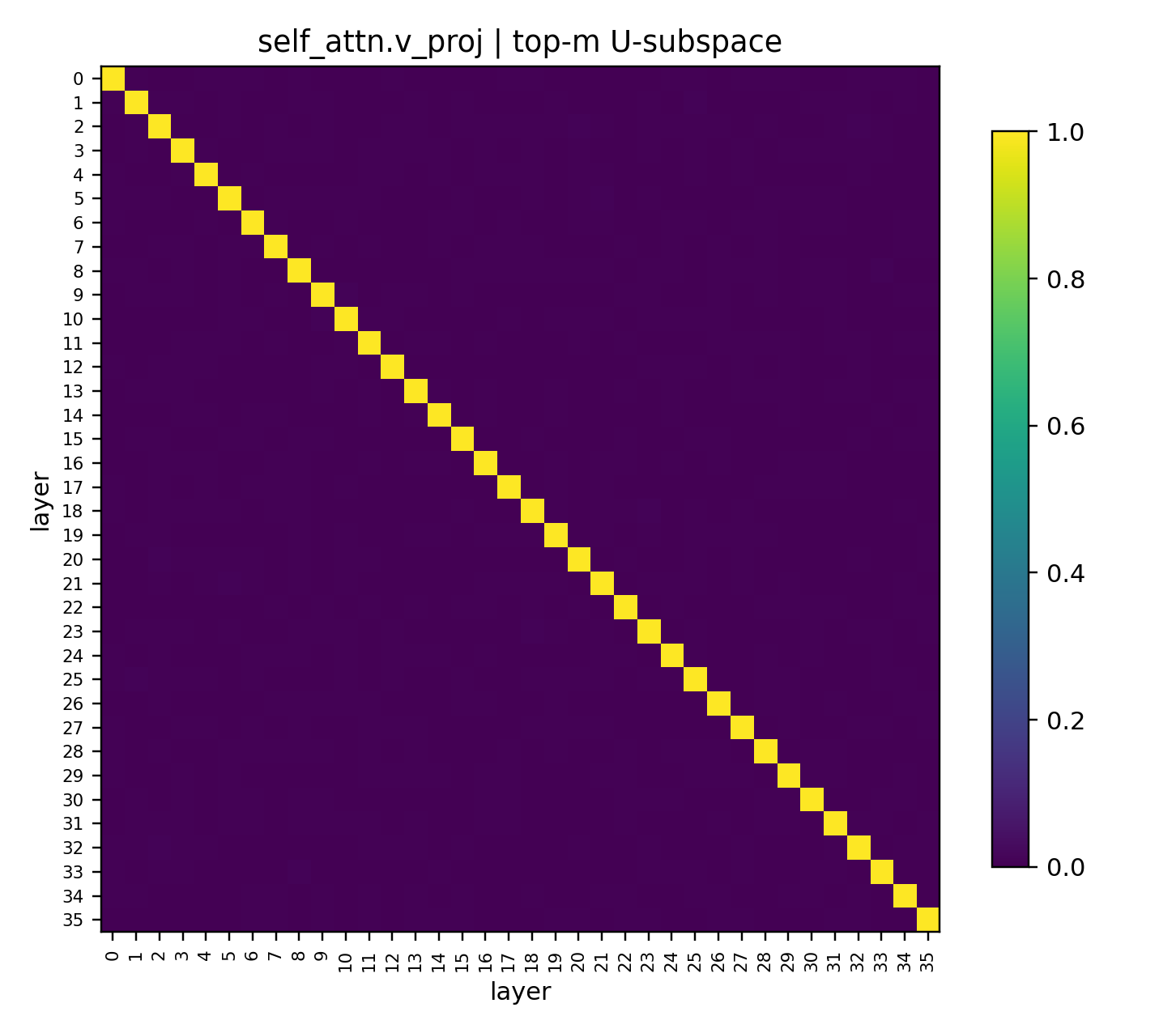} &
\includegraphics[width=0.195\textwidth]{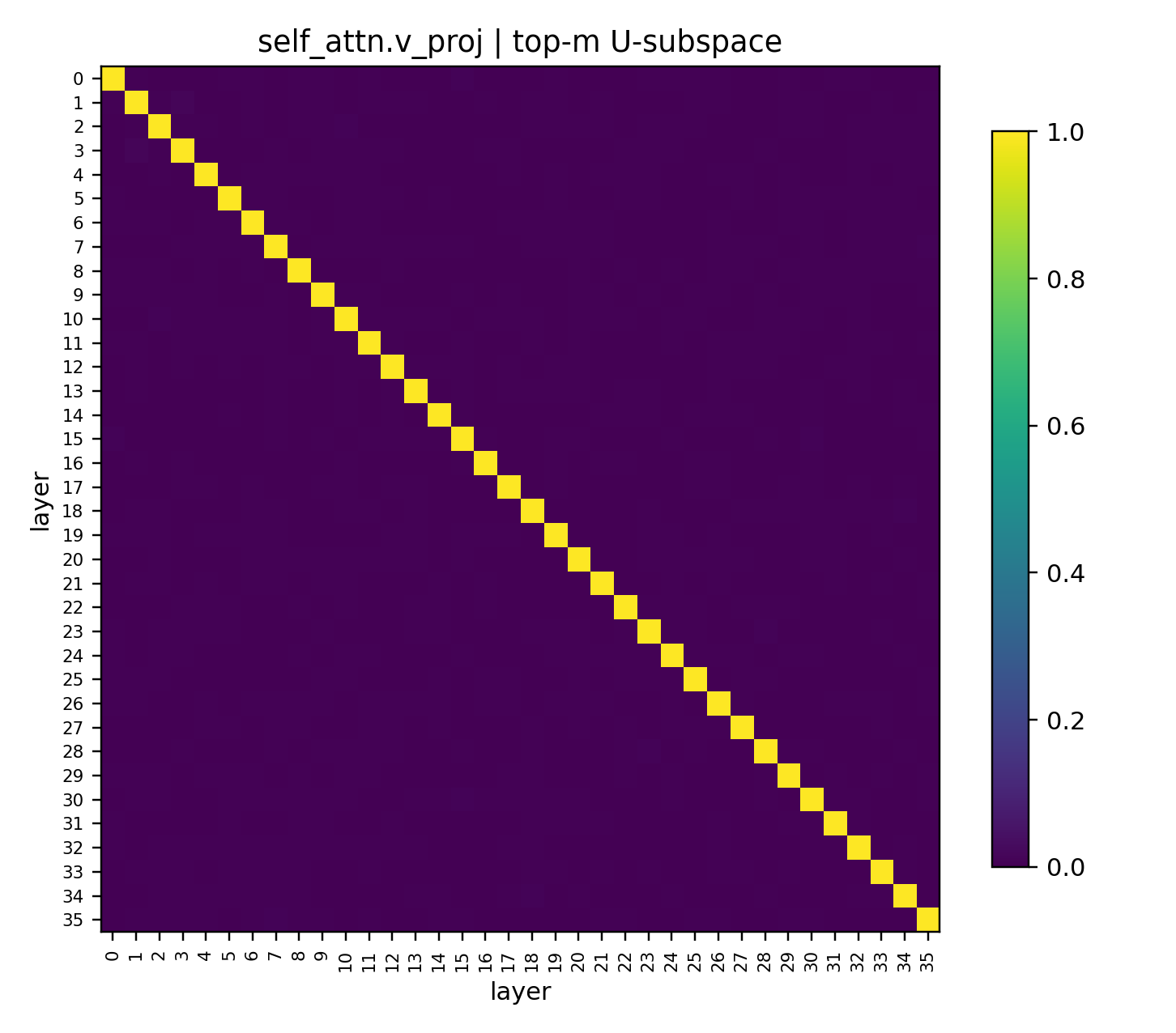} &
\includegraphics[width=0.195\textwidth]{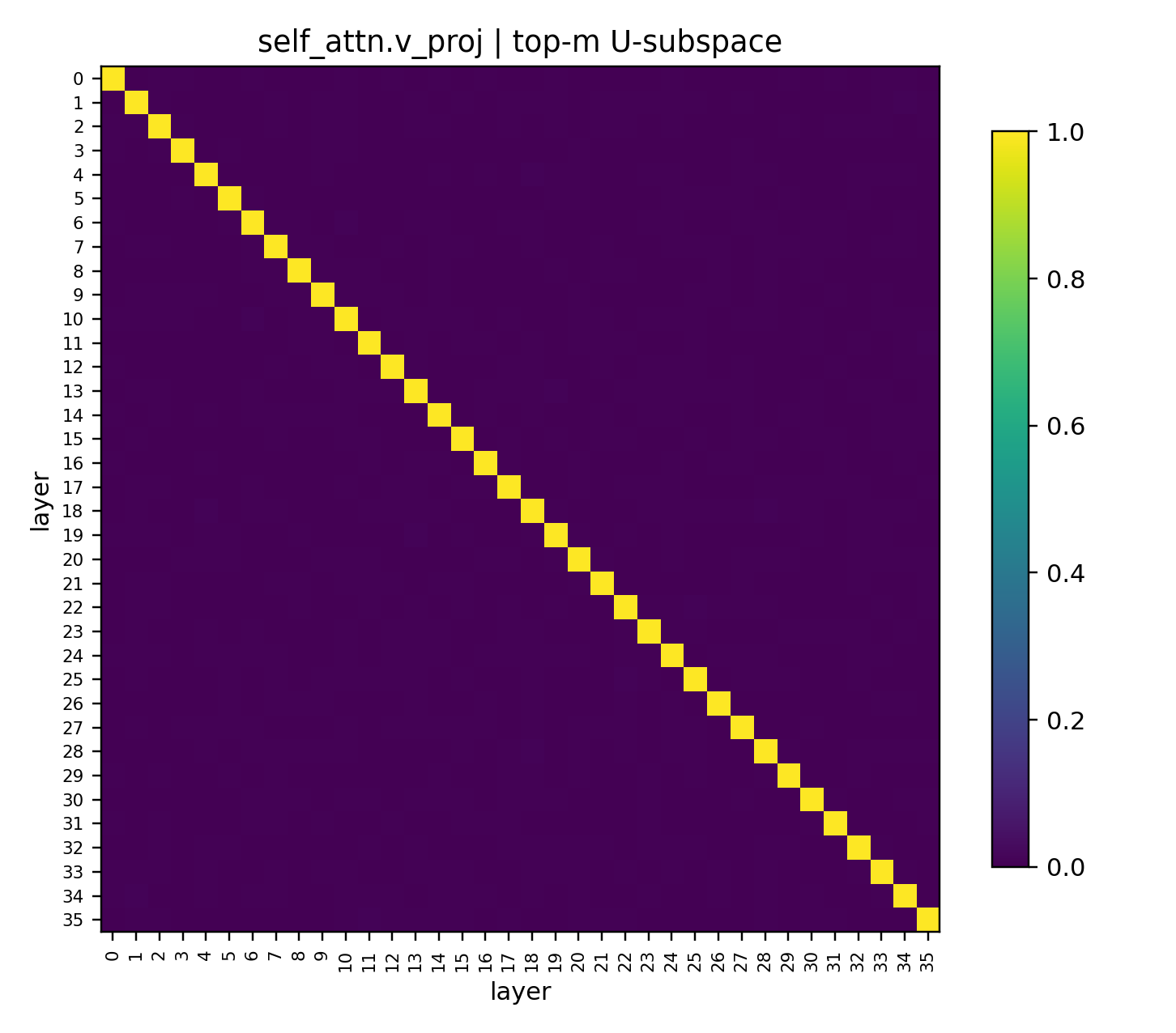} &
\includegraphics[width=0.195\textwidth]{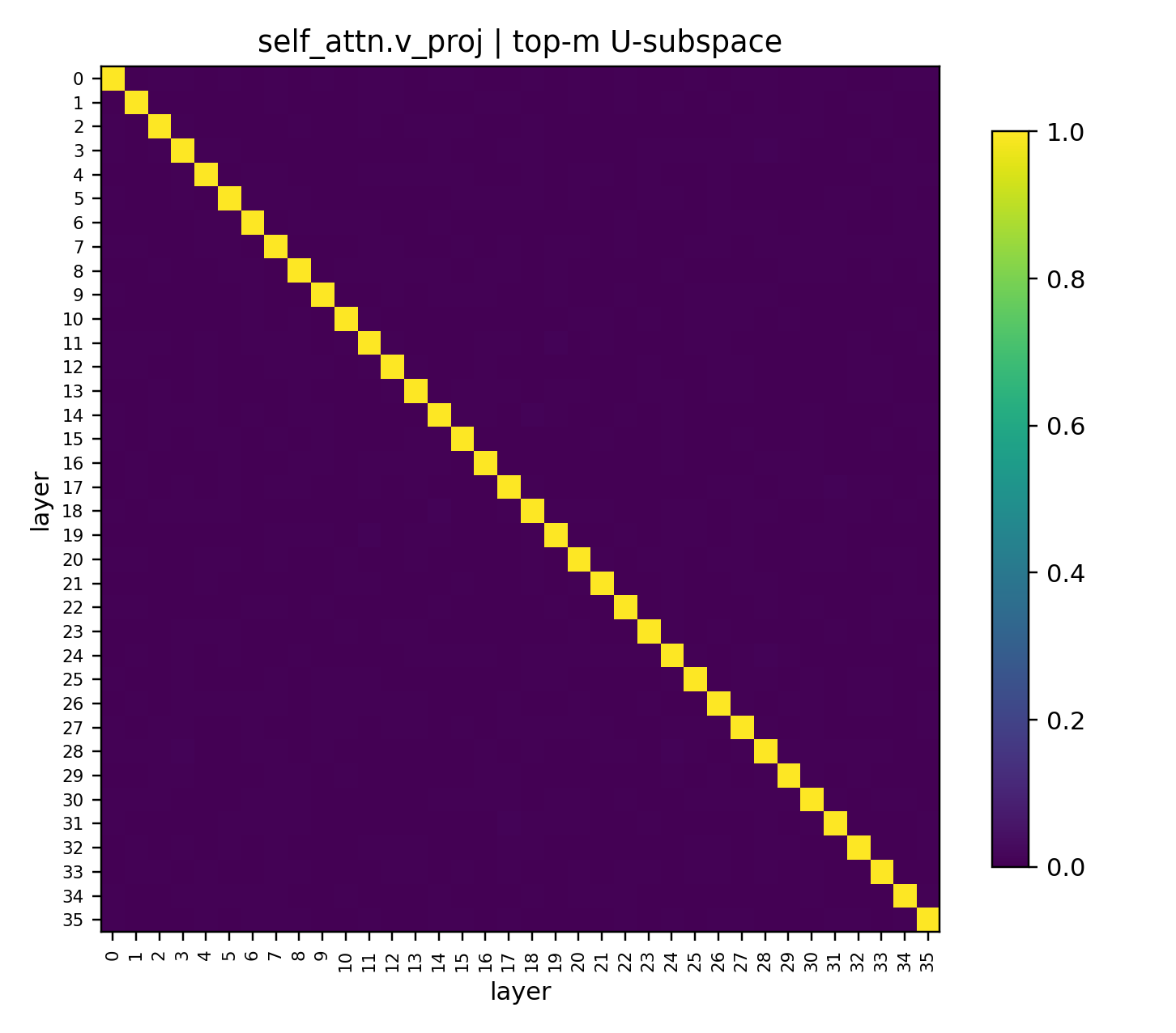} \\
\texttt{o\_proj} &
\includegraphics[width=0.195\textwidth]{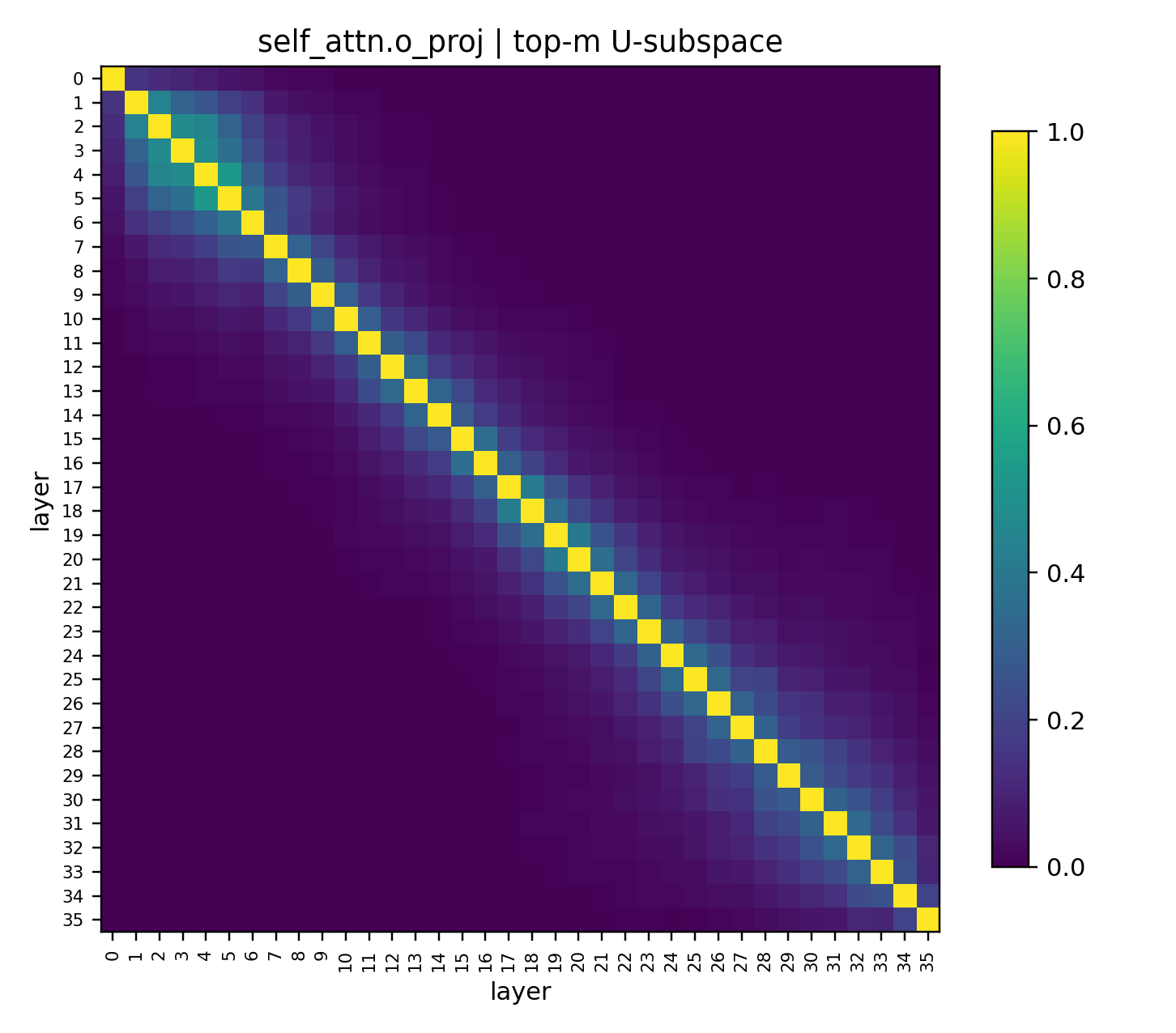} &
\includegraphics[width=0.195\textwidth]{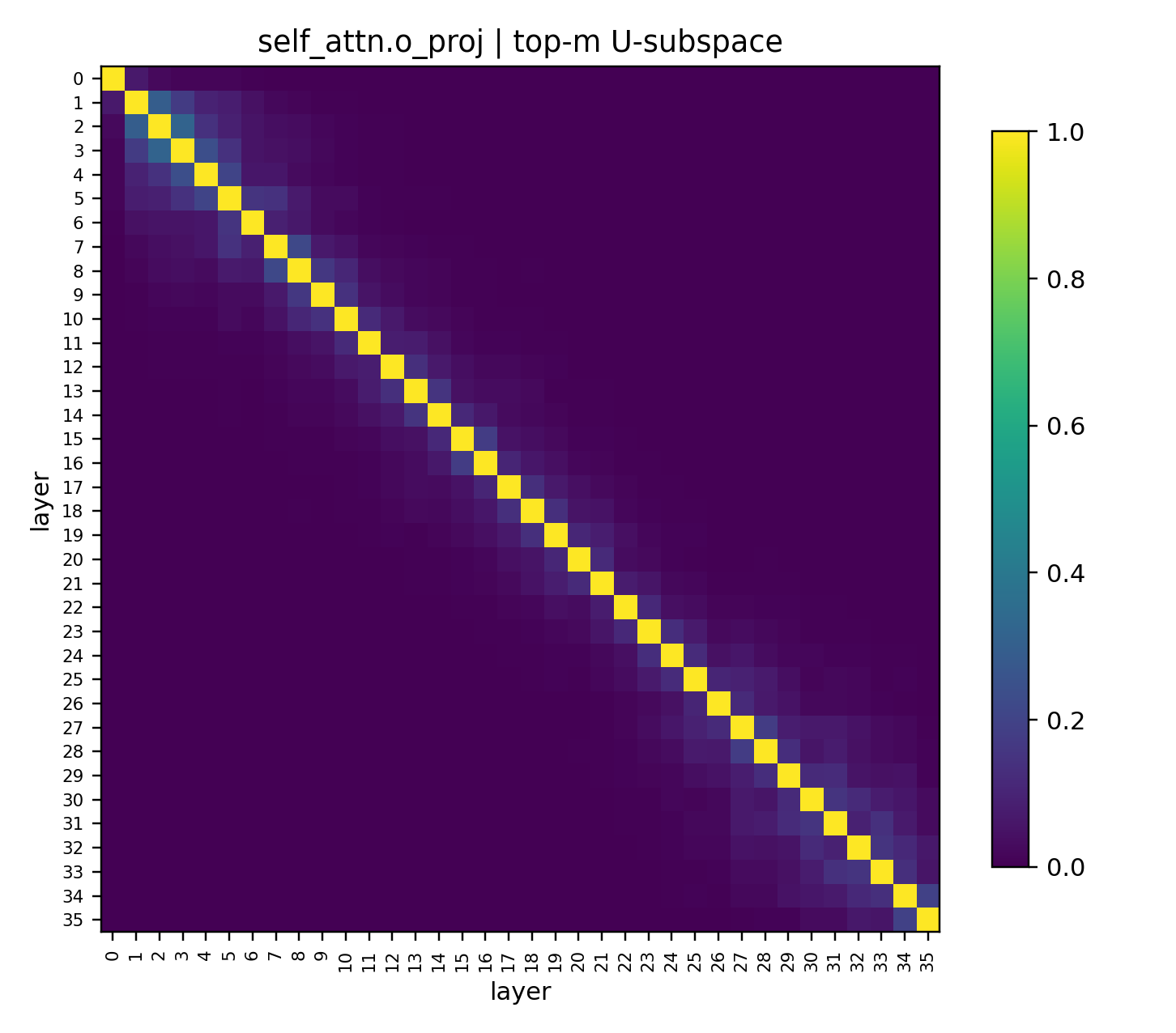} &
\includegraphics[width=0.195\textwidth]{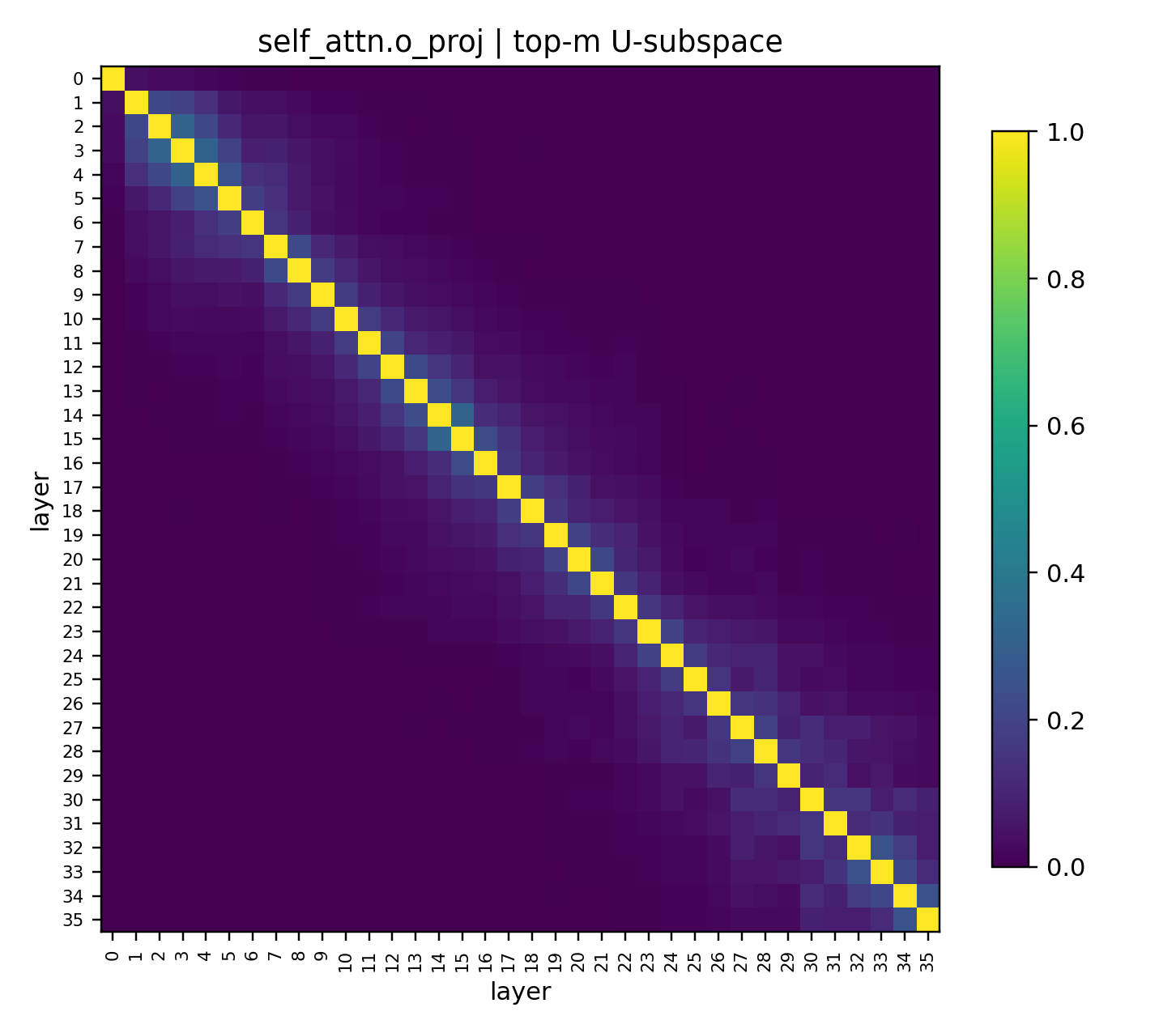} &
\includegraphics[width=0.195\textwidth]{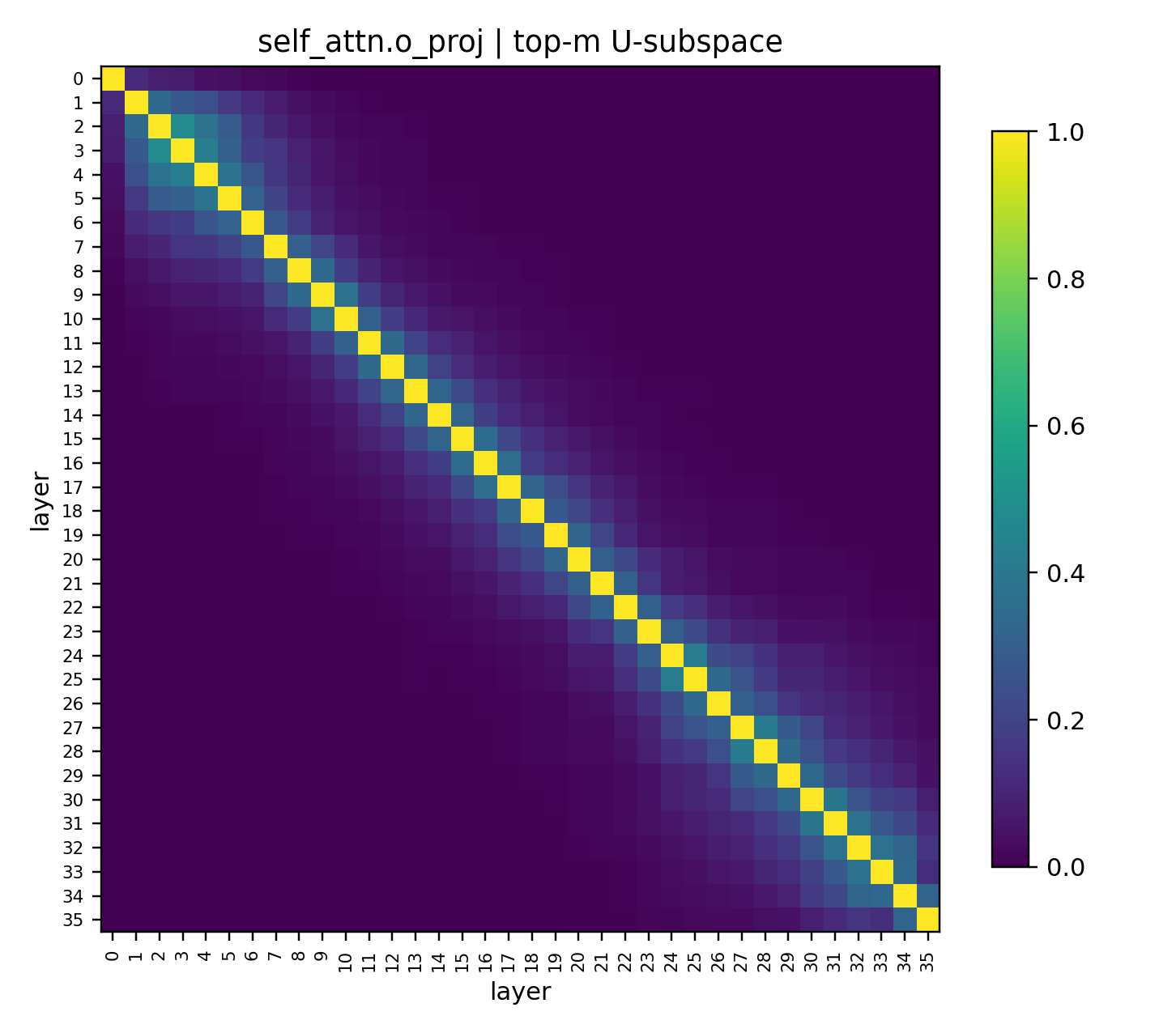} \\
\texttt{gate\_proj} &
\includegraphics[width=0.195\textwidth]{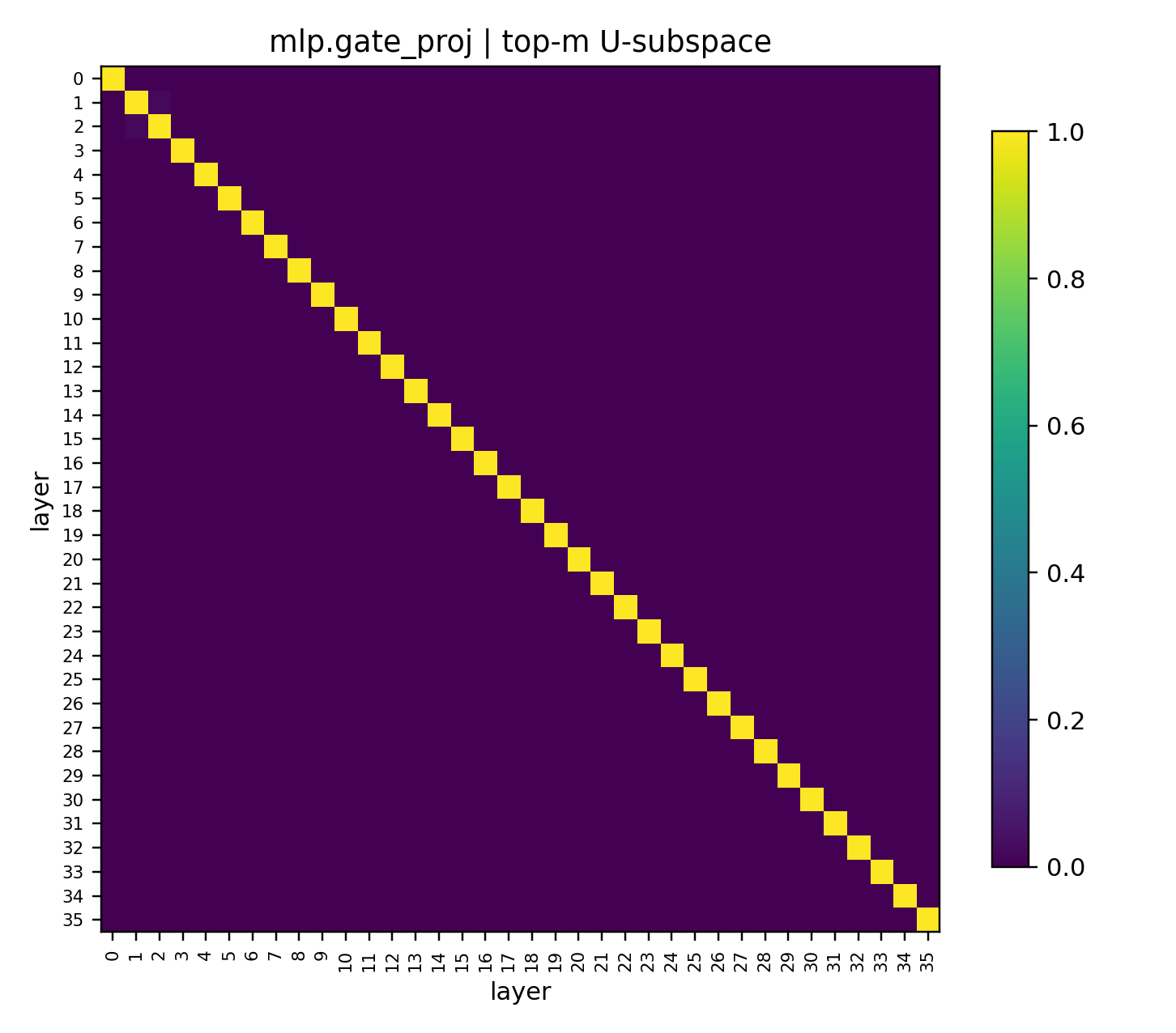} &
\includegraphics[width=0.195\textwidth]{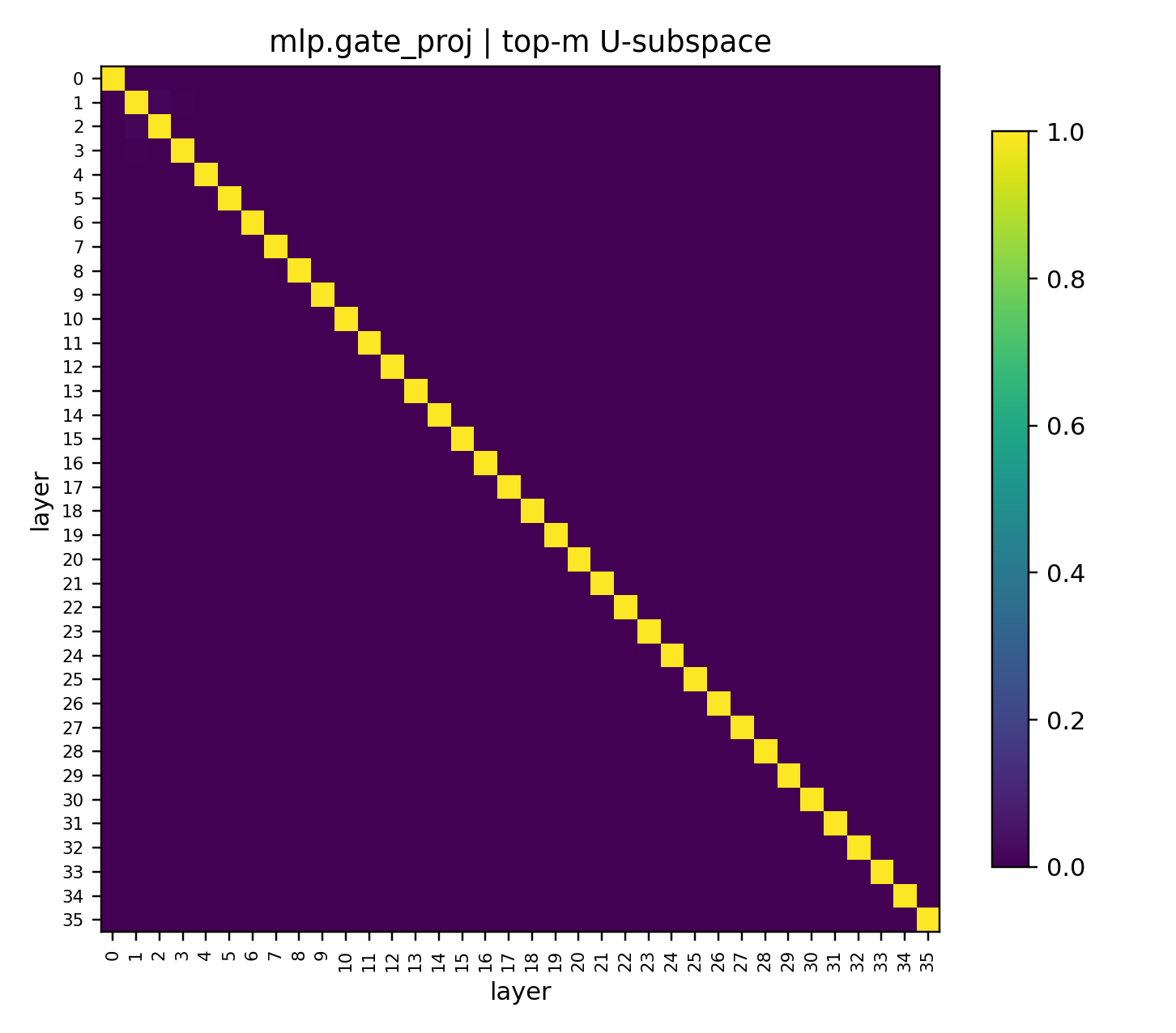} &
\includegraphics[width=0.195\textwidth]{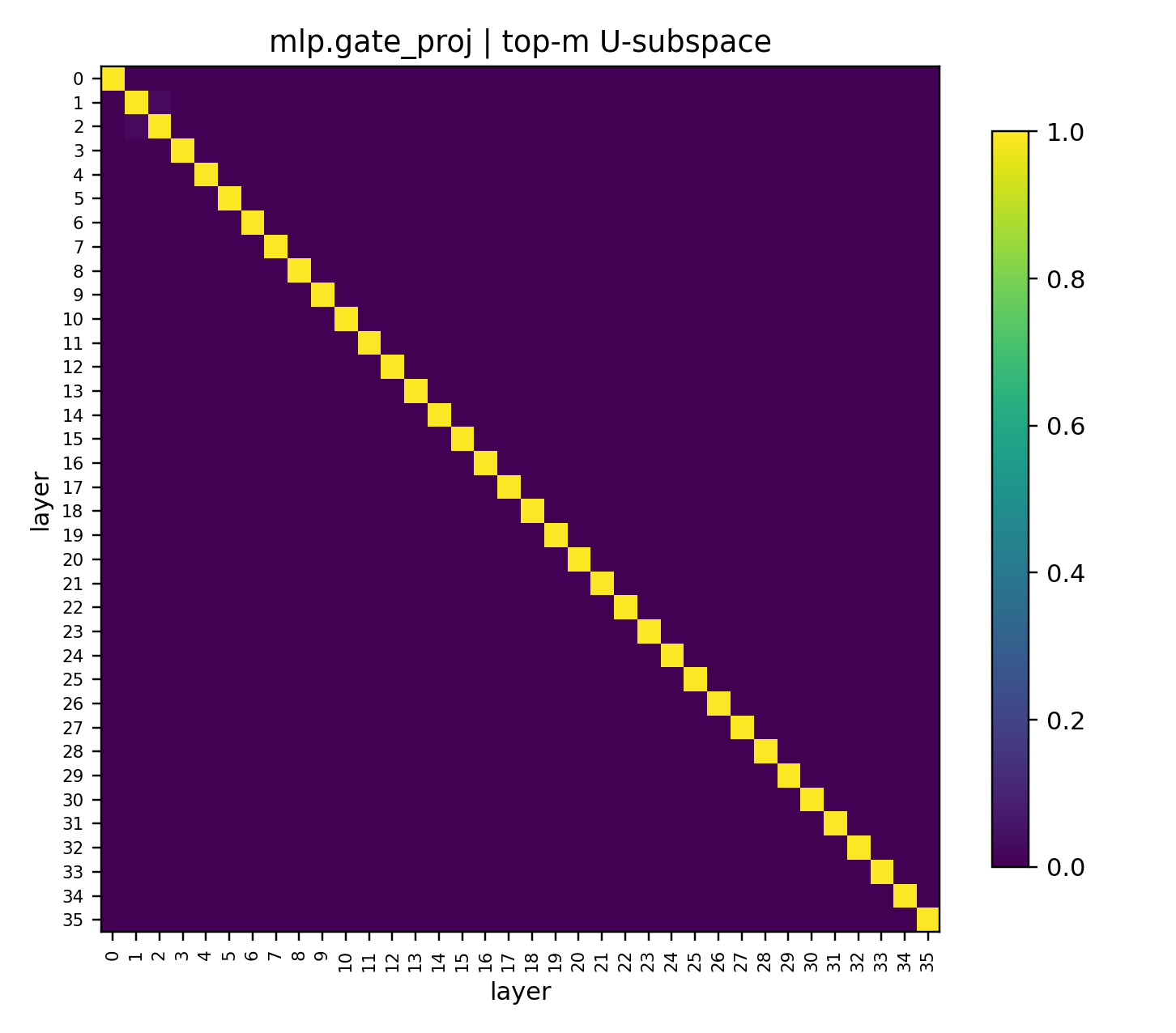} &
\includegraphics[width=0.195\textwidth]{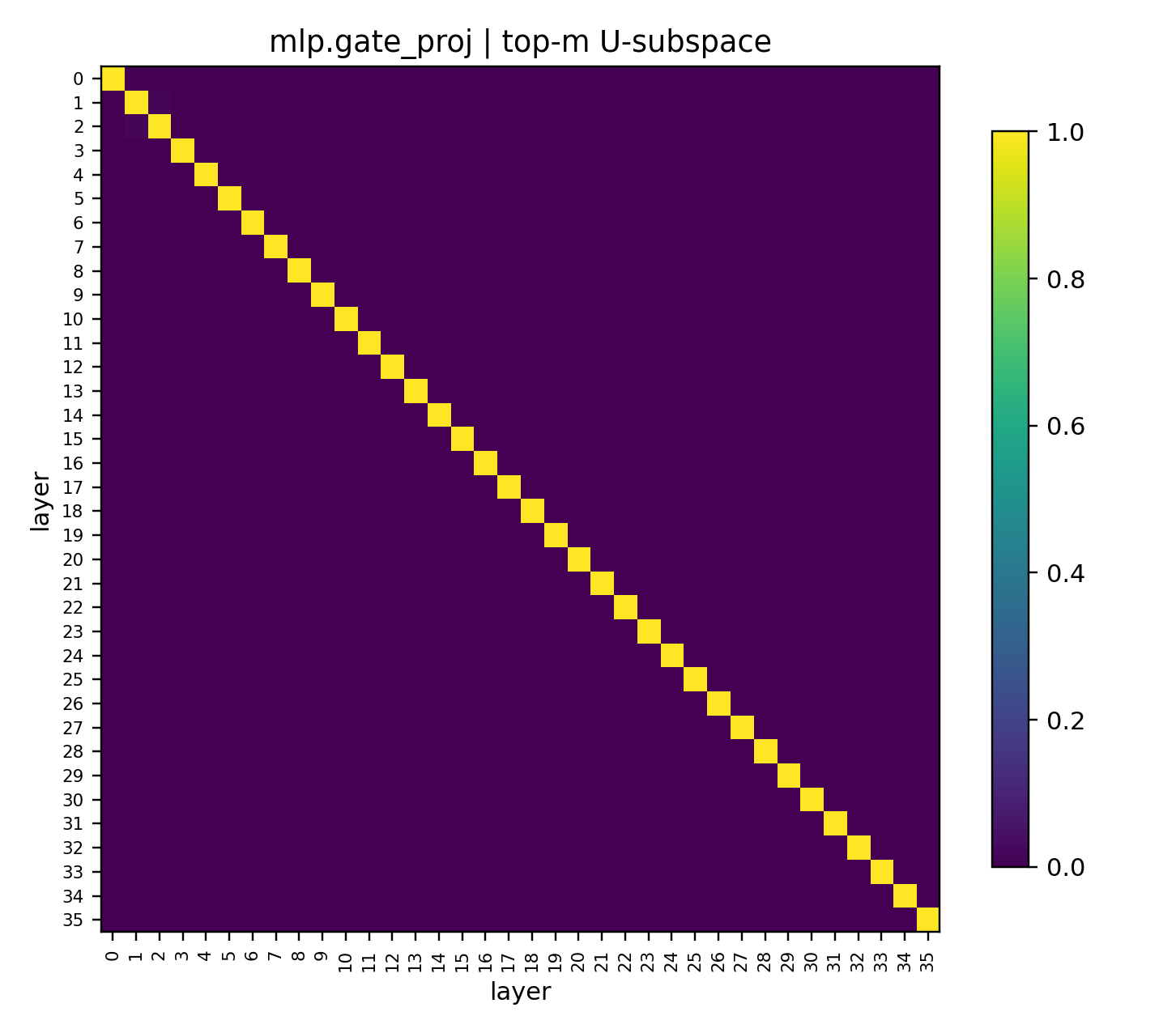} \\
\texttt{up\_proj} &
\includegraphics[width=0.195\textwidth]{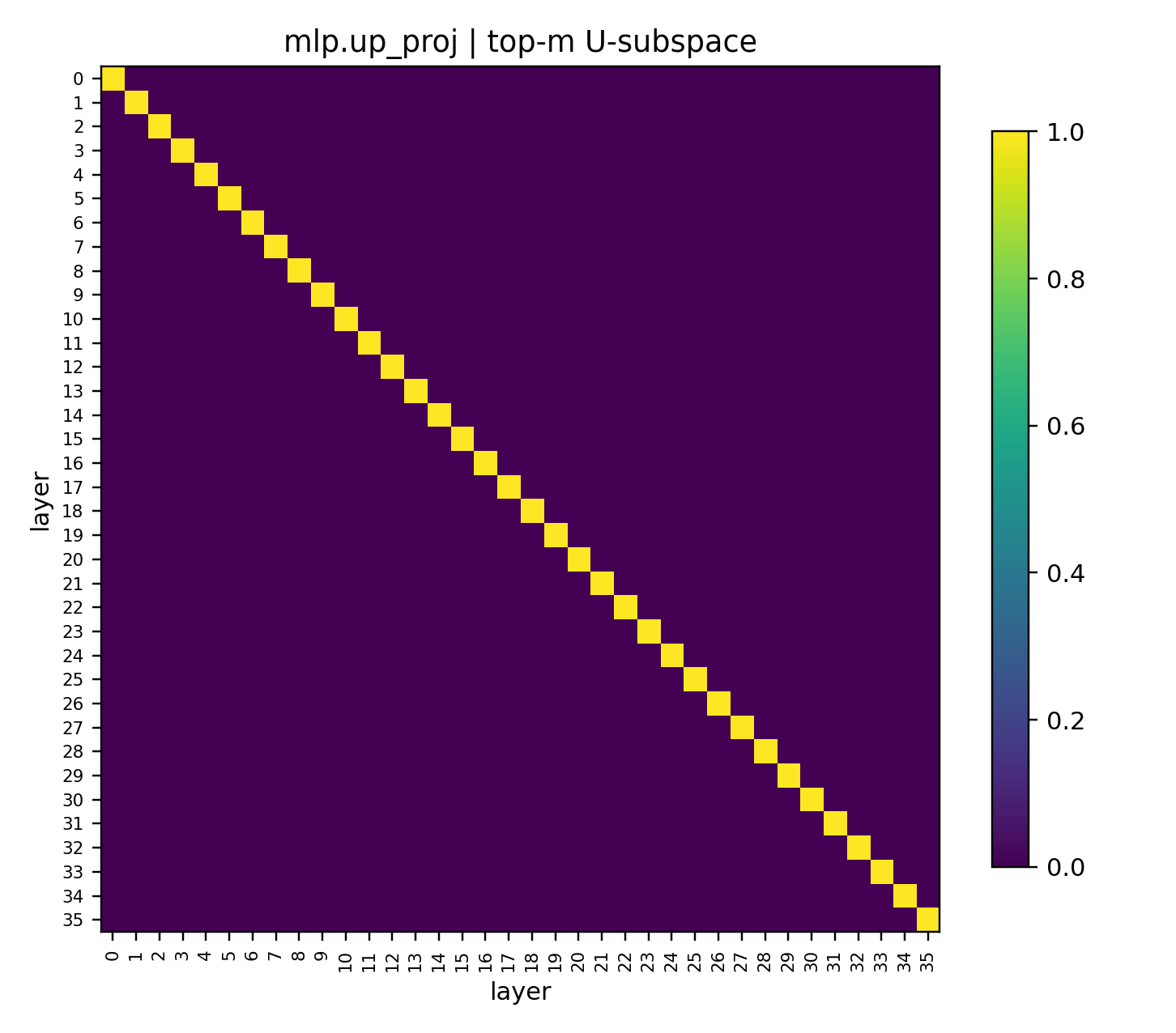} &
\includegraphics[width=0.195\textwidth]{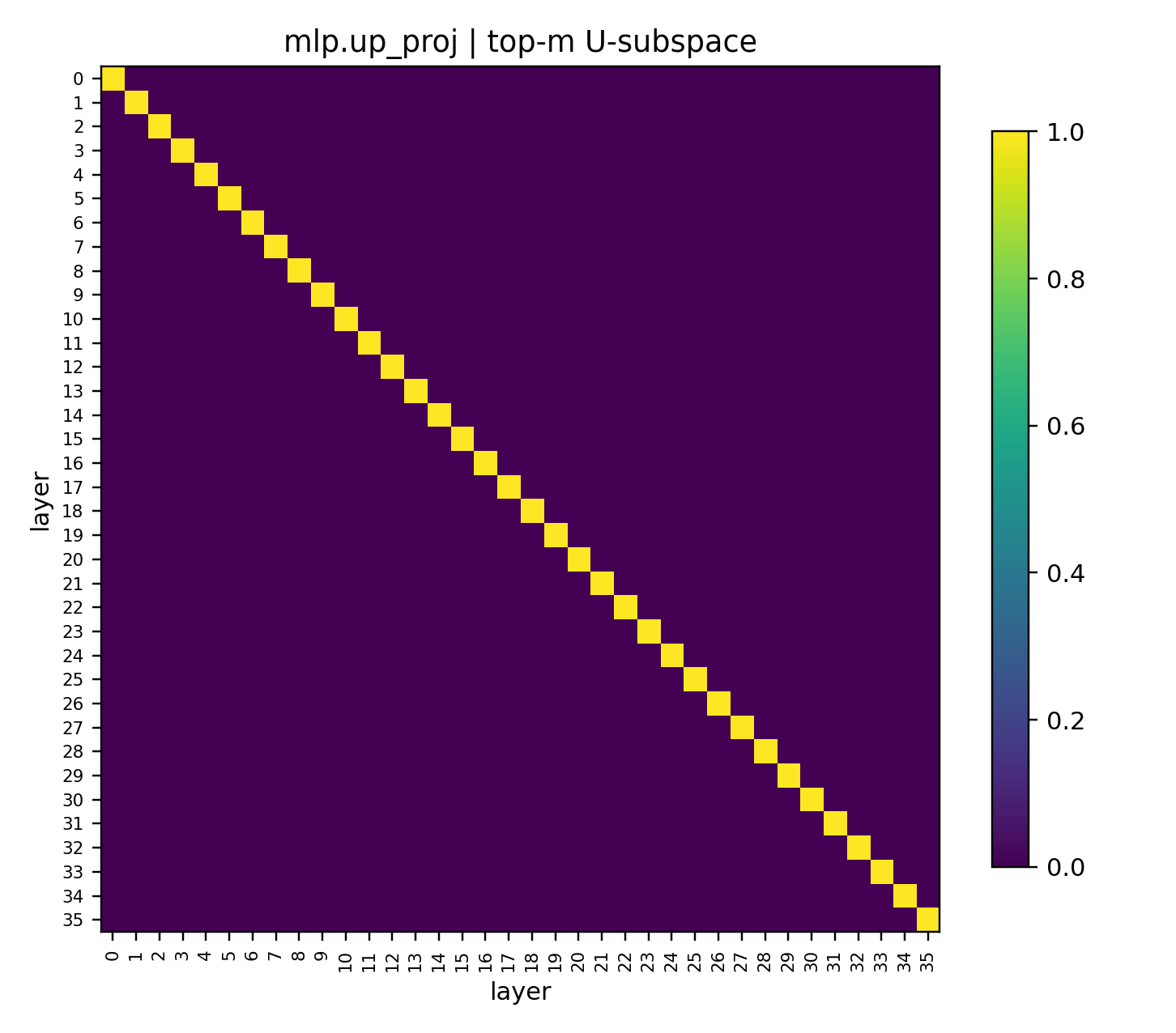} &
\includegraphics[width=0.195\textwidth]{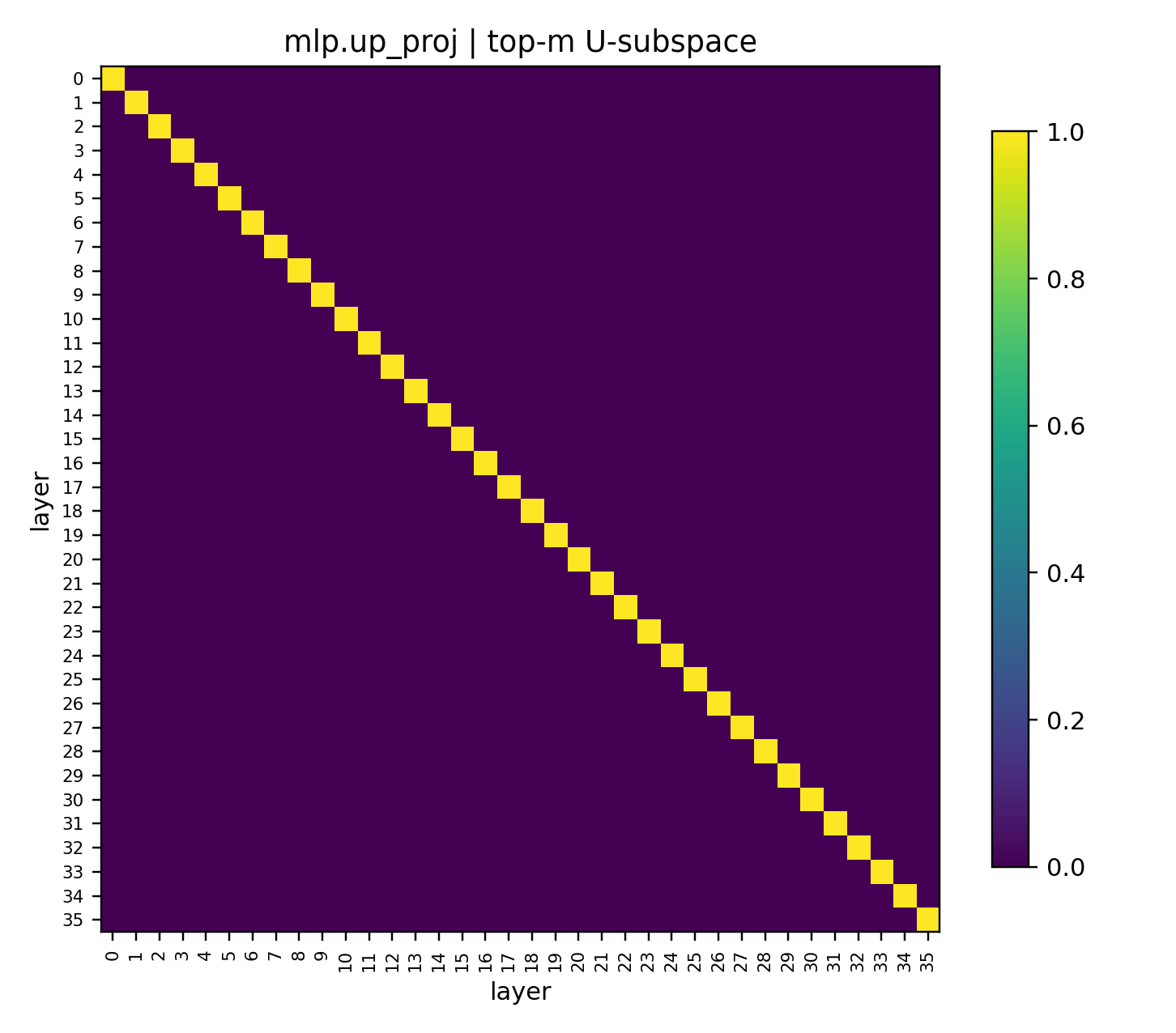} &
\includegraphics[width=0.195\textwidth]{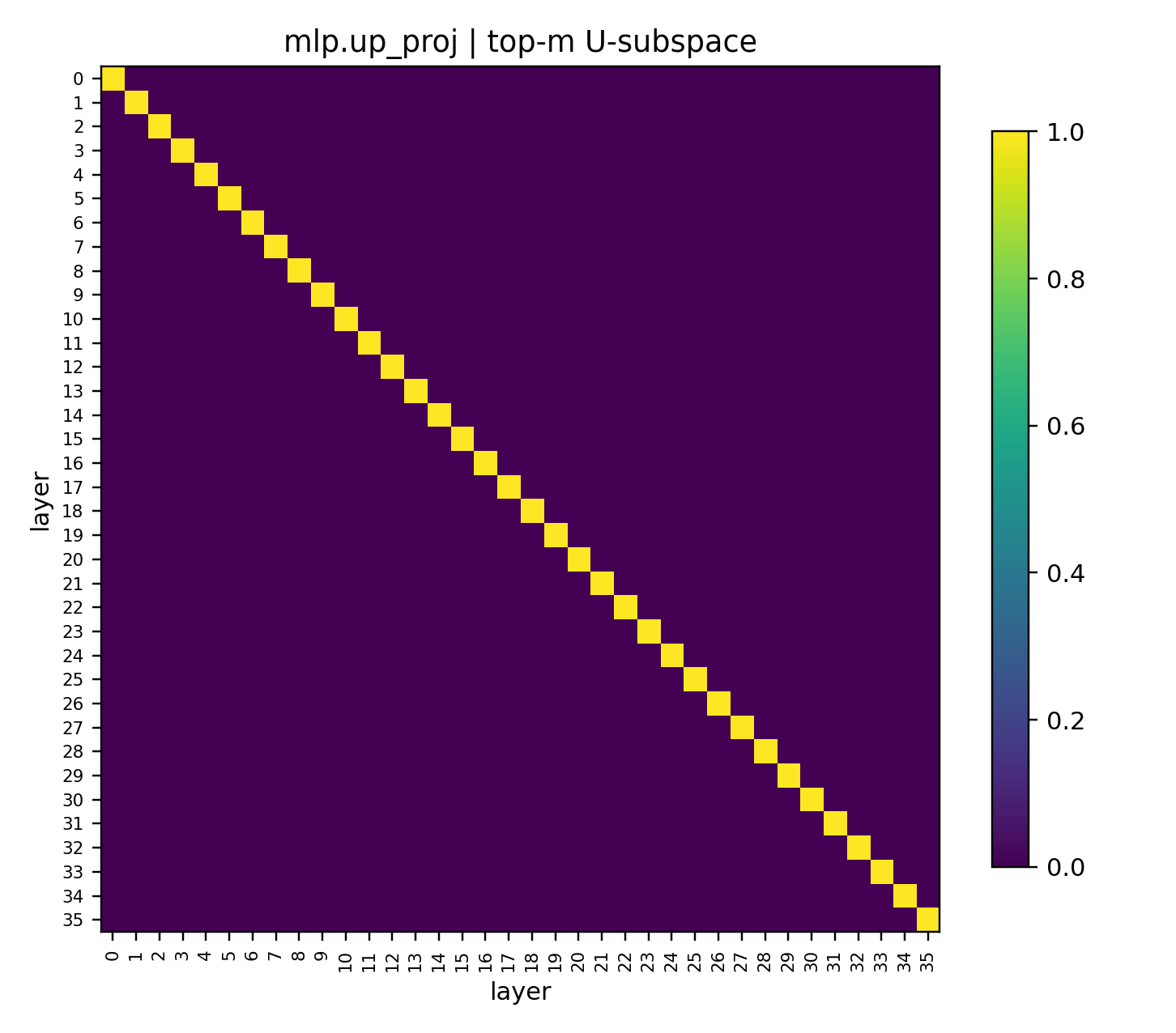} \\
\texttt{down\_proj} &
\includegraphics[width=0.195\textwidth]{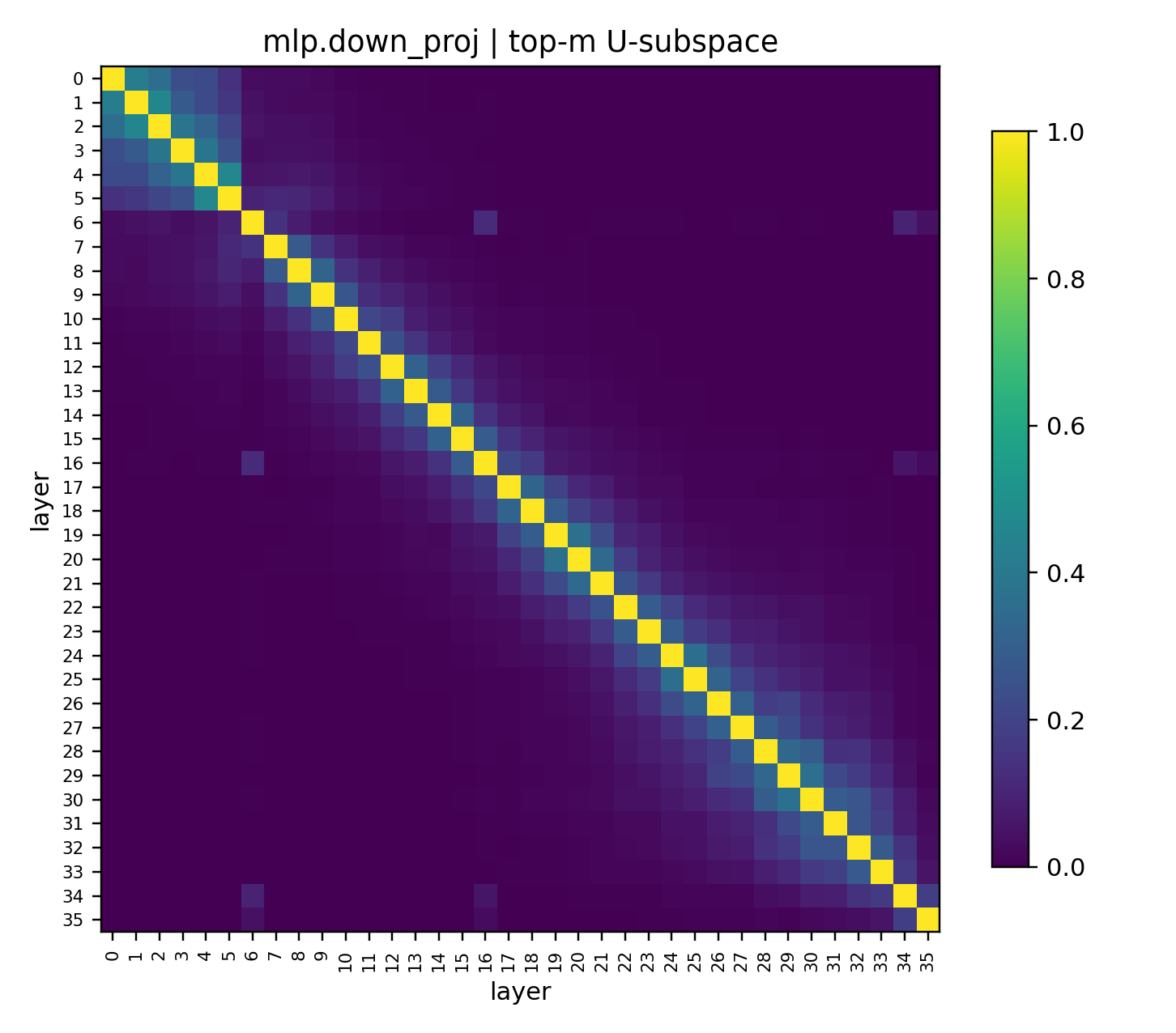} &
\includegraphics[width=0.195\textwidth]{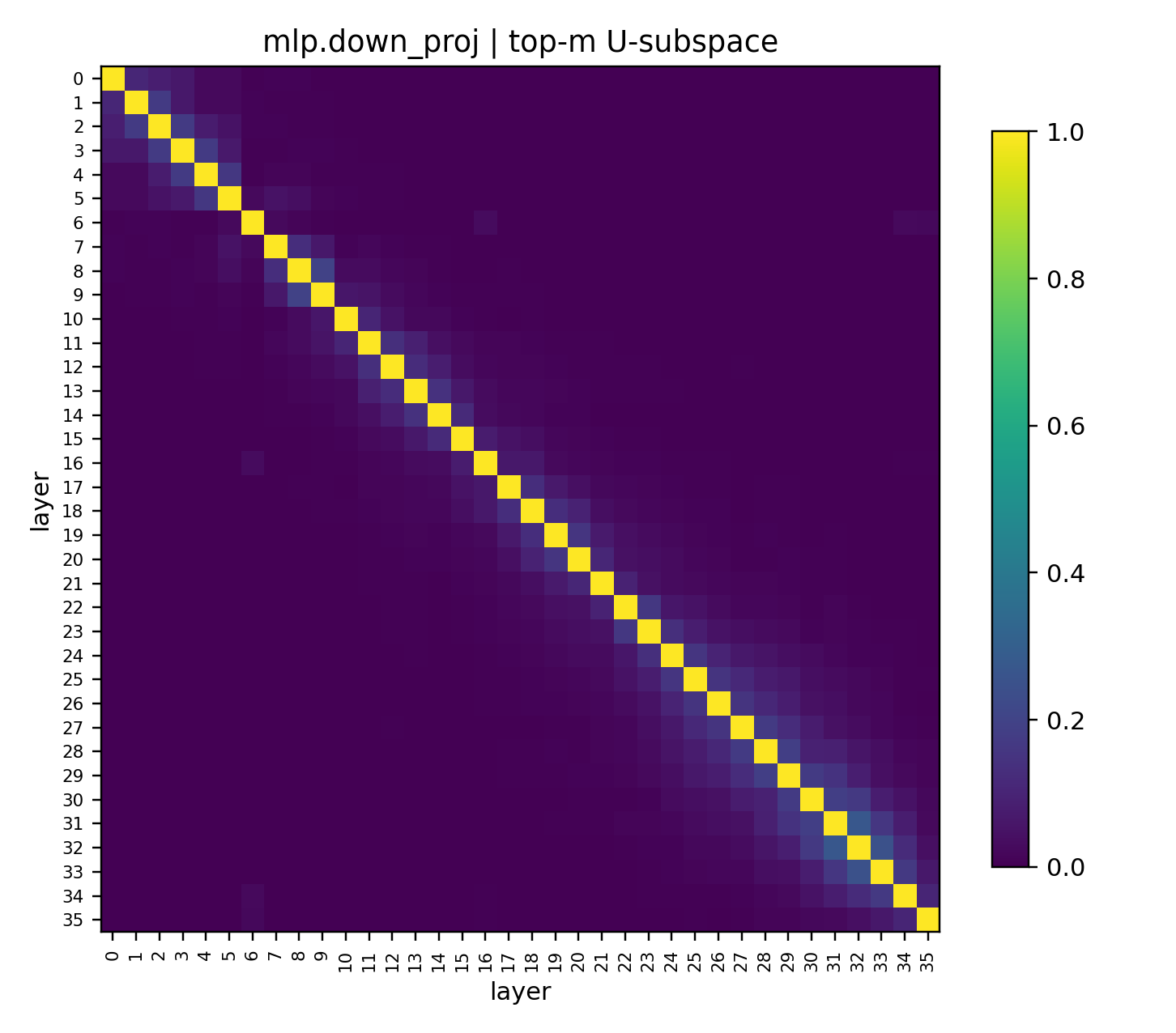} &
\includegraphics[width=0.195\textwidth]{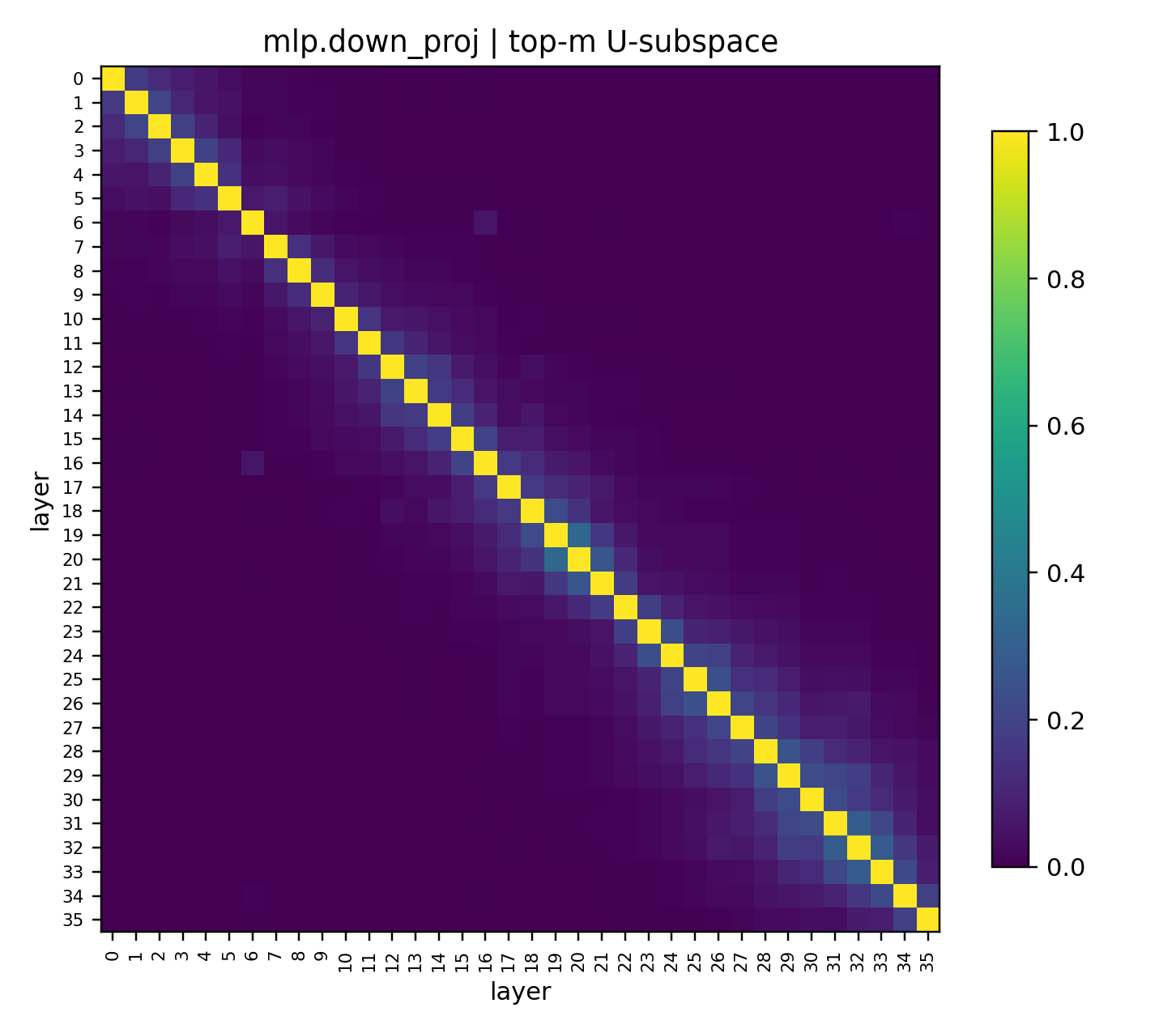} &
\includegraphics[width=0.195\textwidth]{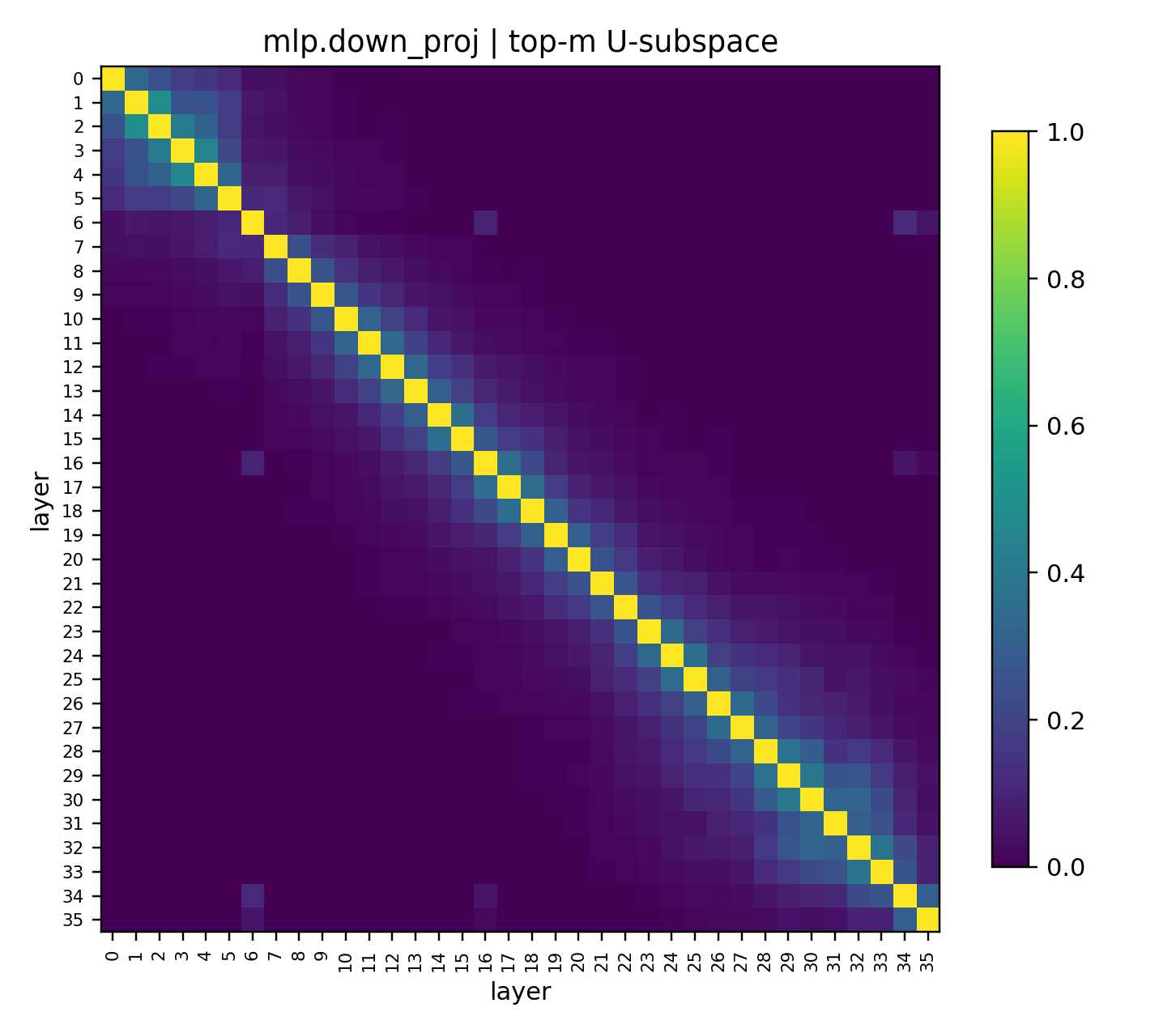} \\
\bottomrule
\end{tabular}
\caption{\textbf{Qwen3-8B: Top-$m$ output-subspace overlap heatmap wall ($\mathrm{Align}_U$, $m{=}4$).}
Each cell shows the inter-layer subspace-overlap heatmap for a specific (task, module).}
\label{fig:heatwall_qwen_subU}
\end{figure*}

\end{document}